\documentclass[aps,reprint,superscriptaddress]{revtex4-2}

\usepackage{graphicx}
\usepackage{array}
\usepackage{amssymb}
\usepackage{amsfonts}
\usepackage{amsmath}
\usepackage{mathrsfs}
\usepackage{multirow}
\usepackage{times}
\usepackage{chngpage}

\begin{document}
\preprint{APS/123-QED}
\title{Data organization limits the predictability of binary classification}

\author{Fei Jing}
\affiliation{Musketeers Foundation Institute of Data Science, The University of Hong Kong, Hong Kong SAR, China}

\author{Zi-Ke Zhang}\email{zkz@zju.edu.cn}
\affiliation{Center for Digital Communication Studies, Zhejiang University, Hangzhou 310058, China}

\author{Yi-Cheng Zhang}\email{yi-cheng.zhang@unifr.ch}
\affiliation{Department of Physics, University of Fribourg, Chemin du Mus{\'e}e 3, 1700 Fribourg, Switzerland}

\author{Qingpeng Zhang}\email{qpzhang@hku.hk}
\affiliation{Musketeers Foundation Institute of Data Science, The University of Hong Kong, Hong Kong SAR, China}

\begin{abstract}
The structure of data organization is widely recognized as having a substantial influence on the efficacy of machine learning algorithms, particularly in binary classification tasks. Our research provides a statistical physics framework from the perspective of statistical physics, revealing that the upper bound of binary classification predictability on a given dataset is equivalent to obtaining the ground state energy of the canonical ensemble of classifiers on fixed data patterns, primarily constrained by the inherent qualities of the data. Through both theoretical reasoning and empirical examination on dozens of real datasets, we employed standard objective functions, evaluative metrics, and binary classifiers to arrive at two principal conclusions.
Firstly, we show that the theoretical upper bound of binary classification on actual datasets can be theoretically attained as the ground state energy of corresponding partition functions, regardless of any specific algorithms. This upper boundary represents a calculable equilibrium between the learning loss and the metric of evaluation. Secondly, we have computed the precise upper bounds for three commonly used evaluation metrics, uncovering a fundamental uniformity with our overarching thesis: the upper bound is intricately linked to the organization of dataset's characteristics, independent of the classifier in use.
Furthermore, our subsequent analysis uncovers a detailed relationship between the upper limit and the level of class overlap within the binary classification data. This relationship is instrumental for pinpointing the most effective feature subsets for use in feature engineering.
\end{abstract}

\maketitle

The majority of machine learning algorithms has predominantly concentrated on enhancing model accuracy, yet leaving the underlying mechanism of the predictability of machine learning problems remains a \emph{black-box} \cite{luchnikov2022probing} and insufficiently understood~\cite{hastie2009elements,agresti2012categorical,noble2006support,chen2016xgboost}. A thorough quantification of predictability is paramount for the thoughtful design, training, and deployment of machine learning models, shedding light on the capabilities and limitations of AI in practical scenarios.
	
Conventional machine learning has often favored error-based metrics as objective functions. However, these metrics have proven to have a negligible impact on optimization outcomes~\cite{rosasco2004loss} and are prone to complications arising from imbalanced datasets in classification challenges~\cite{cortes2003auc}. This recognition has sparked a surge in research directed at AUC maximization within classification tasks~\cite{yang2022auc}, spawning a variety of learning frameworks~\cite{rebentrost2014quantum,joachims2005support}, methodologies~\cite{ying2016stochastic}, optimization techniques~\cite{norton2019maximization}, and their successful implementation in a myriad of fields~\cite{yuan2021large}. Theoretical insights have also been advanced regarding consistency~\cite{Gao2012OnTC} and bounding generalization errors~\cite{barbier2019optimal,opper1991generalization}.
	
	Although research has made strides in optimizing binary classifiers and demonstrating that an optimal classifier produces a convex ROC curve~\cite{fawcett2006introduction}, the maximal AUC remains undisclosed. While efforts to probe link predictability using methods such as information entropy~\cite{sun2020revealing} and random turbulence~\cite{lu2015toward} are notable, their reliance on stringent assumptions limits their applicability to extensive, real-world datasets. A gap persists in analytically defining the general predictability of machine learning in practical applications.

\begin{table*}

	\caption{Comprehensive experimental results and theoretical upper bound analysis for various binary classifiers on four real datasets. The classifiers evaluated include XGBoost, Multilayer Perceptron (MLP), Support Vector Machine (SVM), Logistic Regression (LR), Decision Tree (DT), Random Forest (RF), K-Nearest Neighbors (KNN), and Naive Bayes (NB). Performance metrics for each classifier are presented in terms of area under the ROC curve (AUC-ROC, abbreviated as AR), area under the Precision-Recall curve (AUC-PR, abbreviated as AP), and Accuracy (AC). 
    }
 \begin{ruledtabular}
	\begin{tabular}{lccccccccccccc}
		\multirow{2}*{Data}&\multirow{2}*{Metrics} &  \multicolumn{4}{c}{XGBoost}& \multirow{2}*{MLP}& \multirow{2}*{SVM}& \multirow{2}*{LR}& \multirow{2}*{DT}& \multirow{2}*{RF}& \multirow{2}*{KNN}& \multirow{2}*{NB}&\multirow{2}*{Boundaries}\\
		\cline{3-6}
		~&~& Square&Logistic& Hinge&Softmax&~&~&~&~&~&~&~&~\\
		\hline
		\multirow{3}*{AID}&AR&0.6941&0.6917&0.6096&0.6223&0.5777&0.5&0.4921&0.7863&0.7961&0.7464&0.5228&0.9112\\
		~&AP&0.6517&0.6491&0.5245&0.5373&0.4939&0.4454&0.4424&0.7232&0.7065&0.6483&0.4606&0.8806\\
		~&AC&0.6491&0.6476&0.634&0.6469&0.5957&0.5546&0.5546&0.8015&0.8015&0.7533&0.5692&0.8015\\
		\hline
		\multirow{3}*{HED}&AR&0.7959&0.7948&0.7249&0.7337&0.6127&0.595&0.619&0.9998&0.9997&0.7149&0.6103&1.0\\
		~&AP&0.7702&0.769&0.6541&0.6652&0.5767&0.5562&0.574&0.9998&0.9996&0.6523&0.5667&1.0\\
		~&AC&0.7349&0.7347&0.7249&0.7337&0.6126&0.595&0.619&0.9998&0.9997&0.7149&0.6103&0.9998\\
		\hline
		\multirow{3}*{INE}&AR&0.9275&0.9239&0.7543&0.7788&0.7467&0.6229&0.6737&0.9728&0.9638&0.794&0.7336&0.9983\\
		~&AP&0.9751&0.9738&0.8659&0.8778&0.8627&0.8071&0.829&0.9844&0.9783&0.886&0.8577&0.9993\\
		~&AC&0.8734&0.8699&0.857&0.8704&0.839&0.8008&0.8043&0.9764&0.9764&0.8635&0.7913&0.9764\\
		\hline
		\multirow{3}*{SUD}&AR&0.9645&0.9322&0.8807&0.8603&0.6855&0.7206&0.665&0.8742&0.8807&0.8105&0.6773&0.9649\\
		~&AP&0.9396&0.9031&0.7973&0.7848&0.5309&0.5822&0.514&0.8028&0.7973&0.6845&0.5038&0.9381\\
		~&AC&0.9038&0.8942&0.9038&0.8942&0.7596&0.7885&0.75&0.9038&0.9038&0.8462&0.7404&0.9038\\
	\end{tabular}
 \end{ruledtabular}
	\label{tab1}
\end{table*}

	Consider a binary dataset $\mathcal{S}=\{(x_i, y_i) \mid i=1,2,\ldots,m \}$, where $m$ is the number of samples, and each consists of a $k$-dimensional feature vector $x_i = (x_{i,1}, x_{i,2}, \ldots, x_{i,k}) \in \mathcal{X}$ and a corresponding class label $y_i \in \{1, -1\}$, where $\mathcal{X}$ is the domain comprising all samples in $\mathcal{S}$. For a given sample $x_i$, $\mathcal{P}(x_i)$ and $\mathcal{N}(x_i)$ indicate the count of positive and negative samples, respectively. Additionally, the entire dataset contains $n_+$ positive and $n_-$ negative samples. The proportion of positive instances for a particular feature vector $x_i$ is expressed as $p_+(x_i)=\mathcal{P}(x_i)/(\mathcal{P}(x_i)+\mathcal{N}(x_i))$.

     The objective of binary classification models is to assign a sample $x_i$ to a binary class by employing a real-valued classifier $f: \mathcal{X} \rightarrow \mathbb{R}$ to gauge the probability that an instance belongs to each class. Nowadays, physical principles have been widely acknowledged to be capable to provide essential understanding of machine learning algorithms \cite{hou2020statistical, wang2024tensor, liu2024kan, Kording2004BayesianII,Darlington2018NeuralIO, Carleo2017Solving,Levin2007PRL}. Essentially, such process is typically an optimization problem if one can traverse all potential classifiers, and can be naturally expressed by statistical mechanics \cite{mezard1987spin}. That is, the configuration of all possible classifiers is weighted by the Boltzmann factor $e^{-\beta E(f)}$, where $T=1/\beta$ is the artificial temperature and $E(f)$ is the energy function. Then, the corresponding partition function can be written as $Z=\sum_f e^{-\beta E(f)}$, and the ground state energy, i.e., the minimum cost, is $E_0=\lim_{\beta \rightarrow \infty} -\frac{1}{\beta} \ln{Z}$. Apparently, binary classifiers are akin to the configurations of a canonical ensemble and its outcome space is \emph{m}! once dataset is fixed. In this Letter, we further demonstrate that it achieves the optimal ROC and PR curves across general datasets. This upper bound is an intrinsic property of the dataset itself, independent of the specific classifier employed.  To assess the predictability, in the following, we investigate three prominent evaluation metrics: the ROC curve ($AR$), the PR curve ($AP$), and accuracy ($AC$), respectively with three energy functions, $E^{AR}$, $E^{AP}$, and $E^{AC}$ and partition functions, $Z^{AR}$, $Z^{AP}$, and $Z^{AC}$. Each of these metrics evaluates classifier predictability from distinct perspectives.




{\bf ROC curve.}
	The ROC curve is a fundamental tool for displaying the discriminative capacity of a binary classifier across various threshold settings~\cite{fawcett2006introduction,metz1978basic}. It delineates the trade-off between sensitivity (true positive rate) and specificity (true negative rate), assisting in the selection of an optimal threshold. The area under the ROC curve (AUC-ROC) quantifies the probability that a randomly chosen positive instance is ranked higher than a negative instance, serving as a summary measure of classifier performance across all thresholds~\cite{hanley1982meaning,bradley1997use}. As proved in the Supplemental Material, we have established the exact upper bound of the AUC ($AR^{u}$) for a given binary dataset $\mathcal{S}$ as follows, 
	
	\begin{equation}\label{eq_auc}
		\text{AR}^{u}=\frac{1}{2n_+n_-}\sum\limits_{x_i,x_j}\max\Big\{\mathcal{P}(x_i)\mathcal{N}(x_j) ,\mathcal{P}(x_j)\mathcal{N}(x_i) \Big\},
	\end{equation} 
	where $x_i$ and $x_j$ are two arbitrary feature vectors from $\mathcal{S}$. This upper bound is attainable by a binary classifier if and only if it satisfies $f^*_{\text{AR}}(x_i) \sim p_+(x_i)$, where $f(x_i) \sim f'(x_i)$ denotes that $f(x_i) > f(x_j)$ if and only if $f'(x_i) > f'(x_j)$ for every pair $(x_i,x_j)$ in $\mathcal{S}$. For the classifier ensemble , the partition function is $ Z^{AR} = \sum_{\pi^f \in \Pi(\mathcal{S})} \exp(-\beta E(\pi^f))$ where   $ E(\pi^f) = \sum_{i > j} p_+(x^f_{i}) p_-(x^f_{j}) + \frac{1}{2}\sum_i p_+(x^f_{i}) p_-(x^f_{i})$ is the energy function of $AR$, and $\pi^f$ is the set of ordered samples under a specific binary classifier $f$. As a consequence, the upper bound of the predictability for observed dataset $\mathcal{S}$ is exactly the complementarity of corresponding ground state energy $E^{AR}_0=\lim\limits_{\beta \rightarrow \infty} -\frac{1}{\beta} \ln{Z^{AR}} =\frac{1}{2n_+n_-}\sum\limits_{x_i,x_j}\min\Big\{\mathcal{P}(x_i)\mathcal{N}(x_j) ,\mathcal{P}(x_j)\mathcal{N}(x_j) \Big\}$ (details see Supplemental Material). Note that, it shows that the ground state energy, yielding the binary classification predictability, only relies on the organization of given dataset $\mathcal{S}$, independent of the concrete form of binary classifiers.

\begin{figure*}[ht]
		\centering
		\includegraphics[width=.85\linewidth]{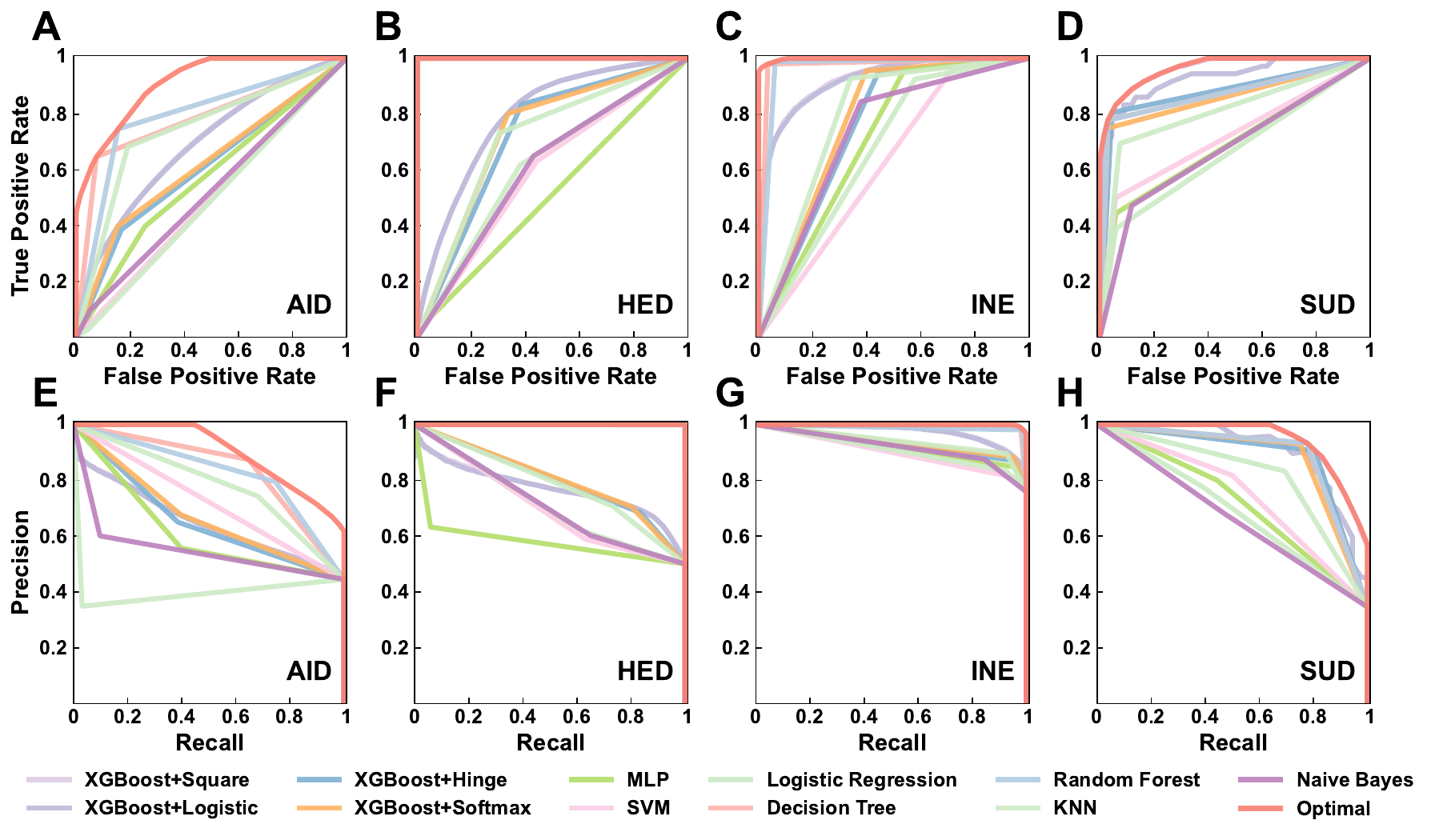}
		\caption{Demonstrating the precise upper bound of AUC with accompanying optimal ROC curves (A-D) and PR curves (E-H) across four real-world datasets utilizing a suite of binary classifiers. The classifiers include XGBoost, MLP, SVM, Logistic Regression, Decision Tree, Random Forest, KNN, and Naive Bayes. The red curves symbolize the theoretical optimal PR curves. Detailed values for each classifier-dataset combination are documented in Table \ref{tab1}. 
  }
		\label{fig1}
	\end{figure*}

Fig. \ref{fig1}A-\ref{fig1}D depicts the results of in-sample experiments featuring a suite of well-known binary classifiers, including XGBoost, Multilayer Perceptron (MLP), Support Vector Machine (SVM), Logistic Regression, Decision Tree, Random Forest, KNN, and Naive Bayes. These classifiers were tested with a variety of objective functions. In addition, the theoretical upper limits were computed in accordance with Eq.~\ref{eq_auc}. Across all cases, the theoretical upper bounds exceeded the performance of each classifier as measured by the ROC curves, thereby confirming the accuracy of our theoretical predictions.

{\bf PR curve.}
Unlike the ROC curve, the PR (precision-recall) curve is particularly useful for evaluating performance on imbalanced datasets where one class significantly outnumbers the other \cite{tran2017novo}. The area under PR curve (AUC-PR, $\text{AP}$) illustrates the precision-recall trade-off as the classification threshold is adjusted. However, the ground state energy of its corresponding partition function $Z^{AP} = \sum_{\pi^f \in \Pi(\mathcal{S})} \exp\Big(-\beta \left(1-AP(\pi^f)\right)\Big)$, can not be explicitly derived as the expression of $\text{AP}$ is mathematically ambiguous. We alternatively demonstrate that the theoretical upper bound of the area under the PR curve ($\text{AP}^{u}$) for a binary dataset $\mathcal{S}$ is given by (see Supplemental Material for the mathematical proof):

\begin{equation}\label{eq_pr}
\text{AP}^{u} \!=\!\sum\limits_{i=1}^m    \frac{p_+(x^*_i)}{2n_+}
\!\left(\!\frac{1}{i-1}\sum\limits_{j=1}^{i-1} p_+(x^*_j) \!+\!   \frac{1}{i} \sum\limits_{j=1}^i  p_+(x^*_j)\!  \right)\!,
\end{equation}
where $x^*_i$ and $x^*_j$ are feature vectors sorted according to $f^*_{\text{AP}}(x_i)$. It is achievable by a binary classifier if and only if it satisfies $f^*_{\text{AP}}(x_i) \sim p_+(x_i)$ for every $x_i \in \mathcal{S}$, mirroring the condition for $\text{AR}^u$.

Fig.~\ref{fig1}E-\ref{fig1}H presents the experimental findings with aforementioned representative binary classifiers against the theoretical upper bounds derived using Eq.~\ref{eq_pr}. As with the ROC curve, the experimental results for binary classification do not exceed the theoretical upper bounds ($\text{AP}^u$), thereby supporting the validity of our theoretical derivation.

{\bf Accuracy.}
Accuracy (AC) is the fraction of instances correctly classified by the classifier~\cite{congalton1991review,madani2018fast}. The ground state energy of its corresponding partition function $Z^{AC} = \sum_{\pi^f\in \Pi(\mathcal{S})} \exp\left\{-\beta\left( \frac{1}{2} +  \frac{1}{2m} \sum_{i} x^f_i \Big(\mathcal{N}(x_i) - \mathcal{P}(x_i)\Big) \right) \right\}$, can be obtained as $E^{AC}_0 = \frac{1}{m} \sum\limits_i \min\Big\{\mathcal{P}(x_i),\mathcal{N}(x_i)  \Big\}$. And the theoretical upper bound (detailed in Supplemental Material) reads

\begin{equation}\label{eq_ac}
	\text{AC}^u =\frac{1}{m} \sum\limits_{x_i} \max\Big\{ \mathcal{P}(x_i),\mathcal{N}(x_i) \Big\}   .
\end{equation}
This is again exactly the complementarity of its ground state energy $AC^u=1-E^{AC}_0$, indicating the statistical classifiers ensemble classifiers on a fixed data organization indeed governs the practicability of binary classification. In addition, This upper limit is attainable if, and only if, the optimal classifier dictates that:

\begin{eqnarray}\nonumber
f^*_{\text{AC}}(x_i) = \left\{
\begin{aligned}
	1 & & \mathcal{P}(x_i) \geq \mathcal{N}(x_i) \\
	-1& & \mathcal{P}(x_i) < \mathcal{N}(x_i)
\end{aligned}
\right.,
\end{eqnarray}
for each $x_i$ in $\mathcal{S}$. Table~\ref{tab1} compares the accuracy achieved by various classifier configurations against the theoretical upper bounds, further substantiating the rigour of our theoretical derivation. 

{\bf A universal optimal classifier.}
The theoretical foundation of optimal classifiers demonstrates their alignment with the common objective functions including Square loss, Logistic loss, Hinge loss, and Softmax loss~\cite{hastie2009elements} (refer to Supplemental Material for additional details). Specifically, the optimal classifiers for both the ROC and PR curves are congruent, denoted as $f^*_{\text{AR}} \equiv f^*_{\text{AP}}$. This implies that continuous binary classifiers converge to a unified optimal form for each feature vector $x_i \in \mathcal{S}$, symbolized as $g^*_{\text{Square}} \equiv g^*_{\text{Logistic}} \sim p_+(x_i) \equiv f^*_{\text{AR}} \equiv f^*_{\text{AP}}$, where $g^*$ represents the objective function. Similarly, discrete binary classifiers are equivalent, denoted as $g^*_{\text{Hinge}} \equiv g^*_{\text{Softmax}} \equiv f^*_{\text{AC}}$.

From this, we can deduce a universal representation for any optimal classifier, encompassing both continuous and discrete forms, that achieves the best performance across various objective functions and evaluation metrics:

\begin{equation}\label{eq_universal}
	\mathcal{F}^*(x_i)=\left\{
	\begin{aligned}
		1&  & f^*(x_i)\geq 0 \\
		-1&  & f^*(x_i) < 0
	\end{aligned}
	\right. 	,
\end{equation}
where $f^*(x_i) = \frac{\mathcal{P}(x_i) - \mathcal{N}(x_i)}{\mathcal{P}(x_i) + \mathcal{N}(x_i)}$ is the derived optimal classifier that minimizes loss and maximizes performance across various objective functions and evaluation metrics. In addition, it clearly shows that $f^*(x_i)$ only depends on the data organization rather than algorithms which is the focus of previous study \cite{kim2021fermi}. 

Eq.~\ref{eq_universal} underscores the synergy between the learning process, which is steered by objective functions, and the evaluation process within the realm of binary classification. The efficacy of a binary classifier, irrespective of its computational complexity or efficiency, is inherently bound by the selected objective function and the intrinsic properties of the dataset. This establishes a performance ceiling dictated by both the chosen evaluation metric and the data's innate characteristics.  

\begin{figure*}[ht]
	\centering
	\includegraphics[width=.75\linewidth]{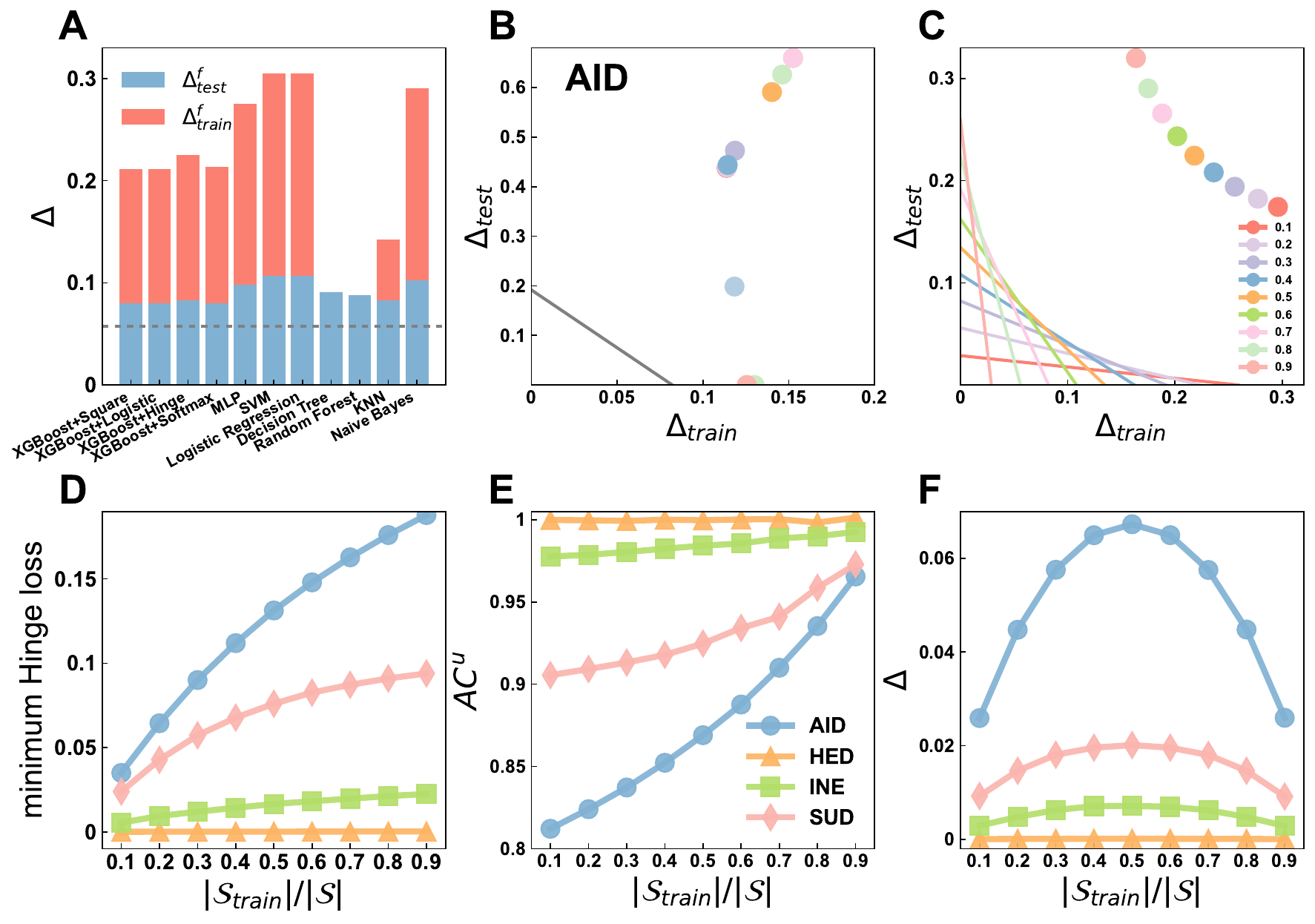}
	\caption{Detailed evaluation of expected errors ($\Delta$) across all feature vectors for each dataset. Panels A and B showcase the error dynamics for the AID dataset during training ($\Delta_{train}^{f}$) and testing ($\Delta_{test}^{f}$) phases, considering a random division ratio where the training set size $|\mathcal{S}_{train}|$ constitutes 70\% of the total set size $|\mathcal{S}|$. (A) The dashed line indicates the expected error of the optimal classifier as determined by Eq.~\ref{eq_delta}. (B) Illustrates the correlation between $\Delta_{train}^{f}$ and $\Delta_{test}^{f}$, contrasting the theoretical solution outlined in Eq.~\ref{eq_delta} (gray line) against empirical observations from diverse binary classifiers (represented by dots). The legend is identical to that of Fig.~\ref{fig1}. (C) Explores the interplay between $\Delta_{train}^{f}$ and $\Delta_{test}^{f}$ under a spectrum of random division ratios $|\mathcal{S}_{train}|/|\mathcal{S}|$, varying from 10\% to 90\%. The classifier we used here is XGBoost with Square loss. (D-F) depict the expected errors or performance metrics for each dataset under different random divisions: (D) Hinge loss, (E) upper bound of accuracy ($AC^{u}$), and (F) anticipated optimal errors ($\Delta$). In these panels, dots correspond to numerical results derived from the data divisions, while lines represent the theoretical predictions adjusted for the respective division ratios (as detailed in Eqs. 93 and 95 within the Supplemental Material). More detailed experimental results on 41 datasets are shown in Supplemental Material).
	}
	\label{fig_delta}
\end{figure*}

Note that, the aforementioned binary boundary theory is derived in an in-sample context, where the classification and validation are based on the full dataset. As a consequence, a fundamental question remains: To what extent can the boundary theory derived from full dataset be applied to the out-of-sample scenario? We further perform out-of-sample validation, and sensitivity analysis with various random separation ratio on four
real datasets and synthetic data (see Supplemental Material). 

Specifically, we divide the data to a training set ($\mathcal{S}_{train}$) and a test set ($\mathcal{S}_{test}$). The model is trained on the training set and validated on the test set. 
The performance of the learned classifier varies given different statistical characteristics of the data in the test set, even through it is trained guided by the objective function based on the data organization in the training set. Hence, the generalization ability of the classifier is determined by the gap between data distributions of training and test sets~\cite{10239439,6881630}.  
Here, we perform extensive sensitivity analysis of the optimization and evaluation process.  Without loss of generality, we focus on the analysis of Hinge loss and accuracy (\text{AC}).

We start by defining $\mathcal{P}_{train}(x_i)$ and $\mathcal{N}_{train}(x_i)$ as the number of positive and negative instances with feature vector $x_i$ in $\mathcal{S}_{train}$, $\mathcal{P}_{test}(x_i)$ and $\mathcal{N}_{test}(x_i)$ as the number of positive and negative instances with feature vector $x_i$ in $\mathcal{S}_{test}$. Given a discrete classifier $f(x_i)$, we further define $\Delta_{train}^{f}$ as the bias between the actual Hinge loss and the optimal loss in $\mathcal{S}_{train}$. Similarly, $\Delta_{test}^{f}$ is defined as the bias between the actual accuracy and optimal accuracy in $\mathcal{S}_{test}$. Therefore, the question is converted to minimizing the summation of the two errors  
\begin{equation}\label{eq_delta}
	\Delta = \min_{f} (\Delta_{train}^{f}+\Delta_{test}^{f})=\sum\limits_{x_i}\Delta(x_i)
\end{equation}
Here,
\begin{eqnarray}
    \Delta(x_i)=\begin{cases} 
    0 &  \text{if } \mathcal{Q}_{train}(x_i)\mathcal{Q}_{test}(x_i) \geq 0\\
    \epsilon(x_i) &  \text{if } \mathcal{Q}_{train}(x_i)\mathcal{Q}_{test}(x_i) < 0
     \end{cases},
\end{eqnarray}
where $\mathcal{Q}_{train}(x_i)$ and $\mathcal{Q}_{test}(x_i)$ are respectively given by $\mathcal{P}_{train}(x_i)-\mathcal{N}_{train}(x_i)$ and $\mathcal{P}_{test}(x_i)-\mathcal{N}_{test}(x_i)$, and 
\begin{equation*}
    \epsilon(x_i)=\min\{|\mathcal{Q}_{train}(x_i)|, |\mathcal{Q}_{test}(x_i)| \}.
\end{equation*}

Apparently, there must exist a \emph{perfect} classifier that reaches the minimum Hinge loss and highest accuracy for both training set and test set when $\Delta=0$. In addition, $ \Delta=0$ if and only if (detailed mathematical proof see Sensitivity Analysis in Supplemental Material)
\begin{eqnarray}\label{eq_delta_condition}
	\left\{ 
	\begin{aligned}
		& \mathcal{P}_{train}(x_i)\geq \mathcal{N}_{train}(x_i) \\
		&  \mathcal{P}_{test}(x_i)\geq \mathcal{N}_{test}(x_i) \\
	\end{aligned}
	\right.
	\ \text{or} \
	\left\{ 
	\begin{aligned}
		& \mathcal{P}_{train}(x_i) < \mathcal{N}_{train}(x_i) \\
		& \mathcal{P}_{test}(x_i) < \mathcal{N}_{test}(x_i) \\
	\end{aligned}
	\right.
\end{eqnarray}
for every $x_i \in \mathcal{S}$. Eq.~\ref{eq_delta_condition} shows that the upper bound of binary classification on a specific data, i.e. $\Delta=0$, can be reached only if the distribution of data between the training and test sets are consistent. Hence, when the training set and the test set share a similar data structure, the trained classifier can exhibit a phenomenon known as \emph{benign overfitting}~\cite{bartlett2020benign}, showcasing exceptional generalization capability. 
However, the trained classifier in real scenarios often can approach the upper limit of corresponding loss function ($\Delta_{train}^{f}$=0), but difficult to simultaneously achieve the upper bound of performance on the test set ($\Delta_{test}^{f}$=0) because of the discrepencies between the distributions of data in training and test sets.

In Fig.~\ref{fig_delta}, we investigate at length the characteristics of both training and test sets in order to comprehensively understand the generalization ability of the proposed boundary theory. It shows that, the theoretical optimal loss for $\Delta$ is smaller than that of any representative classier, suggesting that the predictability of available classifiers are still a certain distance from the theoretical upper limit, both for a specific division (i.e. $|\mathcal{S}_{train}|/|\mathcal{S}| = 70\%$, Fig. \ref{fig_delta}A) or all possible random divisions (Fig. \ref{fig_delta}C). Therefore, $\Delta_{train}^{f}$ and $\Delta_{test}^{f}$ are respectively equivalent to the learning and generalization differences in supervised machine learning. If one aims at obtaining the upper bound of a given dataset (corresponds to $\Delta$), he or she needs to minimize simultaneously both $\Delta_{train}^{f}$ and $\Delta_{test}^{f}$ \cite{jiang2018predicting}. However, on one hand, it requires fitting training data as much as possible to obtain minimal $\Delta_{train}^{f}$, often leading to the overfitting problem \cite{ying2019,bartlett2020benign}, which further dilutes the generalization ability (corresponds to greater $\Delta_{test}^{f}$ and more distant from the upper bound). On the other hand, one may try to solve such dilemma by early-stopping, dropout or adding regularization items \cite{srivastava2014dropout,ying2019}, resulting in a greater learning loss (corresponds to larger $\Delta_{train}^{f}$), leaving an unstable prediction.

Fig.~\ref{fig_delta}D shows a general increase of learning loss with the size of training set ($|\mathcal{S}_{train}|$), which converges to a certain limit if the feeding schema is sufficient enough. Comparatively, the upper bound ($\text{AC}^u$) monotonically increases with $|\mathcal{S}_{train}|$ (Fig.~\ref{fig_delta}E), which agrees with the common sense that more knowledge can be extracted by adding more sources. The good agreements are additional proved theoretically by giving an arbitrary parameter governing the ratio of random divisions (represented by solid lines in Fig.~\ref{fig_delta}D-\ref{fig_delta}F, see also Sensitivity Analysis, Supplemental Material). Furthermore, Fig. \ref{fig_delta}F shows that the optimal error ($\Delta$) will be achieved when the training set is either adequate large or small enough as its expectation is symmetric based on  Eq. 95 in Supplemental Material.  

\begin{figure}[ht]
	\centering
	\includegraphics[width=\linewidth]{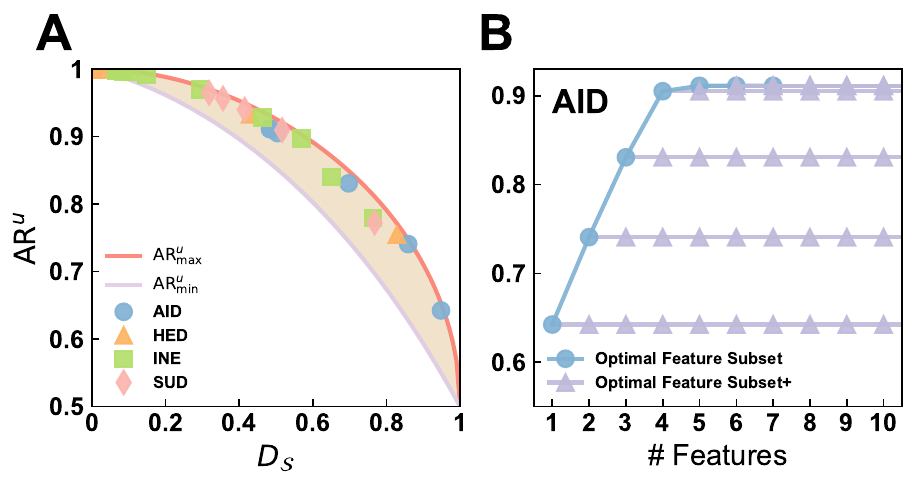}
	\caption{Characterization of the upper bound for AUC-ROC ($\text{AR}^u$) as influenced by key dataset properties on the \emph{AID} dataset, with extended results for additional datasets available in the Supplemental Material, Fig. S6. (A) Illustrates how $\text{AR}^u$ varies with the degree of overlap ($D_{\mathcal{S}}$) between positive and negative sample distributions. The red and purple curves denote the theoretically expected maximum and minimum bounds, respectively. The plotted dots represent the empirically obtained optimal AUC scores corresponding to different overlapping scenarios facilitated by feature selection. (B) Depicts the relationship between $\text{AR}^u$ and the number of features, where the blue curve indicates the maximum upper bounds achievable with a specified number of features within the \emph{optimal feature subset}. The purple curves denote the maximal upper bounds achievable when the \emph{optimal feature subset} is expanded by the inclusion of additional arbitrary feature transformations, referred to as the \emph{optimal feature subset+}. Additional results about feature selection can be seen in Figs. S6, S7, S60, S61 in Supplemental Material.
 }
\label{fig_overlapping}
\end{figure}
The relationship between the distribution pattern of feature vectors~\cite{fajardo2023fundamental}, specifically the overlapping ratio of data samples, and the expected upper bounds ($\text{AR}^u$) is investigated. To ascertain the minimum and maximum values of $\text{AR}^u$, we formulate two optimization problems as follows:
\begin{equation}\label{eq:overlapping_min}
     \text{AR}^u_{\min} (D_{\mathcal{S}}) = \min \Big\{ \text{AR}^u (\mathcal{P}, \mathcal{N}) \mid  D_{\mathcal{S}} (\mathcal{P}, \mathcal{N})=D_{\mathcal{S}} \Big\}
\end{equation}
and
\begin{equation}\label{eq:overlapping_max}
     \text{AR}^u_{\max} (D_{\mathcal{S}}) = \max \Big\{ \text{AR}^u (\mathcal{P}, \mathcal{N}) \mid  D_{\mathcal{S}} (\mathcal{P}, \mathcal{N})=D_{\mathcal{S}} \Big\}.
\end{equation}
In these equations, $D_{\mathcal{S}} (\mathcal{P} , \mathcal{N}) = 1 - J (\mathcal{P} || \mathcal{N})$ denotes the overlapping ratio between positive and negative samples, with $J (\mathcal{P} || \mathcal{N})$ being the Jensen-Shannon divergence~\cite{menendez1997jensen} between the normalized distributions $\mathcal{P}$ and $\mathcal{N}$. At the extremes, $D_{\mathcal{S}}=1$ signifies complete overlap, whereas $D_{\mathcal{S}}=0$ indicates complete separation.

By employing numerical simulations and heuristic methods (see Supplemental Material for details), we deduce the optimal solutions for these optimization problems as $\text{AR}^u_{\min}$ and $\text{AR}^u_{\max}$. Fig.~\ref{fig_overlapping}A illustrates a monotonic decrease in both $\text{AR}^u_{\min}$ and $\text{AR}^u_{\max}$ as $D_{\mathcal{S}}$ increases, with the upper bounds for all four datasets analyzed resting between these values. This suggests that greater overlap between positive and negative samples correlates with a reduction in the achievable upper bound for binary classification performance. In the limit, the upper bound can attain $1$ (perfect accuracy) when $D_{\mathcal{S}}=0$, and $0.5$ (equivalent to random guessing) when $D_{\mathcal{S}}=1$. 

The influence of feature selection on the upper bound is scrutinized. Fig.~\ref{fig_overlapping}B presents a case study utilizing the AID dataset. It is observed that the upper bound tends to rise with the inclusion of additional features. Nevertheless, there exists a specific subset of features\textemdash the optimal feature subset\textemdash that is adequate to reach the potential upper bound for the majority of datasets. Beyond this subset, incorporating more features does not enhance the upper bound. It is also noted that integrating transformations of features from the optimal subset fails to provide any further increase in the upper bound. Collectively, these insights underscore the efficacy of our methodology in streamlining the binary classification process by pinpointing the most effective subset of features for feature engineering purposes. 

This research delves into the impact that characteristics of features have on the upper limit of predictive accuracy in binary classification endeavors. Through a combination of theoretical framework of statistical physics and numerical experimentation, we have shed light on the pivotal roles played by data organization and feature engineering in determining the ceiling of a model's performance. Additionally, we introduce a quantitative assessment of error balance ($\Delta$), which offers insights into the practicality of reaching these upper bounds in both training and testing scenarios.

The implications of our findings for machine learning (ML) are substantial. We have empirically confirmed that datasets with greater overlap among samples pose more significant challenges for accurate classification. Our theoretical framework convincingly demonstrates that expanding the feature set can reduce overlap between classes and push the boundaries of predictive performance higher. Therefore, in the context of raw data, the discovery and curation of new features, along with the strategic selection of an optimal subset, are crucial for maximizing the efficacy of model training. On the other hand, relying solely on transformations of existing features does not contribute to an improvement of the upper performance limit.

Nonetheless, although the underlying physical principle for the predictability of binary classification is comprehensively discovered, our study still has limitations. Firstly, the theoretical upper bound of prediction performance is attainable only in scenarios where all distinct samples are correctly classified, which does not account for the balance between data availability and model complexity. In real-world applications, achieving this upper bound is often challenging due to constraints on data and computational resources. The degree to which the upper bound can be approached is contingent upon the data distribution, and a theoretical framework to quantify this degree is an avenue for future investigation. Secondly, our analysis presumes that the data space is discretized. The practical utility of the derived upper bound would benefit from considering the distance and similarity between data samples, which we have not addressed. We advocate for future studies that explore the implications of limited model complexity and the behavior in a continuous data space.

Expanding beyond the realm of binary classification, the physical principles of our boundary theory could also enhance our understanding of multiclass classification issues, which can be segmented into a series of binary classification challenges. Looking ahead, our forthcoming endeavors will concentrate on dynamic data organization, such as time-series classification or variable data sizes, where the grand canonical ensemble is a promising approach \cite{bianconi2022grand}, as well as on predictability and interpretability within a more applied framework, particularly concerning feature engineering, classifiers, optimization techniques, and energy landscapes. In subsequent research, we aim to advance the predictability of ML and AI systems, enhancing the trustworthiness of AI by improving our ability to explain, manage, and regulate these systems. 

This work was supported by the National Natural Science Foundation of China (Grant Nos. 72371224 and 71972164), the Research Grants Council of the Hong Kong Special Administrative Region (Grant No. 11218221) and the Fundamental Research Funds for the Central Universities.

\bibliographystyle{apsrev4-2}
\bibliography{aps-main}

\end{document}


\title{Supplemental Material for \\ Data organization limits the predictability of binary classification}

\author{Fei Jing}
\affiliation{Musketeers Foundation Institute of Data Science, The University of Hong Kong, Hong Kong SAR, China}

\author{Zi-Ke Zhang}\email{zkz@zju.edu.cn}
\affiliation{Center for Digital Communication Studies, Zhejiang University, Hangzhou 310058, China}

\author{Yi-Cheng Zhang}\email{yi-cheng.zhang@unifr.ch}
\affiliation{Department of Physics, University of Fribourg, Chemin du Mus{\'e}e 3, 1700 Fribourg, Switzerland}

\author{Qingpeng Zhang}\email{qpzhang@hku.hk}
\affiliation{Musketeers Foundation Institute of Data Science, The University of Hong Kong, Hong Kong SAR, China}

 \maketitle
\tableofcontents

\newpage
\section{Preliminaries}

Binary classification involves assigning elements of a set into one of two groups, or classes, based on a classification rule. This process is crucial in various applications such as medical diagnostics, quality control, and information retrieval. In machine learning and data analysis, binary classification is a supervised learning task aimed at predicting one of two possible outcomes for a given input.

In this context, the data is labeled as positive (often denoted by 1) or negative (denoted by -1). The objective is to construct a predictive model capable of accurately classifying new, unseen instances into these categories by learning from patterns in the training data. Initially, a labeled dataset is gathered, where each instance is associated with a known class label. The dataset is then divided into a training set for model development and a test set for evaluating performance on new data. During training, the model identifies patterns that distinguish between the classes. Techniques such as XGBoost, MLP, SVM, LR, DT, RF, KNN, and Naive Bayes can be utilized for binary classification.

Let's consider a binary classification scenario where the goal is to predict a binary label. An input-output pair is represented as $z=(x,y)$, where $x\in\mathcal{X}$ is the feature vector (input data) and $y\in\{1,-1\}$ is the class label. Given a training set $\mathcal{S}=\{(x_i,y_i) | i=1,2,\ldots,m\}$, we define $\mathcal{S}_+$ as the subset containing $n_+$ positive samples and $\mathcal{S}_-$ as the subset with $n_-$ negative samples, such that the total number of instances is $m=|\mathcal{S}|=n_+ + n_-$. Let $\mathcal{P}(x_i)$ and $\mathcal{N}(x_i)$ denote the counts of positive and negative instances for a given feature vector $x_i$.

Classifiers are mappings that assign instances to specific classes. Some produce a continuous output, allowing various thresholds to define class membership—these are known as **continuous classifiers**. They combine a classification function $f(x): \mathcal{X} \rightarrow \mathbb{R}$ with a threshold $t$ to translate scores into binary classes. Others yield a discrete class label—these are **discrete classifiers** and are described by the function $f(x): \mathcal{X} \rightarrow \{1,-1\}$. Logistic regression and neural networks are examples of continuous classifiers, while SVM, decision trees, and random forests are discrete classifiers.

Objective functions gauge the alignment between model predictions and actual labels. During training, the aim is to minimize the difference between these predictions and true labels. We discuss several objective functions for binary classification problems:
\begin{itemize}
    \item Square loss function: $\min\limits_f \ \frac{1}{m} \sum\limits_{x_i} \Big( f(x_i) - y_i \Big)^2$;
    \item Logistic loss function: $\min\limits_f \ \frac{1}{m} \sum\limits_{x_i} -y_i\log f(x_i) - (1-y_i)\log \Big(1-f(x_i)\Big)$;
    \item Hinge loss function: $\min\limits_f \ \frac{1}{2m} \sum\limits_{x_i} \max\Big\{ 0, 1 - f(x_i)y_i \Big\}$;
    \item Softmax function: $\min\limits_f \ \frac{1}{m} \sum\limits_{x_i} -\log f(x_i) - \log \Big(1-f(x_i)\Big)$.
\end{itemize}
Square and Logistic loss functions are typically suited for continuous classifiers, while Hinge and Softmax losses can be applied to both continuous and discrete classifiers.

For binary classification outcomes, we consider the instances to be either positive or negative, leading to four potential results from the classifier:

\begin{itemize}
    \item TP (True Positive): $\sum\limits_{x_i \in \mathcal{S}} \frac{\mathcal{P}(x_i)}{2} \Big( f(x_i) + 1 \Big)$;
    \item FP (False Positive): $\sum\limits_{x_i \in \mathcal{S}} \frac{\mathcal{N}(x_i)}{2} \Big( f(x_i) + 1 \Big)$;
    \item FN (False Negative): $\sum\limits_{x_i \in \mathcal{S}} \frac{\mathcal{P}(x_i)}{2} \Big( 1 - f(x_i) \Big)$;
    \item TN (True Negative): $\sum\limits_{x_i \in \mathcal{S}} \frac{\mathcal{N}(x_i)}{2} \Big(1 - f(x_i)\Big)$.
\end{itemize}

\section{Objective Functions}

The objective function in the training process of a classification model serves as a guide for parameter adjustment to fit the dataset. Here, we discuss four commonly used objective functions and show a direct correlation between their optimal solutions~\cite{rosasco2004loss} and the dataset through discrete analysis.

\subsection{Optimal square loss and statistical ensemble on square cost function}
\subsubsection{Optimal square loss}
\noindent
Given a dataset \( \mathcal{S} \) with feature domain \( \mathcal{X} \), and assuming \( f(x) \) as a continuous classification function with parameters to be trained, the square error loss function is given by:
\begin{align}
    \begin{aligned}
        \min_f \ \frac{1}{m}\sum\limits_{x_i} \Big( f(x_i) - y_i\Big)^2.
    \end{aligned}
\end{align}
The optimal solution for this objective function is expressed as:
\begin{align}
    \begin{aligned}
        f^*_{\text{Square}} = \arg  \min_f \ \frac{1}{m}\sum\limits_{x_i} \Big( f(x_i) - y_i\Big)^2.
    \end{aligned}
\end{align}
It's evident that:
\begin{align}
    \begin{aligned}
        \min_f \ \sum\limits_{x_i} \Big( f(x_i) - y_i\Big)^2 \geq \sum\limits_{x_i} \min_{f(x_i)} \Big( f(x_i) - y_i\Big)^2.
    \end{aligned}
\end{align}
Moreover, equality holds:
\begin{align}
    \begin{aligned}
        \min_f \ \sum\limits_{x_i} \Big( f(x_i) - y_i\Big)^2 = \sum\limits_{x_i} \min_{f(x_i)} \Big( f(x_i) - y_i\Big)^2.
    \end{aligned}
\end{align}
if and only if $f(x_i)$ and $f(x_j)$ are  nearly independent for any $x_i \neq x_j \in\mathcal{S}$. In the binary classification context, the square loss function can be transformed to:
\begin{align}
    \begin{aligned}
        \min_f \ \sum\limits_{x_i} \Big( f(x_i) - y_i\Big)^2 
        =& \sum\limits_{x_i} \min_{f(x_i)} \Big( f(x_i) - y_i\Big)^2\\
        =& \sum\limits_{x_i} \min_{f(x_i)} \mathcal{P}(x_i)  \Big( f(x_i) - 1\Big)^2 + \mathcal{N}(x_i)\Big( f(x_i) + 1\Big)^2.
    \end{aligned}
\end{align}
The optimal solution becomes:
\begin{align}
    \begin{aligned}
        f^*_{\text{Square}}(x_i) = \arg\min_{f(x_i)}  \mathcal{P}(x_i)  \Big( f(x_i) - 1\Big)^2 + \mathcal{N}(x_i)\Big( f(x_i) + 1\Big)^2
    \end{aligned}
\end{align}
and is simply:
\begin{align}
    \begin{aligned}
        f^*_{\text{Square}}(x_i) = \frac{\mathcal{P}(x_i) - \mathcal{N}(x_i)}{\mathcal{P}(x_i)+ \mathcal{N}(x_i)}
    \end{aligned}
\end{align}
for each $x_i\in\mathcal{S}$. Finally, we have
\begin{align}
    \text{minimum square loss} = \frac{4}{m}&\sum\limits_{x_i} \frac{\mathcal{P}(x_i)\mathcal{N}(x_i)}{\mathcal{P}(x_i)+\mathcal{N}(x_i)}.
\end{align}

\subsubsection{Statistical ensemble on square loss function}
\noindent
In this context, every classifier $f$ can be transformed into a $n$-dimensional vector
\begin{align}
    \pi^f=(x_1^f, x_2^f,\cdots,x_n^f)
\end{align}
in which $x_i^f=f(x_i)$ for each $x_i\in\mathcal{S}$.
Next, the energy of $\pi^f$ is defined as
\begin{align}
\begin{aligned}
     E^{Square}(\pi^f) =& \frac{1}{m}\sum\limits_{x_i} \Big( \pi_i^f - y_i\Big)^2 \\
    =&\frac{1}{m}\sum\limits_{x_i} \mathcal{P}(x_i)  \Big( \pi_i^f  - 1\Big)^2 + \mathcal{N}(x_i)\Big( \pi_i^f  + 1\Big)^2.
\end{aligned}
\end{align}
Let $\Pi(\mathcal{S})=\mathbb{R}^n$. Then, the partition function is 
\begin{align}
\begin{aligned}
     Z^{Square} =& \sum_{\pi_f\in \Pi(\mathcal{S})} \exp(-\beta E^{Square}(\pi^f))\\
    =& \int \prod_{x^f_i} \mathrm{d}\pi^f_i \exp\left\{-\beta\left( \frac{1}{m}\sum\limits_{x_i} \mathcal{P}(x_i)  \Big( \pi^f_i - 1\Big)^2 + \mathcal{N}(x_i)\Big( \pi^f_i + 1\Big)^2 \right) \right\}
\end{aligned}
\end{align}
And, the ground state is calculated as
\begin{align}
    \begin{aligned}
        E_0^{Square} =&  \lim\limits_{\beta \rightarrow \infty} -\frac{1}{\beta} \ln{Z^{Square}} \\
        =& \lim\limits_{\beta \rightarrow \infty} -\frac{1}{\beta} \ln{\int \prod_{x^f_i} \mathrm{d}\pi^f_i  \exp\left\{-\beta\left( \frac{1}{m}\sum\limits_{x_i} \mathcal{P}(x_i)  \Big(\pi^f_i - 1\Big)^2 + \mathcal{N}(x_i)\Big( \pi^f_i + 1\Big)^2 \right) \right\}} \\
        =& \lim\limits_{\beta \rightarrow \infty} -\frac{1}{\beta} \ln{ \int \prod_{x^f_i} \mathrm{d}\pi^f_i \prod\limits_{x_i} \exp\left\{-\frac{\beta}{m} \left\{  \mathcal{P}(x_i)  \Big( \pi^f_i - 1\Big)^2 + \mathcal{N}(x_i)\Big( \pi^f_i + 1\Big)^2 \right\} \right\} }\\
        =& \lim\limits_{\beta \rightarrow \infty} -\frac{1}{\beta} \ln{  \prod\limits_{x^f_i} \int  \mathrm{d}\pi^f_i \exp\left\{-\frac{\beta}{m} \left\{  \mathcal{P}(x_i)  \Big( \pi^f_i - 1\Big)^2 + \mathcal{N}(x_i)\Big( \pi^f_i + 1\Big)^2 \right\} \right\} } \\
        =& \sum\limits_{x_i} \lim\limits_{\beta \rightarrow \infty} -\frac{1}{\beta} \ln{  \int  \mathrm{d}\pi^f_i \exp\left\{-\frac{\beta}{m} \left\{  \mathcal{P}(x_i)  \Big( \pi^f_i - 1\Big)^2 + \mathcal{N}(x_i)\Big( \pi^f_i + 1\Big)^2 \right\} \right\} } .
    \end{aligned}
\end{align}
Utilizing Gaussian transformation, we finally obtain
\begin{align}
    \begin{aligned}
        E_0^{Square} =\frac{4}{m}&\sum\limits_{x_i} \frac{\mathcal{P}(x_i)\mathcal{N}(x_i)}{\mathcal{P}(x_i)+\mathcal{N}(x_i)}.
    \end{aligned}
\end{align}

\subsection{Optimal logistic loss and statistical ensemble on logistic cost function}
\subsubsection{Optimal logistic loss}
\noindent
The Logistic loss function, frequently used for binary classification, is defined for a dataset \( \mathcal{S} \) and feature domain \( \mathcal{X} \) as:
\begin{align}
    \begin{aligned}
        \min_f \ \frac{1}{m}\sum\limits_{x_i} -y_i\log f(x_i) -\Big( 1-y_i\Big) \log\Big( 1-f(x_i)\Big),
    \end{aligned}
\end{align}
with the optimal solution:
\begin{align}
    \begin{aligned}
        f^*_{\text{Logistic}} = \arg \min_f \  \frac{1}{m}\sum\limits_{x_i} -y_i\log f(x_i) -\Big( 1-y_i\Big) \log\Big( 1-f(x_i)\Big).
    \end{aligned}
\end{align}
Close to the optimal, the logistic loss function can be approximated as:
\begin{align}
     \min_f \ \frac{1}{m}\sum\limits_{x_i} \Big(\mathcal{N}(x_i)- \mathcal{P}(x_i)\Big) \log f(x_i) - 2\mathcal{N}(x_i)\log\Big( 1-f(x_i)\Big).
\end{align}
The corresponding optimal solution then is:
\begin{align}
    \begin{aligned}
         f^*_{\text{Logistic}}(x_i) = \arg\min_{f(x_i)} \Big(\mathcal{N}(x_i)- \mathcal{P}(x_i)\Big) \log f(x_i) - 2\mathcal{N}(x_i)\log\Big( 1-f(x_i)\Big)
    \end{aligned}
\end{align}
for each $x_i\in\mathcal{S}$. Finally, it can be simplified as 
\begin{align}
    \begin{aligned}
        f^*_{\text{Logistic}}(x_i) = \frac{\mathcal{P}(x_i) - \mathcal{N}(x_i)}{\mathcal{P}(x_i)+ \mathcal{N}(x_i)}
    \end{aligned}
\end{align}
for each $x_i\in\mathcal{S}$.

\subsubsection{Statistical ensemble on logistic cost function}
\noindent
In this context, the energy of $\pi^f$ is defined as
\begin{align}
\begin{aligned}
     E^{Logistic}(\pi^f) =& \frac{1}{m}\sum\limits_{x_i} \Big(\mathcal{N}(x_i)- \mathcal{P}(x_i)\Big) \log \pi_i^f - 2\mathcal{N}(x_i)\log\Big( 1-\pi_i^f\Big).
\end{aligned}
\end{align}
Let $\Pi(\mathcal{S})=\mathbb{R}^n$. Then, the partition function is 
\begin{align}
\begin{aligned}
     Z^{Logistic} =& \sum_{\pi_f\in \Pi(\mathcal{S})} \exp(-\beta E^{Logistic}(\pi^f))\\
    =& \int \prod_{x^f_i} \mathrm{d}\pi^f_i \exp\left\{-\beta\left( \frac{1}{m}\sum\limits_{x_i} \Big(\mathcal{N}(x_i)- \mathcal{P}(x_i)\Big) \log \pi_i^f - 2\mathcal{N}(x_i)\log\Big( 1-\pi_i^f\Big) \right) \right\}.
\end{aligned}
\end{align}
And, the ground state is calculated as
\begin{align}
    \begin{aligned}
        E_0^{Logistic} =&  \lim\limits_{\beta \rightarrow \infty} -\frac{1}{\beta} \ln{Z^{Logistic}} \\
        =& \lim\limits_{\beta \rightarrow \infty} -\frac{1}{\beta} \ln{\int_0^1 \prod_{x^f_i} \mathrm{d}\pi^f_i \exp\left\{-\beta\left( \frac{1}{m}\sum\limits_{x_i} \Big(\mathcal{N}(x_i)- \mathcal{P}(x_i)\Big) \log \pi_i^f - 2\mathcal{N}(x_i)\log\Big( 1-\pi_i^f\Big) \right) \right\}} \\
        =& \lim\limits_{\beta \rightarrow \infty} -\frac{1}{\beta} \ln{ \int_0^1 \prod_{x^f_i} \mathrm{d}\pi^f_i \prod\limits_{x_i} \exp\left\{-\frac{\beta}{m} \left\{  \Big(\mathcal{N}(x_i)- \mathcal{P}(x_i)\Big) \log \pi_i^f - 2\mathcal{N}(x_i)\log\Big( 1-\pi_i^f\Big) \right\} \right\} }\\
        =& \lim\limits_{\beta \rightarrow \infty} -\frac{1}{\beta} \ln{ \prod_{x_i} \int_0^1  \mathrm{d}\pi^f_i  \exp\left\{-\frac{\beta}{m} \left\{  \Big(\mathcal{N}(x_i)- \mathcal{P}(x_i)\Big) \log \pi_i^f - 2\mathcal{N}(x_i)\log\Big( 1-\pi_i^f\Big) \right\} \right\}  } \\
        =& \sum\limits_{x_i} \lim\limits_{\beta \rightarrow \infty} -\frac{1}{\beta} \ln{ \int_0^1  \mathrm{d}\pi^f_i  \exp\left\{-\frac{\beta}{m} \left\{  \Big(\mathcal{N}(x_i)- \mathcal{P}(x_i)\Big) \log \pi_i^f - 2\mathcal{N}(x_i)\log\Big( 1-\pi_i^f\Big) \right\} \right\} } \\
        =& \sum\limits_{x_i} \min\limits_{\pi_i^f \in (0,1)} \frac{1}{m} \left\{  \Big(\mathcal{N}(x_i)- \mathcal{P}(x_i)\Big) \log \pi_i^f - 2\mathcal{N}(x_i)\log\Big( 1-\pi_i^f\Big) \right\} \\
        =& \frac{1}{m} \sum\limits_{x_i} \Big(\mathcal{N}(x_i)- \mathcal{P}(x_i)\Big) \ln{\frac{\mathcal{P}(x_i)- \mathcal{N}(x_i)}{\mathcal{P}(x_i)+ \mathcal{N}(x_i)}} - 2 \mathcal{N}(x_i) \ln \frac{2\mathcal{N}(x_i)}{\mathcal{P}(x_i)+ \mathcal{N}(x_i)},
    \end{aligned}
\end{align}
in which 
\begin{align}
    \begin{aligned}
        f^*_{\text{Logistic}}(x_i) = \frac{\mathcal{P}(x_i) - \mathcal{N}(x_i)}{\mathcal{P}(x_i)+ \mathcal{N}(x_i)}
    \end{aligned}
\end{align}
for each $x_i\in\mathcal{S}$.

\subsection{Optimal Hinge loss and statistical ensemble on Hinge cost function}
\subsubsection{Optimal Hinge loss}
\noindent
The hinge loss function is commonly used in Support Vector Machines (SVM) and other models employing maximum-margin classifiers, regardless of whether the classification task is discrete or continuous. It penalizes misclassified instances based on their distance from the decision boundary. The Hinge loss function for binary classification is described as follows:
\begin{align}
    \min_f \ \frac{1}{2m} \sum_{i=1}^m \max\{0, 1 - y_i f(x_i)\},
\end{align}
and its optimal solution is
\begin{align}
    f^*_{\text{Hinge}} = \arg \min_f \ \frac{1}{2m} \sum_{i=1}^m \max\{0, 1 - y_i f(x_i)\}.
\end{align}
Unlike other objective functions, the output for any solution of the hinge loss function is discrete. We employ a similar technique to the hinge loss function as before,
\begin{align}
    \min_{f(x_i) \in \{1, -1\}} \mathcal{P}(x_i) \max\{0, 1 - f(x_i)\} + \mathcal{N}(x_i) \max\{0, 1 + f(x_i)\}
\end{align}
for any \(x_i\),
where its optimal solution can also be expressed as
\begin{align}
    f^*_{\text{Hinge}} (x_i) = \arg \max_{f(x_i) \in \{1, -1\}} \mathcal{P}(x_i) \max\{0, 1 - f(x_i)\} + \mathcal{N}(x_i) \max\{0, 1 + f(x_i)\}
\end{align}
for each \(x_i \in \mathcal{S}\).
Since \(f(x_i)\) must be equal to \(1\) or \(-1\), we can rewrite the optimal solution as
\begin{align}
f^*_{\text{Hinge}}(x_i) =
\begin{cases}
1,  & \text{if } \mathcal{P}(x_i) \geq \mathcal{N}(x_i), \\
-1, & \text{if } \mathcal{P}(x_i) < \mathcal{N}(x_i),
\end{cases}
\end{align}
for each \(x_i \in \mathcal{S}\). And,
\begin{align}
    \text{minimum Hinge loss} = \frac{1}{m}\sum\limits_{x_i} \min\Big\{\mathcal{P}(x_i), \mathcal{N}(x_i)\Big\}.
\end{align}

\subsubsection{Statistical ensemble on Hinge cost function}
\noindent
In this context, the energy of $\pi^f$ is defined as
\begin{align}
\begin{aligned}
     E^{Hinge}(\pi^f) =&\frac{1}{2m}\sum\limits_{x_i} \mathcal{P}(x_i) \max\{0, 1 - \pi^f_i\} + \mathcal{N}(x_i) \max\{0, 1 + \pi^f_i\},
\end{aligned}
\end{align}
where $\pi^f_i \in \{1,-1\}$ for each $x_i$.
Let $\Pi(\mathcal{S})=\{1,-1\}^n$. Then, the partition function is 
\begin{align}
\begin{aligned}
     Z^{Hinge} =& \sum_{\pi_f\in \Pi(\mathcal{S})} \exp(-\beta E^{Hinge}(\pi^f))\\
    =&  \sum_{\pi_f\in \Pi(\mathcal{S})}  \exp\left\{-\beta\left( \frac{1}{2m}\sum\limits_{x_i} \mathcal{P}(x_i) \max\{0, 1 - \pi^f_i\} + \mathcal{N}(x_i) \max\{0, 1 + \pi^f_i\} \right) \right\}.
\end{aligned}
\end{align}
And, the ground state is calculated as
\begin{align}
    \begin{aligned}
        E_0^{Hinge} =&  \lim\limits_{\beta \rightarrow \infty} -\frac{1}{\beta} \ln{Z^{Hinge}} \\
        =& \lim\limits_{\beta \rightarrow \infty} -\frac{1}{\beta} \ln{\sum_{\pi_f\in \Pi(\mathcal{S})}  \exp\left\{-\beta\left( \frac{1}{2m}\sum\limits_{x_i} \mathcal{P}(x_i) \max\{0, 1 - \pi^f_i\} + \mathcal{N}(x_i) \max\{0, 1 + \pi^f_i\} \right) \right\}} \\
        =& \lim\limits_{\beta \rightarrow \infty} -\frac{1}{\beta} \ln{\sum_{\pi_f\in \Pi(\mathcal{S})} \prod_{x_i} e^{-\frac{\beta}{2m} \Big( \mathcal{P}(x_i) \max\{0, 1 - \pi^f_i\} + \mathcal{N}(x_i) \max\{0, 1 + \pi^f_i\}\Big) } } \\
        =& \lim\limits_{\beta \rightarrow \infty} -\frac{1}{\beta} \ln { \prod_{x_i} \Big( e^{-\beta \mathcal{N}(x_i)/m} + e^{-\beta \mathcal{P}(x_i)/m} \Big) } \\
        =& \sum\limits_{x_i} \lim\limits_{\beta \rightarrow \infty} -\frac{1}{\beta} \ln{\Big( e^{-\beta \mathcal{N}(x_i)/m} + e^{-\beta \mathcal{P}(x_i)/m} \Big)} \\
        =& \frac{1}{m}  \sum\limits_{x_i}\min\Big\{\mathcal{P}(x_i), \mathcal{N}(x_i)\Big\}.
    \end{aligned}
\end{align}

\subsection{Optimal Softmax loss and statistical ensemble on Softmax cost function}
\subsubsection{Optimal Softmax loss}
\noindent
The Softmax loss function, also known as the cross-entropy loss or log-likelihood loss, is essential in machine learning and deep learning, particularly for classification tasks. It measures the dissimilarity between predicted class probabilities and true class labels. The softmax function is typically used in conjunction with this loss function to convert raw model outputs into probability distributions over multiple classes.
Mathematically, for a feature \(x\), the softmax function computes the probability of the positive class as follows:
\begin{align}
    f(x) = \frac{e^{z_+}}{e^{z_+} + e^{z_-}}.
\end{align}
Here, \(z_+\) (\(z_-\)) denotes the score that measures the likelihood of the data with feature \(x\) belonging to the positive (negative) class.

The softmax loss function is defined as the negative log-likelihood of the true class in binary classification tasks:
\begin{align}
    \min_f \ \frac{1}{m} \sum_{i=1}^m \big[-y_i \log f(x_i) - (1 - y_i) \log (1 - f(x_i))\big],
\end{align}
with the optimal solution being
\begin{align}
    f^*_{\text{Softmax}} = \arg \min_f \ \frac{1}{m} \sum_{i=1}^m \big[-y_i \log f(x_i) - (1 - y_i) \log (1 - f(x_i))\big].
\end{align}
We can simplify the objective function as
\begin{align}
    \min_{f(x_i)} \ -\mathcal{P}(x_i) \log f(x_i) - \mathcal{N}(x_i) \log (1 - f(x_i)),
\end{align}
and its optimal solution as
\begin{align}
     f^*_{\text{Softmax}} (x_i) = \arg \min_{f(x_i)} -\mathcal{P}(x_i) \log
f(x_i) - \mathcal{N}(x_i) \log (1 - f(x_i)), 
\end{align}
for every \(x_i \in \mathcal{S}\).
By performing the necessary derivations, we find the optimal solution for the Softmax loss function as
\begin{align}
    f^*_{\text{Softmax}}(x_i) = \frac{\mathcal{P}(x_i)}{\mathcal{P}(x_i) + \mathcal{N}(x_i)},
\end{align}
for each \(x_i \in \mathcal{S}\).

\subsubsection{Statistical ensemble on Softmax cost function}
\noindent
In this context, the energy of $\pi^f$ is defined as
\begin{align}
\begin{aligned}
     E^{Softmax}(\pi^f) =&\frac{1}{m} \sum_{i=1}^m -\mathcal{P}(x_i) \log \pi_i^f - \mathcal{N}(x_i) \log (1 - \pi_i^f),
\end{aligned}
\end{align}
Let $\Pi(\mathcal{S})=\mathbb{R}^n$. Then, the partition function is 
\begin{align}
\begin{aligned}
     Z^{Softmax} =& \sum_{\pi_f\in \Pi(\mathcal{S})} \exp(-\beta E^{Softmax}(\pi^f))\\
    =& \int \prod_{x^f_i} \mathrm{d}\pi^f_i \exp\left\{-\beta\left( \frac{1}{m} \sum_{i=1}^m -\mathcal{P}(x_i) \log \pi_i^f - \mathcal{N}(x_i) \log (1 - \pi_i^f) \right) \right\}.
\end{aligned}
\end{align}
And, the ground state is calculated as
\begin{align}
    \begin{aligned}
        E_0^{Softmax} =&  \lim\limits_{\beta \rightarrow \infty} -\frac{1}{\beta} \ln{Z^{Softmax}} \\
        =& \lim\limits_{\beta \rightarrow \infty} -\frac{1}{\beta} \ln{\int_0^1 \prod_{x^f_i} \mathrm{d}\pi^f_i \exp\left\{-\beta\left( \frac{1}{m}\sum\limits_{x_i} -\mathcal{P}(x_i) \log \pi_i^f - \mathcal{N}(x_i) \log (1 - \pi_i^f) \right) \right\}} \\
        =& \lim\limits_{\beta \rightarrow \infty} -\frac{1}{\beta} \ln{ \int_0^1 \prod_{x^f_i} \mathrm{d}\pi^f_i \prod\limits_{x_i} \exp\left\{-\frac{\beta}{m} \left\{  -\mathcal{P}(x_i) \log \pi_i^f - \mathcal{N}(x_i) \log (1 - \pi_i^f) \right\} \right\} }\\
        =& \lim\limits_{\beta \rightarrow \infty} -\frac{1}{\beta} \ln{ \prod_{x_i} \int_0^1  \mathrm{d}\pi^f_i  \exp\left\{-\frac{\beta}{m} \left\{ -\mathcal{P}(x_i) \log \pi_i^f - \mathcal{N}(x_i) \log (1 - \pi_i^f) \right\} \right\}  } \\
        =& \sum\limits_{x_i} \lim\limits_{\beta \rightarrow \infty} -\frac{1}{\beta} \ln{ \int_0^1  \mathrm{d}\pi^f_i  \exp\left\{-\frac{\beta}{m} \left\{  -\mathcal{P}(x_i) \log \pi_i^f - \mathcal{N}(x_i) \log (1 - \pi_i^f) \right\} \right\} } \\
        =& \sum\limits_{x_i} \min\limits_{\pi_i^f \in (0,1)} \frac{1}{m} \left\{  -\mathcal{P}(x_i) \log \pi_i^f - \mathcal{N}(x_i) \log (1 - \pi_i^f) \right\} \\
        =& \frac{1}{m} \sum\limits_{x_i} - \mathcal{P}(x_i)\ln{\frac{\mathcal{P}(x_i)}{\mathcal{P}(x_i)+ \mathcal{N}(x_i)}} -  \mathcal{N}(x_i) \ln \frac{\mathcal{N}(x_i)}{\mathcal{P}(x_i)+ \mathcal{N}(x_i)},
    \end{aligned}
\end{align}
in which 
\begin{align}
    \begin{aligned}
        f^*_{\text{Logistic}}(x_i) = \frac{\mathcal{P}(x_i) }{\mathcal{P}(x_i)+ \mathcal{N}(x_i)}
    \end{aligned}
\end{align}
for each $x_i\in\mathcal{S}$.

\section{Evaluation Measurements and Statistical Ensembles}\label{sec3}

The evaluation of the performance of binary classifiers involves various metrics, among which the Receiver Operating Characteristic (ROC) curve and the Precision-Recall (PR) curve are prominent. These metrics offer insights into the effectiveness of a classifier at different threshold settings. This note delves into the optimal ROC and PR curves~\cite{fawcett2006introduction,yang2022auc}, providing a mathematical exposition of their derivation and interpretation in the context of a given dataset.

\subsection{Optimal ROC Curve and statistical ensemble on \text{AR} cost function}

\subsubsection{Optimal ROC Curve}
The ROC curve represents the trade-off between the true positive rate (TPR) and the false positive rate (FPR) of a classifier. The curve is constructed by plotting TPR against FPR at various threshold levels. An optimal ROC curve approaches the top-left corner of the plot, indicating both high TPR and low FPR.

Consider a dataset $\mathcal{S} = \{x_1, x_2, \ldots, x_m\}$ with a feature domain $\mathcal{X}$. A classifier $f$ assigns a score to each instance in $\mathcal{S}$, resulting in a sorted sequence $\mathcal{S}^f = \{x_1^f, x_2^f, \ldots, x_m^f\}$, where $f(x_i^f) \geq f(x_j^f)$ for $i < j$. The real label corresponding to $x_i^f$ is denoted as $y_i^f$. To construct the ROC curve, one must calculate the TPR and FPR at various thresholds $t \in \{1, \ldots, m\}$ where each $t=k$ indicates that instances $x_1^f,\ldots,x_k^f$ are classified as positive. The TPR and FPR are given by:
\begin{equation}
\text{TPR} = \frac{\text{TP}}{n_+} = \frac{1}{n_+} \sum_{i=1}^k \frac{1}{2} (1 + y^f_i),
\end{equation}

\begin{equation}
\text{FPR} = \frac{\text{FP}}{n_-} = \frac{1}{n_-} \sum_{i=1}^k \frac{1}{2} (1 - y^f_i).
\end{equation}

To find the optimal ROC curve, the objective is to maximize TPR and minimize FPR for each threshold $k$. This bi-objective optimization can be expressed as:
\begin{equation}\label{eq:bi-objective-optimization}
\begin{aligned}
& \underset{f}{\text{max}} \quad \sum_{i=1}^k \frac{1}{2} (1 + y^f_i), \\
& \underset{f}{\text{min}} \quad \sum_{i=1}^k \frac{1}{2} (1 - y^f_i).
\end{aligned}
\end{equation}
Considering that the sum of true positives and false positives equals $k$, the optimization problem in Eq.~\eqref{eq:bi-objective-optimization} simplifies to:

\begin{equation}\label{eq:optimization-problem}
\max \quad \sum_{i=1}^k \frac{1}{2} (1 + y^f_i).
\end{equation}

The calculation of FPR for a fixed $k$ includes instances that either exceed or equal the score $f(x_k^f)$. However, the optimization is primarily concerned with the maximization of true positives. Therefore we have
\begin{align}
    \sum_{i=1}^k \frac{1}{2} \left(1+ y^f_i \right) &= \sum_{i=1}^m \frac{1}{2}\mathbbm{1}\left( f(x_i) > f(x_k^f) \right) \left(1+ y_i \right) + \frac{\alpha}{2} \mathbbm{1}\left( f(x_i) = f(x_k^f) \right) \left(1+ y_i \right),
\end{align}
in which~$\alpha = \frac{k-\sum_{i=1}^m\mathbbm{1}\left( f(x_i) > f(x_k^f) \right) }{\sum_{i=1}^m \mathbbm{1}\left( f(x_i) = f(x_k^f) \right) }$~is the ratio of instances with score $f(x_k^f)$ in the subset $\{ x_1^f,x_2^f,\cdots,x_k^f\}$ to all instances with score $f(x_k^f)$. Without loss of generality, we assume that $f(x_i) = f(x_j) \iff x_i=x_j$ for every $x_i,x_j \in \mathcal{S}$. Then we obtain that
\begin{align}
    \begin{aligned}
        &\sum\limits_{i=1}^k \frac{1}{2} \Big(1+ y^f_i \Big) \\
        =&\sum\limits_{x_i\in\mathcal{S}} \mathbb{I}\left( f(x_i) > f(x_k^f) \right) \mathcal{P}(x_i) + \alpha\sum\limits_{x_i\in\mathcal{S}} \mathbb{I}\left( f(x_i) = f(x_k^f) \right)   \mathcal{P}(x_i) \\
        =& \sum\limits_{i=1}^m \mathbb{I}\left( f(x_i) > f(x_k^f) \right) \frac{\mathcal{P}(x_i)}{\mathcal{P}(x_i)+\mathcal{N}(x_i)}+ \alpha\sum\limits_{i=1}^m \mathbb{I}\left( f(x_i) = f(x_k^f) \right) \frac{\mathcal{P}(x_k^f)}{\mathcal{P}(x_k^f)+\mathcal{N}(x_k^f)} \\
        =& \sum\limits_{i=1}^m \mathbb{I}\left( f(x_i) > f(x_k^f) \right) \frac{\mathcal{P}(x_i)}{\mathcal{P}(x_i)+\mathcal{N}(x_i)}+ \left( k-\sum\limits_{i=1}^m \mathbb{I}\left( f(x_i) > f(x_k^f) \right) \right) \frac{\mathcal{P}(x_k^f)}{\mathcal{P}(x_k^f)+\mathcal{N}(x_k^f)} \\
        =& \sum\limits_{i=1}^m \frac{\mathcal{P}(x_i^f)}{\mathcal{P}(x_i^f)+\mathcal{N}(x_i^f)}.
    \end{aligned}
\end{align}
Denote $w_i=\frac{\mathcal{P}(x_i)}{\mathcal{P}(x_i) + \mathcal{N}(x_i)}$ as the non-negative weight of instance $x_i$. Hence, the optimization problem~(\ref{eq:optimization-problem}) can be regarded as a problem of how to choose a $k$-elements subset of $\mathcal{S}$ which satisfies that the total weight is maximum. And it is naturally rewritten in the form of combinatorial optimization as follows,
\begin{align}
    \begin{aligned}
        \max & \quad \sum\limits_{i=1}^m  w_iz_i\\
        \text{s.t.}& \quad \sum\limits_{i=1}^m z_i=k\\
        & \quad   z_i\in\{0,1\}, \  i =1,2,\cdots,m
    \end{aligned}
\end{align}
Actually, this combinatorial optimization problem belongs to the classical 0-1 knapsack problems, which is the most common problem being solved. Noting that all weights are non-negative, simple greedy algorithm can reach the optimal solution if we select the top $k$ instances with the highest weight. As a rule of how to select optimal $k$-element subset with arbitrary $k$, the optimal classifier for best ROC curve is the weight function, that is,
\begin{align}
    \begin{aligned}
        f^*_{\text{ROC}} (x_i) = \frac{\mathcal{P}(x)}{\mathcal{P}(x_i) + \mathcal{N}(x_i)}
    \end{aligned}
\end{align}
for every $x_i\in\mathcal{S}$.
Here, we denote that $\{x^*_1,x^*_2,\cdots,x^*_n\}$ is the sorted sequence of $\mathcal{S}$ in descending order of weight.

Naturally, the best ROC curve can be drawn sequentially from a series of data points in (FPR, TPR)-plane as follows,
\begin{align}
\begin{aligned}
\left\{\left(\frac{1}{n_-}\sum\limits_{i=1}^k  \frac{\mathcal{N}(x^*_i)}{\mathcal{P}(x^*_i) + \mathcal{N}(x^*_i)},\frac{1}{n_+}\sum\limits_{i=1}^k  \frac{\mathcal{P}(x^*_i)}{\mathcal{P}(x^*_i) + \mathcal{N}(x^*_i)}\right)\right\}_{k=0,1,\cdots,n}
\end{aligned}.
\end{align}
The optimal ROC curve can be regarded as a combination of $n$ linear piecewise functions, whose derivatives are composed of the following sequence
\begin{align}
    \begin{aligned}
        \left\{  \frac{n_- \mathcal{P}(x^*_i)}{n_+\mathcal{N}(x^*_i)}\right\}_{k=1,2,\cdots,n}
    \end{aligned}.
\end{align}
It is easy to check that, the above sequence is monotonically decreasing, since $\{x^*_1,x^*_2,\cdots,x^*_n\}$ is sorted by descending order of weight and
\begin{align}
\begin{aligned}
\frac{\mathcal{P}(x^*_i)}{\mathcal{P}(x^*_i) + \mathcal{N}(x^*_i)}=\frac{1}{1+\frac{1}{\frac{\mathcal{P}(x^*_i)}{\mathcal{N}(x^*_i)}}}
\end{aligned}.
\end{align}
Therefore, the best ROC curve is \textbf{concave}. Moreover, the area under the best ROC curve is the upper bound of AR ($text{AR}^u$). In other words, $\text{AR}^u$ is equal to the area under the curve described as $ f^*_{\text{ROC}}$ in the following:
\begin{align}
    \begin{aligned}
        \text{AR}^u =& \sum\limits_{i=1}^m  \frac{1}{n_-} \frac{\mathcal{N}(x^*_i)}{\mathcal{P}(x^*_i) + \mathcal{N}(x^*_i)} \left( \frac{1}{n_+}\sum\limits_{j<i} \frac{\mathcal{P}(x^*_j)}{\mathcal{P}(x^*_j) + \mathcal{N}(x^*_j)} +   \frac{1}{2n_+} \frac{\mathcal{P}(x^*_i)}{\mathcal{P}(x^*_i) + \mathcal{N}(x^*_i)}   \right) \\
        =& \frac{1}{n_-n_+} \sum\limits_{i>j} \frac{\mathcal{N}(x^*_i)\mathcal{P}(x^*_j)}{\Big( \mathcal{P}(x^*_i) + \mathcal{N}(x^*_i)\Big) \left( \mathcal{P}(x^*_j) + \mathcal{N}(x^*_j)\right)} + \frac{1}{2} \sum\limits_{i=1}^m \frac{\mathcal{N}(x^*_i)\mathcal{P}(x^*_i)}{\Big( \mathcal{P}(x^*_i) + \mathcal{N}(x^*_i)\Big)^2} \\
        =& \frac{1}{2n_-n_+} \sum\limits_{i,j} \frac{\max\Big\{ \mathcal{N}(x^*_i)\mathcal{P}(x^*_j), \mathcal{N}(x^*_j)\mathcal{P}(x^*_i)\Big\}}{\Big( \mathcal{P}(x^*_i) + \mathcal{N}(x^*_i)\Big)\left( \mathcal{P}(x^*_j) + \mathcal{N}(x^*_j)\right)}  \\
        =& \frac{1}{2n_-n_+} \sum\limits_{i,j} \frac{\max\Big\{ \mathcal{N}(x_i)\mathcal{P}(x_j), \mathcal{N}(x_j)\mathcal{P}(x_i)\Big\}}{\Big( \mathcal{P}(x_i) + \mathcal{N}(x_i)\Big)\Big( \mathcal{P}(x_j) + \mathcal{N}(x_j)\Big)}  \\
        =& \frac{1}{2n_-n_+} \sum\limits_{x_i,x_j} \max\Big\{ \mathcal{P}(x_i)\mathcal{N}(x_j) , \mathcal{P}(x_j)\mathcal{N}(x_i) \Big\}.
    \end{aligned}
\end{align}

\subsubsection{Statistical ensemble on \text{AR} cost function}
\label{sec:ar ensemble}
For a given dataset $\mathcal{S}:=\{x_1,x_2,\cdots,x_m\}$, one classifier $f$ can be transformed into a ranking of $S$ according to the score given by $f$: 
\begin{align}
    \pi_f:=\{x^f_{1},x^f_{2},\cdots,x^f_{m}\},
\end{align}
where $x^f_{i}$ is the $i$-th sample of the ordered set ranked by $f$ in $\mathcal{S}$. Here, we define that $\Pi(\mathcal{S})$ as the rankings of all possible classifiers for $\mathcal{S}$. Then, we define the energy of $\pi^f$ as follows,
\begin{align}
    E^{AR}(\pi^f) =& 1 - AR(\pi^f) \\
    =& 1- \sum_{i < j} p_+(x^f_{i}) p_-(x^f_{j}) - \frac{1}{2}\sum_i p_+(x^f_{i}) p_-(x^f_{i}) \\
    =&  \sum_{i > j} p_+(x^f_{i}) p_-(x^f_{j}) + \frac{1}{2}\sum_i p_+(x^f_{i}) p_-(x^f_{i})
\end{align}

Treating $\Pi(\mathcal{S})$ as an ensemble, we can write the following partition function,
\begin{align}
\begin{aligned}
    Z^{AR} =& \sum_{\pi^f \in \Pi(\mathcal{S})} \exp(-\beta E^{AR}(\pi))\\ 
=& \sum_{\pi^f \in \Pi(\mathcal{S})} \exp \left\{ -\beta\left(  \sum_{i > j} p_+(x^f_{i}) p_-(x^f_{j}) + \frac{1}{2}\sum_i p_+(x^f_{i}) p_-(x^f_{i}) \right) \right\}.
\end{aligned}
\end{align}
And, the ground state energy $E^{AR}_0$ ($1-AR^u$) can be defined as 
\begin{align}
    E^{AR}_0 = \lim\limits_{\beta \rightarrow \infty} -\frac{1}{\beta} \ln{Z^{AR}} .
\end{align}

Next, we can solve $E^{AR}_0$ as follows.
\begin{align}
    \begin{aligned}
         E^{AR}_0 =& \lim\limits_{\beta \rightarrow \infty} -\frac{1}{\beta} \ln{Z^{AR}} \\
         =&\lim\limits_{\beta \rightarrow \infty} -\frac{1}{\beta} \ln \sum_{\pi_f \in \Pi(\mathcal{S})} \exp \left\{ -\beta\left(  \sum_{i > j} p_+(x^f_{i}) p_-(x^f_{j}) + \frac{1}{2}\sum_i p_+(x^f_{i}) p_-(x^f_{i}) \right) \right\} \\
         =& \lim\limits_{\beta \rightarrow \infty} -\frac{1}{\beta} \ln \sum_{\pi_f \in \Pi(\mathcal{S})}  \prod_{i>j} \exp\Big(-\beta p_+(x^f_{i}) p_-(x^f_{j})\Big) \prod_{i=1}^n \exp\Big(-\frac{\beta}{2} p_+(x^f_{i}) p_-(x^f_{i})\Big) \\
         =& \lim\limits_{\beta \rightarrow \infty} -\frac{1}{\beta} \ln \prod\limits_{i,j} \Big(  e^{-\frac{\beta}{2} p_+(x^f_{i}) p_-(x^f_{j})} + e^{-\frac{\beta}{2} p_+(x^f_{j}) p_-(x^f_{i})} \Big) \\
         =& \lim\limits_{\beta \rightarrow \infty} -\frac{1}{\beta} \sum\limits_{i,j} \ln \Big(  e^{-\frac{\beta}{2} p_+(x^f_{i}) p_-(x^f_{j})} + e^{-\frac{\beta}{2} p_+(x^f_{j}) p_-(x^f_{i})} \Big)  \\
         =& \sum\limits_{i,j} \lim\limits_{\beta \rightarrow \infty} -\frac{1}{\beta} \ln \Big(  e^{-\frac{\beta}{2} p_+(x^f_{i}) p_-(x^f_{j})} + e^{-\frac{\beta}{2} p_+(x^f_{j}) p_-(x^f_{i})} \Big) \\
         =&\frac{1}{2} \sum\limits_{i,j} \min\Big\{ p_+(x_{j}) p_-(x_{k}), p_+(x_{k}) p_-(x_{j})  \Big\}  \\
         =& \frac{1}{2n_+n_-}\sum\limits_{x_i,x_j}\min\Big\{\mathcal{P}(x_i)\mathcal{N}(x_j) ,\mathcal{P}(x_j)\mathcal{N}(x_i) \Big\}.
    \end{aligned}
\end{align}
Then, we know that
\begin{align}
    AR^u = 1-E^{AR}_0 = \frac{1}{2n_+n_-}\sum\limits_{x_i,x_j}\max\Big\{\mathcal{P}(x_i)\mathcal{N}(x_j) ,\mathcal{P}(x_j)\mathcal{N}(x_i) \Big\}.
\end{align}

Further, it is worthy noting that
\begin{align}
\begin{aligned}
    AR^u =&\frac{1}{2} \sum\limits_{i, j} \max\Big\{ p_+(x_{i}) p_-(x_{j}), p_+(x_{j}) p_-(x_{i})  \Big\}  \\
    =&\sum\limits_{i < j}  \max\Big\{ p_+(x_{i}) p_-(x_{j}), p_+(x_{j}) p_-(x_{i})  \Big\}  + \frac{1}{2}\sum\limits_{i} p_+(x_{i}) p_-(x_{i})\\
    =& \sum\limits_{i < j} \max \left\{ p_+(x_{i}) \Big(1-p_+(x_{j}) \Big), p_+(x_{j}) \Big(1-p_+(x_{i}) \Big)   \right\} + \frac{1}{2}\sum\limits_{i} p_+(x_{i}) p_-(x_{i})  \\
    =&\sum\limits_{i < j} \left( \max \Big\{ p_+(x_{i}) , p_+(x_{j})    \Big\}  -  p_+(x_{i}) p_+(x_{j}) \right) + \frac{1}{2}\sum\limits_{i} p_+(x_{i}) p_-(x_{i})\\
    =& \sum\limits_{i < j: \ p_+(x_{i}) > p_+(x_{j}) } \Big( p_+(x_{i}) -  p_+(x_{i}) p_+(x_{j}) \Big) + \frac{1}{2}\sum\limits_{i} p_+(x_{i}) p_-(x_{i})\\
    =& \sum\limits_{i < j: \ p_+(x_{i}) > p_+(x_{j}) }  p_+(x_{i})  p_-(x_{j}) + \frac{1}{2}\sum\limits_{i} p_+(x_{i}) p_-(x_{i}).
\end{aligned}
\end{align}
Naturally, we construct a ranking $\pi^*=\{ x_1^*,x^*_2,\cdots,x^*_m \}$ for $\mathcal{S}$ satisfying that $p_+(x^*_i) > p_+(x^*_j)$ for any $i < j$, whose AUC is equal to $E^{AR}_0$ as follows,
\begin{align}
    E^{AR}(\pi_0) =& \sum\limits_{j < k: \ p_+(x'_{j}) > p_+(x'_{k}) }  p_+(x'_{j})  p_-(x'_{k}) + \frac{1}{2}\sum\limits_{j} p_+(x'_{j}) p_-(x'_{j})=E_0
\end{align}
It also means that, the optimal classifier $f^*_{AR}$ corresponding to $E^{AR}_0$ is 
\begin{align}
    f^*_{AR}(x_i) = p_+(x_{i}) , \ \  \forall x_i \in \mathcal{S}.
\end{align}

\subsection{Optimal PR Curve and statistical ensemble on \text{AP} cost function}
\subsubsection{Optimal PR Curve}\label{sec:pr}
The precision-recall (PR) curve is a widely utilized metric for evaluating binary classifiers, emphasizing the trade-off between precision and recall. It provides a detailed view of a classifier's performance at various decision thresholds. To plot a PR curve, one must compute precision and recall at numerous thresholds. This concept is analogous to the receiver operating characteristic (ROC) curve, which allows adjustment of the threshold to balance precision and recall.

For each threshold value, precision and recall are determined using the following formulas:
\begin{align}
    \text{precision} = \frac{\text{TP}}{\text{TP} + \text{FP}},
\end{align}
where TP is the number of true positives and FP is the number of false positives. Recall, also known as true positive rate (TPR), is defined as:
\begin{align}
    \text{recall} = \frac{\text{TP}}{\text{TP} + \text{FN}},
\end{align}
where FN is the number of false negatives. Both the ROC and PR curves share a commonality in that they evaluate the classifier's performance by computing metrics at various thresholds. At a specific threshold \( k \), the precision and recall can be expressed as:
\begin{align}
    \text{precision} &= \frac{\text{TP}}{k} = \frac{n_+}{k} \times \text{TPR}, \\
    \text{recall} &= \text{TPR},
\end{align}
where \( n_+ \) is the total number of positive samples.

The objective in seeking the optimal PR curve is to maximize both precision and recall for a given threshold \( k \). This can be formulated as the following optimization problem:
\begin{align}
    \max_f \ \text{TPR},
\end{align}
which we have addressed in the previous section. The optimal classifier that achieves the best PR curve is identical to the one that optimizes the ROC curve and is given by:
\begin{align}
    f^*_{\text{PR}}(x) = f^*_{\text{ROC}}(x) = \frac{\mathcal{P}(x)}{\mathcal{P}(x) + \mathcal{N}(x)},
\end{align}
where \( x \) belongs to the sample space \( \mathcal{S} \). The optimal PR curve is constructed by plotting a series of data points in the (recall, precision)-plane, which are calculated as follows:
\begin{align}
\left\{\left(\frac{1}{n_+}\sum\limits_{i=1}^k \frac{\mathcal{P}(x^*_i)}{\mathcal{P}(x^*_i) + \mathcal{N}(x^*_i)}, \frac{1}{k}\sum\limits_{i=1}^k \frac{\mathcal{P}(x^*_i)}{\mathcal{P}(x^*_i) + \mathcal{N}(x^*_i)}\right)\right\}_{k=0,1,\cdots,n}
\end{align}

Furthermore, the upper bound of the area under the PR curve ($\text{AP}^u $) can be calculated as follows:
\begin{align}
    \text{AP}^u = \frac{1}{2n_+} \sum\limits_{x^*_i} p_+(x^*_i) \left( \frac{1}{i-1}\sum\limits_{j=1}^{i-1} p_+(x^*_j) + \frac{1}{i}\sum\limits_{j=1}^{i} p_+(x^*_j) \right),
\end{align}
where \( p_+(x_i) \) is the probability of a sample \( x_i \) being positive:
\begin{align}
    p_+(x_i) = \frac{\mathcal{P}(x_i)}{\mathcal{P}(x_i) + \mathcal{N}(x_i)}.
\end{align}

\subsubsection{statistical ensemble on \text{AP} cost function}
Akin to $AR$ ensemble, we define that $\Pi(\mathcal{S})$ includes all possible classifiers for $\mathcal{S}$. For any classifier $\pi_f  \in \Pi(\mathcal{S})$, we define its energy as $1-AP(\pi_f)$. As a consequence, the partition function and ground state of $AP$ ensemble can be written as 
\begin{align}
   Z^{AP} = \sum_{\pi^f \in \Pi(\mathcal{S})} \exp\Big(-\beta \left(1-AP(\pi^f)\right)\Big)
\end{align}
and
\begin{align}
   E^{AP}_0 =  \lim\limits_{\beta \rightarrow \infty} -\frac{1}{\beta} \ln{Z^{AP}}.
\end{align}

However, it is very difficult to directly calculate the above ground state because there is no explicit expression for $AP$, unlike $AR$. Alternatively, we have proved in Sec. \ref{sec:pr} that, the optimal $AR$ shares the same classifier (ranking) with the optimal $AP$, that is,
\begin{align}
    f^*_{AP}(x_i) =f^*_{AR}(x_i) = p_+(x_{i}) , \ \  \forall x_i \in \mathcal{S}.
\end{align}
Furthermore, we can write the $AP^u$ in an indirect expression:
\begin{align}\label{eq_pr}
\text{AP}^{u} \!=\!\sum\limits_{i=1}^m    \frac{p_+(x^*_i)}{2n_+}
\!\left(\!\frac{1}{i-1}\sum\limits_{j=1}^{i-1} p_+(x^*_j) \!+\!   \frac{1}{i} \sum\limits_{j=1}^i  p_+(x^*_j)\!  \right)\!,
\end{align}
where $\pi^*=\{ x_1^*,x^*_2,\cdots,x^*_m \}$ has been mentioned in Sec. \ref{sec:ar ensemble}.

\subsection{Optimal Accuracy and statistical ensemble on \text{AC} cost function}
\subsubsection{Optimal Accuracy}
Accuracy (AC) is a fundamental metric that quantifies the proportion of correctly predicted instances against the total number of instances within a dataset. This metric is universally applicable across classifiers. Given a dataset $\mathcal{S}$ and a feature space $\mathcal{X}$, accuracy can be computed with the following expression:
\begin{align}
    \text{AC} = \frac{1}{2m} \sum\limits_{x_i \in \mathcal{S}} \Big( 1 + f(x_i)y_i \Big),
\end{align}
where $m$ represents the total number of instances, $f(x_i)$ is the predicted label for instance $x_i$, and $y_i$ is the true label. The accuracy increases by $\frac{1}{2m}$ for each correctly classified instance $x_i$; it remains unchanged for incorrect predictions. 

While accuracy itself is not a conventional loss function, optimizing for the highest possible accuracy and determining the optimal classifier functions are critical endeavors in machine learning. The mathematical upper limit of accuracy, denoted as $\text{AC}^u$, can be formalized as:
\begin{align}
    \text{AC}^u = \max_{f} \frac{1}{2m} \sum\limits_{x_i \in \mathcal{S}} \Big( 1 + f(x_i)y_i \Big),
\end{align}
with the corresponding optimal classifier being:
\begin{align}
    f^*_{\text{AC}} = \arg \max_{f} \frac{1}{2m} \sum\limits_{x_i \in \mathcal{S}} \Big( 1 + f(x_i)y_i \Big).
\end{align}
Through further analysis, we can expand and simplify the equation by considering the relationship between probabilities associated with positive and negative instances:
\begin{align}
    \max_{f} \sum\limits_{x_i \in \mathcal{S}} \frac{1}{2m}\Big( 1 + f(x_i)y_i \Big) &= \max_{f} \sum\limits_{x \in \mathcal{X}} \frac{\mathcal{P}(x)}{2}\Big( 1 + f(x) \Big) + \frac{\mathcal{N}(x)}{2}\Big( 1 - f(x) \Big) \nonumber \\ 
    &\leq \sum\limits_{x_i \in \mathcal{S}} \max_{f(x)} \frac{\mathcal{P}(x_i)}{2}\Big( 1 + f(x_i) \Big) + \frac{\mathcal{N}(x_i)}{2}\Big( 1 - f(x_i) \Big),
\end{align}
which leads us to redefine $\text{AC}^u$ as:
\begin{align}
    \text{AC}^u = \frac{1}{m} \sum\limits_{x_i} \max_{f(x_i)} \frac{\mathcal{P}(x_i)}{2}\Big( 1 + f(x_i) \Big) + \frac{\mathcal{N}(x_i)}{2}\Big( 1 - f(x_i) \Big),
\end{align}
and the optimal classifier for each instance $x_i \in \mathcal{S}$ becomes:
\begin{align}
    f^*_{\text{AC}}(x_i) = \arg \max_{f(x_i)} \frac{\mathcal{P}(x_i)}{2}\Big( 1 + f(x_i) \Big) + \frac{\mathcal{N}(x_i)}{2}\Big( 1 - f(x_i) \Big).
\end{align}
By enumeration, the solution for the optimal classifier is:
\begin{align}
    f^*_{\text{AC}}(x_i) = 
    \begin{cases}
        1 & \text{if } \mathcal{P}(x_i) \geq \mathcal{N}(x_i), \\
        -1 & \text{if } \mathcal{P}(x_i) < \mathcal{N}(x_i),
    \end{cases}
\end{align}
resulting in the upper limit of accuracy being:
\begin{align}
    \text{AC}^u = \frac{1}{m} \sum\limits_{x_i \in \mathcal{S}} \max\Big\{ \mathcal{P}(x_i), \mathcal{N}(x_i) \Big\}.
\end{align}

\subsubsection{statistical ensemble on \text{AC} cost function}
Unlike $AR$ and $AP$, $AC$ focuses on the predicted binary label for every sample, rather than a ranking of all samples in $\mathcal{S}$. In this context, every classifier $f$ can be transformed into a $m$-dimensional binary vector
\begin{align}
    \pi^f=(x_1^f, x_2^f,\cdots,x_m^f),
\end{align}
where $x_i^f\in\{-1,1\}$ for $i=1,2,\cdots,m$. This also means that $\Pi(\mathcal{S}) = \{1,-1\}^m$ for $AC$ cost function. And, the energy of $\pi^f$ is defined as
\begin{align}
\begin{aligned}
     E^{AC}(\pi^f) =& 1-AC(\pi^f) \\
    =&1- \left(\frac{1}{m} \sum_{i}  \frac{\mathcal{P}(x_i)}{2}\Big( 1 + x^f_i \Big) + \frac{\mathcal{N}(x_i)}{2}\Big( 1 - x^f_i \Big) \right)\\
    =& \frac{1}{m} \sum_{i} \frac{\mathcal{P}(x_i)}{2}\Big( 1 - x^f_i \Big) + \frac{\mathcal{N}(x_i)}{2}\Big( 1 + x^f_i \Big)\\
    =& \frac{1}{2} +  \frac{1}{2m} \sum_{i} x^f_i \Big(\mathcal{N}(x_i) - \mathcal{P}(x_i)\Big).
\end{aligned}
\end{align}
Then, the partition function is 
\begin{align}
\begin{aligned}
     Z^{AC} =& \sum_{\pi_f\in \Pi(\mathcal{S})} \exp(-\beta E^{AC}(\pi^f))\\
    =& \sum_{\pi^f\in \Pi(\mathcal{S})} \exp\left\{-\beta\left( \frac{1}{2} +  \frac{1}{2m} \sum_{i} x^f_i \Big(\mathcal{N}(x_i) - \mathcal{P}(x_i)\Big) \right) \right\}
\end{aligned}
\end{align}
And, the ground state is calculated as
\begin{align}
    \begin{aligned}
        E^{AC}_0 = &  \lim\limits_{\beta \rightarrow \infty} -\frac{1}{\beta} \ln{Z^{AC}} \\
        =& \lim\limits_{\beta \rightarrow \infty} -\frac{1}{\beta} \ln \sum_{\pi_f\in \Pi(\mathcal{S})} \exp\left\{-\beta\left( \frac{1}{2} +  \frac{1}{2m} \sum_{i} x^f_i \Big(\mathcal{N}(x_i) - \mathcal{P}(x_i)\Big) \right) \right\} \\
        =& \lim\limits_{\beta \rightarrow \infty} -\frac{1}{\beta} \ln \sum_{\pi_f\in \Pi(\mathcal{S})} e^{-\beta/2} \prod_{i} \exp\left(-\frac{\beta}{2m}x^f_i \Big(\mathcal{N}(x_i) - \mathcal{P}(x_i)\Big)\right) \\
        =& \lim\limits_{\beta \rightarrow \infty} -\frac{1}{\beta} \ln\left\{ e^{-\beta/2}  \prod_{i}  \Big( e^{-\frac{\beta}{2m}(\mathcal{N}(x_i) - \mathcal{P}(x_i)) } + e^{-\frac{\beta}{2m}(\mathcal{P}(x_i) - \mathcal{N}(x_i)) } \Big)  \right\} \\
        =& \frac{1}{2} -\lim\limits_{\beta \rightarrow \infty} \frac{1}{\beta} \ln\left\{   \prod_{i}  \Big( e^{-\frac{\beta}{2m}(\mathcal{N}(x_i) - \mathcal{P}(x_i)) } + e^{-\frac{\beta}{2m}(\mathcal{P}(x_i) - \mathcal{N}(x_i)) } \Big)  \right\} \\
        = & \frac{1}{2} - \sum\limits_i  \lim\limits_{\beta \rightarrow \infty} \frac{1}{\beta} \ln\left\{ e^{-\frac{\beta}{2m}(\mathcal{N}(x_i) - \mathcal{P}(x_i)) } + e^{-\frac{\beta}{2m}(\mathcal{P}(x_i) - \mathcal{N}(x_i)) } \right\} \\
        =& \frac{1}{2} + \frac{1}{2m} \sum\limits_i \min\Big\{\mathcal{N}(x_i) - \mathcal{P}(x_i),  \mathcal{P}(x_i) - \mathcal{N}(x_i) \Big\} \\
        =& \frac{1}{m} \sum\limits_i \min\Big\{\mathcal{P}(x_i),\mathcal{N}(x_i)  \Big\}
    \end{aligned}
\end{align}
Indeed, 
\begin{align}
    AC^u = 1-E^{AC}_0 = \frac{1}{m} \sum\limits_i \max\Big\{\mathcal{P}(x_i),\mathcal{N}(x_i)  \Big\}.
\end{align}
Naturally, the optimal classifier corresponding to $AC^u$ is
\begin{align}
    f^*_{AC}(x_i)=\left\{
    \begin{aligned}
        1& & \text{if} \ \mathcal{P}(x_i) \geq \mathcal{N}(x_i) \\
        -1& & \text{if} \ \mathcal{P}(x_i) < \mathcal{N}(x_i) 
    \end{aligned}
    \right. ,
\end{align}
for any $x_i\in\mathcal{S}$.

\section{Sensitivity Analysis}

In Sections~2 and~3, we established the mathematical relationships between the lower bound of the objective function with respect to the training set and the upper bound of the evaluation metrics related to the test set. This section extends that discussion within the out-of-sample framework by combining these relationships to investigate the dual concerns of training loss and generalizability in classification tasks.

Let us denote the training and test sets by $\mathcal{S}_{train}$ and $\mathcal{S}_{test}$, respectively. Furthermore, we denote $\mathcal{P}_{train}(x_i)$ and $\mathcal{N}_{train}(x_i)$ as the respective counts of positive and negative instances with feature $x_i$ in the training set, and similarly, $\mathcal{P}_{test}(x_i)$ and $\mathcal{N}_{test}(x_i)$ for the test set.

For a discrete classifier $f(x)$, the boundary hinge loss within the training set $\mathcal{S}_{train}$ is defined as:
\begin{align}
    \sum_{x_i} \min\{\mathcal{P}_{train}(x_i), \mathcal{N}_{train}(x_i)\}.
\end{align}
Here, $\Delta_{train}$ represents the discrepancy between the classifier's hinge loss and the boundary hinge loss, expressed as:
\begin{align}\nonumber
    \Delta_{train}=\sum_{x_i} \Delta_{train}(x_i),
\end{align}
where
$$
\Delta_{train}(x_i)=
\begin{cases}
    \mathcal{N}_{train}(x_i) - \min\{\mathcal{P}_{train}(x_i), \mathcal{N}_{train}(x_i)\} & \text{if } f(x_i)=1, \\
    \mathcal{P}_{train}(x_i) - \min\{\mathcal{P}_{train}(x_i), \mathcal{N}_{train}(x_i)\} & \text{if } f(x_i)=-1.
\end{cases}
$$

In parallel, the maximum accuracy achievable on the test set $\mathcal{S}_{test}$ is:
$$
\sum_{x_i} \max\{\mathcal{P}_{test}(x_i), \mathcal{N}_{test}(x_i)\}.
$$
Similarly, $\Delta_{test}$ quantifies the error between the classifier's actual accuracy and the maximum possible accuracy:
$$
\Delta_{test}=\sum_{x_i} \Delta_{test}(x_i),
$$
where
$$
\Delta_{test}(x_i)=
\begin{cases}
    \max\{\mathcal{P}_{test}(x_i), \mathcal{N}_{test}(x_i)\} - \mathcal{P}_{test}(x_i) & \text{if } f(x_i)=1, \\
    \max\{\mathcal{P}_{test}(x_i), \mathcal{N}_{test}(x_i)\} - \mathcal{N}_{test}(x_i) & \text{if } f(x_i)=-1.
\end{cases}
$$

Therefore, the summation of these two discrepancies represents their respective optimization potential and performance evaluation capacity. For any given feature $x_i$, the combined error is:
\begin{align} \nonumber
    &(\Delta_{train} + \Delta_{test})(x_i) =\begin{cases} \nonumber
   \max\{\mathcal{N}_{train}(x_i)-\mathcal{P}_{train}(x_i),0 \} + \max\{\mathcal{N}_{test}(x_i)-\mathcal{P}_{test}(x_i), 0\}   & \text{if } f(x_i)=1, \\
    \max\{\mathcal{P}_{train}(x_i)-\mathcal{N}_{train}(x_i),0 \} + \max\{\mathcal{P}_{test}(x_i)-\mathcal{N}_{test}(x_i), 0\} & \text{if } f(x_i)=-1.
\end{cases}
\end{align}

We now introduce a lower bound for $(\Delta_{train} + \Delta_{test})$, defined as:
\begin{align}\label{min_delta}
    \Delta = (\Delta_{train} + \Delta_{test})_{\min} = \sum_{x_i} \Delta(x_i),
\end{align}
where
\begin{align}\nonumber
    &\Delta(x_i)=\begin{cases} 
    0 & \text{if } \mathcal{P}_{train}(x_i) \geq \mathcal{N}_{train}(x_i), \ \mathcal{P}_{test}(x_i) \geq \mathcal{N}_{test}(x_i)\\
    0 & \text{if } \mathcal{P}_{train}(x_i) < \mathcal{N}_{train}(x_i), \ \mathcal{P}_{test}(x_i) < \mathcal{N}_{test}(x_i)\\
    \epsilon(x_i) & \text{if } \mathcal{P}_{train}(x_i) \geq \mathcal{N}_{train}(x_i), \ \mathcal{P}_{test}(x_i) < \mathcal{N}_{test}(x_i)\\
    \epsilon(x_i) & \text{if } \mathcal{P}_{train}(x_i) < \mathcal{N}_{train}(x_i), \ \mathcal{P}_{test}(x_i) \geq \mathcal{N}_{test}(x_i)
    \end{cases}
\end{align}
and
$$\epsilon(x_i) = \min\{|\mathcal{P}_{train}(x_i)-\mathcal{N}_{train}(x_i)|, |\mathcal{P}_{test}(x_i)-\mathcal{N}_{test}(x_i)|\} .$$
As we observed, $\Delta$ is a general lower bound for the sum of training error and evaluation error regardless of the specific classifier in use. Naturally, $ \Delta=0$ if and only if
\begin{align}\nonumber
    \left\{ 
    \begin{aligned}
        &\mathcal{P}_{train}(x_i) \geq \mathcal{N}_{train}(x_i) \\
        &\mathcal{P}_{test}(x_i) \geq \mathcal{N}_{test}(x_i) 
    \end{aligned}
    \right.
   \ \text{or} \
     \left\{ 
    \begin{aligned}
       & \mathcal{P}_{train}(x_i) < \mathcal{N}_{train}(x_i) \\
        & \mathcal{P}_{test}(x_i) < \mathcal{N}_{test}(x_i) 
    \end{aligned}
    \right.
\end{align}
for each $x_i\in\mathcal{S}$.

\section{Random Division}

\emph{Random division} method is a standard approach to splitting a dataset into training and testing subsets. In this method, each instance is independently assigned to the training subset with probability $p$ and to the testing subset with probability $1-p$. For a given feature vector $x_i$, the counts $\mathcal{P}_{train}(x_i)$ and $\mathcal{N}_{train}(x_i)$ follow binomial distributions $\mathcal{B}(\mathcal{P}(x_i), p)$ and $\mathcal{B}(\mathcal{N}(x_i), p)$ respectively. Considering this, we can investigate the discrepancy $\Delta$ within the context of probabilistic partitioning. We define the expected value of $\Delta$ as $\mathbb{E}[\Delta] = \frac{1}{m} \sum_{x_i} \mathbb{E}[\Delta(x_i; p)]$, which is calculated as:
\begin{equation}
\begin{split}
\mathbb{E}[\Delta(x_i; p)] 
&= \mathbb{E}\left[\min\left\{\mathcal{N}_{train}(x_i) - \mathcal{P}_{test}(x_i), \mathcal{P}_{train}(x_i) - \mathcal{N}_{test}(x_i)\right\}\right] + \\
&\quad \mathbb{E}\left[\max\left\{\mathcal{P}_{test}(x_i), \mathcal{N}_{test}(x_i)\right\}\right] - \mathbb{E}\left[\min\left\{\mathcal{P}_{train}(x_i), \mathcal{N}_{train}(x_i)\right\}\right] \\
&= \mathbb{E}\left[\max\left\{\mathcal{P}_{test}(x_i), \mathcal{N}_{test}(x_i)\right\}\right] + \mathbb{E}\left[\max\left\{\mathcal{P}_{train}(x_i), \mathcal{N}_{train}(x_i)\right\}\right] - \\
&\quad \max\left\{\mathcal{P}(x_i), \mathcal{N}(x_i)\right\},
\end{split}
\end{equation}
where the expected maxima are determined by:
\begin{equation}\nonumber
\begin{split}
\mathbb{E}[\max\{\mathcal{P}_{train}(x_i), \mathcal{N}_{train}(x_i)\}] &= \sum_{i=0}^{\mathcal{P}(x_i)} \sum_{j=0}^{\mathcal{N}(x_i)} \max\{i, j\} \binom{\mathcal{P}(x_i)}{i} \binom{\mathcal{N}(x_i)}{j} p^{\mathcal{P}(x_i) + \mathcal{N}(x_i) - i - j} (1-p)^{i+j},
\end{split}
\end{equation}
and
\begin{equation}\nonumber
\begin{split}
\mathbb{E}[\max\{\mathcal{P}_{test}(x_i), \mathcal{N}_{test}(x_i)\}] &= \sum_{i=0}^{\mathcal{P}(x_i)} \sum_{j=0}^{\mathcal{N}(x_i)} \max\{i, j\} \binom{\mathcal{P}(x_i)}{i} \binom{\mathcal{N}(x_i)}{j} p^{i+j} (1-p)^{\mathcal{P}(x_i) + \mathcal{N}(x_i) - i - j}.
\end{split}
\end{equation}
It's important to note that $\Delta$ is symmetric; its value is invariant if we exchange the roles of the training and testing subsets $\mathcal{S}_{train}$ and $\mathcal{S}_{test}$. This symmetry is apparent in the equality $\mathbb{E}[\Delta(x_i; p)] = \mathbb{E}[\Delta(x_i; 1-p)]$, as both expressions yield the same result.

We observe that $\Delta$ exhibits symmetrical behavior, as its value remains invariant under the interchange of the training set $\mathcal{S}_{train}$ and the test set $\mathcal{S}_{test}$. This symmetry property is further substantiated by the equality in expected values for complementary probabilities in the random partitioning process, expressed mathematically as $\mathbb{E}[\Delta(x_i; p)] $.

To enhance the interpretability of experimental outcomes, we derive mathematical representations for two key metrics: the expected maximum accuracy of the test set, denoted as $\text{AC}^u$, and the expected minimum hinge loss for the training set. These metrics, in the context of random partitioning, are defined by the following equations:
\begin{itemize}
    \item For the expected maximum accuracy of the test set:
\begin{equation}
\mathbb{E} \text{AC}^u = \frac{1}{m(1-p)} \sum_{x_i} \mathbb{E}\max\Big\{ \mathcal{P}_{test}(x_i), \mathcal{N}_{test}(x_i)\Big\}
\end{equation}
\item For the expected minimum hinge loss of the training set:
\begin{equation}
\mathbb{E} [\text{minimum hinge loss}] = \frac{1}{mp} \sum_{x_i} \mathbb{E}\min\Big\{ \mathcal{P}_{train}(x_i), \mathcal{N}_{train}(x_i)\Big\}
\end{equation}
\end{itemize}

The expected minimum, present in the equation for hinge loss, is elucidated as follows:
\begin{equation}
\mathbb{E}[\min\{ \mathcal{P}_{train}(x_i), \mathcal{N}_{train}(x_i)\} ]= \sum_{i=0}^{\mathcal{P}(x_i)} \sum_{j=0}^{\mathcal{N}(x_i)} \min\{ i,j \} \binom{\mathcal{P}(x_i)}{i} \binom{\mathcal{N}(x_i)}{j} p^{\mathcal{P}(x_i)+\mathcal{N}(x_i)-i-j} (1-p)^{i+j}.
\end{equation}
In these expressions, $\mathcal{P}_{train}(x_i)$ and $\mathcal{N}_{train}(x_i)$ denote the number of positive and negative instances of $x_i$ in the training set, respectively, and analogously for $\mathcal{P}_{test}(x_i)$ and $\mathcal{N}_{test}(x_i)$ in the test set. The binomial coefficients reflect the combinatorial possibilities for selecting $i$ positives out of $\mathcal{P}(x_i)$ and $j$ negatives out of $\mathcal{N}(x_i)$, factoring in the probability $p$ of an instance belonging to the training set.

\section{Overlapping and Boundary}

Indeed, training loss and evaluation metrics inherently have a lower bound (0) and an upper bound (1). However, as analyzed in Suppleentary Notes~2 and~3, the precise boundaries do not always align with these natural limits. The reason is that positive and negative samples sometimes overlap, making it impossible for any classifier to achieve 100\% prediction accuracy. Consequently, this section will further explore the quantifiable correlation between the degree of overlap among positive and negative samples in a dataset and the performance boundaries. However, prior to this exploration, our first step will be to establish a definition for the term ``overlap'' within the context of a dataset.

Given positive data distribution $\{\mathcal{P}(x_i) / n_+\}_{x_i\in\mathcal{X}}$ and negative data distribution $\{\mathcal{N}(x_i) / n_-\}_{x_i\in\mathcal{X}}$, these two probability distributions can be abbreviated as $$\mathcal{P}:=\{\hat{p}(x_i)|x_i\in\mathcal{S}\}$$ and $$\mathcal{Q}:=\{\hat{n}(x_i)|x_i\in\mathcal{S}\}$$ respectively, in which $\hat{p}(x_i)=\mathcal{P}(x_i)/n_+$ and $\hat{n}(x_i)=\mathcal{N}(x_i)/n_-$ for every $x_i\in\mathcal{S}$. Considering that Jensen-Shannon divergence is a symmetry and bounded measures to quantify the divergence degree between two probability distributions, we define the overlapping measures between $\mathcal{P}$ and $\mathcal{Q}$ as the complement of $\emph{J}(\mathcal{P}||\mathcal{N})$, that is,
\begin{align}\label{eq_overlap}
\begin{aligned}
    D_{\mathcal{S}} =&1- \emph{J}(\mathcal{P}||\mathcal{N})  \\
   =&1- \frac{1}{2}\Big(\text{KL} (\mathcal{P}||\mathcal{M}) + \text{KL} (\mathcal{N}||\mathcal{M})\Big)\\
   =&1-\frac{1}{2} \left( \sum\limits_{x_i\in\mathcal{S}} \hat{p}(x_i) \log_2\left(\frac{2\hat{p}(x_i)}{\hat{p}(x_i) + \hat{n}(x_i)}\right) + \hat{n}(x_i)\log_2\left(\frac{2\hat{n}(x_i)}{\hat{p}(x_i) + \hat{n}(x_i)}\right) \right)\\
   =& -\frac{1}{2} \left( \sum\limits_{x_i\in\mathcal{S}} \hat{p}(x_i) \log_2\left(\frac{\hat{p}(x_i)}{\hat{p}(x_i) + \hat{n}(x_i)}\right) + \hat{n}(x_i)\log_2\left(\frac{\hat{n}(x_i)}{\hat{p}(x_i) + \hat{n}(x_i)}\right) \right)
\end{aligned}
\end{align}
where $\mathcal{M}=\frac{1}{2}\Big(\mathcal{P}+\mathcal{N}\Big)$ is a mixture distribution of $\mathcal{P}$ and $\mathcal{N}$, and $\text{KL}(\mathcal{P}||\mathcal{N})$ is the Kullback–Leibler divergence of any two distributions $\mathcal{P}$ and $\mathcal{N}$. Here, $D_{\mathcal{S}}$ is also bounded by zero and one, in particular, $D_{\mathcal{S}}=0$ means that they are completely separated;  $D_{\mathcal{S}}=1$ means that they are totally overlapped. Therefore, the specific value of $D_{\mathcal{S}}\in[0,1]$ quantitatively characterizes the degree of overlap between $\mathcal{P}$ and $\mathcal{N}$. 

Next, we plan to discover the quantitative relationship between $D_{\mathcal{S}}$ and $\text{AR}^u$ through describing the fluctuation of $\text{AR}^u$ given a fixed overlapping $D_{\mathcal{S}}$. At first, we define $\text{AR}^u_{\max}(D_{\mathcal{S}})$ and $\text{AR}^u_{\min}(D_{\mathcal{S}})$ as the maximum and minimum values of upper bound of AUC ($\text{AR}^u$) of dataset $\mathcal{S}$ with overlapping $D_{\mathcal{S}}$, respectively. In other words, $\text{AR}^u_{\max}(D_{\mathcal{S}})$ and $\text{AR}^u_{\min}(D_{\mathcal{S}})$ can be obtained through solving the two following optimization problems:
\begin{align}\label{eq_min}
\begin{aligned}
    \min& \quad \sum\limits_{x_i,x_j\in\mathcal{S}}\max\Big\{\hat{p}(x_i)\hat{n}(x_j),\hat{n}(x_i)\hat{p}(x_j) \Big\} \\
    \text{s.t.}& \quad \sum\limits_{x_i\in\mathcal{S}} \hat{p}(x_i) \log_2\left(\frac{\hat{p}(x_i)}{\hat{p}(x_i) + \hat{n}(x_i)}\right) + \hat{n}(x_i)\log_2\left(\frac{\hat{n}(x_i)}{\hat{p}(x_i) + \hat{n}(x_i)}\right)+2D_{\mathcal{S}}=0 \\
    &\quad \sum\limits_{x_i\in\mathcal{S}} \hat{p}(x_i) = 1 \\
    &\quad \sum\limits_{x_i\in\mathcal{S}} \hat{n}(x_i) = 1 \\
    &\quad \hat{p}(x_i)\geq 0 \quad i=1,2,\cdots,m\\
    &\quad\hat{n}(x_i)\geq 0  \quad i=1,2,\cdots,m\\
\end{aligned}
\end{align}
and
\begin{align}\label{eq_max}
\begin{aligned}
    \max& \quad \sum\limits_{x_i,x_j\in\mathcal{S}}\max\Big\{\hat{p}(x_i)\hat{n}(x_j),\hat{n}(x_i)\hat{p}(x_j) \Big\} \\
    \text{s.t.}& \quad \sum\limits_{x_i\in\mathcal{S}} \hat{p}(x_i) \log_2\left(\frac{\hat{p}(x_i)}{\hat{p}(x_i) + \hat{n}(x_i)}\right) + \hat{n}(x_i)\log_2\left(\frac{\hat{n}(x_i)}{\hat{p}(x_i) + \hat{n}(x_i)}\right)+2D_{\mathcal{S}}=0 \\
    &\quad \sum\limits_{x_i\in\mathcal{S}} \hat{p}(x_i) = 1 \\
    &\quad \sum\limits_{x_i\in\mathcal{S}} \hat{n}(x_i) = 1 \\
    &\quad \hat{p}(x_i)\geq 0 \quad i=1,2,\cdots,m\\
    &\quad\hat{n}(x_i)\geq 0  \quad i=1,2,\cdots,m\\
\end{aligned}
\end{align}

Now, we observed an interesting phenomenon that both $\text{AR}^u$ and $D_{\mathcal{S}}$ remain unchanged if we swap the values of $\Big(\hat{p}(x_i), \hat{n}(x_i)\Big)$ and $\Big(\hat{p}(x_j), \hat{n}(x_j)\Big)$. Considering the computational process of $\text{AR}^u$, we assume that $\hat{p}(x_1) / \hat{n}(x_1) \leq \hat{p}(x_2)/\hat{n}(x_2) \leq \cdots \leq \hat{p}(x_m)/\hat{n}(x_m)$ without loss of generality. Under this assumption, the $\text{AR}^u$ can be simplified as
\begin{align}
    \text{AR}^u = \sum\limits_{i=1}^m\sum\limits_{j=1}^{i-1} \hat{p}(x_i)\hat{n}(x_j) + \frac{1}{2}\sum\limits_{i=1}^m \hat{p}(x_i)\hat{n}(x_i).
\end{align}
Combined with the above assumption and equation, Eq.~\ref{eq_min} and Eq.~\ref{eq_max} can be rewritten as
\begin{align}\label{eq_min1}
\begin{aligned}
    \min& \quad \sum\limits_{i=1}^m\sum\limits_{j=1}^{i-1} \hat{p}(x_i)\hat{n}(x_j) + \frac{1}{2}\sum\limits_{i=1}^m \hat{p}(x_i)\hat{n}(x_i) \\
    \text{s.t.}& \quad \sum\limits_{i=1}^m \hat{p}(x_i) \log_2\left(\frac{\hat{p}(x_i)}{\hat{p}(x_i) + \hat{n}(x_i)}\right) + \hat{n}(x_i)\log_2\left(\frac{\hat{n}(x_i)}{\hat{p}(x_i) + \hat{n}(x_i)}\right)+2D_{\mathcal{S}}=0 \\
    &\quad \hat{p}(x_1) / \hat{n}(x_1) \leq \hat{p}(x_2)/\hat{n}(x_2) \leq \cdots \leq \hat{p}(x_m)/\hat{n}(x_m) \\
    &\quad \sum\limits_{i=1}^m \hat{p}(x_i) = 1 \\
    &\quad \sum\limits_{i=1}^m \hat{n}(x_i) = 1 \\
    &\quad \hat{p}(x_i)\geq 0 \quad i=1,2,\cdots,m\\
    &\quad \hat{n}(x_i) \geq 0 \quad i=1,2,\cdots,m 
\end{aligned}
\end{align}
and
\begin{align}\label{eq_max1}
\begin{aligned}
    \max& \quad \sum\limits_{i=1}^m\sum\limits_{j=1}^{i-1} \hat{p}(x_i)\hat{n}(x_j) + \frac{1}{2}\sum\limits_{i=1}^m \hat{p}(x_i)\hat{n}(x_i) \\
    \text{s.t.}& \quad \sum\limits_{i=1}^m \hat{p}(x_i) \log_2\left(\frac{\hat{p}(x_i)}{\hat{p}(x_i) + \hat{n}(x_i)}\right) + \hat{n}(x_i)\log_2\left(\frac{\hat{n}(x_i)}{\hat{p}(x_i) + \hat{n}(x_i)}\right)+2D_{\mathcal{S}}=0 \\
    &\quad \hat{p}(x_1) / \hat{n}(x_1) \leq \hat{p}(x_2)/\hat{n}(x_2) \leq \cdots \leq \hat{p}(x_m)/\hat{n}(x_m) \\
    &\quad \sum\limits_{i=1}^m \hat{p}(x_i) = 1 \\
    &\quad \sum\limits_{i=1}^m \hat{n}(x_i) = 1 \\
    &\quad \hat{p}(x_i)\geq 0 \quad i=1,2,\cdots,m \\
 & \quad\hat{n}(x_i) \geq 0 \quad i=1,2,\cdots,m 
\end{aligned}
\end{align}
respectively.  According to the symmetry of the ranking assumption
\[
\frac{\hat{p}(x_1)}{\hat{n}(x_1)} \leq \frac{\hat{p}(x_2)}{\hat{n}(x_2)} \leq \cdots \leq \frac{\hat{p}(x_m)}{\hat{n}(x_m)},
\]
it follows that the feasible region is partitioned into \(m!\) pairwise symmetric sub-regions. Within each sub-region, the mathematical formulation of \(\text{AR}^u_{\max}(D_{\mathcal{S}})\) is invariant and possesses an identical maximum value. Consequently, we can disregard the ranking assumption in Eq.~\eqref{eq_max1} and deduce a simplified version as follows:
\begin{align}\label{eq_max2}
\begin{aligned}
    \max& \quad \sum_{i=1}^m\sum_{j=1}^{i-1} \hat{p}(x_i)\hat{n}(x_j) + \frac{1}{2}\sum_{i=1}^m \hat{p}(x_i)\hat{n}(x_i), \\
    \text{\text{s.t.}}& \quad \sum_{i=1}^m \hat{p}(x_i) \log_2\left(\frac{\hat{p}(x_i)}{\hat{p}(x_i) + \hat{n}(x_i)}\right) + \hat{n}(x_i)\log_2\left(\frac{\hat{n}(x_i)}{\hat{p}(x_i) + \hat{n}(x_i)}\right)+2D_{\mathcal{S}}=0, \\
    &\quad \sum_{i=1}^m \hat{p}(x_i) = 1, \\
    &\quad \sum_{i=1}^m \hat{n}(x_i) = 1, \\
    &\quad \hat{p}(x_i)\geq 0, \quad i=1,2,\cdots,m, \\
 &\quad \hat{n}(x_i) \geq 0, \quad i=1,2,\cdots,m.
\end{aligned}
\end{align}

We employ the SLSQP solver~\cite{kraft1988software} to resolve the optimization problem outlined in~\eqref{eq_max2} and acquire the corresponding numerical solution for \(\text{AR}^u_{\max}\). In Fig.~\ref{figS8}A, the \(\text{AR}^u\) curve versus \(D_{\mathcal{S}}\) is depicted. Each datum point on the curve represents the numerical solution of the optimization problem under specific parameters. Notably, the curve converges swiftly as \(m\) increases, and attains the optimal \(\text{AR}^u_{\max}\) curve when \(m \geq 10\).

Regarding the \(\text{AR}^u_{\min}\) curve, a heuristic approach is utilized to construct its optimal solution from the feasible region's boundaries. Specifically, we examine a particular case where \(\hat{p}(x_1)=1-b\), \(\hat{p}(x_2)=b\), \(\hat{n}(x_1)=0\), and \(\hat{n}(x_2)=1\), with \(b\) being a tunable parameter in the interval \([0,1]\). In this scenario, the \(\text{AR}^u\) is expressed as
\begin{align}
    \text{AR}^u = 1-\frac{b}{2},
\end{align}
and \(D_{\mathcal{S}}\) is given by
\begin{align}\label{eq_heuristic}
    D_{\mathcal{S}}=&-\frac{1}{2}\left(b\log_2 \frac{b}{b+1} + \log_2 \frac{1}{b+1}\right) \nonumber\\
    =&-\frac{1}{2}\left(b\log_2 b -(b+1)\log_2 (b+1) \right).
\end{align}
Furthermore, the numerical curve of $\text{AR}^u_{\min}$ obtained through the SLSQP solver aligns closely with this heuristic solution as given in~\eqref{eq_heuristic}, providing partial validation for our heuristic approach (see Fig.~\ref{figS8}B).

\section{Feature Engineering}

In previous sections, we have discussed the boundaries of the objective function and evaluation metrics from the perspective of row data (feature vectors). In fact, column data (features) can also influence the degree of overlap and boundaries in a dataset through their impact on row data. In feature engineering, there are two classic methods of handling column data: feature selection and feature extraction. The former emphasizes adding or removing new features unrelated to existing ones, and the latter is based on extraction and mapping of original features. In this section, we will discuss these two methods separately. But before that, we need to discuss the simplest case first.

Suppose we have an original dataset $\mathcal{S}=\{(x_i,y_i): i =1,2,\cdots,m\}$ with $k$ features. This implies that the feature vector $x_i$ can be expressed as $x_i=(x_{i,1},x_{i,2},\cdots,x_{i,k})$. We also define that there are $\mathcal{P}(x_i)$ positive instances and $\mathcal{N}(x_i)$ negative instances with feature vector $x_i$ in the entire dataset. By introducing a new feature into every instance, we can create a new dataset $\mathcal{S}'$. Using feature vector $x_i$ as an example, these $\mathcal{P}(x_i)+\mathcal{N}(x_i)$ could be added into different feature values, which are included in $\{x_i^s=(x_{i,1}, x_{i,2}, \cdots, x_{i, k}, x^s_{i,k+1}): s=1,2,\cdots,s_i\}$. For the sake of argument, we illustrate that the original feature vector is split into $s_i$ pairwise distinct feature vectors after the addition of one row data, satisfying that
\begin{align}
    \left\{
    \begin{aligned}
        \sum\limits_{j=1}^{s_i} \mathcal{P}(x_i^j)=\mathcal{P}(x_i) \\
        \sum\limits_{j=1}^{s_i} \mathcal{N}(x_i^j)=\mathcal{N}(x_i)
    \end{aligned}
    \right.,
\end{align}
in which $s_i$ is defined as the diversity for $x_i$. Consequently, we will proceed to prove the following lemma.
\begin{lemma}\label{lemma1}
Upon the inclusion of a new feature into the original dataset, $\text{AR}^u$ will either increase or remain constant, while $D_{\mathcal{S}}$ will either decrease or stay the same.
These values will remain unchanged if, and only if, the diversity $s_i=1$ for each $x_i$.
\end{lemma}
\begin{proof}
    The upper bound of AUC in the original dataset $\mathcal{S}$
    \begin{align}
        \text{AR}^u_{original} = \frac{1}{2n_-n_+} \sum\limits_{i,j} \max\{\mathcal{P}(x_i)\mathcal{N}(x_j), \mathcal{P}(x_j)\mathcal{N}(x_i)\},
    \end{align}
    and the new boundary is
    \begin{align}
    \begin{aligned}
        \text{AR}^u_{new}=&\frac{1}{2n_-n_+} \sum\limits_{i,j} \sum\limits_{i'=1}^{s_i} \sum\limits_{j'=1}^{s_j} \max\{\mathcal{P}(x_i^{i'})\mathcal{N}(x_j^{j'}), \mathcal{P}(x_j^{j'})\mathcal{N}(x_i^{i'})\} \\
        \geq & \frac{1}{2n_-n_+} \sum\limits_{i,j} \max\left\{\sum\limits_{i'=1}^{s_i}\mathcal{P}(x_i^{i'})\sum\limits_{j'=1}^{s_j}\mathcal{N}(x_j^{j'}), \sum\limits_{j'=1}^{s_j}\mathcal{P}(x_j^{j'})\sum\limits_{i'=1}^{s_i}\mathcal{N}(x_i^{i'})\right\} \\
        =& \frac{1}{2n_-n_+} \sum\limits_{i,j} \max\{\mathcal{P}(x_i)\mathcal{N}(x_j), \mathcal{P}(x_j)\mathcal{N}(x_i)\} \\
        =& \text{AR}^u_{original}
        \end{aligned}.
    \end{align}
Similarly, the original $D_{\mathcal{S}}$ can be written as
\begin{align}
    D^{original}_{\mathcal{S}}=-\frac{1}{2} \left( \sum\limits_{x_i\in\mathcal{S}} \hat{p}(x_i) \log_2\left(\frac{\hat{p}(x_i)}{\hat{p}(x_i) + \hat{n}(x_i)}\right) + \hat{n}(x_i)\log_2\left(\frac{\hat{n}(x_i)}{\hat{p}(x_i) + \hat{n}(x_i)}\right) \right)
\end{align}
and the new overlapping is
\begin{align}
\begin{aligned}
    D^{new}_{\mathcal{S}}=&-\frac{1}{2} \left( \sum\limits_{x_i\in\mathcal{S}} \sum\limits_{i'=1}^{s_i} \hat{p}(x_i^{i'}) \log_2\left(\frac{\hat{p}(x_i^{i'})}{\hat{p}(x_i^{i'}) + \hat{n}(x_i^{i'})}\right) + \hat{n}(x_i^{i'})\log_2\left(\frac{\hat{n}(x_i^{i'})}{\hat{p}(x_i^{i'}) + \hat{n}(x_i^{i'})}\right) \right) \\
    \leq & -\frac{1}{2} \left( \sum\limits_{x_i\in\mathcal{S}} \sum\limits_{i'=1}^{s_i} \hat{p}(x_i^{i'}) \log_2\left(\frac{\sum\limits_{i'=1}^{s_i}\hat{p}(x_i^{i'})}{\sum\limits_{i'=1}^{s_i}\hat{p}(x_i^{i'}) + \hat{n}(x_i^{i'})}\right) + \sum\limits_{i'=1}^{s_i}\hat{n}(x_i^{i'})\log_2\left(\frac{\sum\limits_{i'=1}^{s_i}\hat{n}(x_i^{i'})}{\sum\limits_{i'=1}^{s_i}\hat{p}(x_i^{i'}) + \hat{n}(x_i^{i'})}\right) \right) \\
    =&-\frac{1}{2} \left( \sum\limits_{x_i\in\mathcal{S}} \hat{p}(x_i) \log_2\left(\frac{\hat{p}(x_i)}{\hat{p}(x_i) + \hat{n}(x_i)}\right) + \hat{n}(x_i)\log_2\left(\frac{\hat{n}(x_i)}{\hat{p}(x_i) + \hat{n}(x_i)}\right) \right) \\
    =&D^{original}_{\mathcal{S}}.
    \end{aligned}
\end{align}
They are equal to each other if and only if there exiSUD $i'$  satisfying that $\hat{p}(x_i^{i'})=\hat{p}(x_i)$ and $\hat{n}(x_i^{i'})=\hat{n}(x_i)$ for every $x_i$, i.e., $s_i=1$. 
\end{proof}
Actually, adding new features will cause the overlapping of the positive and negative samples of the dataset to decrease or remain unchanged, while reducing the original features will cause the overlapping to increase or remain unchanged. The same conclusion is also applicable to the boundaries of various evaluation indicators and loss functions, such as $\text{AR}^u$, $\text{AP}^u$ and $\text{AC}^u$.

\subsection{Feature Selection}

Feature selection is the process of selecting a subset of relevant features (variables, predictors) for use in model construction. A feature selection algorithm can be seen as the combination of a search technique for proposing new feature subsets, along with an evaluation measure which scores the different feature subsets. The simplest algorithm is to test each possible subset of features finding the one which minimizes the error rate to reach the best performance. Actually, the exponential number of potential subset selection and the huge amount of training cost for any classifier make this process difficult. However, the boundary theory we proposed can better select subset of all features with high performance measured by AR, AP (best ranking) and AC and low computational cost (time complexity). 

Given an integer $k_0$, the optimal $k_0$ feature subset is targeted by $\text{AR}^u$. This implies that we can directly compute the $\text{AR}^u$ for each feature subset in the entire dataset $\mathcal{S}$ to assess the data quality and training potential, bypassing the need for a training process. This approach significantly conserves storage space and computational resources. We denote $\text{AR}^u_{k_0}$ and $D^{k_0}_{\mathcal{S}}$ as the optimal $\text{AR}^u$ and $D_{\mathcal{S}}$, respectively, when we traverse all possible $k_0$ feature subsets. The corresponding feature subset is named by \emph{optimal feature subset}. Based on Lemma~\ref{lemma1} and the principle of recursion, it is evident that
\begin{theorem}\label{th1}
$\text{AR}^u_{k_0}$ exhibits a monotonic non-decreasing trend and $D^{k_0}_{\mathcal{S}}$ is monotonic non-increasing when $k_0\in\{1,2,\cdots,k\}$. Here, $k$ represents the total number of features in the entire dataset.
\end{theorem}
\begin{proof}
   Based on the principle of recursion, we only need to prove that $\text{AR}^u_{k_0} \leq \text{AR}^u_{k_0+1}$ and $D^{k_0}_{\mathcal{S}} \geq D^{k_0+1}_{\mathcal{S}}$ for any $k_0\in\{1,2,\cdots,k-1\}$.
    Assumed that $\mathcal{F}^*_{k_0}=\{f^*_1, f_2^*,\cdots,f_{k_0}^*\}$ is the optimal $k_0$-feature subset. Then we construct a new $k_0+1$-feature subset $\mathcal{F}_{k_0+1}=\mathcal{F}^*_{k_0}+\{f_i\}$, in which $f_i$ is any selected feature not belonging to $\mathcal{F}^*_{k_0}$. According to Lemma~\ref{lemma1}, we know that the $\text{AR}^u$ of $\mathcal{F}_{k_0+1}$ is higher than or equal to $\text{AR}^u_{k_0}$. At the same time, we also know that the $\text{AR}^u$ of $\mathcal{F}_{k_0+1}$ is lower than or equal to $\text{AR}^u_{k_0+1}$ since the definition of $\text{AR}^u_{k_0+1}$. Therefore, we successfully proved that $\text{AR}^u_{k_0} \leq \text{AR}^u_{k_0+1}$. In a similar way, we can also prove $D^{k_0}_{\mathcal{S}} \geq D^{k_0+1}_{\mathcal{S}}$ .
\end{proof}

Theorem~\ref{th1} elucidates the direct correlation between the performance bounds and the overlapping index within a given dataset. Specifically, it reveals that an increase in the number of features (raw data) leads to a reduction in the overlap between the positive and negative sample distributions, which in turn enhances the performance boundaries. This insight informs the design of a feature selection algorithm that leverages the overlapping index, $D_{\mathcal{S}}$, to ensure the achievement of the highest possible performance upper limit.

Consider a dataset $\mathcal{S}$ composed of $k$ features $\{F_1, F_2, \ldots, F_k\}$. The optimal feature subset of size $k_0$, denoted as the subset that minimizes $D_{\mathcal{S}}$ (or maximizes $\text{AR}^u$), can be defined where $k_0 = 1, 2, \ldots, k$. However, exhaustively evaluating all $\binom{k}{k_0}$ possible subsets may be computationally prohibitive for large $k$. To address this, approximation techniques like dynamic programming can be employed to devise an efficient approximation algorithm.

For a more practical example, consider the INE dataset with 13 features; it is possible to achieve the dataset's performance boundary using only 8 features, such that $D_{\mathcal{S}}^8 = D_{\mathcal{S}}$. This indicates that the remaining five features are, to some extent, superfluous. We thus define the optimal feature selection dimension $k^*$ as:
\begin{align}
k^* = \min \{ k_0 : D_{\mathcal{S}}^{k_0} = D_{\mathcal{S}} \}.
\end{align}
The feature subset corresponding to $k^*$ is referred to as the \emph{global optimal feature subset}.

\subsection{Feature Extraction}

Feature extraction is the procedure of deriving features (traits, properties, attributes) from raw data. It is seen as an equivalent transformation that creates new features from the original ones. In previous subsections, we deduced that the boundary of performance is dictated by the data structure. From a mathematical standpoint, feature extraction can be viewed as a mapping of original features (row data). If we incorporate new extracted data into the original dataset, the diversity for each feature vector is $1$. In conjunction with Lemma~\ref{lemma1}, we can state,
\begin{theorem}\label{th2}
The inclusion of extracted features into the original dataset does not alter the boundary of training loss, evaluation measures, and overlapping.
\end{theorem}
\begin{proof}
    Any feature extraction process can be regarded as a mapping from existing feature vectors, so the diversity is $1$. Combined with Lemma~\ref{lemma1}, the boundaries and overlapping should be unchanged.
\end{proof}

\subsection{Feature Generated from Other Samples}
Both aforementioned two processes involve manipulating each sample's original feature vector through operations such as selection, transformation, or extraction. Drawing an analogy to clustering algorithms in unsupervised learning, we here explore from the perspective of rows: augmenting samples' features with information derived from other samples, rather than itself, based on agreed rules. E.g. propose a new indicator of a sample,  neighbors' income, by counting all her/his neighborhoods within certain distance \cite{hedefalk2020social}. 

For each sample $x_i\in \mathcal{S}$, here we define its new feature vector is $x'_i=(x_{i,1},x_{i_2}, \cdots,x_{i,2k})$, in which $(x_{i,1},x_{i_2}, \cdots,x_{i,k})$ is its original $k$-feature vector and
\begin{align}
    (x_{i,k+1},x_{i_k+2}, \cdots,x_{i,2k}) = f(\Lambda_r(x_i)).
\end{align}
And, $\Lambda_r(x_i)$ includes all samples whose distance from $x_i$ is less than $r$  in the original feature space, and $f$ represents an arbitrary operator, such as a mean function. 

We then construct a new dataset $\mathcal{S}'_r=\{x'_1,x_2',\cdots, x_n'\}$ with a tunable parameter $r\in[0,\infty]$. Notably, $\mathcal{S}'_0$ is equivalent to $\mathcal{S}$. We can now state the following theorem:
\begin{theorem}
    The $AR^u$ of $\mathcal{S}'_r$ is always equal to $\mathcal{S}$ regardless of $r$.
\end{theorem}
\begin{proof}
    Based on Theorem~\ref{th1}, it is evident that the $AR^u$ of $\mathcal{S}'_r$ is at least as large as that of $\mathcal{S}$. Additionally, for any two samples with identical original feature vectors, their newly added features will also be identical. According to Lemma~\ref{lemma1}, we can say that the $AR^u$ of $\mathcal{S}'_r$ cannot exceed that of $\mathcal{S}$. Hence, the theorem is proven.
\end{proof}

In this section, we use the $AR^u$ measure to characterize the predictability of a given dataset, as demonstrated in Lemma~\ref{lemma1}, Theorem~\ref{th1}, and Theorem~\ref{th2}. Notably, other measures of predictability, such as $AP^u$ and $AC^u$, lead to the same conclusions. The equivalence of these different measures will be thoroughly explained in Section~\ref{sec3}.

\sloppy

\section{Datasets}

We utilized four datasets in main text and thirty-seven additional datasets in Supplemental Material from the Kaggle platform (https://www.kaggle.com). The specifics of four real-world datasets used in main text are as follows:
\begin{itemize}
    \item Airlines Delay Dataset (AID): omprised of 539,383 records across 8 distinct attributes, the objective is to forecast flight delays based on scheduled departure information. 
    \item Heart Disease Dataset (HED): This dataset encompasses a wide range of cardiovascular risk factors, including age, gender, height, weight, blood pressure, cholesterol and glucose levels, smoking status, alcohol intake, physical activity, and presence of cardiovascular diseases, from over 70,000 individuals. It serves as a valuable asset for applying advanced machine learning methods to investigate the link between these factors and cardiovascular health, which could enhance disease understanding and prevention strategies. 
    \item Income Classification Dataset (INE): This dataset features variables such as education, employment, and marital status to predict whether an individual earns more than \$50K annually. 
    \item Student Sleep Study Dataset (SUD): Originating from a survey-based analysis of US students' sleep patterns, this dataset utilizes factors like average sleep duration and phone usage time to infer adequate sleep among students. 
\end{itemize}

The specifics of additional 37 real-world datasets used in Supplemental Material are also as follows:
\begin{itemize}
    \item Adult Census Dataset (ACD): Given a set of demographic and employment attributes such as age, education level, occupation, and hours worked per week, the goal is to build a predictive model that classifies individuals into two categories: 'over threshold' and 'under threshold'. This task is a binary classification problem, where the model's output is a binary decision indicating whether an individual's income exceeds the predetermined threshold or not. 
    \item Android Malware Detection (AMD): This Kaggle dataset focuses on the classification of software applications into two categories : malware (1) and goodware (0). 
    \item ASD questionnairs (ASD): This Kaggle dataset is dedicated to predicting whether a patient has autism using questionnaire data on the topic of autism. 
    \item Asthma Disease Prediction (ADP): The ADP dataset is a comprehensive collection of anonymized health records and patient data, which includes vital patient information, environmental factors, and medical history, enabling the development of advanced machine learning models to forecast asthma treatment outcomes. 
    \item Branch Prediction (BP): Branch prediction is a technique used in CPU design that attempts to guess the outcome of a conditional operation and prepare for the most likely result. A digital circuit that performs this operation is known as a branch predictor. It is an important component of modern CPU architectures, such as the x86. The problem of branch prediction can be treated as a binary classification problem, where target feature will be Branch Taken/ Branch Not. 
    \item Cancer Prediction Dataset (CPDT): This dataset represents a synthetic collection of responses gathered from a university-conducted survey, aimed at studying the potential risk factors for lung cancer. The survey includes a variety of demographic, lifestyle, and health-related questions. This dataset is used to predict whether the respondent has been diagnosed with lung cancer. 
    \item Car Ownership Prediction (COP): The dataset contains information on the occupation, monthly income, credit score, years of employment, finance status, finance history, number of children. This dataset is used to predict whether each individual owns a car or not. 
    \item Cars Purchase Decision (CPDN): This dataset contains details of 1000 customers who intend to buy a car, considering their annual salaries. This dataset is used to predict whether each customer would buy a car. 
    \item Cardiovascular diseases dataset (CDD): This data set predicts whether a patient has cardiovascular disease based on information such as age, height, weight, and whether he or she smokes. 
    \item Contraceptive Method Choice (CMC): This dataset is a subset of the 1987 National Indonesia Contraceptive Prevalence Survey. The samples are married women who were either not pregnant or do not know if they were at the time of interview. This dataset is used to predict the current contraceptive method choice (no use or use). 
    \item College Attending Plan (CAP): The purpose of this dataset is to use the basic information of high school students to predict whether they plan to go to college, in particular, which factors can better distinguish high school students who are willing to go to college from those who are not. 
    \item Symptoms and COVID Presence (SCP):  The purpose of this dataset is to provide symptoms as input and it should be able to predict if COVID is possibly present or not. 
    \item Child Sexual Abuse Awareness Prediction (CSAAP): By asking a series of questions, this dataset predicts a person's level of knowledge about child sexual abuse. 
    \item Cyber Security (CS): This dataset wants to analyze whether the particular URL is prone to phishing (malicious) or not. 
    \item Diabetes Health Indicators (DHI): The Behavioral Risk Factor Surveillance System (BRFSS) is a health-related telephone survey that is collected annually by the CDC. Each year, the survey collects responses from over 400,000 Americans on health-related risk behaviors, chronic health conditions, and the use of preventative services. This dataset predicts whether a patient has diabetes based on their health indicators. 
    \item Diabetes Dataset (DD): This dataset was collected  to predict Type 2 diabetes. An article is also published implementing this dataset~\cite{tigga2020prediction}. 
    \item Employee Dataset (ED): This dataset contains information about employees in a company, including their educational backgrounds, work history, demographics, and employment-related factors. It is used to predict an employee will leave or not. 
    \item Fraud Detection Bank Dataset (FDBD): This dataset is used to predict whether transactions within the bank are fraudulent transactions.
    \item  Game of Thrones (GOT): This dataset is used to predict the next death on Game Of Thrones character. 
    \item Happiness Dataset (HD): This Dataset is based on a survey conducted where people rated different metrics of their city on a scale of 5 and answered if they are happy or unhappy. 
    \item Heart Disease Nowadays (HDN): Annual CDC survey data of 400k adults related to their health status can be used to predict whether an adult is diagnosed as heart disease. 
    \item Heart Disease Health Indicators Dataset (HDHID): This dataset is also cellected from BRFSS survey and used to predict whether an American has heart disease. 
    \item Immigration Madrid (IM): This dataset comes from human resources data from Spain and other countries and is used to predict whether individuals are hired. 
    \item Loan: This dataset contains basic details and loan history from the last 3 months about customers, which is used to predict whether a customer will take out a loan in the future. 
    \item Naive Bayes Classification Data (NBCD): The dataset includes blood glucose and blood pressure data that can be used to classify whether a patient has diabetes. 
    \item Basketball Players' Career Duration (BPCD): The data consists of performance statistics from each player's rookie year. The target variable is a Boolean value that indicates whether a given player will last in the league for five years. 
    \item Phishing Website Detector (PWD): This dataset is used to detect whether a website is a phishing website. 
    \item QSAR Androgen Receptor Dataset (QARD): This dataset was used to develop classification QSAR models for the discrimination of binder/positive (199) and non-binder/negative (1488) molecules by means of different machine learning methods.
    \item Simplified Titanic Dataset (STD): This dataset is a simplified version of the famous Titanic dataset, which contains information about passengers aboard the Titanic ship. It is designed specifically for beginners who are learning about data analysis and classification problems. 
    \item Ad Click Prediction (ACP): This dataset is used to predict whether customer will click Ad and make a purchase. 
    \item Happy or Not (HON): The dataset provides information on various aspects such as housing costs, education quality, transportation facilities, security, healthcare availability, quality of public services, food quality, events, and happiness levels. We use this dataset to predict an individual is happy or not.
    \item Telecom Service Customer (TSC): This dataset predicts whether a customer will continue based on their basic information. 
    \item Blood Transfusion Dataset (BTD): A previous study~\cite{yeh2009knowledge} adopted the donor database of Blood Transfusion Service Center in Hsin-Chu City in Taiwan. The center passes their blood transfusion service bus to one university in Hsin-Chu City to gather blood donated about every three months. To build a FRMTC model, we selected 748 donors at random from the donor database. These 748 donor data, each one included R (Recency - months since last donation), F (Frequency - total number of donation), M (Monetary - total blood donated in c.c.), T (Time - months since first donation), and a binary variable representing whether he/she donated blood in March 2007 (1 stand for donating blood; 0 stands for not donating blood). 
    \item TUNADROMD Malware Detection (TMD): This data set is used to identify whether a piece of software is malware~\cite{borah2021cost}. 
    \item Vehicle Stolen Dataset (VSD): This dataset is used to predict whether a vehicle will be stolen. 
    \item Web Club Recruitment (WCR): Dataset for recruitment contest for intelligence group in Web Club NITK. 
    \item Wine Quality Dataset (WQD): The dataset contains the wine features or ingredients with the quality and the type of wines. So, the data can be used for the prediction of Wine Quality as well as the detection of the type of wines from ingredient analysis. 
\end{itemize}

These datasets are also available for download from ``The Boundary Theory of Binary Classification'' on GitHub (https://github.com/Feijing92/The-boundary-theory-of-binary-classification). Their statistical properties are detailed in Table~\ref{tabS1}.

\bibliographystyle{apsrev4-2}
\bibliography{aps-supp}

\section*{Supplementary Tables and Figures}

\begin{table*}[h]
\caption{Description of 41 real datasets we used in this Letter, including 4 datasets we used in main text and additional 37 datasets we used in Supplemental Materials. It includes the number of instances ($m$), the number of positive instances ($n_+$), the number of negative instances ($n_-$), the number of features ($k$), the overlapping index ($D_{\mathcal{S}}$) and the optimal feature selection dimension ($k^*$). }\label{tabS1}

\begin{ruledtabular}
\begin{tabular}{lrrrrrr}
Name &  $m$ & $n_+$ &$n_-$& $k$ & $D_{\mathcal{S}}$ & $k^*$ \\
\hline
AID&539383&299119&240264&7& 0.4837&5\\
HED&70000&34979&35021&11&0.0015& 5\\
INE&32561&7841&24720&13&0.0657&12 \\
SUD&104&36&68&5&0.3181& 5\\
\hline
ACD&48842&11687&37155&13&0.0799&12\\
AMD&4862&1098&3764&148&0.031&46\\
ASD&3743&1752&1991&17&0.006&16\\
ADP&316800&237600&79200&18&0.5409&2\\
BP&400000&64162&335838&480&0.0391&49\\
CPD&1000&776&224&10&0.0022&7\\
COP&482&181&301&7&0.0084&4\\
CPD&68783&34041&34742&11&0.0645&11\\
CDD&1000&598&402&3&0.0101&3\\
CMC&1472&628&844&9&0.0491&9\\
CAP&8000&5404&2596&4&0.0057&4\\
SCP&5434&4383&1051&20&0.049&13\\
CSAAP&3002&1291&1711&8&0.2444&8\\
CS&11054&6157&4897&30&0.028&23\\
DHI&253680&39977&213703&21&0.0252&21\\
DD&951&685&266&17&0.0737&8\\
ED&4652&1600&3052&8&0.236&8\\
FDBD&20467&5437&15030&112&0.0004&33\\
GOT&1946&495&1451&2&0.5272&2\\
HD&143&77&66&6&0.1197&6\\
HDN&319795&292422&27373&17&0.0135&17\\
HDHID&253680&23893&229787&21&0.0179&21\\
IM&1523&1417&106&5&0.4956&5\\
LOAN&69713&1020&68693&20&0.0007&9\\
NBCD&995&497&498&2&0.1971&2\\
BPCD&1340&509&831&19&0.0333&3\\
PWD&11054&6157&4897&30&0.028&23\\
QARD&1687&199&1488&1024&0.0015&37\\
STD&2240&1662&578&3&0.6963&3\\
ACP&400&143&257&3&0.0051&3\\
HON&143&66&77&8&0.014&7\\
TSC&7043&1869&5174&19&0.0062&10\\
BTD&748&178&570&4&0.2405&3\\
TMD&4464&899&3565&241&0.0126&47\\
VSD&20&13&7&3&0.1026&3\\
WCR&20000&4388&15612&23&0.0007&12\\
WQD&32485&24453&8032&12&0.0006&3\\

\end{tabular}
\end{ruledtabular}
\end{table*}

\begin{table*}
\caption{Comprehensive experimental results and theoretical upper bound analysis of the area under the ROC curve (AUC-ROC, abbreviated as AR) for various binary classifiers on additional 37 real datasets. The classifiers evaluated include XGBoost, Multilayer Perceptron (MLP), Support Vector Machine (SVM), Logistic Regression (LR), Decision Tree (DT), Random Forest (RF), K-Nearest Neighbors (KNN), and Naive Bayes (NB).}\label{tabS2}

\begin{ruledtabular}
\begin{tabular}{lcccccccccccc}
\multirow{2}*{Data} &  \multicolumn{4}{c}{XGBoost}& \multirow{2}*{MLP}& \multirow{2}*{SVM}& \multirow{2}*{LR}& \multirow{2}*{DT}& \multirow{2}*{RF}& \multirow{2}*{KNN}& \multirow{2}*{NB}&\multirow{2}*{Boundaries}\\
\cline{2-5}
~& Square&Logistic& Hinge&Softmax&~&~&~&~&~&~&~&~\\
\hline
ACD&0.9229&0.921&0.738&0.7779&0.7386&0.5869&0.6155&0.9664&0.9565&0.7863&0.6891&0.9975\\
AMD&0.9978&0.9976&0.9907&0.9881&0.994&0.984&0.9643&0.9946&0.9946&0.9795&0.5239&0.9994\\
ASD&0.999&0.999&0.9917&0.9852&0.9881&0.8538&0.815&0.9979&0.9977&0.9345&0.8196&1.0\\
ADP&0.8333&0.8333&0.8333&0.5&0.538&0.5&0.5&0.5&0.6712&0.666&0.8333&0.8333\\
BP&0.9989&0.9989&0.9927&0.9926&0.9926&0.9917&0.9914&0.9937&0.9936&0.985&0.8907&0.9992\\
CPDT&0.9939&0.9753&0.8282&0.8854&0.6006&0.5&0.5907&0.9978&0.9978&0.6343&0.6037&1.0\\
COP&0.9997&0.9981&0.9945&0.9834&0.7502&0.5751&0.575&0.9967&0.9956&0.7711&0.6236&1.0\\
CPDN&0.809&0.8078&0.7298&0.7381&0.6742&0.6119&0.6001&0.9728&0.9726&0.7565&0.5968&0.9984\\
CDD&0.9859&0.9587&0.849&0.8887&0.5533&0.5&0.5&0.9938&0.9946&0.7428&0.5352&0.9999\\
CMC&0.9471&0.9014&0.7847&0.8254&0.6792&0.582&0.5843&0.9794&0.9762&0.7228&0.5935&0.999\\
CAP&0.9287&0.9108&0.8259&0.8441&0.7892&0.5&0.7889&0.9958&0.9963&0.7074&0.7889&1.0\\
SCP&0.9986&0.9986&0.9755&0.9755&0.9813&0.9755&0.9442&0.9755&0.9755&0.9802&0.8491&0.9986\\
CSAAP&0.9788&0.9776&0.9353&0.9344&0.9214&0.9276&0.8588&0.9361&0.9353&0.9179&0.8196&0.9797\\
CS&0.995&0.9946&0.9635&0.9585&0.9839&0.9554&0.9227&0.989&0.9894&0.975&0.6425&0.9997\\
DHI&0.8237&0.8228&0.5137&0.5711&0.5969&0.5&0.5439&0.9881&0.9787&0.6286&0.6667&0.9997\\
DD&0.9978&0.9975&0.9678&0.9678&0.9454&0.8388&0.8499&0.9609&0.9632&0.9513&0.8366&0.9978\\
ED&0.9349&0.9241&0.8296&0.8369&0.7897&0.586&0.623&0.9131&0.9024&0.787&0.6421&0.9806\\
FDBD&0.9816&0.9812&0.9178&0.9208&0.5692&0.5&0.7658&0.9999&0.9997&0.7794&0.7962&1.0\\
GOT&0.8413&0.8041&0.602&0.5955&0.5&0.5&0.5&0.7287&0.7061&0.6181&0.5&0.895\\
HD&0.994&0.964&0.9372&0.9426&0.7684&0.7392&0.645&0.9481&0.9481&0.6948&0.6429&0.9942\\
HDN&0.8469&0.8471&0.5319&0.543&0.5281&0.5&0.5022&0.9791&0.9804&0.5494&0.6521&0.9999\\
HDHID&0.8554&0.8549&0.5199&0.5494&0.5322&0.5&0.5423&0.9892&0.9787&0.6003&0.685&0.9999\\
IM&0.8859&0.8552&0.5138&0.5&0.5&0.5&0.5&0.565&0.5693&0.5251&0.5&0.9052\\
LOAN&0.8981&0.8938&0.5147&0.5093&0.5051&0.5&0.5&0.9994&0.9941&0.5005&0.5132&1.0\\
NBCD&0.9856&0.9822&0.9417&0.9397&0.6895&0.7165&0.5729&0.9427&0.9427&0.9246&0.591&0.9869\\
BPCD&0.9962&0.9874&0.9503&0.9571&0.7495&0.5&0.5299&0.9868&0.9823&0.6698&0.6055&0.9995\\
PWD&0.9956&0.9954&0.962&0.9614&0.9795&0.9574&0.911&0.989&0.9893&0.9731&0.628&0.9997\\
QARD&1.0&0.9992&0.9667&0.9796&0.9975&0.8005&0.9925&0.9997&0.9997&0.7136&0.8218&1.0\\
STD&0.8385&0.8329&0.7036&0.687&0.6402&0.5894&0.5932&0.6857&0.6857&0.6474&0.5997&0.8396\\
ACP&0.9977&0.9772&0.9507&0.9246&0.5147&0.5&0.5128&0.9981&0.9965&0.6918&0.5174&1.0\\
HON&0.9999&0.9979&0.9924&0.9924&0.8431&0.7511&0.605&0.9935&0.9924&0.7273&0.7013&0.9999\\
TSC&0.9296&0.9138&0.7314&0.7823&0.5407&0.5&0.6364&0.9977&0.9962&0.6511&0.7484&1.0\\
BTD&0.9413&0.9255&0.802&0.8172&0.5314&0.5&0.5&0.8954&0.8722&0.6268&0.5&0.9796\\
TMD&0.9998&0.9997&0.9954&0.996&0.9837&0.9849&0.986&0.9965&0.9965&0.9848&0.5284&0.9999\\
VSD&0.9945&0.8022&0.9615&0.8187&0.7088&0.8187&0.7088&0.9286&0.9615&0.8132&0.7088&0.9945\\
WCR&0.8485&0.8529&0.6914&0.6999&0.5872&0.5&0.5973&0.9998&0.9996&0.6138&0.6835&1.0\\
WQD&0.9962&0.9922&0.9307&0.9248&0.6951&0.5&0.5084&0.9998&0.9998&0.9884&0.5227&1.0\\

\end{tabular}
\end{ruledtabular}
\end{table*}

\begin{table*}
\caption{Comprehensive experimental results and theoretical upper bound analysis of the area under the PR curve (AUC-PR, abbreviated as AP) for various binary classifiers on additional 37 real datasets. The classifiers evaluated include XGBoost, Multilayer Perceptron (MLP), Support Vector Machine (SVM), Logistic Regression (LR), Decision Tree (DT), Random Forest (RF), K-Nearest Neighbors (KNN), and Naive Bayes (NB).}\label{tabS3}

\begin{ruledtabular}
\begin{tabular}{lcccccccccccc}
\multirow{2}*{Data} &  \multicolumn{4}{c}{XGBoost}& \multirow{2}*{MLP}& \multirow{2}*{SVM}& \multirow{2}*{LR}& \multirow{2}*{DT}& \multirow{2}*{RF}& \multirow{2}*{KNN}& \multirow{2}*{NB}&\multirow{2}*{Boundaries}\\
\cline{2-5}
~& Square&Logistic& Hinge&Softmax&~&~&~&~&~&~&~&~\\
\hline
ACD&0.9731&0.9727&0.859&0.8783&0.8597&0.7938&0.8054&0.9808&0.9742&0.883&0.8373&0.999\\
AMD&0.9993&0.9992&0.9954&0.9941&0.9972&0.9919&0.9807&0.9975&0.9975&0.9892&0.7845&0.9997\\
ASD&0.9991&0.9991&0.9872&0.9783&0.9816&0.836&0.7818&0.9971&0.996&0.9136&0.7838&1.0\\
ADP&0.5&0.5&0.5&0.25&0.2785&0.25&0.25&0.25&0.3784&0.3745&0.5&0.5\\
BP&0.9997&0.9997&0.9974&0.9974&0.9974&0.997&0.9969&0.9978&0.9977&0.9945&0.9603&0.9998\\
CPDT&0.9826&0.9223&0.649&0.7754&0.3232&0.224&0.3124&0.9965&0.9965&0.3643&0.3192&1.0\\
COP&0.9998&0.9988&0.9934&0.9805&0.7733&0.6619&0.6621&0.9975&0.9954&0.7903&0.6892&1.0\\
CPDN&0.7966&0.7947&0.6638&0.6741&0.6155&0.5697&0.5629&0.9652&0.9566&0.6941&0.5599&0.9979\\
CDD&0.9817&0.9407&0.7964&0.8365&0.4473&0.402&0.402&0.9926&0.9906&0.6218&0.4393&0.9998\\
CMC&0.9537&0.9101&0.7587&0.7978&0.6797&0.6167&0.6183&0.9785&0.9683&0.7126&0.6238&0.9988\\
CAP&0.8798&0.8392&0.6079&0.6661&0.5479&0.3244&0.5467&0.9943&0.9937&0.5216&0.5467&0.9999\\
SCP&0.9936&0.9937&0.9203&0.9203&0.9191&0.9203&0.8707&0.9203&0.9203&0.9109&0.4427&0.9933\\
CSAAP&0.9843&0.9833&0.9281&0.9267&0.9123&0.9158&0.8555&0.9307&0.9281&0.9098&0.8091&0.9849\\
CS&0.9941&0.9936&0.9366&0.9362&0.9748&0.9307&0.8809&0.9843&0.9827&0.9641&0.5266&0.9996\\
DHI&0.9597&0.9598&0.8461&0.8617&0.869&0.8424&0.8542&0.9957&0.9921&0.878&0.8897&0.9999\\
DD&0.9944&0.9937&0.9329&0.9329&0.8886&0.6882&0.691&0.9384&0.9365&0.9043&0.655&0.9927\\
ED&0.9587&0.9508&0.8495&0.856&0.822&0.6975&0.7173&0.9217&0.9095&0.8198&0.7291&0.9885\\
FDBD&0.9927&0.9926&0.9457&0.9478&0.7624&0.7344&0.8569&0.9999&0.9998&0.864&0.8817&1.0\\
GOT&0.9337&0.9156&0.7864&0.7837&0.7456&0.7456&0.7456&0.8445&0.8332&0.7936&0.7456&0.958\\
HD&0.9923&0.9617&0.8973&0.9159&0.6772&0.6478&0.5605&0.9361&0.9361&0.6189&0.5552&0.9908\\
HDN&0.3751&0.3714&0.1194&0.1346&0.1141&0.0856&0.0866&0.9609&0.9599&0.1477&0.1638&0.9985\\
HDHID&0.9814&0.9815&0.9092&0.9143&0.9113&0.9058&0.9131&0.9978&0.9956&0.9233&0.9387&1.0\\
IM&0.3857&0.2898&0.0889&0.0696&0.0696&0.0696&0.0696&0.1692&0.1715&0.0883&0.0696&0.4433\\
LOAN&0.998&0.9979&0.9858&0.9856&0.9855&0.9854&0.9854&1.0&0.9998&0.9854&0.9857&1.0\\
NBCD&0.9868&0.9845&0.913&0.9095&0.632&0.6514&0.5425&0.9219&0.9163&0.8889&0.5556&0.9877\\
BPCD&0.9976&0.9926&0.9465&0.9543&0.7728&0.6201&0.6346&0.9886&0.98&0.7148&0.6755&0.9995\\
PWD&0.9948&0.9946&0.9318&0.9364&0.9665&0.9317&0.8577&0.9843&0.9831&0.961&0.5167&0.9996\\
QARD&1.0&0.9999&0.9912&0.9946&0.9993&0.9494&0.998&0.9999&0.9999&0.9289&0.9559&1.0\\
STD&0.6545&0.6476&0.4616&0.4565&0.4017&0.3433&0.3495&0.4566&0.4566&0.3603&0.3419&0.6551\\
ACP&0.9987&0.9882&0.9491&0.9276&0.6493&0.6425&0.6484&0.9986&0.9961&0.7475&0.6506&1.0\\
HON&0.9998&0.9981&0.9872&0.9872&0.8043&0.7137&0.5998&0.994&0.9872&0.6904&0.6705&0.9998\\
TSC&0.9738&0.9682&0.8383&0.8657&0.7508&0.7346&0.7924&0.9987&0.9974&0.7991&0.8535&1.0\\
BTD&0.9774&0.9719&0.8902&0.8981&0.7736&0.762&0.762&0.9402&0.9266&0.8112&0.762&0.9922\\
TMD&0.9999&0.9999&0.9978&0.9981&0.9921&0.993&0.9936&0.9984&0.9984&0.9927&0.8097&1.0\\
VSD&0.9821&0.697&0.875&0.6952&0.531&0.6952&0.531&0.9071&0.875&0.6214&0.531&0.9821\\
WCR&0.9441&0.9476&0.8524&0.8559&0.8148&0.7806&0.8155&0.9999&0.9998&0.8217&0.8498&1.0\\
WQD&0.9901&0.9772&0.8505&0.8557&0.4233&0.2473&0.2525&0.9992&0.9992&0.9694&0.2669&1.0\\

\end{tabular}
\end{ruledtabular}
\end{table*}

\newpage
\begin{table*}
\caption{Comprehensive experimental results and theoretical upper bound analysis of Accuracy (AC) for various binary classifiers on additional 37 real datasets. The classifiers evaluated include XGBoost, Multilayer Perceptron (MLP), Support Vector Machine (SVM), Logistic Regression (LR), Decision Tree (DT), Random Forest (RF), K-Nearest Neighbors (KNN), and Naive Bayes (NB).}\label{tabS4}

\begin{ruledtabular}
\begin{tabular}{lcccccccccccc}
\multirow{2}*{Data} &  \multicolumn{4}{c}{XGBoost}& \multirow{2}*{MLP}& \multirow{2}*{SVM}& \multirow{2}*{LR}& \multirow{2}*{DT}& \multirow{2}*{RF}& \multirow{2}*{KNN}& \multirow{2}*{NB}&\multirow{2}*{Boundaries}\\
\cline{2-5}
~& Square&Logistic& Hinge&Softmax&~&~&~&~&~&~&~&~\\
\hline
ACD&0.8697&0.8684&0.8555&0.8685&0.839&0.7874&0.7852&0.9717&0.9717&0.8603&0.7934&0.9717\\
AMD&0.9877&0.985&0.9891&0.9866&0.9918&0.9827&0.9733&0.9922&0.9922&0.9813&0.2668&0.9922\\
ASD&0.985&0.9842&0.992&0.9856&0.9885&0.8496&0.8135&0.9979&0.9979&0.9348&0.8191&0.9979\\
ADP&0.75&0.75&0.75&0.75&0.75&0.75&0.75&0.75&0.75&0.75&0.75&0.75\\
BP&0.9936&0.9936&0.9936&0.9936&0.9934&0.993&0.9928&0.994&0.994&0.991&0.9516&0.994\\
CPDT&0.969&0.936&0.901&0.938&0.799&0.776&0.796&0.999&0.999&0.812&0.794&0.999\\
COP&0.9917&0.9834&0.9959&0.9876&0.7842&0.6701&0.6494&0.9959&0.9959&0.7967&0.6909&0.9959\\
CPDN&0.7394&0.7395&0.7308&0.7387&0.6756&0.614&0.6015&0.9727&0.9726&0.7569&0.5991&0.9727\\
CDD&0.942&0.89&0.873&0.903&0.633&0.598&0.598&0.995&0.995&0.767&0.625&0.995\\
CMC&0.8872&0.8315&0.8132&0.8444&0.7079&0.6352&0.6277&0.9783&0.9783&0.7452&0.6332&0.9783\\
CAP&0.8666&0.8462&0.8042&0.8482&0.7493&0.6756&0.7473&0.9972&0.9972&0.7732&0.7473&0.9972\\
SCP&0.9827&0.9827&0.9827&0.9827&0.9827&0.9827&0.9707&0.9827&0.9827&0.9809&0.7565&0.9827\\
CSAAP&0.934&0.933&0.934&0.9334&0.9204&0.9284&0.8511&0.934&0.934&0.9161&0.8165&0.934\\
CS&0.9631&0.9631&0.9636&0.9602&0.9845&0.957&0.9254&0.9897&0.9897&0.9765&0.6019&0.9897\\
DHI&0.8524&0.8515&0.8444&0.8511&0.843&0.8424&0.8436&0.9925&0.9925&0.8669&0.7717&0.9925\\
DD&0.9769&0.9769&0.9769&0.9769&0.9611&0.8854&0.8864&0.9769&0.9769&0.9664&0.8707&0.9769\\
ED&0.8852&0.8788&0.8768&0.8779&0.8364&0.6975&0.7109&0.9256&0.9256&0.8366&0.6909&0.9256\\
FDBD&0.9411&0.9406&0.9427&0.9436&0.7579&0.7344&0.8377&0.9998&0.9998&0.8518&0.7647&0.9998\\
GOT&0.8114&0.7939&0.7955&0.7888&0.7456&0.7456&0.7456&0.8376&0.8376&0.76&0.7456&0.8376\\
HD&0.951&0.9021&0.9371&0.9441&0.7692&0.7413&0.6503&0.951&0.951&0.7063&0.6434&0.951\\
HDN&0.9177&0.9175&0.916&0.9175&0.9154&0.9144&0.9144&0.9963&0.9963&0.9192&0.8612&0.9963\\
HDHID&0.9102&0.9097&0.9077&0.9098&0.9086&0.9058&0.9066&0.9955&0.9955&0.9164&0.8248&0.9955\\
IM&0.9343&0.9304&0.9317&0.9304&0.9304&0.9304&0.9304&0.9376&0.9376&0.9284&0.9304&0.9376\\
LOAN&0.9883&0.9865&0.9858&0.9856&0.9812&0.9854&0.9854&0.9998&0.9998&0.9854&0.9809&0.9998\\
NBCD&0.9417&0.9367&0.9417&0.9397&0.6894&0.7166&0.5729&0.9427&0.9427&0.9246&0.591&0.9427\\
BPCD&0.9739&0.9507&0.9567&0.9619&0.7672&0.6201&0.6261&0.9851&0.9851&0.709&0.6485&0.9851\\
PWD&0.9637&0.9634&0.9617&0.9622&0.98&0.9586&0.9127&0.9897&0.9897&0.9747&0.5856&0.9897\\
QARD&0.9994&0.9917&0.9911&0.9947&0.9994&0.9514&0.9982&0.9994&0.9994&0.9247&0.8162&0.9994\\
STD&0.8098&0.8085&0.8071&0.8094&0.7893&0.7674&0.7705&0.8098&0.8098&0.7246&0.7585&0.8098\\
ACP&0.98&0.935&0.9625&0.935&0.6475&0.6425&0.6425&0.9975&0.9975&0.7375&0.635&0.9975\\
HON&0.993&0.986&0.993&0.993&0.8462&0.7552&0.6084&0.993&0.993&0.7343&0.7063&0.993\\
TSC&0.8697&0.8495&0.8316&0.8533&0.7477&0.7346&0.7617&0.9974&0.9974&0.7806&0.7312&0.9974\\
BTD&0.9118&0.8944&0.8984&0.9011&0.7687&0.762&0.762&0.9318&0.9318&0.7874&0.762&0.9318\\
TMD&0.9966&0.9953&0.9966&0.9969&0.9906&0.9866&0.9877&0.9971&0.9971&0.9904&0.2507&0.9971\\
VSD&0.95&0.8&0.95&0.85&0.75&0.85&0.75&0.95&0.95&0.8&0.75&0.95\\
WCR&0.8516&0.8497&0.845&0.8496&0.4293&0.7806&0.803&0.9997&0.9997&0.8057&0.7925&0.9997\\
WQD&0.9759&0.964&0.9547&0.9555&0.7938&0.7527&0.7527&0.9998&0.9998&0.9913&0.7566&0.9998\\

\end{tabular}
\end{ruledtabular}
\end{table*}

\begin{figure}
    \centering
    \includegraphics[width=.9\linewidth]{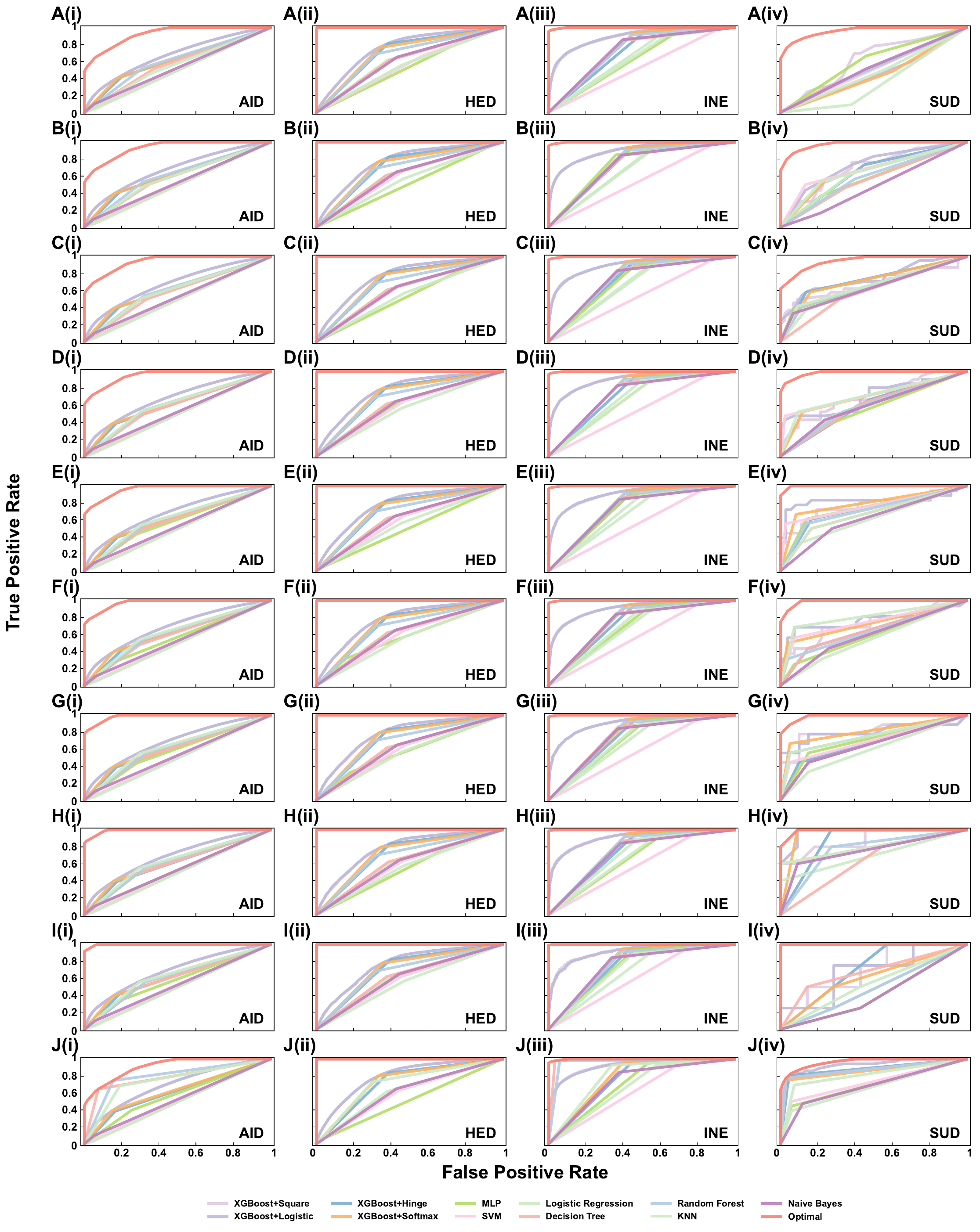}
    \caption{Exact upper bound of AUC and corresponding optimal ROC curves for four real-world datasets when $|\mathcal{S}_{train}|/|\mathcal{S}|=0.1$ (A), $|\mathcal{S}_{train}|/|\mathcal{S}|=0.2$ (B), $|\mathcal{S}_{train}|/|\mathcal{S}|=0.3$ (C), $|\mathcal{S}_{train}|/|\mathcal{S}|=0.4$ (D), $|\mathcal{S}_{train}|/|\mathcal{S}|=0.5$ (E), $|\mathcal{S}_{train}|/|\mathcal{S}|=0.6$ (F), $|\mathcal{S}_{train}|/|\mathcal{S}|=0.7$ (G), $|\mathcal{S}_{train}|/|\mathcal{S}|=0.8$ (H), $|\mathcal{S}_{train}|/|\mathcal{S}|=0.9$ (I), $|\mathcal{S}_{train}|/|\mathcal{S}|=1$ (J). The binary classifiers we used in this experiment include XGBoost, MLP, SVM, Logistic Regresion, Decision Tree, Random Forest, KNN and Naive Bayes. Red curves represent the theoretical optimal ROC curves.}
    \label{figS1}
\end{figure}

\begin{figure}
    \centering
    \includegraphics[width=.9\linewidth]{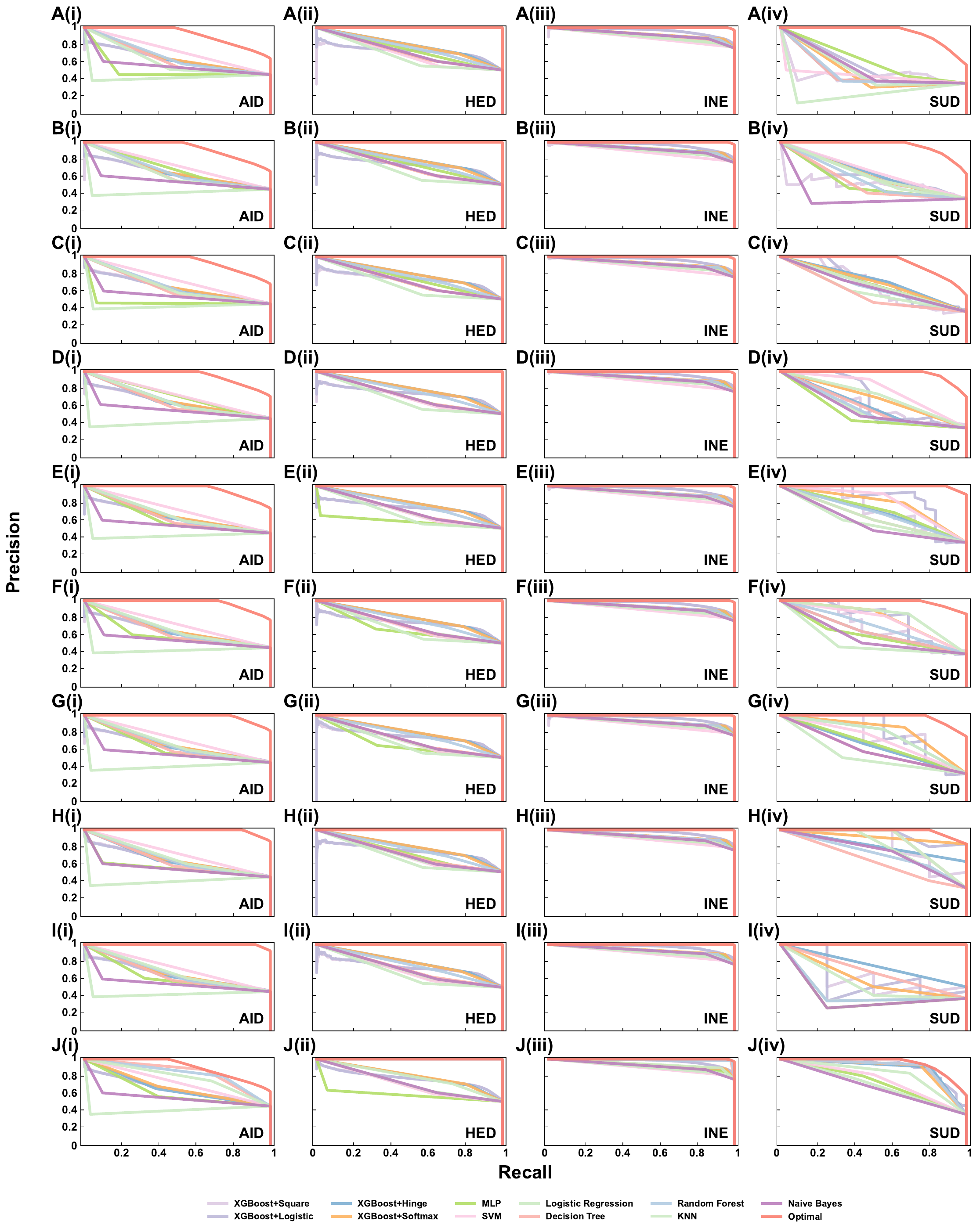}
    \caption{Exact upper bound of AP and corresponding optimal PR curves for four real-world datasets when $|\mathcal{S}_{train}|/|\mathcal{S}|=0.1$ (A), $|\mathcal{S}_{train}|/|\mathcal{S}|=0.2$ (B), $|\mathcal{S}_{train}|/|\mathcal{S}|=0.3$ (C), $|\mathcal{S}_{train}|/|\mathcal{S}|=0.4$ (D), $|\mathcal{S}_{train}|/|\mathcal{S}|=0.5$ (E), $|\mathcal{S}_{train}|/|\mathcal{S}|=0.6$ (F), $|\mathcal{S}_{train}|/|\mathcal{S}|=0.7$ (G), $|\mathcal{S}_{train}|/|\mathcal{S}|=0.8$ (H), $|\mathcal{S}_{train}|/|\mathcal{S}|=0.9$ (I), $|\mathcal{S}_{train}|/|\mathcal{S}|=1$ (J). The binary classifiers we used in this experiment include XGBoost, MLP, SVM, Logistic Regresion, Decision Tree, Random Forest, KNN and Naive Bayes. Red curves represent the theoretical optimal PR curves.}
    \label{figS2}
\end{figure}

\begin{figure}
    \centering
    \includegraphics[width=.8\linewidth]{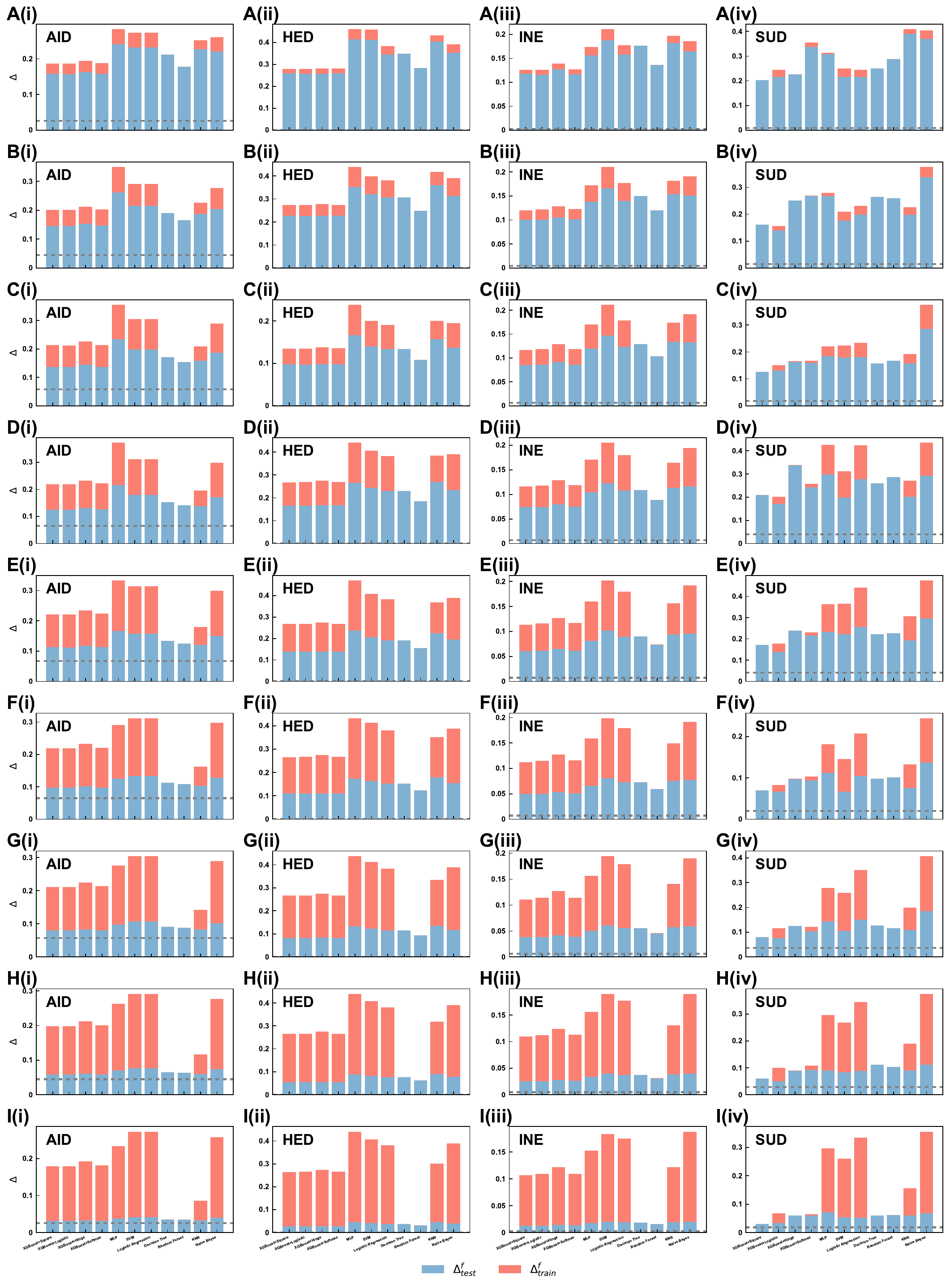}
    \caption{The loss errors of for four datasets in training ($\Delta_{train}^{f}$) and test sets ($\Delta_{test}^{f}$) when $|\mathcal{S}_{train}|/|\mathcal{S}|=0.1$ (A), $|\mathcal{S}_{train}|/|\mathcal{S}|=0.2$ (B), $|\mathcal{S}_{train}|/|\mathcal{S}|=0.3$ (C), $|\mathcal{S}_{train}|/|\mathcal{S}|=0.4$ (D), $|\mathcal{S}_{train}|/|\mathcal{S}|=0.5$ (E), $|\mathcal{S}_{train}|/|\mathcal{S}|=0.6$ (F), $|\mathcal{S}_{train}|/|\mathcal{S}|=0.7$ (G), $|\mathcal{S}_{train}|/|\mathcal{S}|=0.8$ (H), $|\mathcal{S}_{train}|/|\mathcal{S}|=0.9$ (I). Dash line represents the expected error of optimal classier based on Eq. \ref{min_delta}. }
    \label{figS3}
\end{figure}

\begin{figure}
    \centering
    \includegraphics[width=.85\linewidth]{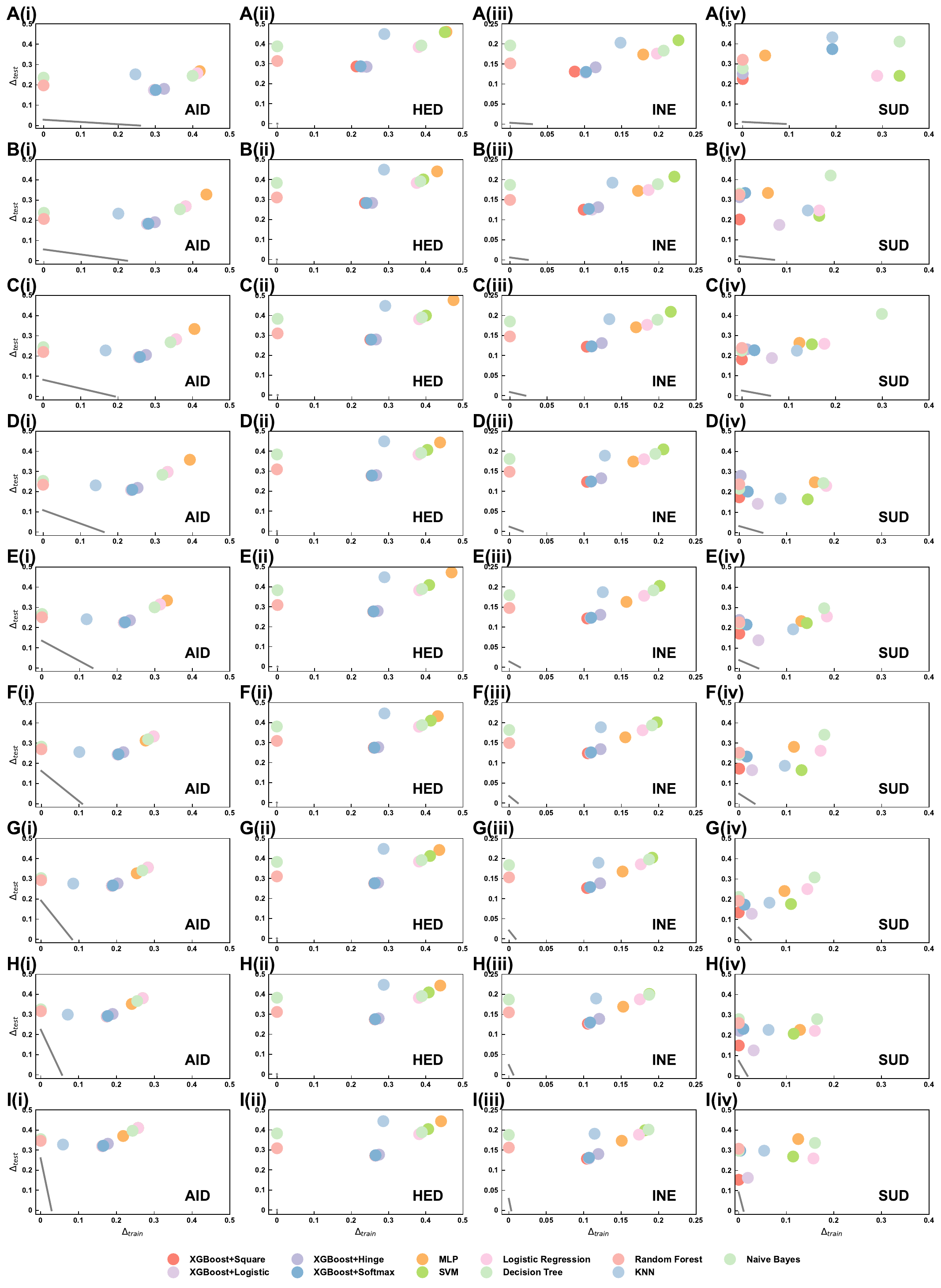}
    \caption{The loss errors of four datasets in training ($\Delta_{train}^{f}$) and test sets ($\Delta_{test}^{f}$) when $|\mathcal{S}_{train}|/|\mathcal{S}|=0.1$ (A), $|\mathcal{S}_{train}|/|\mathcal{S}|=0.2$ (B), $|\mathcal{S}_{train}|/|\mathcal{S}|=0.3$ (C), $|\mathcal{S}_{train}|/|\mathcal{S}|=0.4$ (D), $|\mathcal{S}_{train}|/|\mathcal{S}|=0.5$ (E), $|\mathcal{S}_{train}|/|\mathcal{S}|=0.6$ (F), $|\mathcal{S}_{train}|/|\mathcal{S}|=0.7$ (G), $|\mathcal{S}_{train}|/|\mathcal{S}|=0.8$ (H), $|\mathcal{S}_{train}|/|\mathcal{S}|=0.9$ (I). Gray line represents the expected error of optimal classier based on Eq. \ref{min_delta}. }
    \label{figS4}
\end{figure}

\begin{figure}
    \centering
    \includegraphics[width=.85\linewidth]{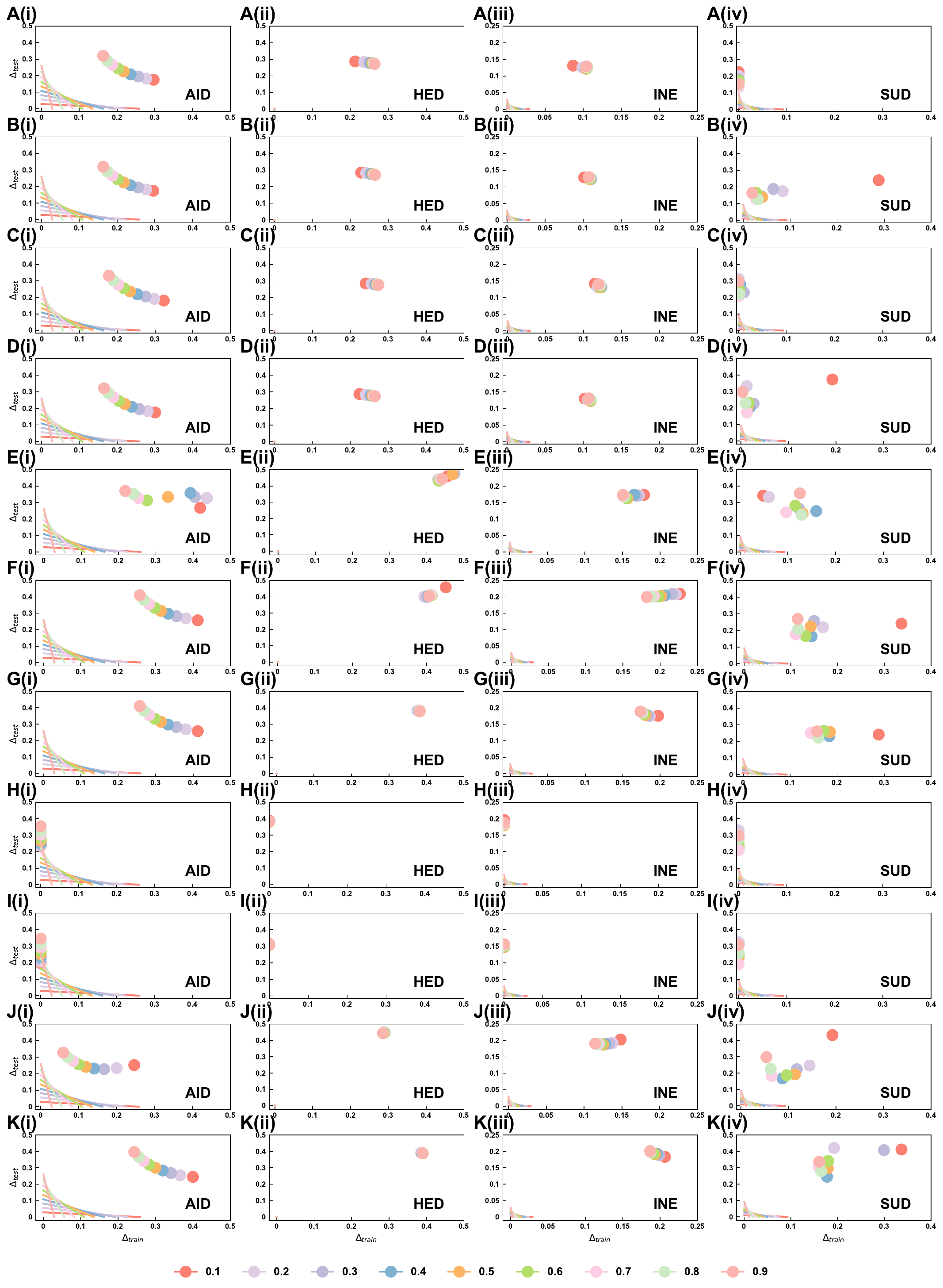}
    \caption{The loss errors of four datasets (AID, HED, INE and SUD) in training ($\Delta_{train}^{f}$) and test sets ($\Delta_{test}^{f}$) of different binary classifiers, including XGBoost with four classical objectives (A-D), MLP (E), SVM (F), Logistic Regression (G), Decision Tree (H), Random Forest (I), KNN (J). Colorful dots and lines represent different $|\mathcal{S}_{train}|/|\mathcal{S}|$ ranging from $0.1$ to $0.9$. }
    \label{figS5}
\end{figure}

\begin{figure}
    \centering
    \includegraphics[width=\linewidth]{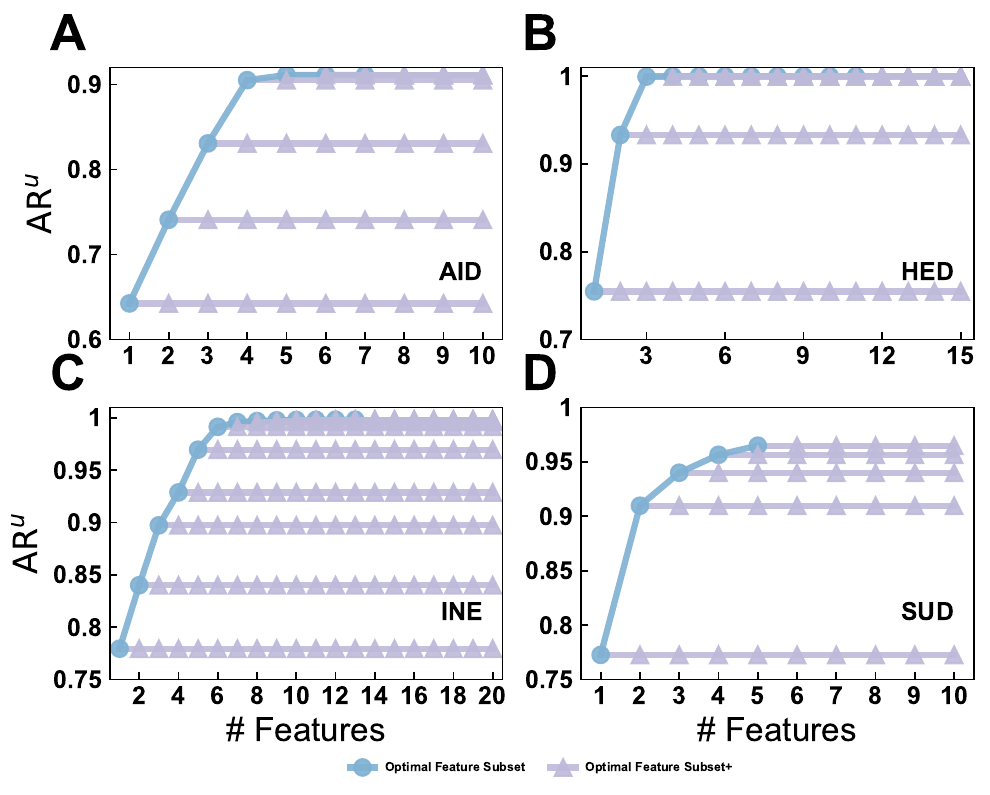}
    \caption{The $\text{AR}^u_{k_0}$ versus the optimal $k_0$ feature subset in feature selection (blue lines and dots). After we selected the optimal $k_0$ feature subset, we would use the feature extraction skill (LDA) to create new extracted features and add them into the original $k_0$ feature one by one (see red lines and dots). The datasets we used in this experiment includes AID (A), HED (B), INE (C) and SUD (D). }
    \label{figS6}
\end{figure}

\begin{figure}
    \centering
    \includegraphics[width=\linewidth]{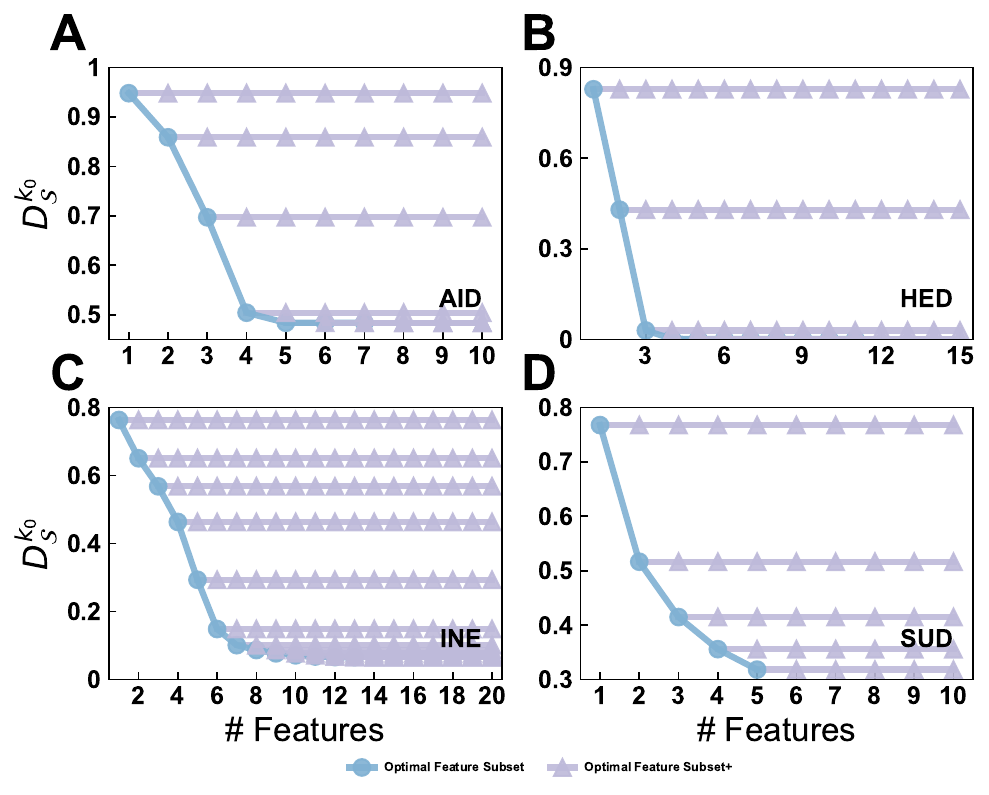}
    \caption{The $D_{\mathcal{S}}^{k_0}$ versus the optimal $k_0$ feature subset in feature selection (blue lines and dots). After we selected the optimal $k_0$ feature subset, we would use the feature extraction skill (LDA) to create new extracted features and add them into the original $k_0$ feature one by one (see red lines and dots). The datasets we used in this experiment includes AID (A), HED (B), INE (C) and SUD (D).}
    \label{figS7}
\end{figure}

\begin{figure}
    \centering
    \includegraphics[width=\linewidth]{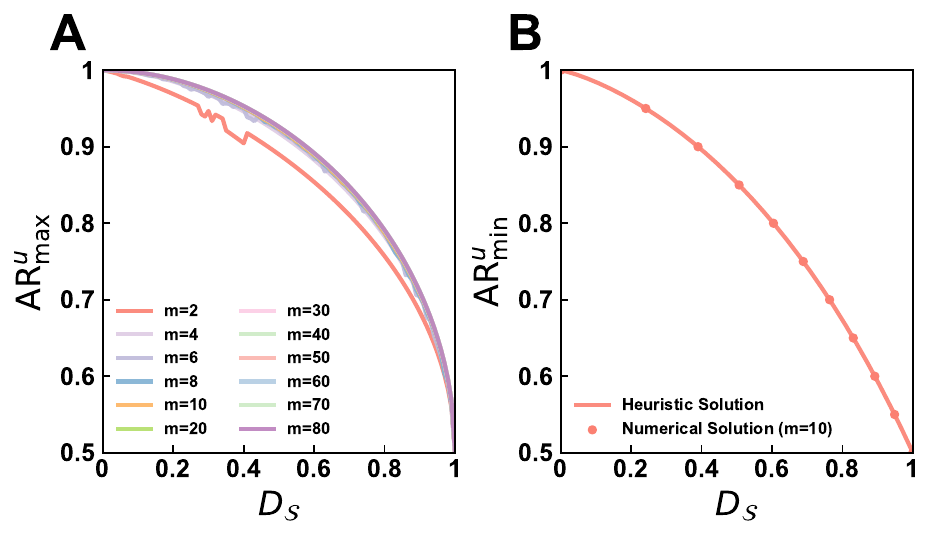}
    \caption{$\text{AR}^u_{\max}(D_{\mathcal{S}})$ curve (A) and $\text{AR}^u_{\min}(D_{\mathcal{S}})$ curve (B). (A) $\text{AR}^u_{\max}(D_{\mathcal{S}})$ curves is numerically solved by the SLSQP solver when $m=2,4,6,8,10,20,30,40,50,60,70,80$. We find that these curves are quickly converged as $m$ increases. Therefore the final $\text{AR}^u_{\max}(D_{\mathcal{S}})$ curve can be approximated to the numerically solved curve when $m=10$. (B) The heuristic curve for $\text{AR}^u_{\min}(D_{\mathcal{S}})$ is calculated by~\eqref{eq_heuristic}. And the numerically approximated $\text{AR}^u_{\min}(D_{\mathcal{S}})$ curve is derived by  the SLSQP solver. }
    \label{figS8}
\end{figure}

\begin{figure}
    \centering
    \includegraphics[width=\linewidth]{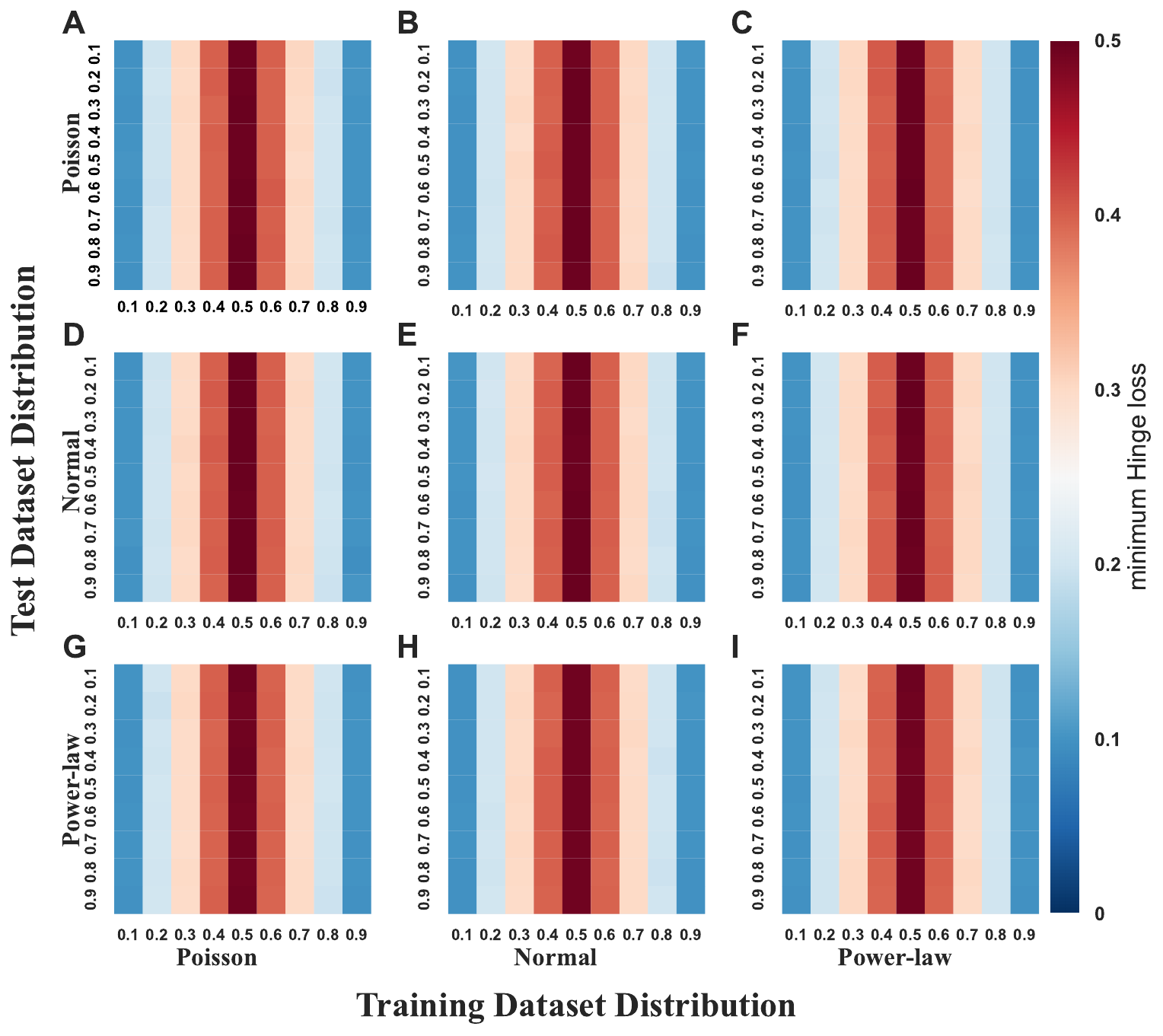}
    \caption{Minimum Hinge loss across synthesized datasets with varied label distributions. Each dataset is bifurcated into distinct training and testing subsets, synthesized using a trio of statistical distributions: Poisson, Normal (Gaussian), and Power-law. Feature vectors for training and testing are generated accordingly. The binary labels are assigned with a probability $p$ for class $1$ and $1-p$ for class $0$, where $p$ spans the set $\{0.1, 0.2, \ldots, 0.9\}$. This process yields nine unique dataset configurations, denoted as: Poisson \& Poisson (A), Normal \& Poisson (B), Power-law \& Poisson (C), Poisson \& Normal (D), Normal \& Normal (E), Power-law \& Normal (F), Poisson \& Power-law (G), Normal \& Power-law (H) and Power-law \& Power-law (I). Each subset of this matrix is visualized as a heat map, charting the minimal Hinge loss achieved across varying label probabilities within the respective training and testing subsets, under specific distribution pairings. Notably, the minimal Hinge loss for the training subset is computed independently of the testing subset, resulting in identical columns for each heat map.}
    \label{figS9}
\end{figure}

\begin{figure}
    \centering
    \includegraphics[width=\linewidth]{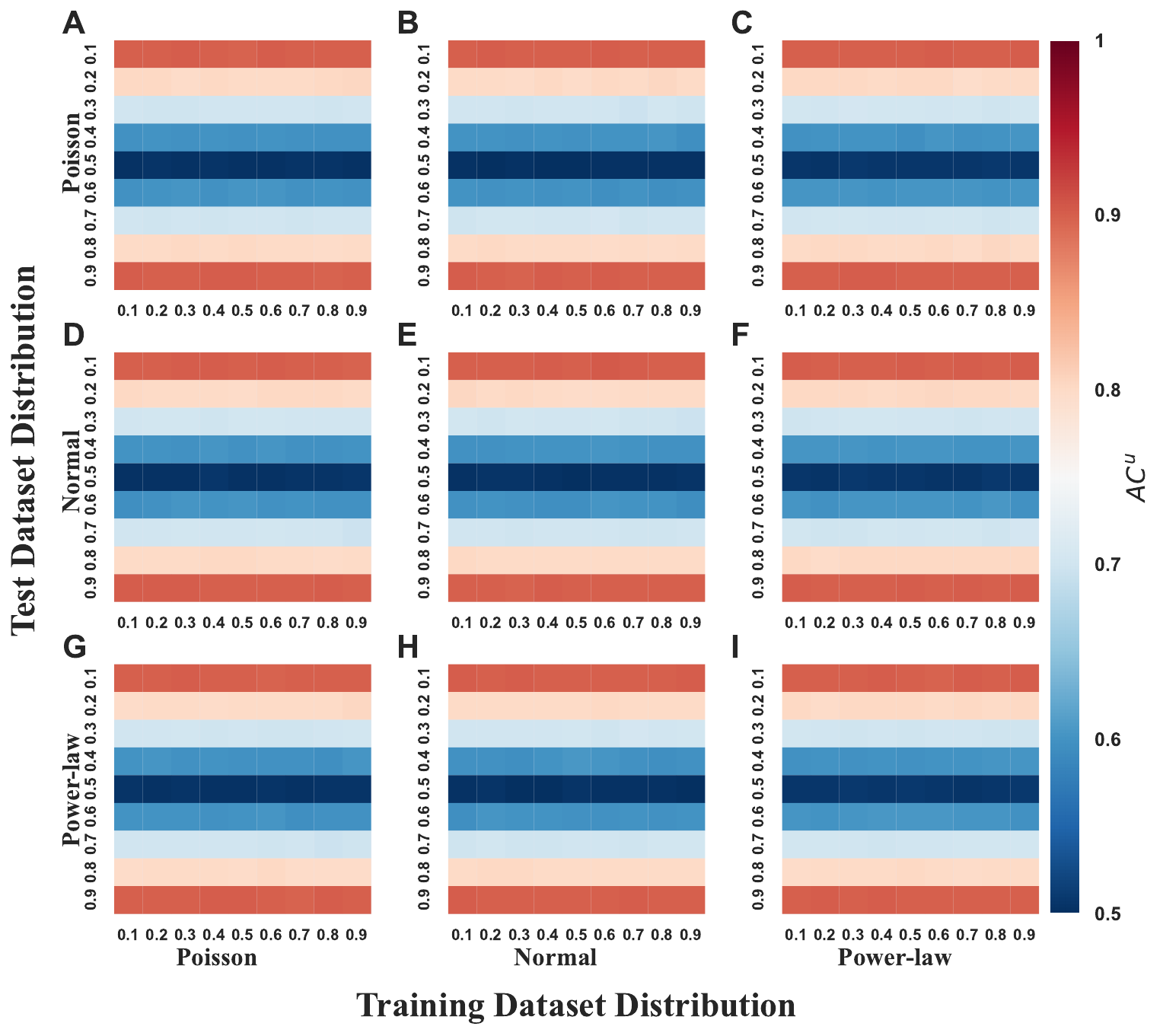}
    \caption{Maximum accuracy ($AC^u$) across synthesized datasets with varied label distributions. Each dataset is bifurcated into distinct training and testing subsets, synthesized using a trio of statistical distributions: Poisson, Normal (Gaussian), and Power-law. Feature vectors for training and testing are generated accordingly. The binary labels are assigned with a probability $p$ for class $1$ and $1-p$ for class $0$, where $p$ spans the set $\{0.1, 0.2, \ldots, 0.9\}$. This process yields nine unique dataset configurations, denoted as: Poisson \& Poisson (A), Normal \& Poisson (B), Power-law \& Poisson (C), Poisson \& Normal (D), Normal \& Normal (E), Power-law \& Normal (F), Poisson \& Power-law (G), Normal \& Power-law (H) and Power-law \& Power-law (I). Each subset of this matrix is visualized as a heat map, charting the maximum accuracy ($AC^u$) achieved across varying label probabilities within the respective training and testing subsets, under specific distribution pairings. Notably, the maximum accuracy for the testing subset is computed independently of the training subset, resulting in identical rows for each heat map.}
    \label{figS10}
\end{figure}

\begin{figure}
    \centering
    \includegraphics[width=\linewidth]{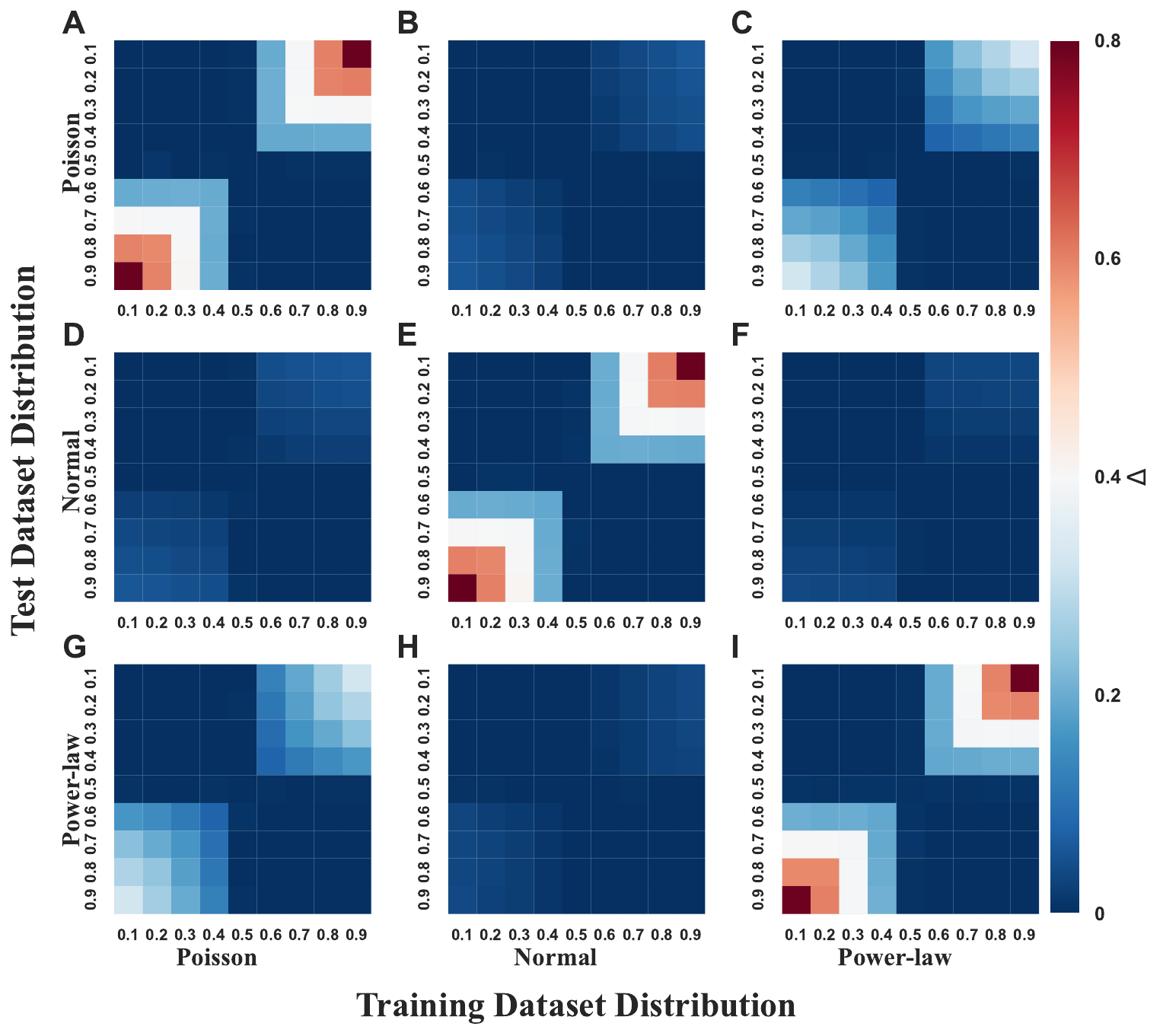}
    \caption{Joint error bound ($\Delta$) across synthesized datasets with varied label distributions. Each dataset is bifurcated into distinct training and testing subsets, synthesized using a trio of statistical distributions: Poisson, Normal (Gaussian), and Power-law. Feature vectors for training and testing are generated accordingly. The binary labels are assigned with a probability $p$ for class $1$ and $1-p$ for class $0$, where $p$ spans the set $\{0.1, 0.2, \ldots, 0.9\}$. This process yields nine unique dataset configurations, denoted as: Poisson \& Poisson (A), Normal \& Poisson (B), Power-law \& Poisson (C), Poisson \& Normal (D), Normal \& Normal (E), Power-law \& Normal (F), Poisson \& Power-law (G), Normal \& Power-law (H) and Power-law \& Power-law (I). Each subset of this matrix is visualized as a heat map, charting the joint error bound ($\Delta$) achieved across varying label probabilities within the respective training and testing subsets, under specific distribution pairings. Notably, $\Delta$ for the training and testing subset is computed based on Eq.~(54) for each heat map.}
    \label{figS11}
\end{figure}

\begin{figure}
\centering
\includegraphics[width=.9\linewidth]{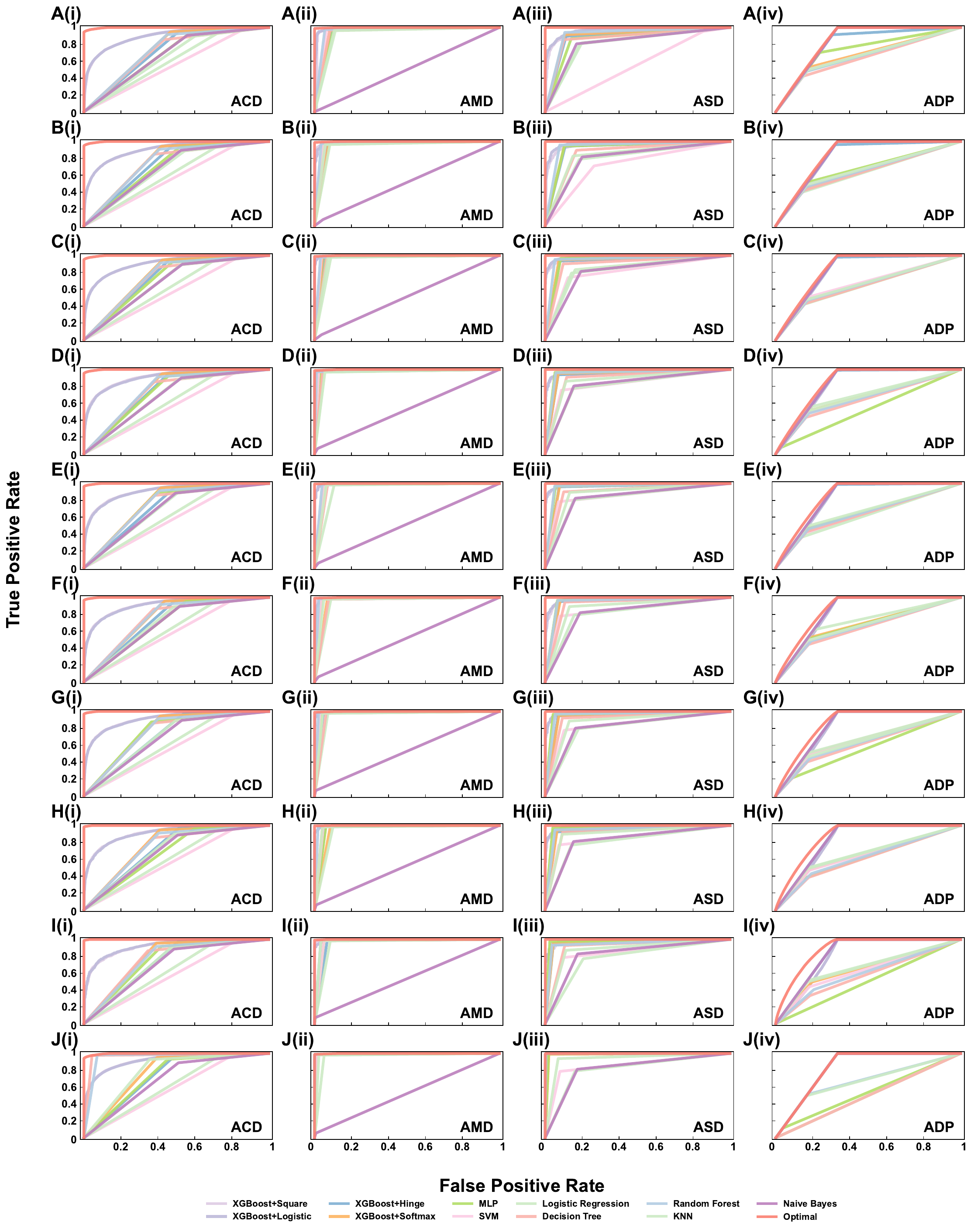}
\caption{Exact upper bound of AUC and corresponding optimal ROC curves for 4 additional real-world datasets (ACD, AMD, ASD and ADP) when $|\mathcal{S}_{train}|/|\mathcal{S}|=0.1$ (A), $|\mathcal{S}_{train}|/|\mathcal{S}|=0.2$ (B), $|\mathcal{S}_{train}|/|\mathcal{S}|=0.3$ (C), $|\mathcal{S}_{train}|/|\mathcal{S}|=0.4$ (D), $|\mathcal{S}_{train}|/|\mathcal{S}|=0.5$ (E), $|\mathcal{S}_{train}|/|\mathcal{S}|=0.6$ (F), $|\mathcal{S}_{train}|/|\mathcal{S}|=0.7$ (G), $|\mathcal{S}_{train}|/|\mathcal{S}|=0.8$ (H), $|\mathcal{S}_{train}|/|\mathcal{S}|=0.9$ (I) and $|\mathcal{S}_{train}|/|\mathcal{S}|=1$ (J). The binary classifiers we used in this experiment include XGBoost, MLP, SVM, Logistic Regresion, Decision Tree, Random Forest, KNN and Naive Bayes. Red curves represent the theoretical optimal ROC curves.}
\label{figS12-1}
\end{figure}

\begin{figure}
\centering
\includegraphics[width=.9\linewidth]{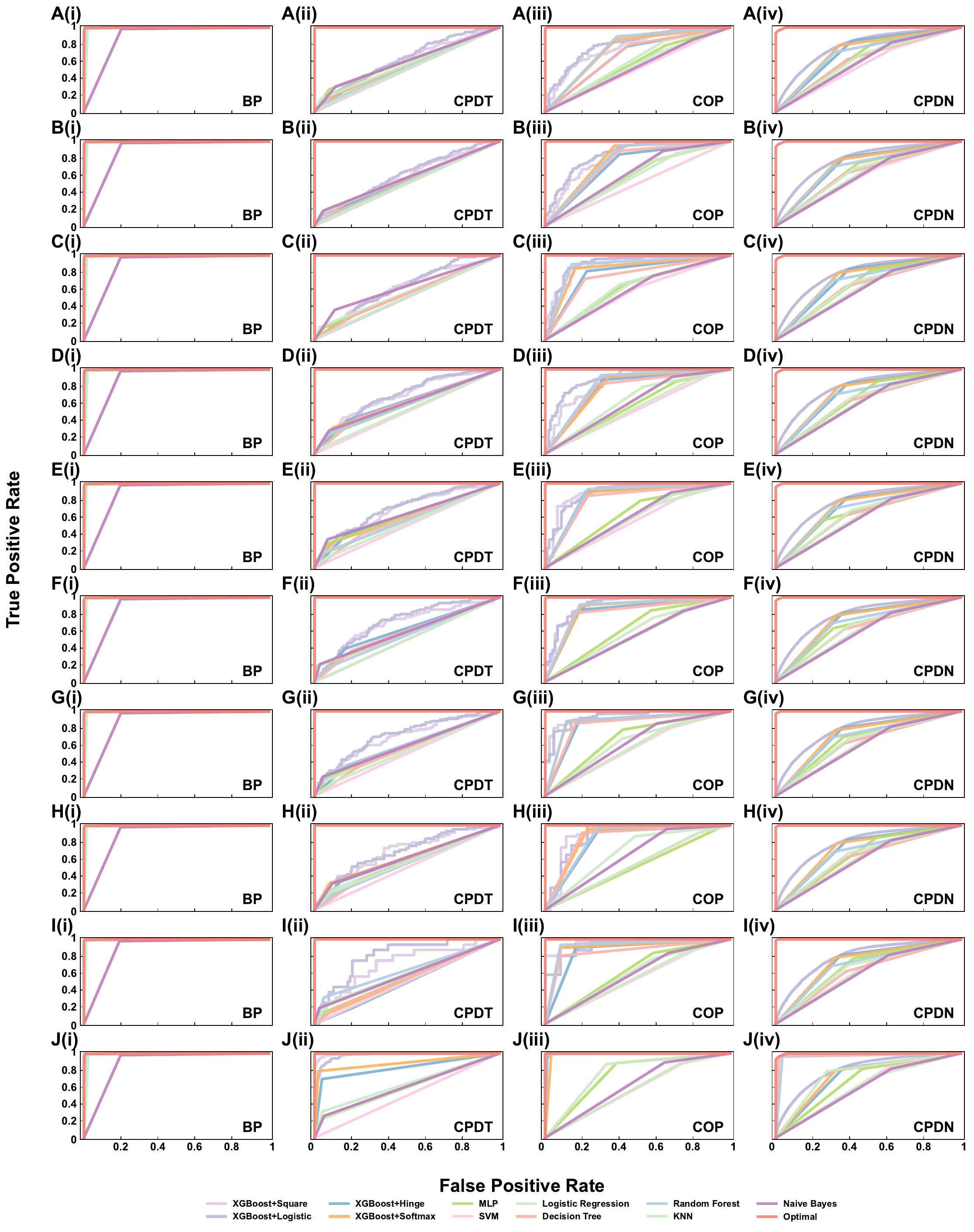}
\caption{Exact upper bound of AUC and corresponding optimal ROC curves for 4 additional real-world datasets (BP, CPDT, COP and CPDN) when $|\mathcal{S}_{train}|/|\mathcal{S}|=0.1$ (A), $|\mathcal{S}_{train}|/|\mathcal{S}|=0.2$ (B), $|\mathcal{S}_{train}|/|\mathcal{S}|=0.3$ (C), $|\mathcal{S}_{train}|/|\mathcal{S}|=0.4$ (D), $|\mathcal{S}_{train}|/|\mathcal{S}|=0.5$ (E), $|\mathcal{S}_{train}|/|\mathcal{S}|=0.6$ (F), $|\mathcal{S}_{train}|/|\mathcal{S}|=0.7$ (G), $|\mathcal{S}_{train}|/|\mathcal{S}|=0.8$ (H), $|\mathcal{S}_{train}|/|\mathcal{S}|=0.9$ (I)  and $|\mathcal{S}_{train}|/|\mathcal{S}|=1$ (J). The binary classifiers we used in this experiment include XGBoost, MLP, SVM, Logistic Regresion, Decision Tree, Random Forest, KNN and Naive Bayes. Red curves represent the theoretical optimal ROC curves.}
\label{figS12-2}
\end{figure}

\begin{figure}
\centering
\includegraphics[width=.9\linewidth]{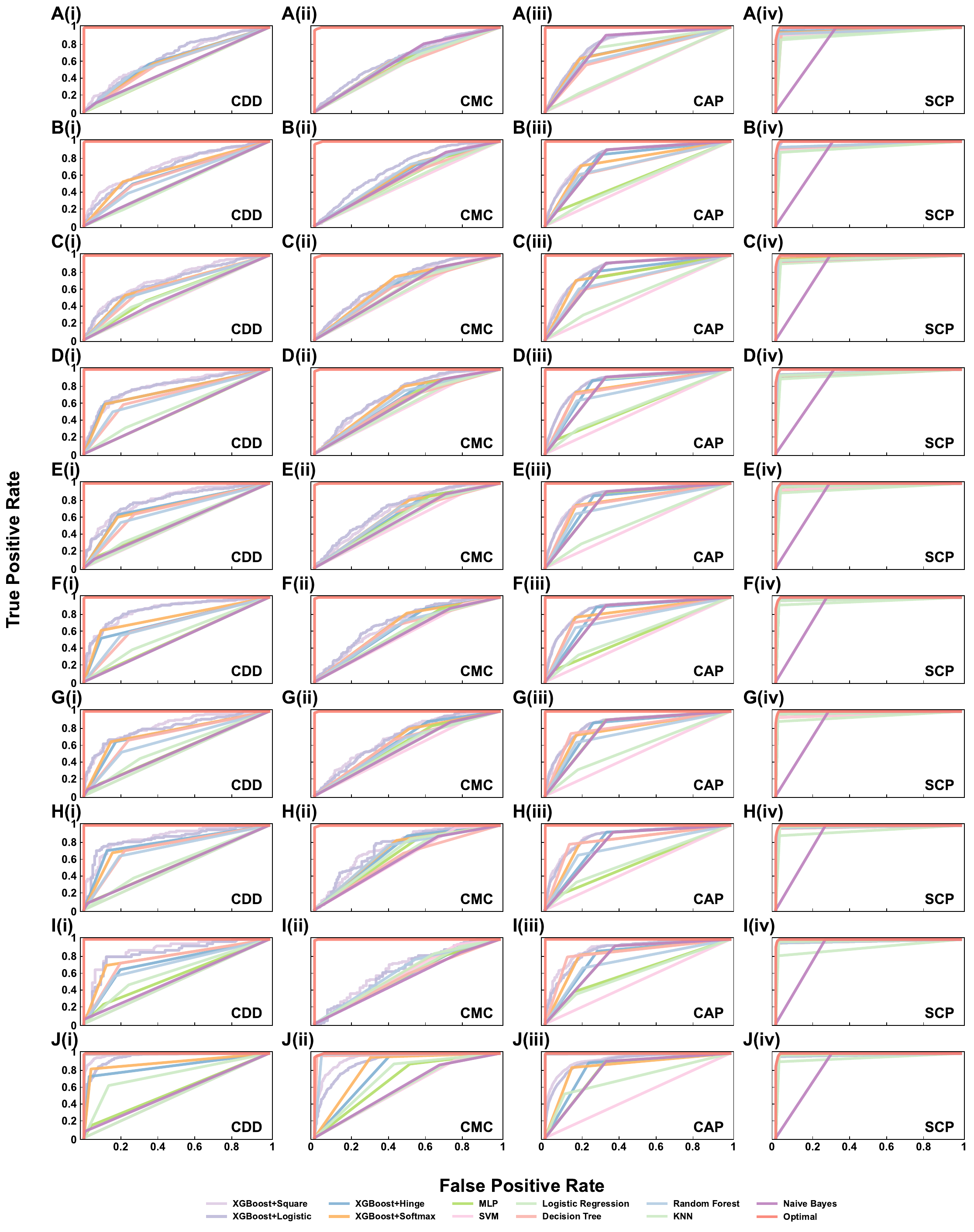}
\caption{Exact upper bound of AUC and corresponding optimal ROC curves for 4 additional real-world datasets (CDD, CMC, CAP and SCP) when $|\mathcal{S}_{train}|/|\mathcal{S}|=0.1$ (A), $|\mathcal{S}_{train}|/|\mathcal{S}|=0.2$ (B), $|\mathcal{S}_{train}|/|\mathcal{S}|=0.3$ (C), $|\mathcal{S}_{train}|/|\mathcal{S}|=0.4$ (D), $|\mathcal{S}_{train}|/|\mathcal{S}|=0.5$ (E), $|\mathcal{S}_{train}|/|\mathcal{S}|=0.6$ (F), $|\mathcal{S}_{train}|/|\mathcal{S}|=0.7$ (G), $|\mathcal{S}_{train}|/|\mathcal{S}|=0.8$ (H), $|\mathcal{S}_{train}|/|\mathcal{S}|=0.9$ (I)  and $|\mathcal{S}_{train}|/|\mathcal{S}|=1$ (J). The binary classifiers we used in this experiment include XGBoost, MLP, SVM, Logistic Regresion, Decision Tree, Random Forest, KNN and Naive Bayes. Red curves represent the theoretical optimal ROC curves.}
\label{figS12-3}
\end{figure}

\begin{figure}
\centering
\includegraphics[width=.9\linewidth]{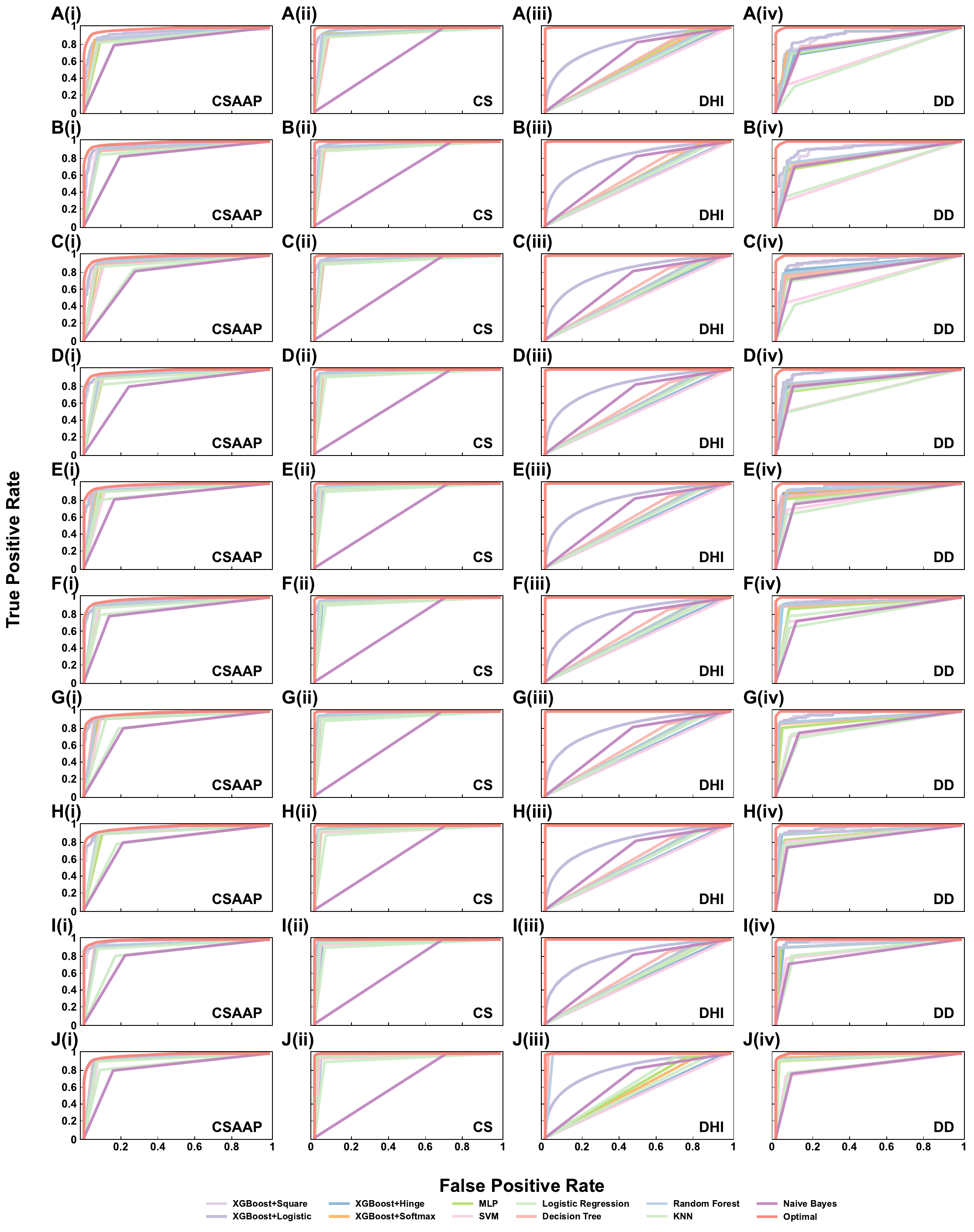}
\caption{Exact upper bound of AUC and corresponding optimal ROC curves for 4 additional real-world datasets (CSAAP, CS, DHI and DD) when $|\mathcal{S}_{train}|/|\mathcal{S}|=0.1$ (A), $|\mathcal{S}_{train}|/|\mathcal{S}|=0.2$ (B), $|\mathcal{S}_{train}|/|\mathcal{S}|=0.3$ (C), $|\mathcal{S}_{train}|/|\mathcal{S}|=0.4$ (D), $|\mathcal{S}_{train}|/|\mathcal{S}|=0.5$ (E), $|\mathcal{S}_{train}|/|\mathcal{S}|=0.6$ (F), $|\mathcal{S}_{train}|/|\mathcal{S}|=0.7$ (G), $|\mathcal{S}_{train}|/|\mathcal{S}|=0.8$ (H), $|\mathcal{S}_{train}|/|\mathcal{S}|=0.9$ (I)  and $|\mathcal{S}_{train}|/|\mathcal{S}|=1$ (J). The binary classifiers we used in this experiment include XGBoost, MLP, SVM, Logistic Regresion, Decision Tree, Random Forest, KNN and Naive Bayes. Red curves represent the theoretical optimal ROC curves.}
\label{figS12-4}
\end{figure}

\begin{figure}
\centering
\includegraphics[width=.9\linewidth]{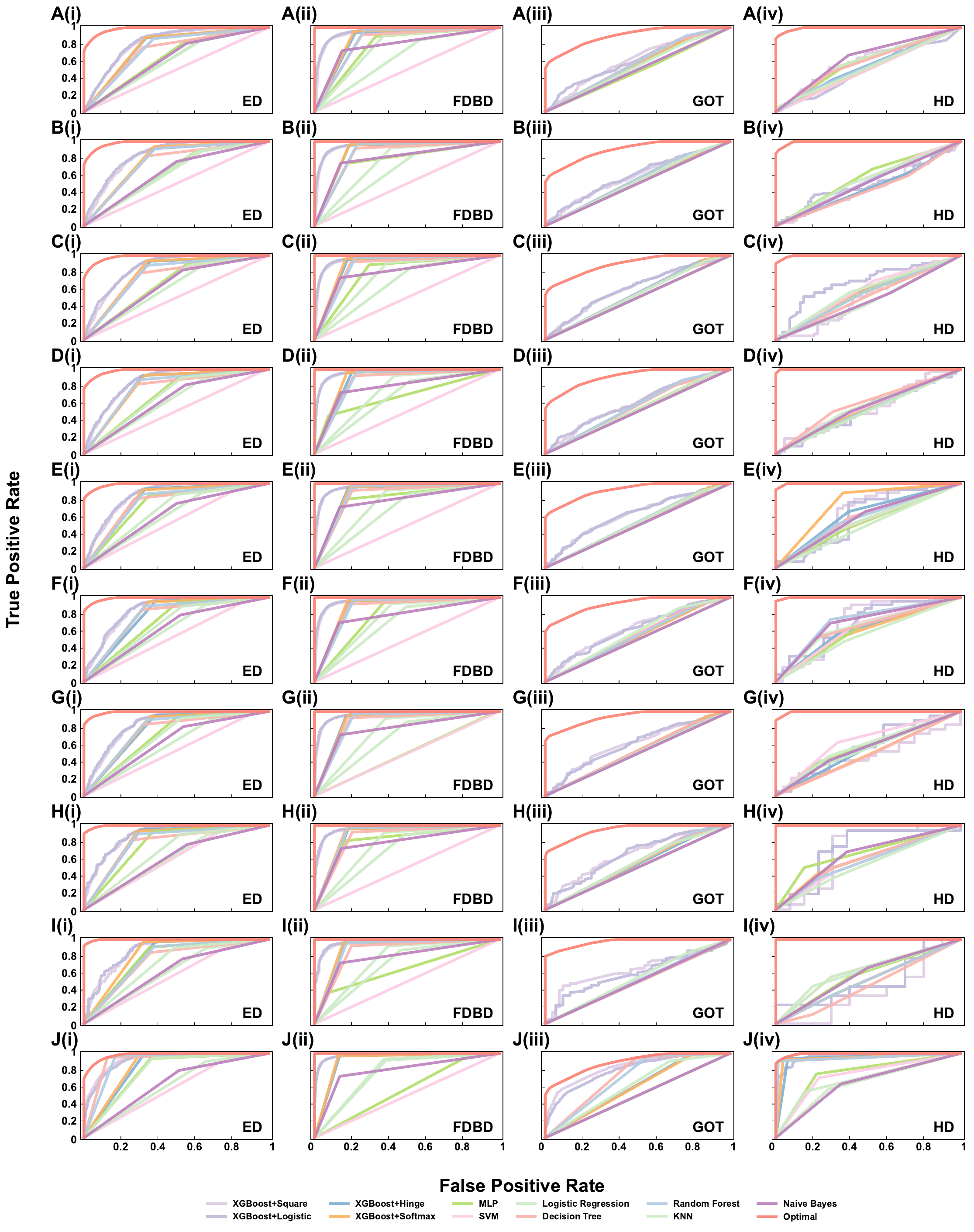}
\caption{Exact upper bound of AUC and corresponding optimal ROC curves for 4 additional real-world datasets (ED, FDBD, GOT and HD) when $|\mathcal{S}_{train}|/|\mathcal{S}|=0.1$ (A), $|\mathcal{S}_{train}|/|\mathcal{S}|=0.2$ (B), $|\mathcal{S}_{train}|/|\mathcal{S}|=0.3$ (C), $|\mathcal{S}_{train}|/|\mathcal{S}|=0.4$ (D), $|\mathcal{S}_{train}|/|\mathcal{S}|=0.5$ (E), $|\mathcal{S}_{train}|/|\mathcal{S}|=0.6$ (F), $|\mathcal{S}_{train}|/|\mathcal{S}|=0.7$ (G), $|\mathcal{S}_{train}|/|\mathcal{S}|=0.8$ (H), $|\mathcal{S}_{train}|/|\mathcal{S}|=0.9$ (I)  and $|\mathcal{S}_{train}|/|\mathcal{S}|=1$ (J). The binary classifiers we used in this experiment include XGBoost, MLP, SVM, Logistic Regresion, Decision Tree, Random Forest, KNN and Naive Bayes. Red curves represent the theoretical optimal ROC curves.}
\label{figS12-5}
\end{figure}

\begin{figure}
\centering
\includegraphics[width=.9\linewidth]{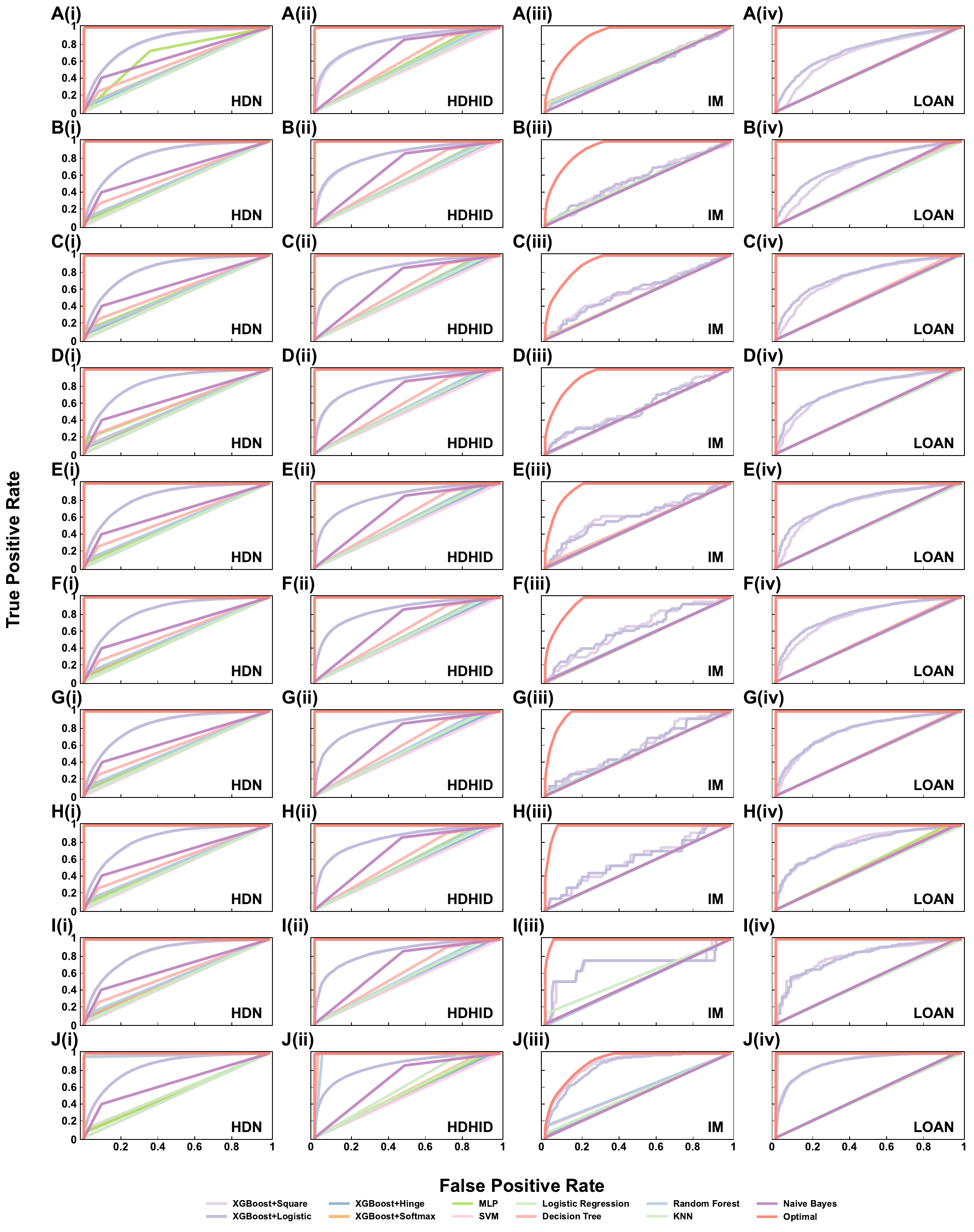}
\caption{Exact upper bound of AUC and corresponding optimal ROC curves for 4 additional real-world datasets (HDN, HDHID, IM and LOAN) when $|\mathcal{S}_{train}|/|\mathcal{S}|=0.1$ (A), $|\mathcal{S}_{train}|/|\mathcal{S}|=0.2$ (B), $|\mathcal{S}_{train}|/|\mathcal{S}|=0.3$ (C), $|\mathcal{S}_{train}|/|\mathcal{S}|=0.4$ (D), $|\mathcal{S}_{train}|/|\mathcal{S}|=0.5$ (E), $|\mathcal{S}_{train}|/|\mathcal{S}|=0.6$ (F), $|\mathcal{S}_{train}|/|\mathcal{S}|=0.7$ (G), $|\mathcal{S}_{train}|/|\mathcal{S}|=0.8$ (H), $|\mathcal{S}_{train}|/|\mathcal{S}|=0.9$ (I)  and $|\mathcal{S}_{train}|/|\mathcal{S}|=1$ (J). The binary classifiers we used in this experiment include XGBoost, MLP, SVM, Logistic Regresion, Decision Tree, Random Forest, KNN and Naive Bayes. Red curves represent the theoretical optimal ROC curves.}
\label{figS12-6}
\end{figure}

\begin{figure}
\centering
\includegraphics[width=.9\linewidth]{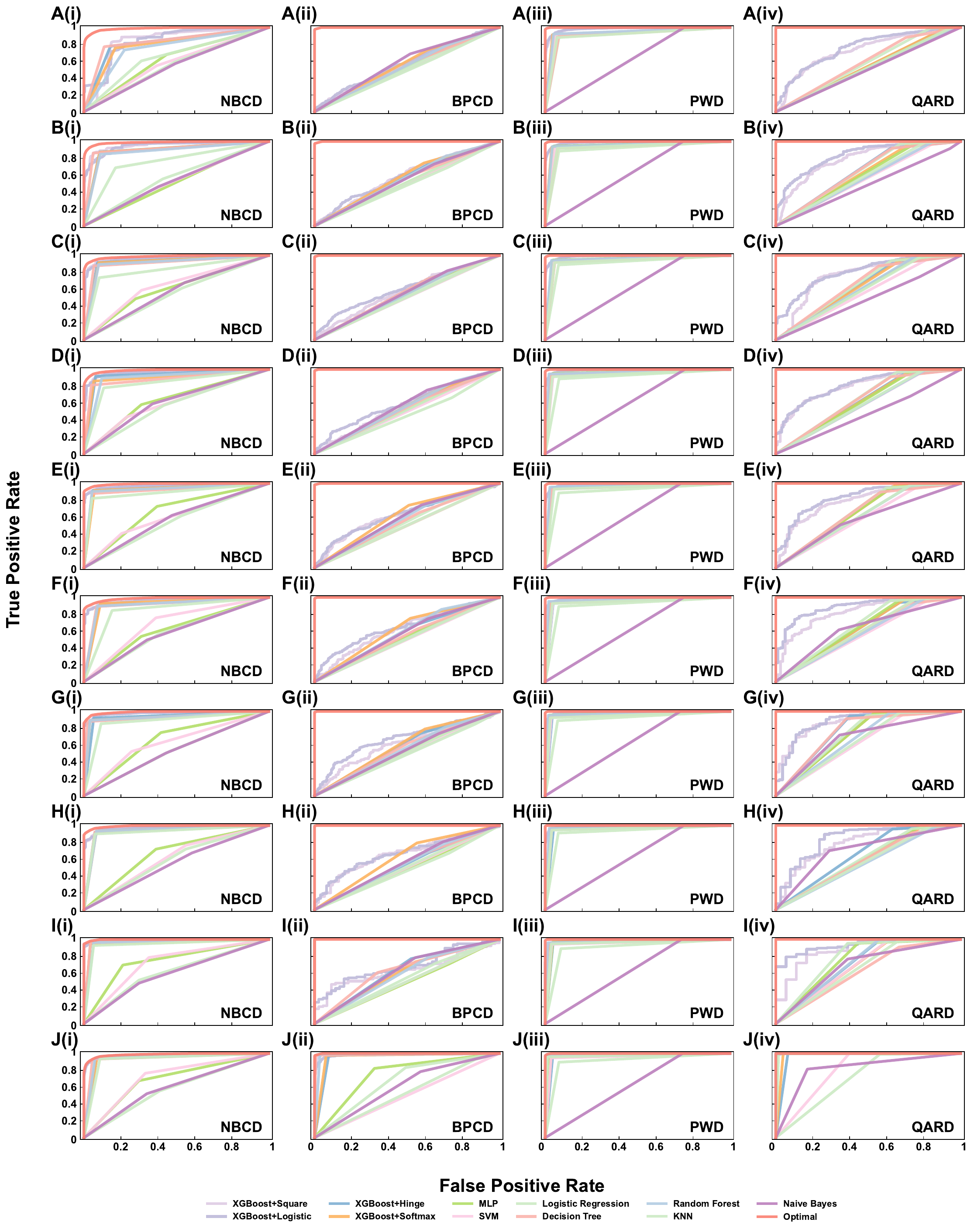}
\caption{Exact upper bound of AUC and corresponding optimal ROC curves for 4 additional real-world datasets (NBCD, BPCD, PWD and QARD) when $|\mathcal{S}_{train}|/|\mathcal{S}|=0.1$ (A), $|\mathcal{S}_{train}|/|\mathcal{S}|=0.2$ (B), $|\mathcal{S}_{train}|/|\mathcal{S}|=0.3$ (C), $|\mathcal{S}_{train}|/|\mathcal{S}|=0.4$ (D), $|\mathcal{S}_{train}|/|\mathcal{S}|=0.5$ (E), $|\mathcal{S}_{train}|/|\mathcal{S}|=0.6$ (F), $|\mathcal{S}_{train}|/|\mathcal{S}|=0.7$ (G), $|\mathcal{S}_{train}|/|\mathcal{S}|=0.8$ (H), $|\mathcal{S}_{train}|/|\mathcal{S}|=0.9$ (I)  and $|\mathcal{S}_{train}|/|\mathcal{S}|=1$ (J). The binary classifiers we used in this experiment include XGBoost, MLP, SVM, Logistic Regresion, Decision Tree, Random Forest, KNN and Naive Bayes. Red curves represent the theoretical optimal ROC curves.}
\label{figS12-7}
\end{figure}

\begin{figure}
\centering
\includegraphics[width=.9\linewidth]{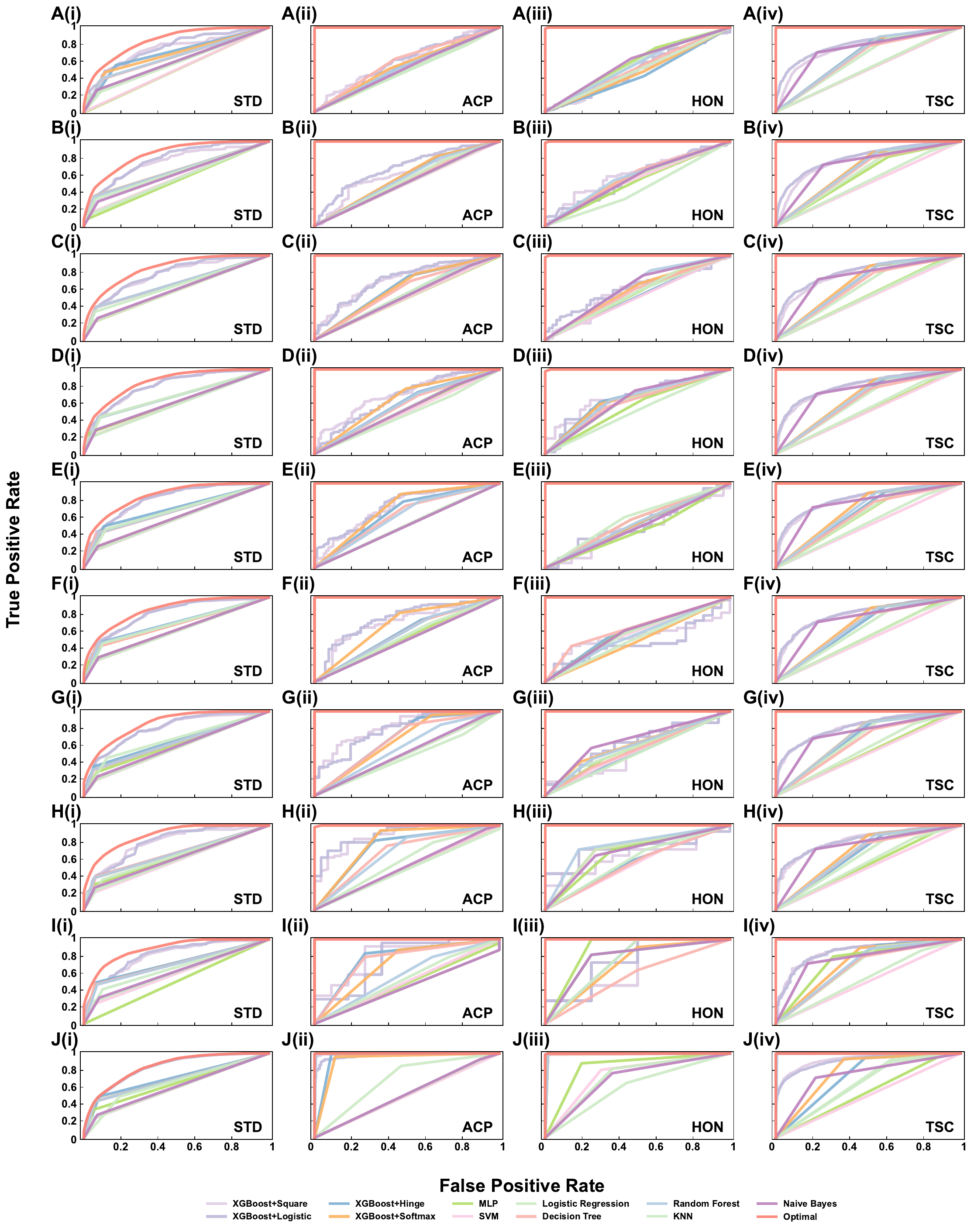}
\caption{Exact upper bound of AUC and corresponding optimal ROC curves for 4 additional real-world datasets (STD, ACP, HON and TSC) when $|\mathcal{S}_{train}|/|\mathcal{S}|=0.1$ (A), $|\mathcal{S}_{train}|/|\mathcal{S}|=0.2$ (B), $|\mathcal{S}_{train}|/|\mathcal{S}|=0.3$ (C), $|\mathcal{S}_{train}|/|\mathcal{S}|=0.4$ (D), $|\mathcal{S}_{train}|/|\mathcal{S}|=0.5$ (E), $|\mathcal{S}_{train}|/|\mathcal{S}|=0.6$ (F), $|\mathcal{S}_{train}|/|\mathcal{S}|=0.7$ (G), $|\mathcal{S}_{train}|/|\mathcal{S}|=0.8$ (H), $|\mathcal{S}_{train}|/|\mathcal{S}|=0.9$ (I)  and $|\mathcal{S}_{train}|/|\mathcal{S}|=1$ (J). The binary classifiers we used in this experiment include XGBoost, MLP, SVM, Logistic Regresion, Decision Tree, Random Forest, KNN and Naive Bayes. Red curves represent the theoretical optimal ROC curves.}
\label{figS12-8}
\end{figure}

\begin{figure}
\centering
\includegraphics[width=.9\linewidth]{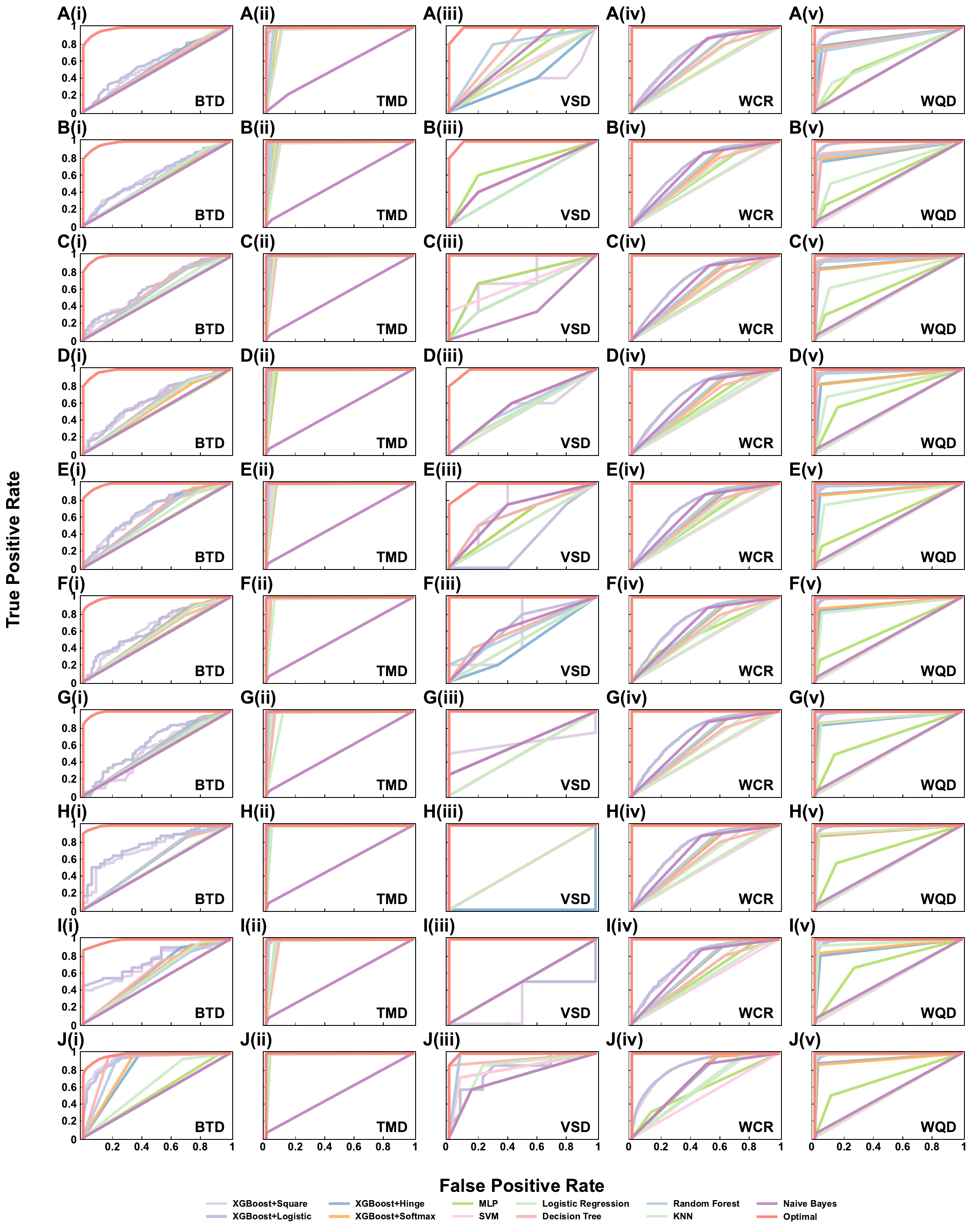}
\caption{Exact upper bound of AUC and corresponding optimal ROC curves for 5 additional real-world datasets (BTD, TMD, VSD, WCR and WQD) when $|\mathcal{S}_{train}|/|\mathcal{S}|=0.1$ (A), $|\mathcal{S}_{train}|/|\mathcal{S}|=0.2$ (B), $|\mathcal{S}_{train}|/|\mathcal{S}|=0.3$ (C), $|\mathcal{S}_{train}|/|\mathcal{S}|=0.4$ (D), $|\mathcal{S}_{train}|/|\mathcal{S}|=0.5$ (E), $|\mathcal{S}_{train}|/|\mathcal{S}|=0.6$ (F), $|\mathcal{S}_{train}|/|\mathcal{S}|=0.7$ (G), $|\mathcal{S}_{train}|/|\mathcal{S}|=0.8$ (H), $|\mathcal{S}_{train}|/|\mathcal{S}|=0.9$ (I)  and $|\mathcal{S}_{train}|/|\mathcal{S}|=1$ (J). The binary classifiers we used in this experiment include XGBoost, MLP, SVM, Logistic Regresion, Decision Tree, Random Forest, KNN and Naive Bayes. Red curves represent the theoretical optimal ROC curves.}
\label{figS12-9}
\end{figure}

\begin{figure}
\centering
\includegraphics[width=.9\linewidth]{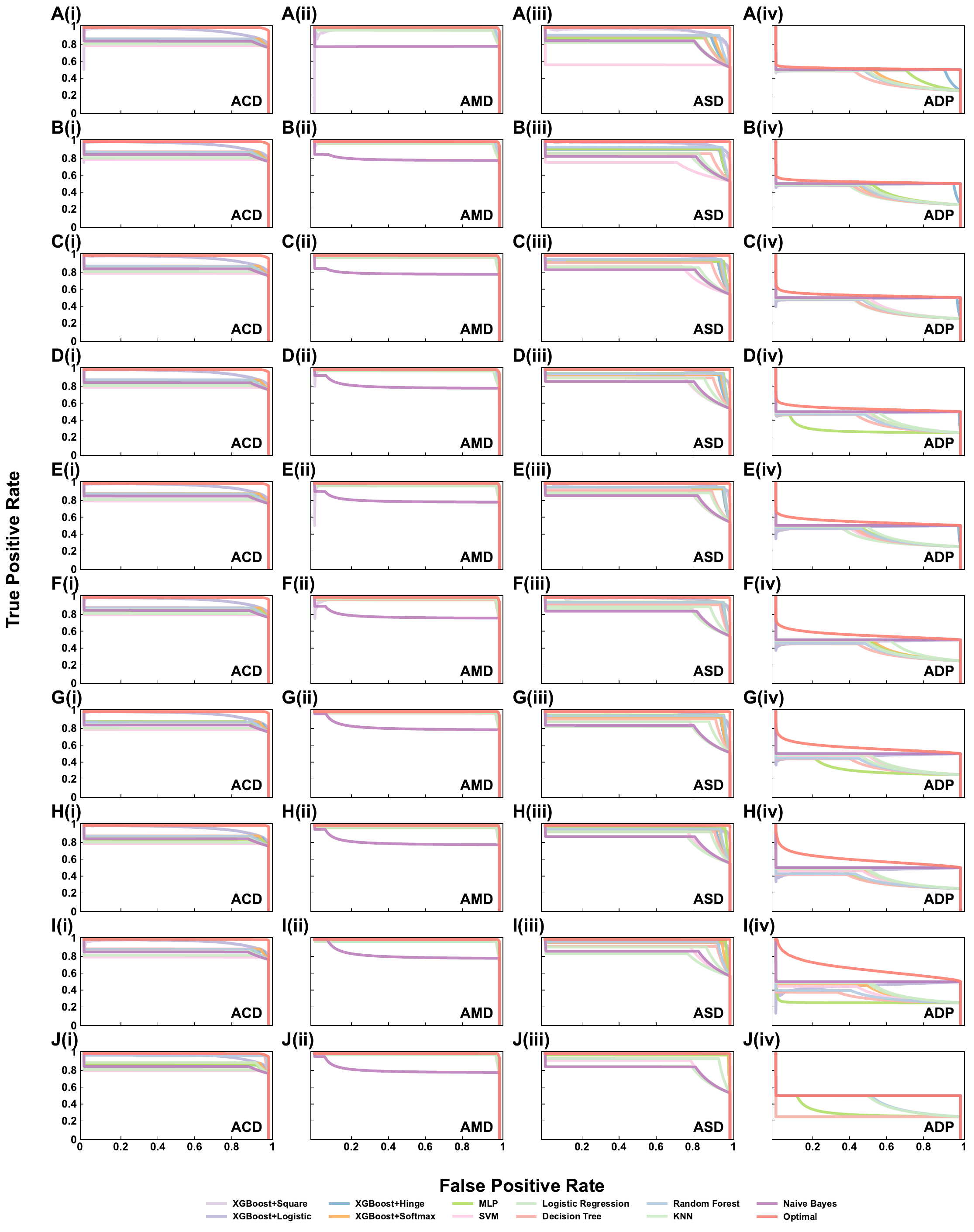}%
\caption{Exact upper bound of AP and corresponding optimal PR curves for 4 additional real-world datasets (ACD, AMD, ASD and ADP) when $|\mathcal{S}_{train}|/|\mathcal{S}|=0.1$ (A), $|\mathcal{S}_{train}|/|\mathcal{S}|=0.2$ (B), $|\mathcal{S}_{train}|/|\mathcal{S}|=0.3$ (C), $|\mathcal{S}_{train}|/|\mathcal{S}|=0.4$ (D), $|\mathcal{S}_{train}|/|\mathcal{S}|=0.5$ (E), $|\mathcal{S}_{train}|/|\mathcal{S}|=0.6$ (F), $|\mathcal{S}_{train}|/|\mathcal{S}|=0.7$ (G), $|\mathcal{S}_{train}|/|\mathcal{S}|=0.8$ (H), $|\mathcal{S}_{train}|/|\mathcal{S}|=0.9$ (I) and $|\mathcal{S}_{train}|/|\mathcal{S}|=1$ (J). The binary classifiers we used in this experiment include XGBoost, MLP, SVM, Logistic Regresion, Decision Tree, Random Forest, KNN and Naive Bayes. Red curves represent the theoretical optimal ROC curves.}
\label{figS13-1}
\end{figure}

\begin{figure}
\centering
\includegraphics[width=.9\linewidth]{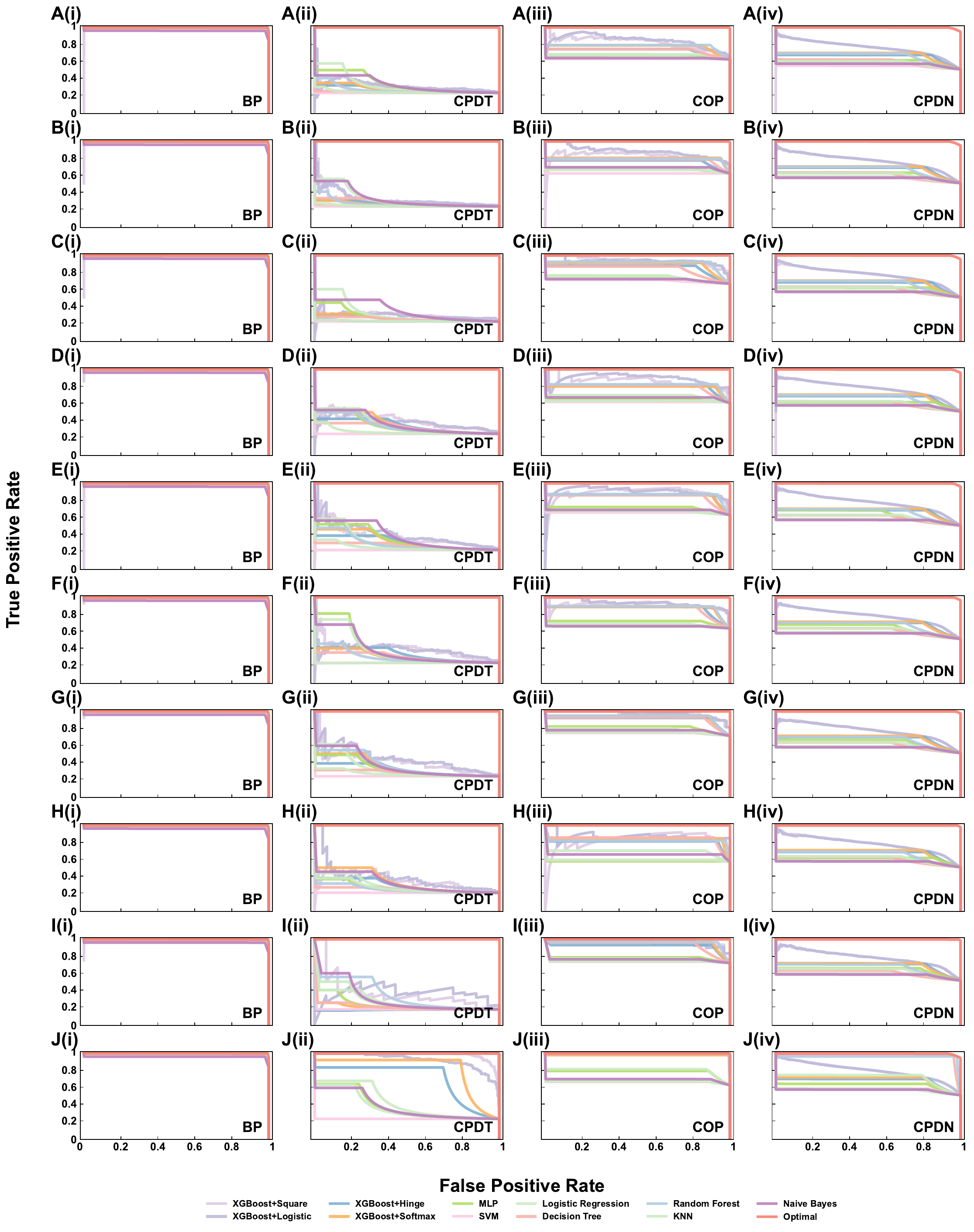}%
\caption{Exact upper bound of AP and corresponding optimal PR curves for 4 additional real-world datasets (BP, CPDT, COP and CPDN) when $|\mathcal{S}_{train}|/|\mathcal{S}|=0.1$ (A), $|\mathcal{S}_{train}|/|\mathcal{S}|=0.2$ (B), $|\mathcal{S}_{train}|/|\mathcal{S}|=0.3$ (C), $|\mathcal{S}_{train}|/|\mathcal{S}|=0.4$ (D), $|\mathcal{S}_{train}|/|\mathcal{S}|=0.5$ (E), $|\mathcal{S}_{train}|/|\mathcal{S}|=0.6$ (F), $|\mathcal{S}_{train}|/|\mathcal{S}|=0.7$ (G), $|\mathcal{S}_{train}|/|\mathcal{S}|=0.8$ (H), $|\mathcal{S}_{train}|/|\mathcal{S}|=0.9$ (I) and $|\mathcal{S}_{train}|/|\mathcal{S}|=1$ (J). The binary classifiers we used in this experiment include XGBoost, MLP, SVM, Logistic Regresion, Decision Tree, Random Forest, KNN and Naive Bayes. Red curves represent the theoretical optimal ROC curves.}
\label{figS13-2}
\end{figure}

\begin{figure}
\centering
\includegraphics[width=.9\linewidth]{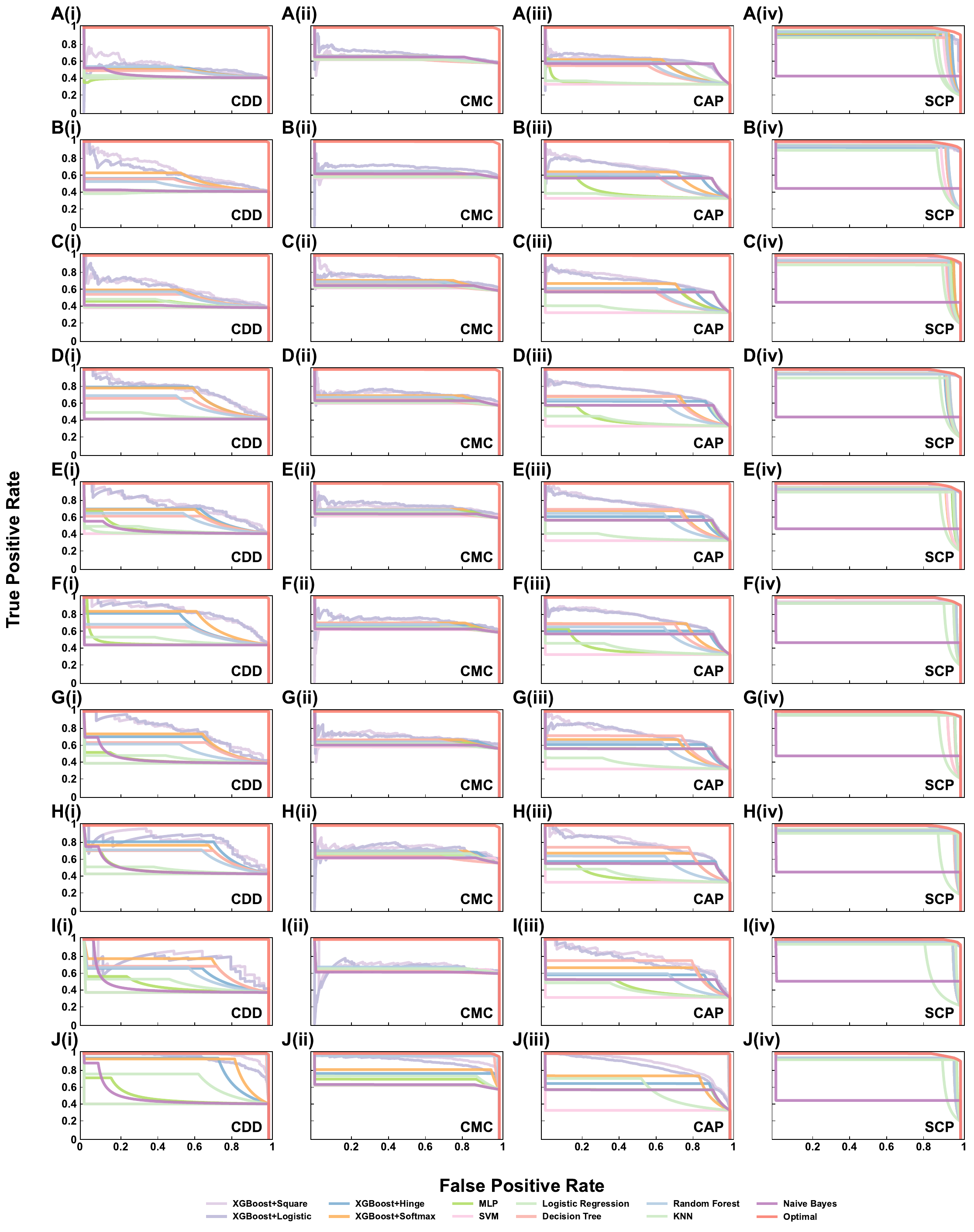}%
\caption{Exact upper bound of AP and corresponding optimal PR curves for 4 additional real-world datasets (CDD, CMC, CAP and SCP) when $|\mathcal{S}_{train}|/|\mathcal{S}|=0.1$ (A), $|\mathcal{S}_{train}|/|\mathcal{S}|=0.2$ (B), $|\mathcal{S}_{train}|/|\mathcal{S}|=0.3$ (C), $|\mathcal{S}_{train}|/|\mathcal{S}|=0.4$ (D), $|\mathcal{S}_{train}|/|\mathcal{S}|=0.5$ (E), $|\mathcal{S}_{train}|/|\mathcal{S}|=0.6$ (F), $|\mathcal{S}_{train}|/|\mathcal{S}|=0.7$ (G), $|\mathcal{S}_{train}|/|\mathcal{S}|=0.8$ (H), $|\mathcal{S}_{train}|/|\mathcal{S}|=0.9$ (I) and $|\mathcal{S}_{train}|/|\mathcal{S}|=1$ (J). The binary classifiers we used in this experiment include XGBoost, MLP, SVM, Logistic Regresion, Decision Tree, Random Forest, KNN and Naive Bayes. Red curves represent the theoretical optimal ROC curves.}
\label{figS13-3}
\end{figure}

\begin{figure}
\centering
\includegraphics[width=.9\linewidth]{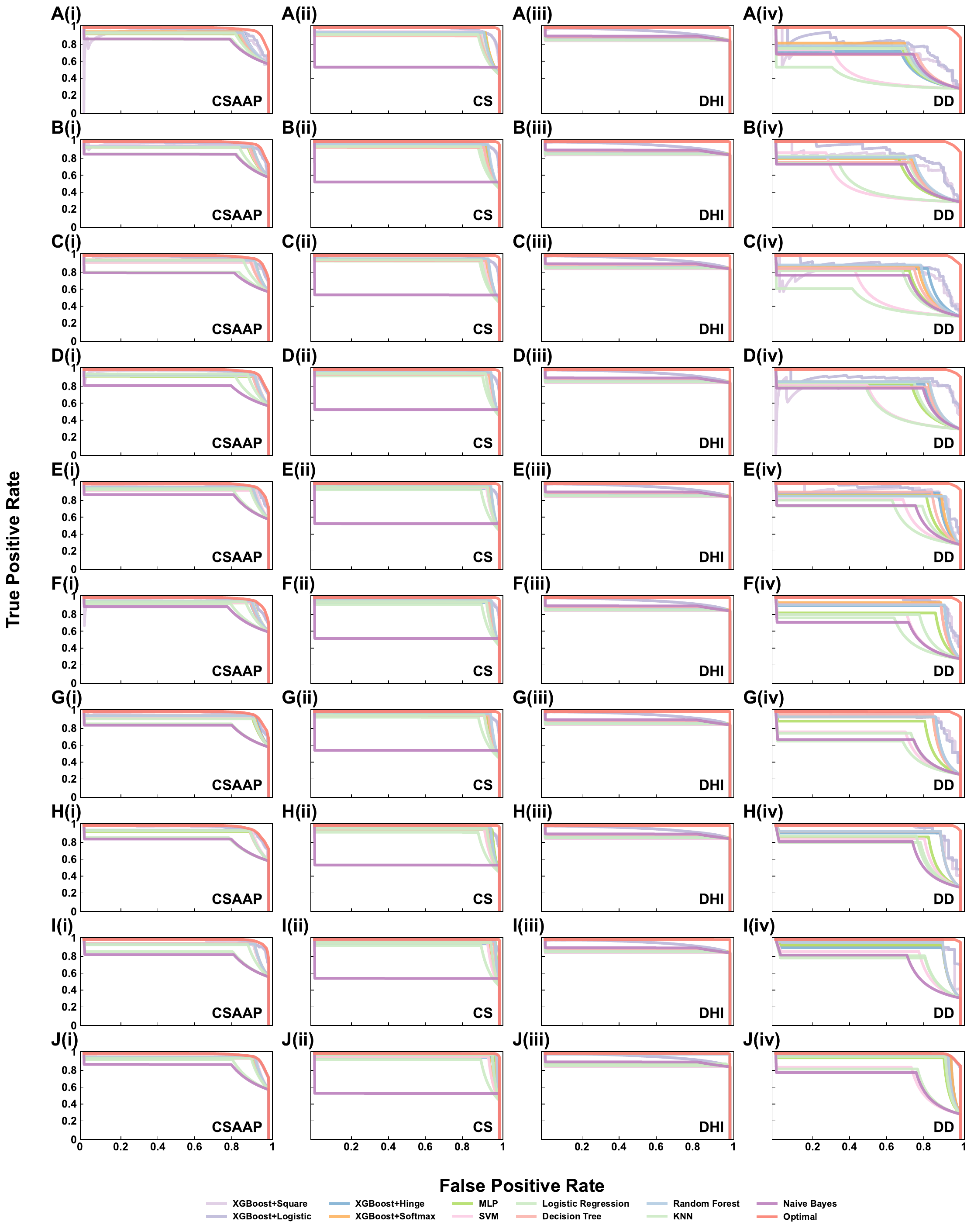}%
\caption{Exact upper bound of AP and corresponding optimal PR curves for 4 additional real-world datasets (CSAAP, CS, DHI and DD) when $|\mathcal{S}_{train}|/|\mathcal{S}|=0.1$ (A), $|\mathcal{S}_{train}|/|\mathcal{S}|=0.2$ (B), $|\mathcal{S}_{train}|/|\mathcal{S}|=0.3$ (C), $|\mathcal{S}_{train}|/|\mathcal{S}|=0.4$ (D), $|\mathcal{S}_{train}|/|\mathcal{S}|=0.5$ (E), $|\mathcal{S}_{train}|/|\mathcal{S}|=0.6$ (F), $|\mathcal{S}_{train}|/|\mathcal{S}|=0.7$ (G), $|\mathcal{S}_{train}|/|\mathcal{S}|=0.8$ (H), $|\mathcal{S}_{train}|/|\mathcal{S}|=0.9$ (I) and $|\mathcal{S}_{train}|/|\mathcal{S}|=1$ (J). The binary classifiers we used in this experiment include XGBoost, MLP, SVM, Logistic Regresion, Decision Tree, Random Forest, KNN and Naive Bayes. Red curves represent the theoretical optimal ROC curves.}
\label{figS13-4}
\end{figure}

\begin{figure}
\centering
\includegraphics[width=.9\linewidth]{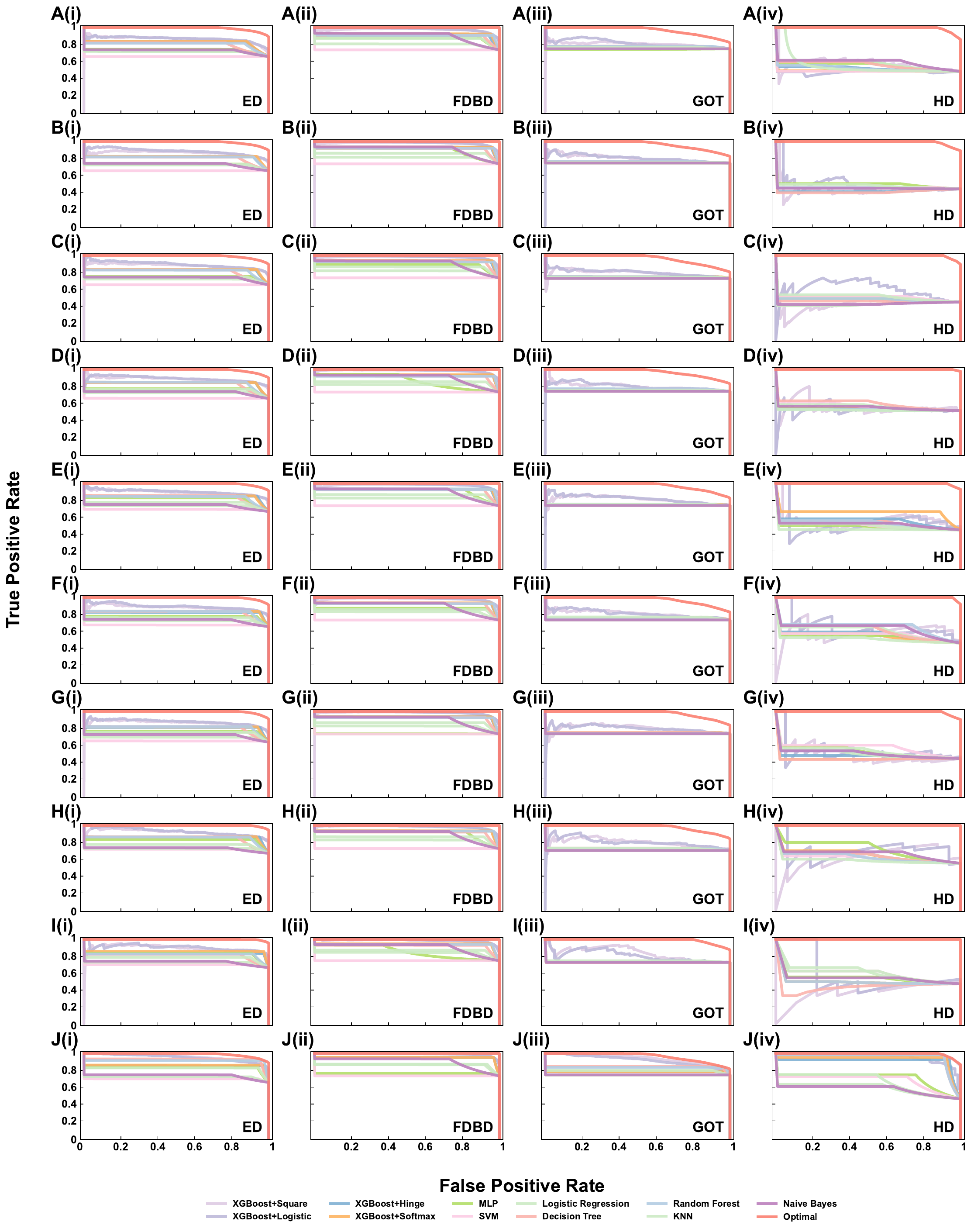}%
\caption{Exact upper bound of AP and corresponding optimal PR curves for 4 additional real-world datasets (ED, FDBD, GOT and HD) when $|\mathcal{S}_{train}|/|\mathcal{S}|=0.1$ (A), $|\mathcal{S}_{train}|/|\mathcal{S}|=0.2$ (B), $|\mathcal{S}_{train}|/|\mathcal{S}|=0.3$ (C), $|\mathcal{S}_{train}|/|\mathcal{S}|=0.4$ (D), $|\mathcal{S}_{train}|/|\mathcal{S}|=0.5$ (E), $|\mathcal{S}_{train}|/|\mathcal{S}|=0.6$ (F), $|\mathcal{S}_{train}|/|\mathcal{S}|=0.7$ (G), $|\mathcal{S}_{train}|/|\mathcal{S}|=0.8$ (H), $|\mathcal{S}_{train}|/|\mathcal{S}|=0.9$ (I) and $|\mathcal{S}_{train}|/|\mathcal{S}|=1$ (J). The binary classifiers we used in this experiment include XGBoost, MLP, SVM, Logistic Regresion, Decision Tree, Random Forest, KNN and Naive Bayes. Red curves represent the theoretical optimal ROC curves.}
\label{figS13-5}
\end{figure}

\begin{figure}
\centering
\includegraphics[width=.9\linewidth]{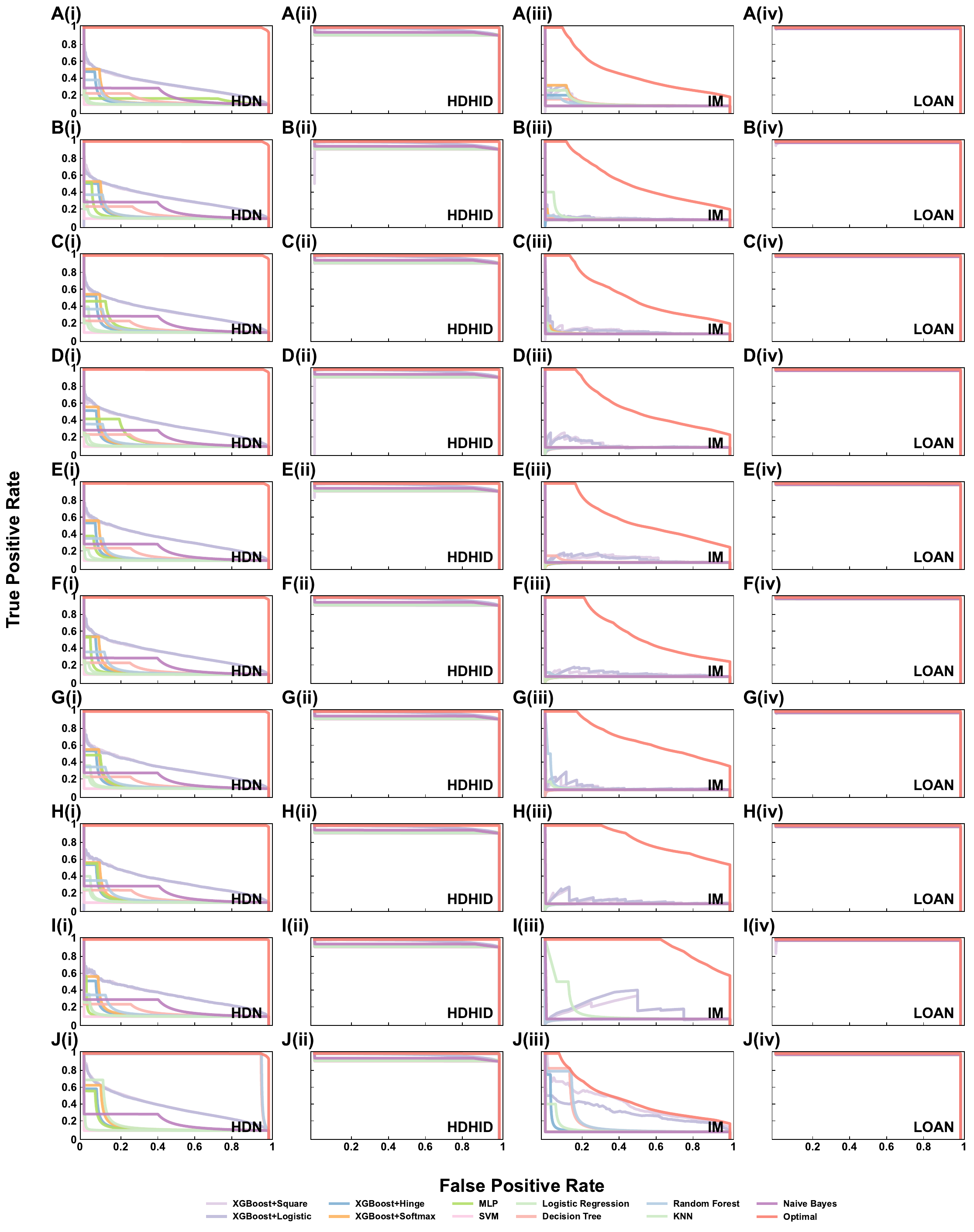}%
\caption{Exact upper bound of AP and corresponding optimal PR curves for 4 additional real-world datasets (HDN, HDHID, IM and LOAN) when $|\mathcal{S}_{train}|/|\mathcal{S}|=0.1$ (A), $|\mathcal{S}_{train}|/|\mathcal{S}|=0.2$ (B), $|\mathcal{S}_{train}|/|\mathcal{S}|=0.3$ (C), $|\mathcal{S}_{train}|/|\mathcal{S}|=0.4$ (D), $|\mathcal{S}_{train}|/|\mathcal{S}|=0.5$ (E), $|\mathcal{S}_{train}|/|\mathcal{S}|=0.6$ (F), $|\mathcal{S}_{train}|/|\mathcal{S}|=0.7$ (G), $|\mathcal{S}_{train}|/|\mathcal{S}|=0.8$ (H), $|\mathcal{S}_{train}|/|\mathcal{S}|=0.9$ (I) and $|\mathcal{S}_{train}|/|\mathcal{S}|=1$ (J). The binary classifiers we used in this experiment include XGBoost, MLP, SVM, Logistic Regresion, Decision Tree, Random Forest, KNN and Naive Bayes. Red curves represent the theoretical optimal ROC curves.}
\label{figS13-6}
\end{figure}

\begin{figure}
\centering
\includegraphics[width=.9\linewidth]{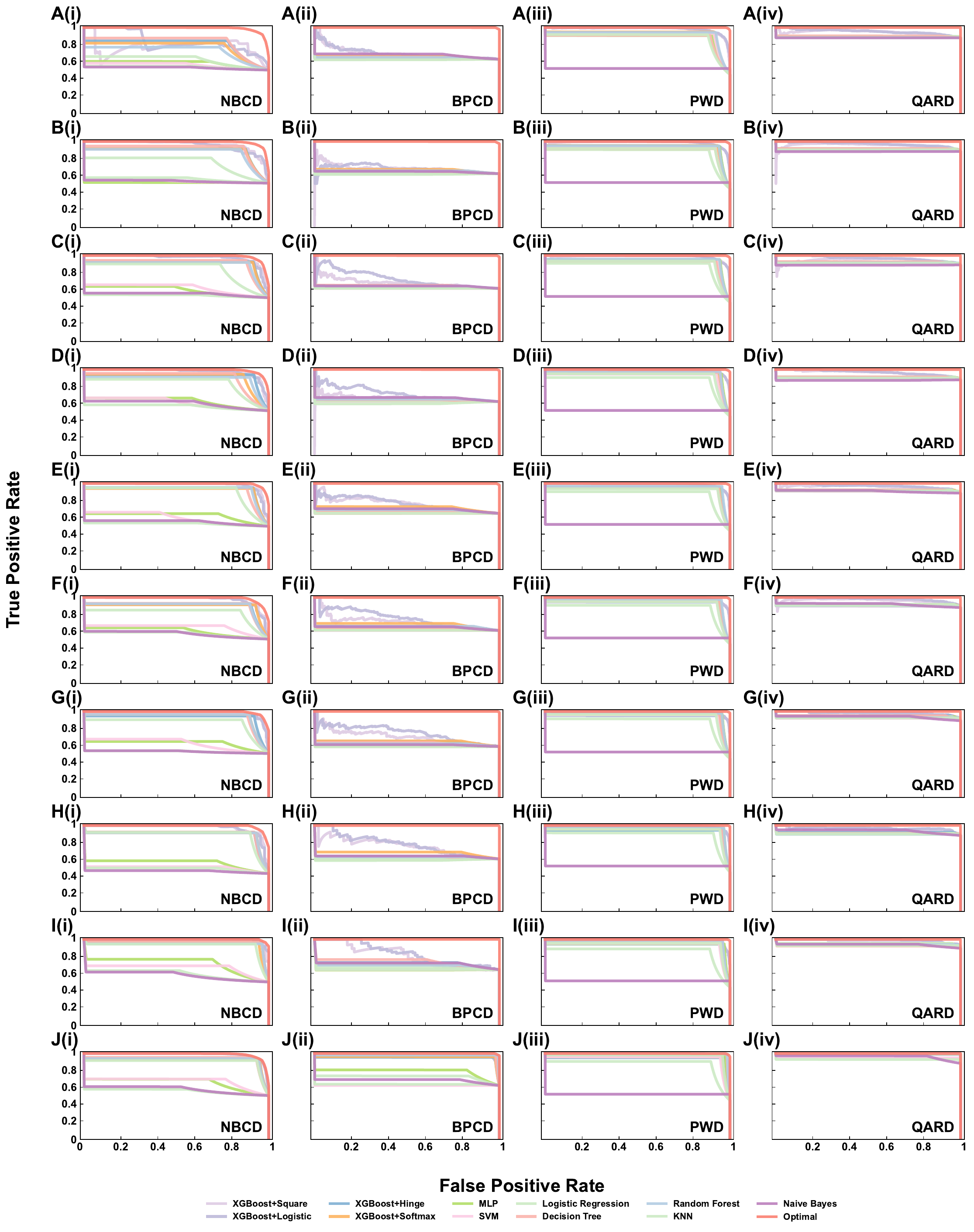}%
\caption{Exact upper bound of AP and corresponding optimal PR curves for 4 additional real-world datasets (NBCD, BPCD, PWD and QARD) when $|\mathcal{S}_{train}|/|\mathcal{S}|=0.1$ (A), $|\mathcal{S}_{train}|/|\mathcal{S}|=0.2$ (B), $|\mathcal{S}_{train}|/|\mathcal{S}|=0.3$ (C), $|\mathcal{S}_{train}|/|\mathcal{S}|=0.4$ (D), $|\mathcal{S}_{train}|/|\mathcal{S}|=0.5$ (E), $|\mathcal{S}_{train}|/|\mathcal{S}|=0.6$ (F), $|\mathcal{S}_{train}|/|\mathcal{S}|=0.7$ (G), $|\mathcal{S}_{train}|/|\mathcal{S}|=0.8$ (H), $|\mathcal{S}_{train}|/|\mathcal{S}|=0.9$ (I) and $|\mathcal{S}_{train}|/|\mathcal{S}|=1$ (J). The binary classifiers we used in this experiment include XGBoost, MLP, SVM, Logistic Regresion, Decision Tree, Random Forest, KNN and Naive Bayes. Red curves represent the theoretical optimal ROC curves.}
\label{figS13-7}
\end{figure}

\begin{figure}
\centering
\includegraphics[width=.9\linewidth]{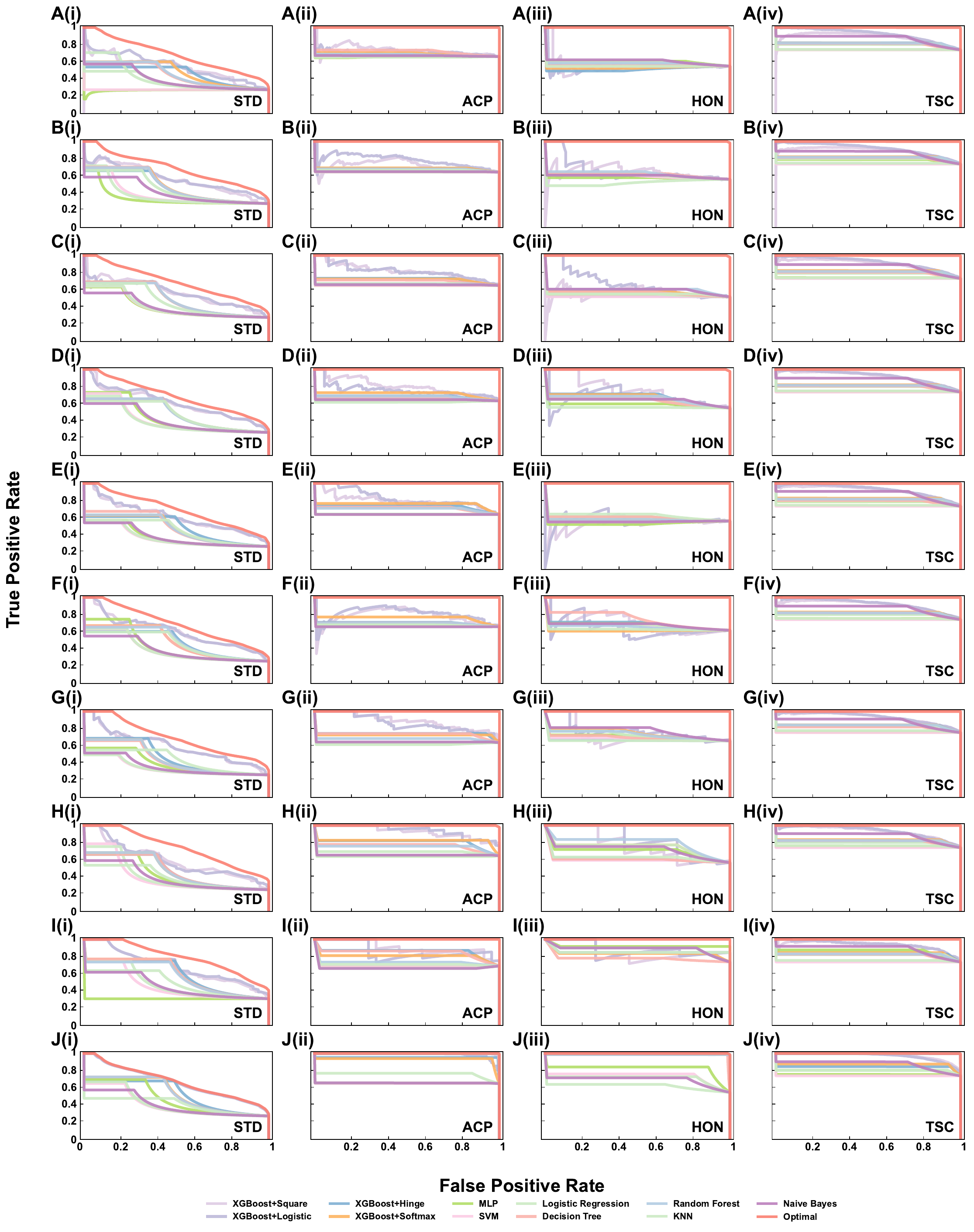}%
\caption{Exact upper bound of AP and corresponding optimal PR curves for 4 additional real-world datasets (STD, ACP, HON and TSC) when $|\mathcal{S}_{train}|/|\mathcal{S}|=0.1$ (A), $|\mathcal{S}_{train}|/|\mathcal{S}|=0.2$ (B), $|\mathcal{S}_{train}|/|\mathcal{S}|=0.3$ (C), $|\mathcal{S}_{train}|/|\mathcal{S}|=0.4$ (D), $|\mathcal{S}_{train}|/|\mathcal{S}|=0.5$ (E), $|\mathcal{S}_{train}|/|\mathcal{S}|=0.6$ (F), $|\mathcal{S}_{train}|/|\mathcal{S}|=0.7$ (G), $|\mathcal{S}_{train}|/|\mathcal{S}|=0.8$ (H), $|\mathcal{S}_{train}|/|\mathcal{S}|=0.9$ (I) and $|\mathcal{S}_{train}|/|\mathcal{S}|=1$ (J). The binary classifiers we used in this experiment include XGBoost, MLP, SVM, Logistic Regresion, Decision Tree, Random Forest, KNN and Naive Bayes. Red curves represent the theoretical optimal ROC curves.}
\label{figS13-8}
\end{figure}

\begin{figure}
\centering
\includegraphics[width=.75\linewidth]{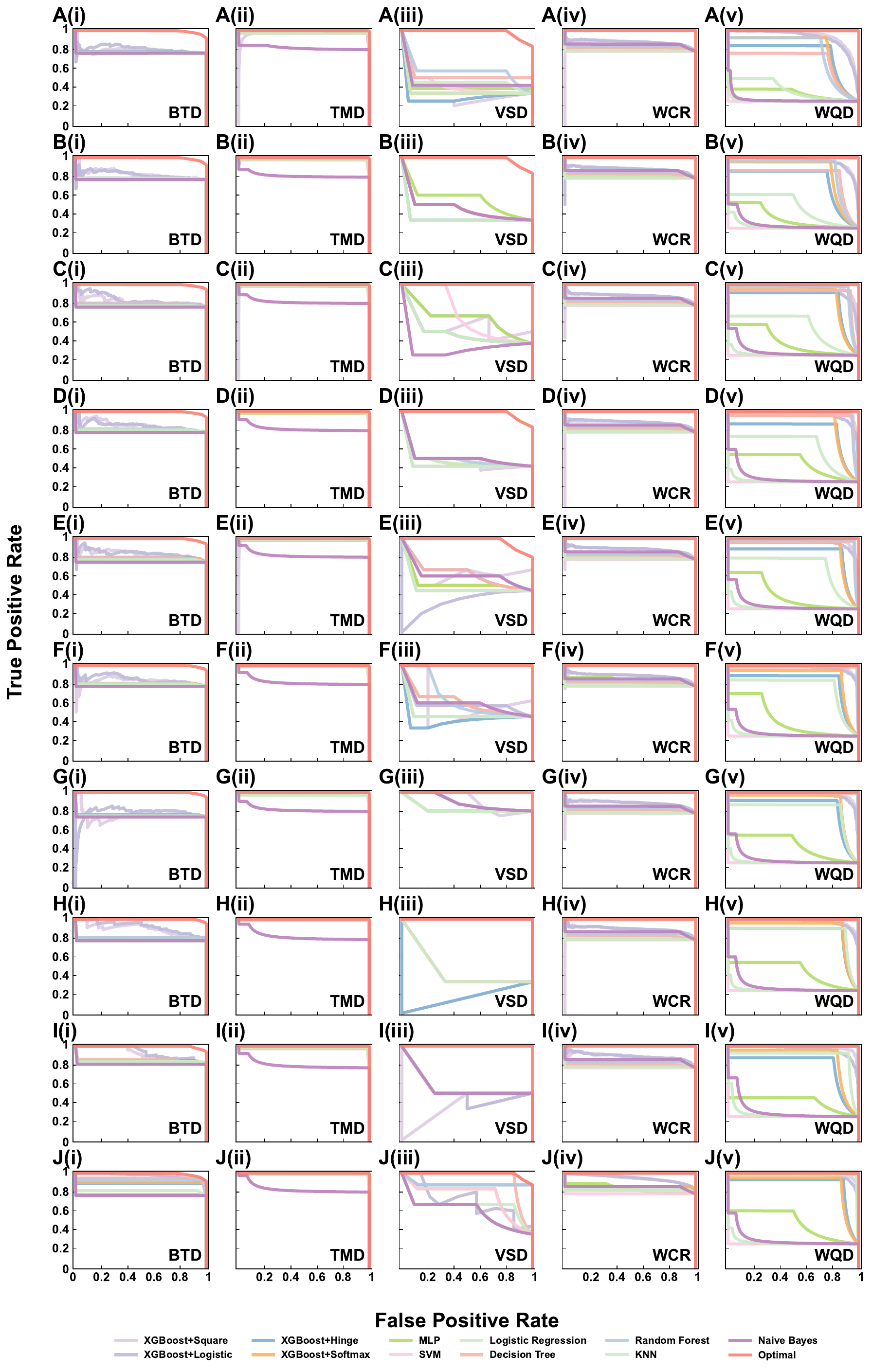}%
\caption{Exact upper bound of AP and corresponding optimal PR curves for 5 additional real-world datasets (BTD, TMD, VSD, WCR and WQD) when $|\mathcal{S}_{train}|/|\mathcal{S}|=0.1$ (A), $|\mathcal{S}_{train}|/|\mathcal{S}|=0.2$ (B), $|\mathcal{S}_{train}|/|\mathcal{S}|=0.3$ (C), $|\mathcal{S}_{train}|/|\mathcal{S}|=0.4$ (D), $|\mathcal{S}_{train}|/|\mathcal{S}|=0.5$ (E), $|\mathcal{S}_{train}|/|\mathcal{S}|=0.6$ (F), $|\mathcal{S}_{train}|/|\mathcal{S}|=0.7$ (G), $|\mathcal{S}_{train}|/|\mathcal{S}|=0.8$ (H), $|\mathcal{S}_{train}|/|\mathcal{S}|=0.9$ (I) and $|\mathcal{S}_{train}|/|\mathcal{S}|=1$ (J). The binary classifiers we used in this experiment include XGBoost, MLP, SVM, Logistic Regresion, Decision Tree, Random Forest, KNN and Naive Bayes. Red curves represent the theoretical optimal ROC curves.}
\label{figS13-9}
\end{figure}

\begin{figure}
\centering
\includegraphics[width=.85\linewidth]{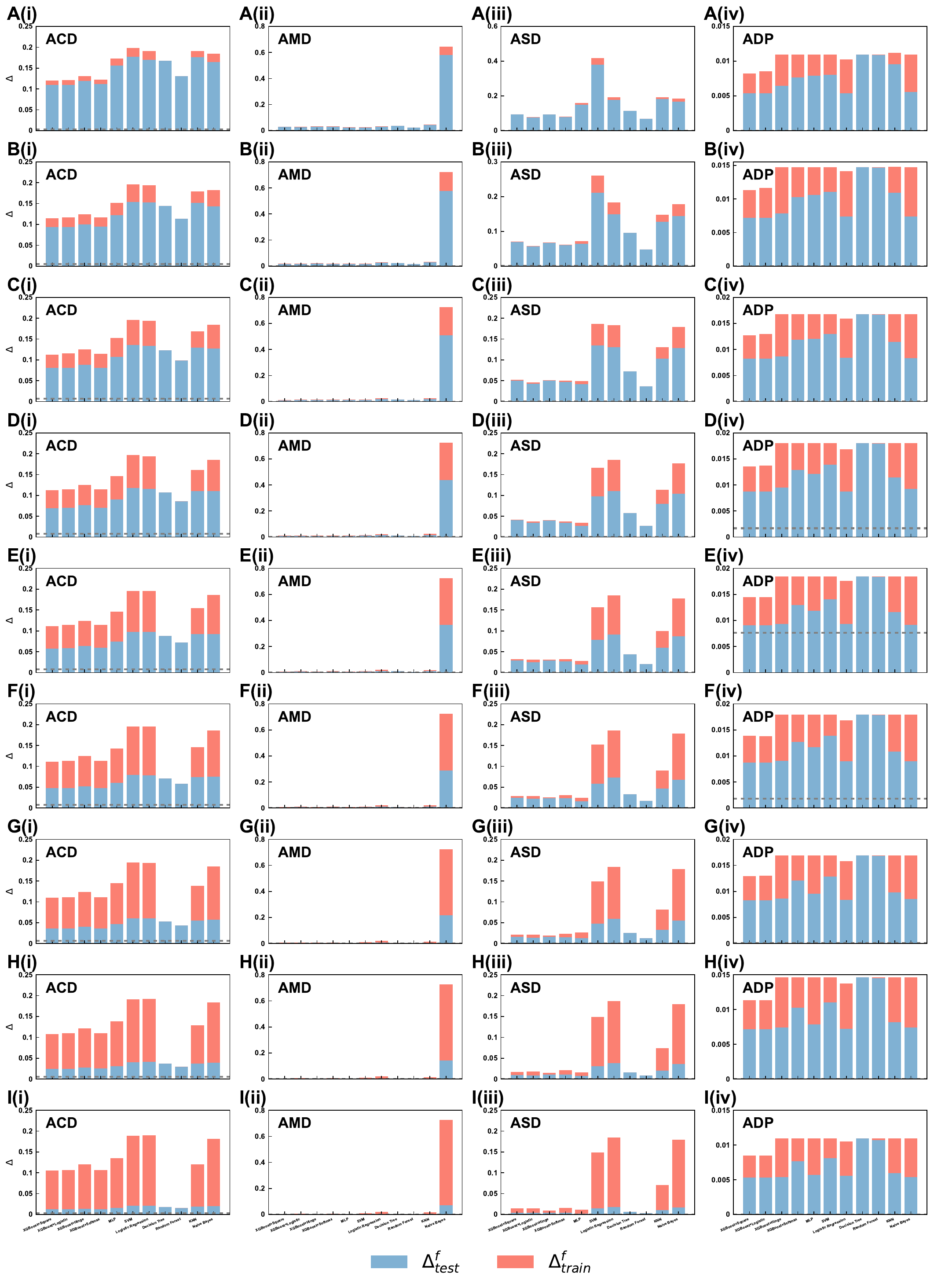}%
\caption{The loss errors of for 4 additional datasets (ACD, AMD, ASD and ADP) in training ($\Delta_{train}^{f}$) and test sets ($\Delta_{test}^{f}$) when $|\mathcal{S}_{train}|/|\mathcal{S}|=0.1$ (A), $|\mathcal{S}_{train}|/|\mathcal{S}|=0.2$ (B), $|\mathcal{S}_{train}|/|\mathcal{S}|=0.3$ (C), $|\mathcal{S}_{train}|/|\mathcal{S}|=0.4$ (D), $|\mathcal{S}_{train}|/|\mathcal{S}|=0.5$ (E), $|\mathcal{S}_{train}|/|\mathcal{S}|=0.6$ (F), $|\mathcal{S}_{train}|/|\mathcal{S}|=0.7$ (G), $|\mathcal{S}_{train}|/|\mathcal{S}|=0.8$ (H), $|\mathcal{S}_{train}|/|\mathcal{S}|=0.9$ (I). Dash line represents the expected error of optimal classier based on Eq. \ref{min_delta}.  }
\label{figS14-1}
\end{figure}

\begin{figure}
\centering
\includegraphics[width=.85\linewidth]{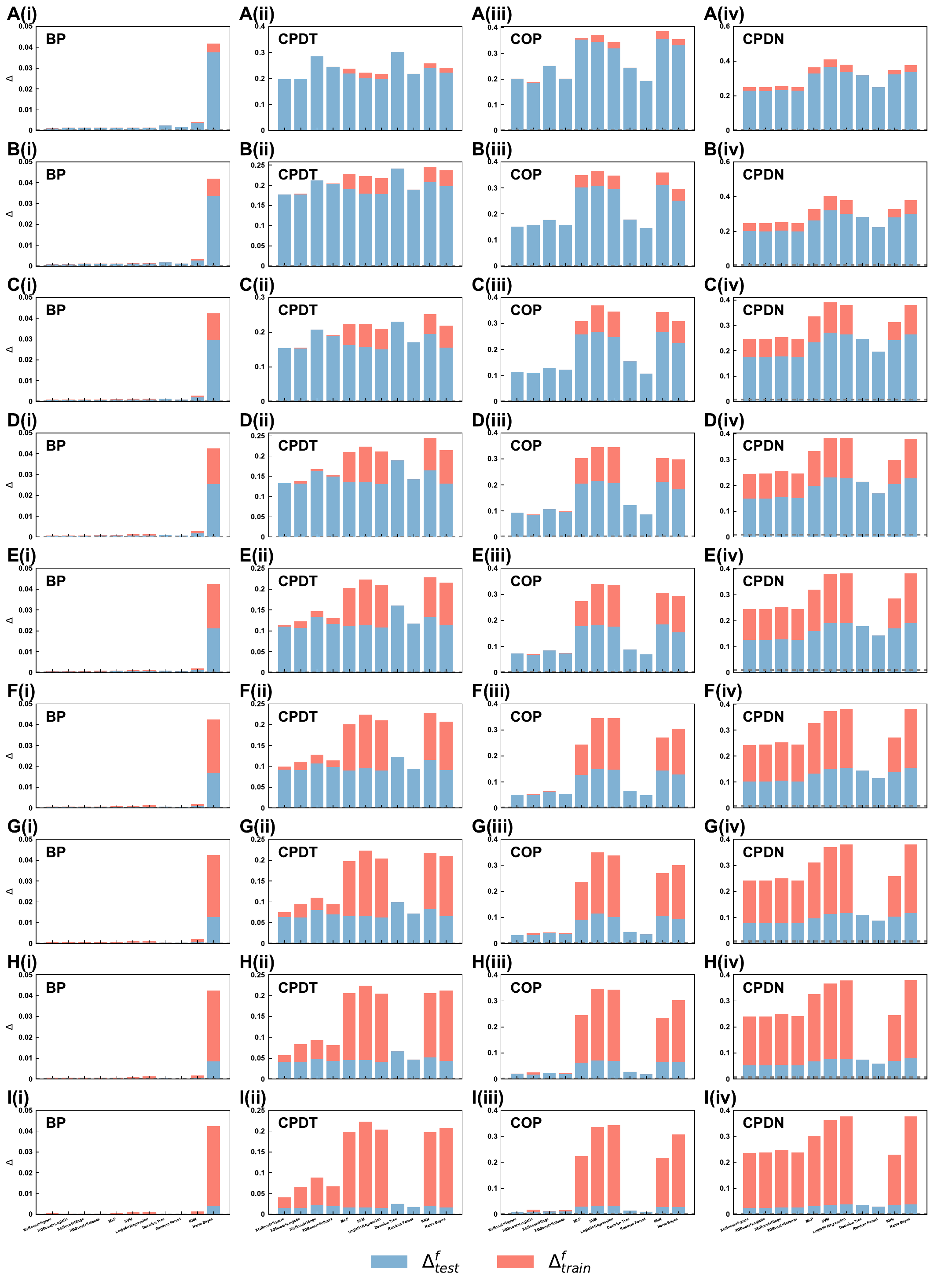}%
\caption{The loss errors of for 4 additional datasets BP, CPDT, COP and CPDN) in training ($\Delta_{train}^{f}$) and test sets ($\Delta_{test}^{f}$) when $|\mathcal{S}_{train}|/|\mathcal{S}|=0.1$ (A), $|\mathcal{S}_{train}|/|\mathcal{S}|=0.2$ (B), $|\mathcal{S}_{train}|/|\mathcal{S}|=0.3$ (C), $|\mathcal{S}_{train}|/|\mathcal{S}|=0.4$ (D), $|\mathcal{S}_{train}|/|\mathcal{S}|=0.5$ (E), $|\mathcal{S}_{train}|/|\mathcal{S}|=0.6$ (F), $|\mathcal{S}_{train}|/|\mathcal{S}|=0.7$ (G), $|\mathcal{S}_{train}|/|\mathcal{S}|=0.8$ (H), $|\mathcal{S}_{train}|/|\mathcal{S}|=0.9$ (I). Dash line represents the expected error of optimal classier based on Eq. \ref{min_delta}.  }
\label{figS14-2}
\end{figure}

\begin{figure}
\centering
\includegraphics[width=.85\linewidth]{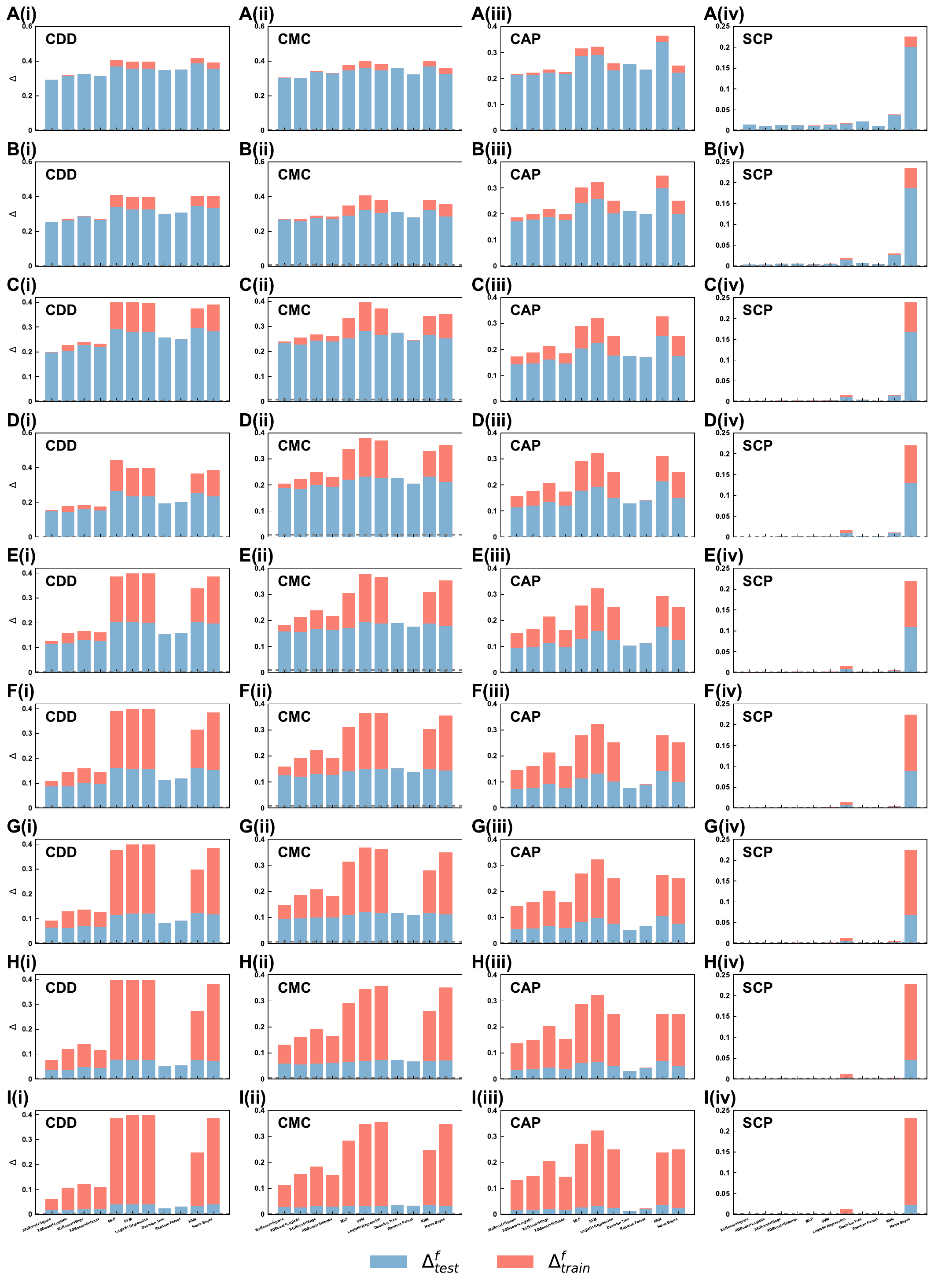}%
\caption{The loss errors of for 4 additional datasets (CDD, CMC, CAP and SCP) in training ($\Delta_{train}^{f}$) and test sets ($\Delta_{test}^{f}$) when $|\mathcal{S}_{train}|/|\mathcal{S}|=0.1$ (A), $|\mathcal{S}_{train}|/|\mathcal{S}|=0.2$ (B), $|\mathcal{S}_{train}|/|\mathcal{S}|=0.3$ (C), $|\mathcal{S}_{train}|/|\mathcal{S}|=0.4$ (D), $|\mathcal{S}_{train}|/|\mathcal{S}|=0.5$ (E), $|\mathcal{S}_{train}|/|\mathcal{S}|=0.6$ (F), $|\mathcal{S}_{train}|/|\mathcal{S}|=0.7$ (G), $|\mathcal{S}_{train}|/|\mathcal{S}|=0.8$ (H), $|\mathcal{S}_{train}|/|\mathcal{S}|=0.9$ (I). Dash line represents the expected error of optimal classier based on Eq. \ref{min_delta}.  }
\label{figS14-3}
\end{figure}

\begin{figure}
\centering
\includegraphics[width=.85\linewidth]{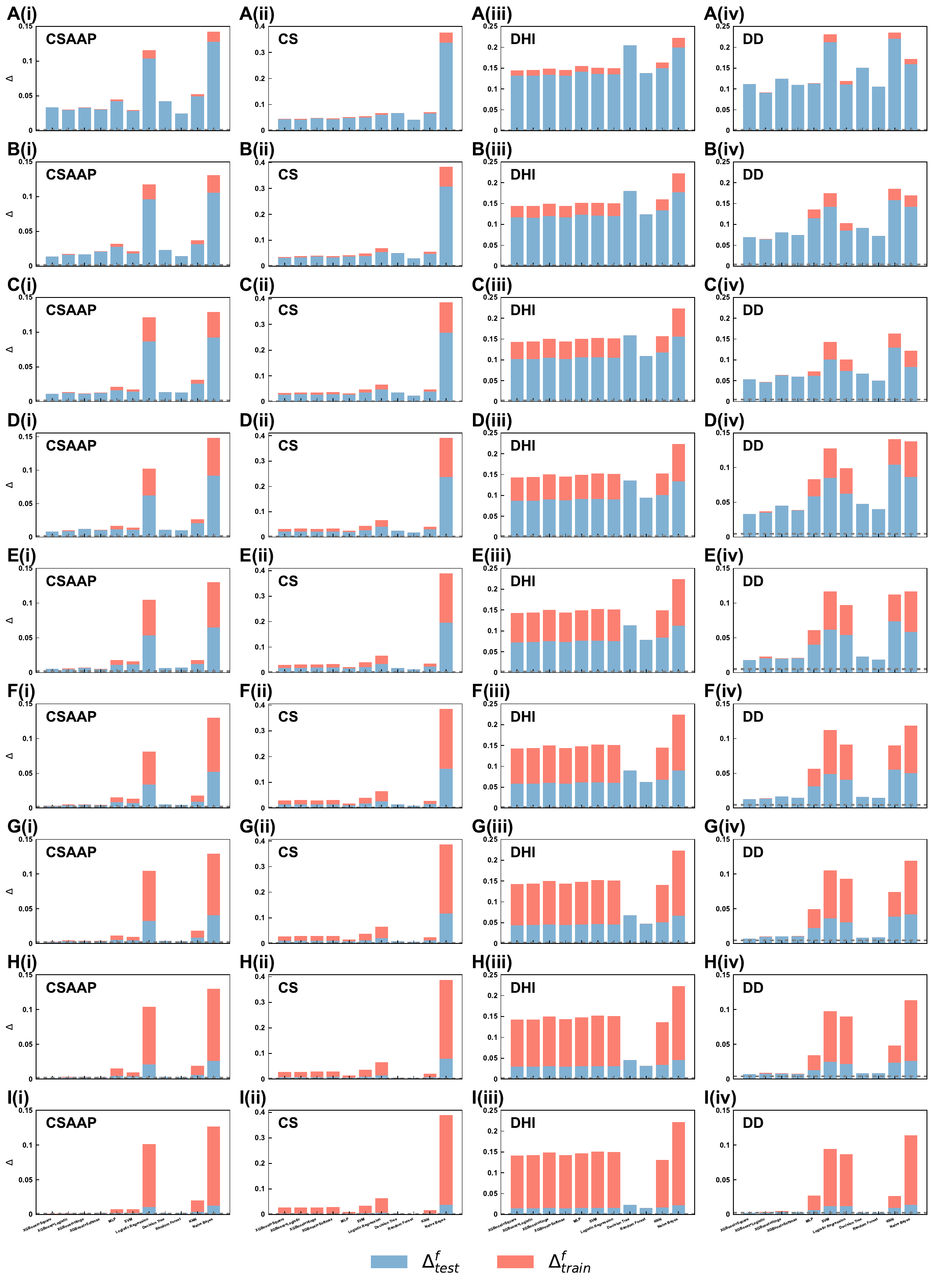}%
\caption{The loss errors of for 4 additional datasets (CSAAP, CS, DHI and DD) in training ($\Delta_{train}^{f}$) and test sets ($\Delta_{test}^{f}$) when $|\mathcal{S}_{train}|/|\mathcal{S}|=0.1$ (A), $|\mathcal{S}_{train}|/|\mathcal{S}|=0.2$ (B), $|\mathcal{S}_{train}|/|\mathcal{S}|=0.3$ (C), $|\mathcal{S}_{train}|/|\mathcal{S}|=0.4$ (D), $|\mathcal{S}_{train}|/|\mathcal{S}|=0.5$ (E), $|\mathcal{S}_{train}|/|\mathcal{S}|=0.6$ (F), $|\mathcal{S}_{train}|/|\mathcal{S}|=0.7$ (G), $|\mathcal{S}_{train}|/|\mathcal{S}|=0.8$ (H), $|\mathcal{S}_{train}|/|\mathcal{S}|=0.9$ (I). Dash line represents the expected error of optimal classier based on Eq. \ref{min_delta}.  }
\label{figS14-4}
\end{figure}

\begin{figure}
\centering
\includegraphics[width=.85\linewidth]{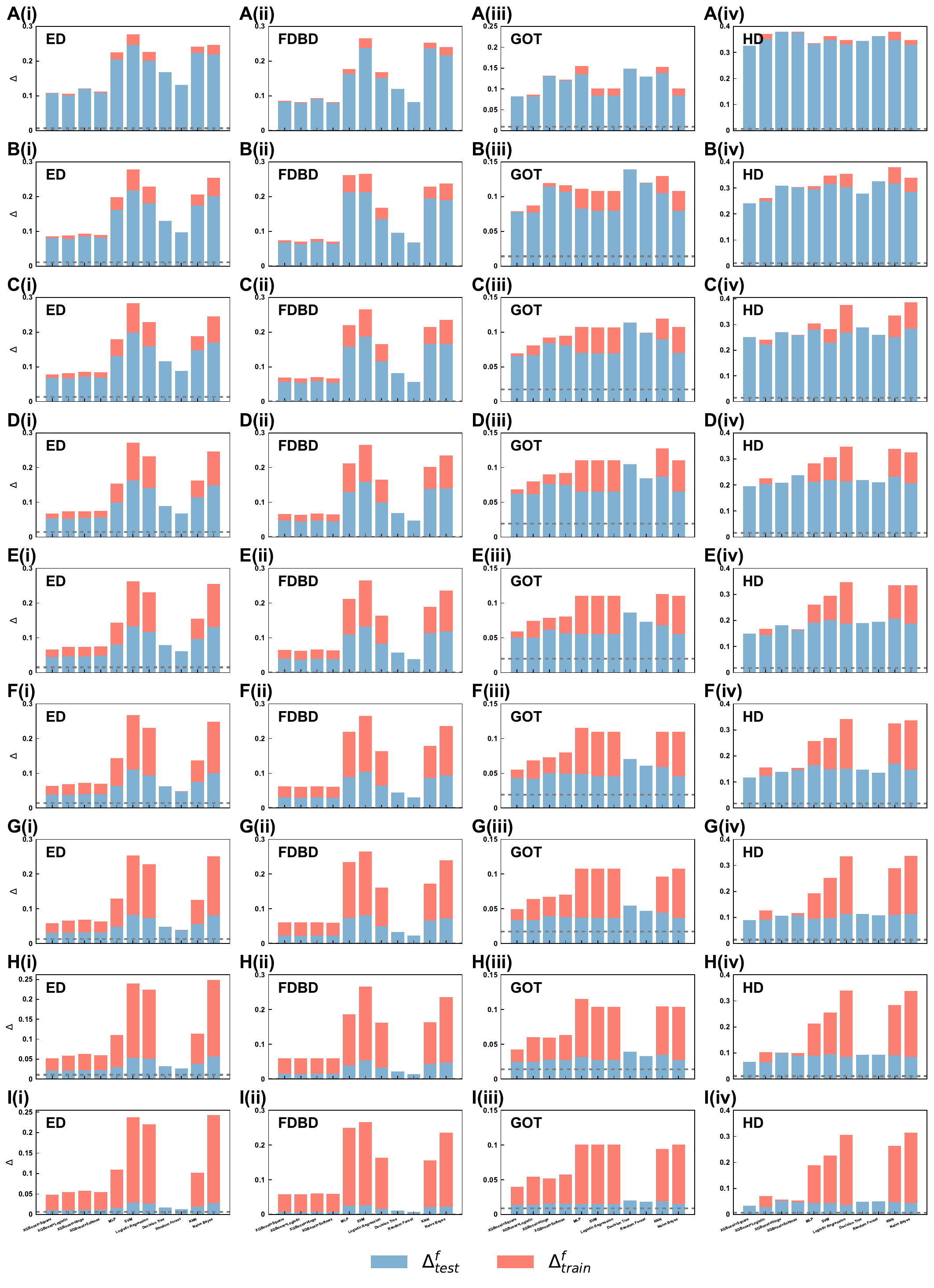}%
\caption{The loss errors of for 4 additional datasets (ED, FDBD, GOT and HD) in training ($\Delta_{train}^{f}$) and test sets ($\Delta_{test}^{f}$) when $|\mathcal{S}_{train}|/|\mathcal{S}|=0.1$ (A), $|\mathcal{S}_{train}|/|\mathcal{S}|=0.2$ (B), $|\mathcal{S}_{train}|/|\mathcal{S}|=0.3$ (C), $|\mathcal{S}_{train}|/|\mathcal{S}|=0.4$ (D), $|\mathcal{S}_{train}|/|\mathcal{S}|=0.5$ (E), $|\mathcal{S}_{train}|/|\mathcal{S}|=0.6$ (F), $|\mathcal{S}_{train}|/|\mathcal{S}|=0.7$ (G), $|\mathcal{S}_{train}|/|\mathcal{S}|=0.8$ (H), $|\mathcal{S}_{train}|/|\mathcal{S}|=0.9$ (I). Dash line represents the expected error of optimal classier based on Eq. \ref{min_delta}.  }
\label{figS14-5}
\end{figure}

\begin{figure}
\centering
\includegraphics[width=.85\linewidth]{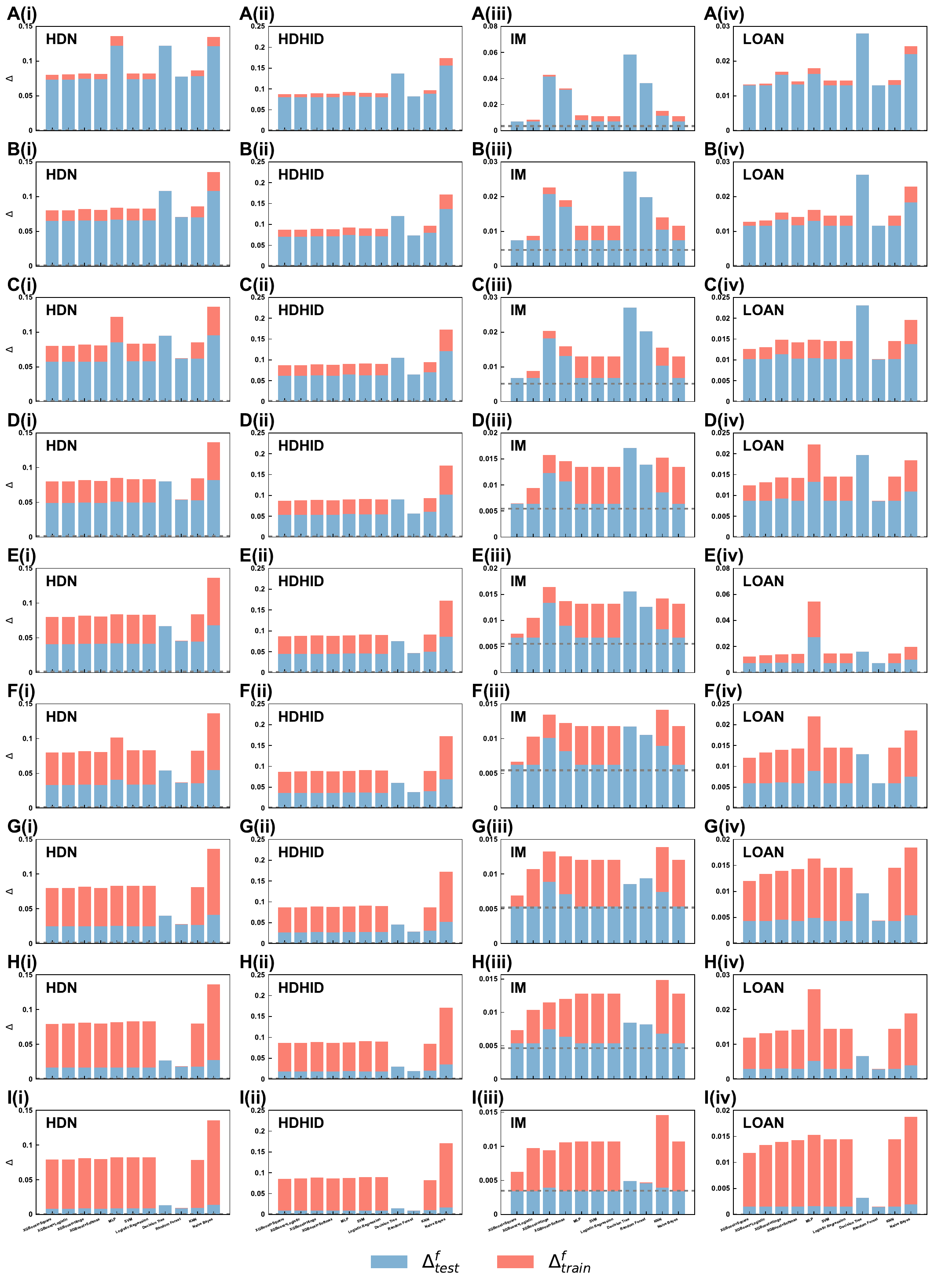}%
\caption{The loss errors of for 4 additional datasets (HDN, HDHID, IM and LOAN) in training ($\Delta_{train}^{f}$) and test sets ($\Delta_{test}^{f}$) when $|\mathcal{S}_{train}|/|\mathcal{S}|=0.1$ (A), $|\mathcal{S}_{train}|/|\mathcal{S}|=0.2$ (B), $|\mathcal{S}_{train}|/|\mathcal{S}|=0.3$ (C), $|\mathcal{S}_{train}|/|\mathcal{S}|=0.4$ (D), $|\mathcal{S}_{train}|/|\mathcal{S}|=0.5$ (E), $|\mathcal{S}_{train}|/|\mathcal{S}|=0.6$ (F), $|\mathcal{S}_{train}|/|\mathcal{S}|=0.7$ (G), $|\mathcal{S}_{train}|/|\mathcal{S}|=0.8$ (H), $|\mathcal{S}_{train}|/|\mathcal{S}|=0.9$ (I). Dash line represents the expected error of optimal classier based on Eq. \ref{min_delta}.  }
\label{figS14-6}
\end{figure}

\begin{figure}
\centering
\includegraphics[width=.85\linewidth]{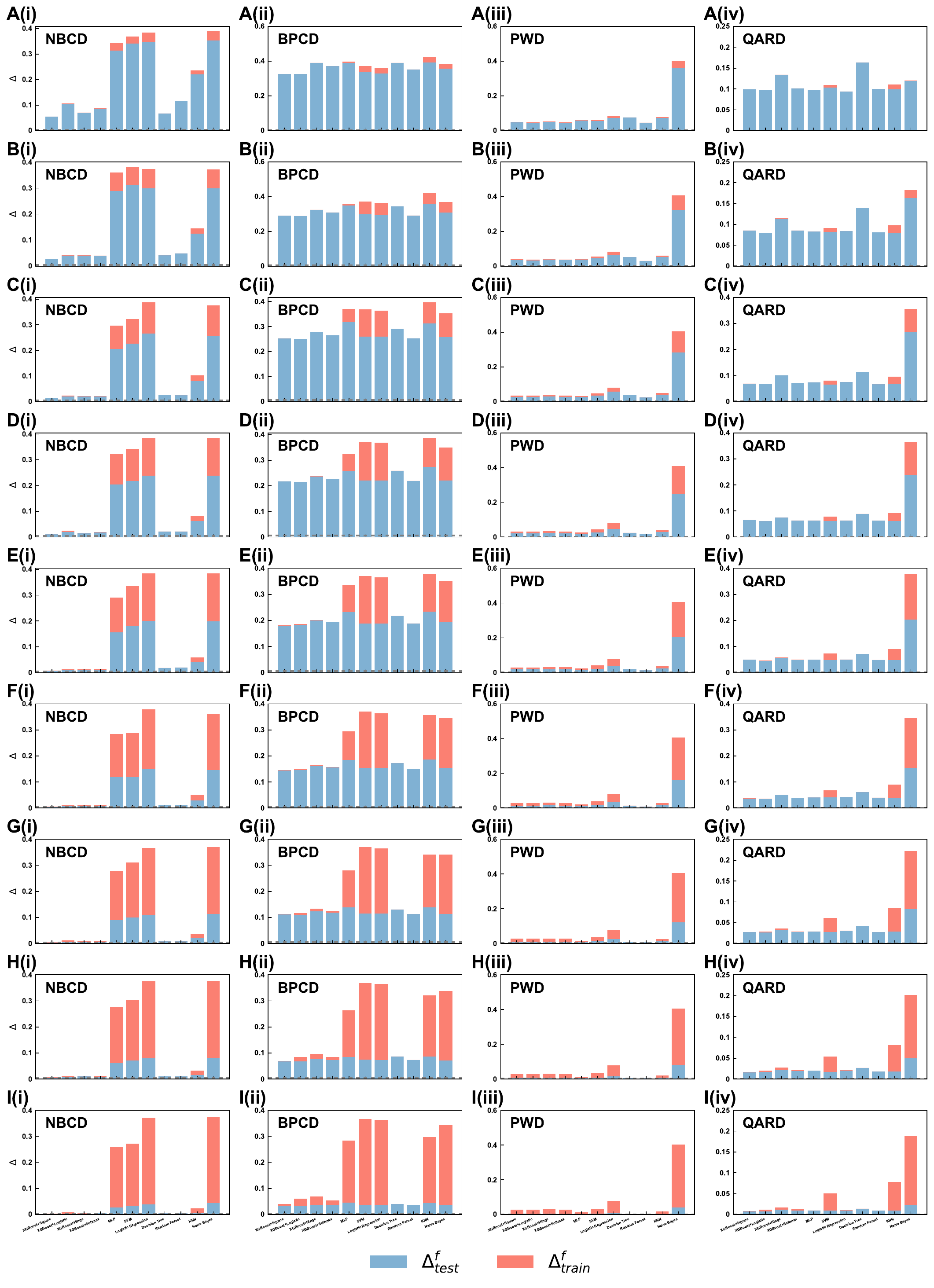}%
\caption{The loss errors of for 4 additional datasets (NBCD, BPCD, PWD and QARD) in training ($\Delta_{train}^{f}$) and test sets ($\Delta_{test}^{f}$) when $|\mathcal{S}_{train}|/|\mathcal{S}|=0.1$ (A), $|\mathcal{S}_{train}|/|\mathcal{S}|=0.2$ (B), $|\mathcal{S}_{train}|/|\mathcal{S}|=0.3$ (C), $|\mathcal{S}_{train}|/|\mathcal{S}|=0.4$ (D), $|\mathcal{S}_{train}|/|\mathcal{S}|=0.5$ (E), $|\mathcal{S}_{train}|/|\mathcal{S}|=0.6$ (F), $|\mathcal{S}_{train}|/|\mathcal{S}|=0.7$ (G), $|\mathcal{S}_{train}|/|\mathcal{S}|=0.8$ (H), $|\mathcal{S}_{train}|/|\mathcal{S}|=0.9$ (I). Dash line represents the expected error of optimal classier based on Eq. \ref{min_delta}.  }
\label{figS14-7}
\end{figure}

\begin{figure}
\centering
\includegraphics[width=.85\linewidth]{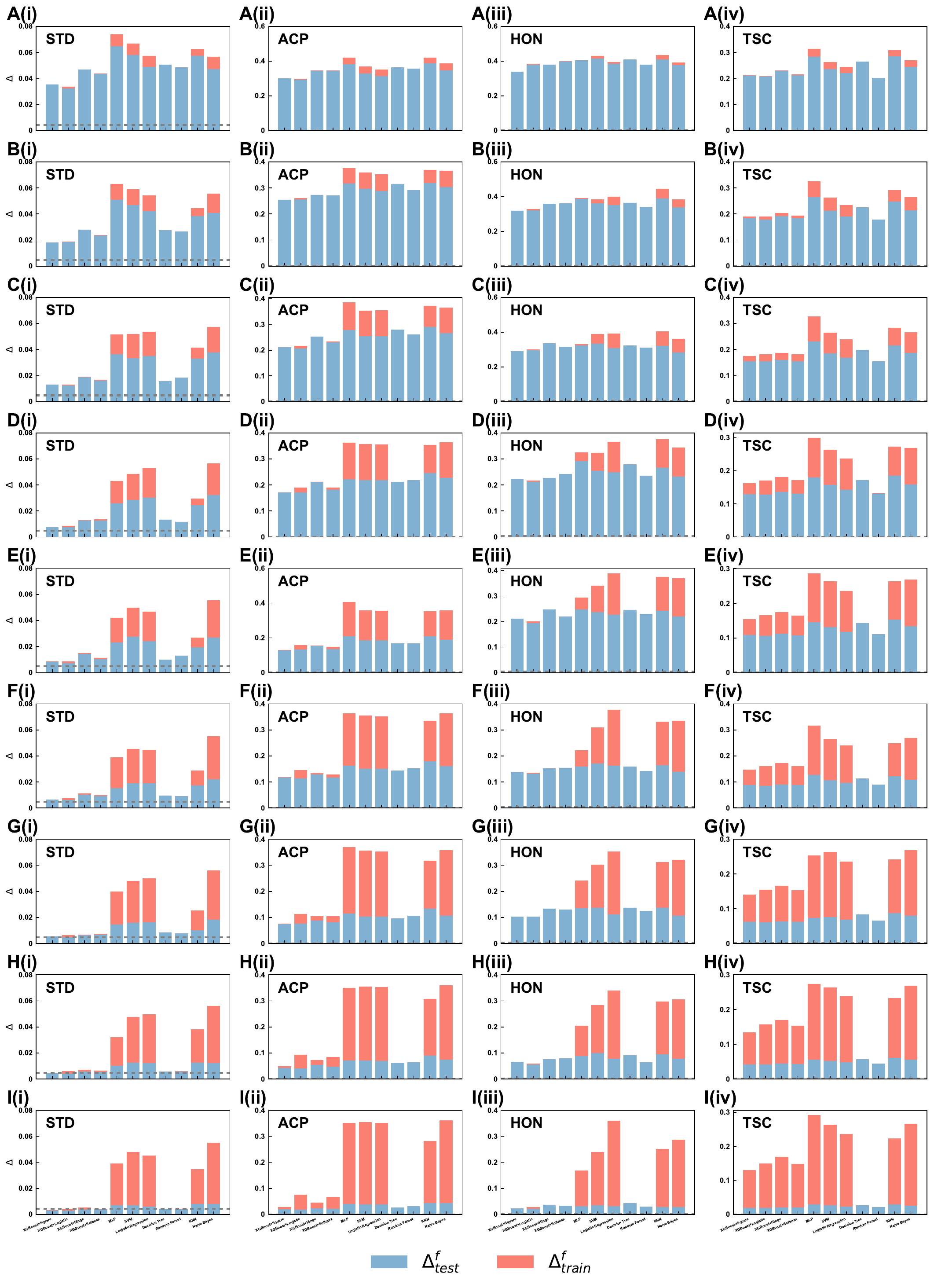}%
\caption{The loss errors of for 4 additional datasets (STD, ACP, HON and TSC) in training ($\Delta_{train}^{f}$) and test sets ($\Delta_{test}^{f}$) when $|\mathcal{S}_{train}|/|\mathcal{S}|=0.1$ (A), $|\mathcal{S}_{train}|/|\mathcal{S}|=0.2$ (B), $|\mathcal{S}_{train}|/|\mathcal{S}|=0.3$ (C), $|\mathcal{S}_{train}|/|\mathcal{S}|=0.4$ (D), $|\mathcal{S}_{train}|/|\mathcal{S}|=0.5$ (E), $|\mathcal{S}_{train}|/|\mathcal{S}|=0.6$ (F), $|\mathcal{S}_{train}|/|\mathcal{S}|=0.7$ (G), $|\mathcal{S}_{train}|/|\mathcal{S}|=0.8$ (H), $|\mathcal{S}_{train}|/|\mathcal{S}|=0.9$ (I). Dash line represents the expected error of optimal classier based on Eq. \ref{min_delta}.  }
\label{figS14-8}
\end{figure}

\begin{figure}
\centering
\includegraphics[width=.8\linewidth]{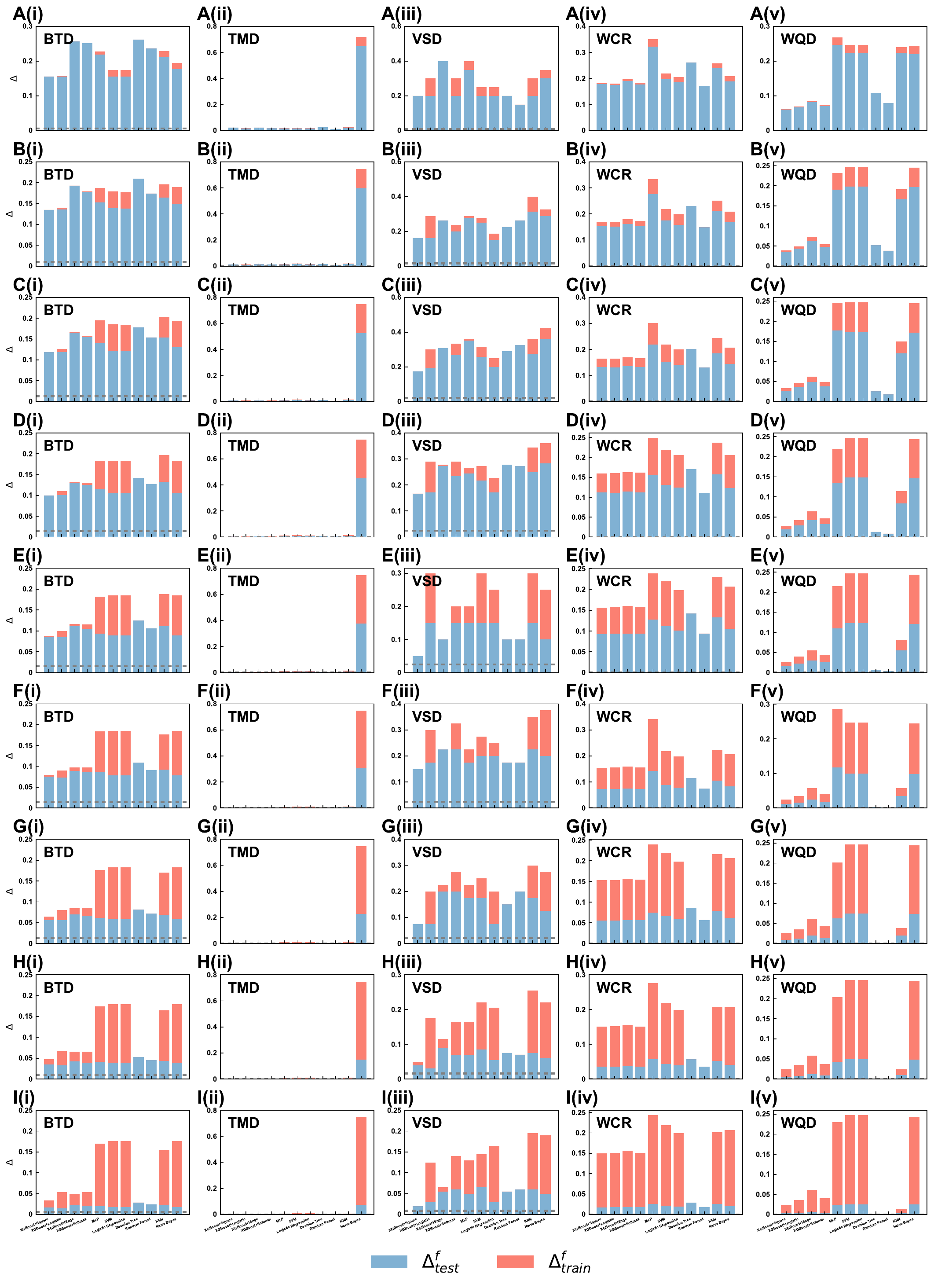}%
\caption{The loss errors of for 5 additional datasets (BTD, TMD, VSD, WCR and WQD) in training ($\Delta_{train}^{f}$) and test sets ($\Delta_{test}^{f}$) when $|\mathcal{S}_{train}|/|\mathcal{S}|=0.1$ (A), $|\mathcal{S}_{train}|/|\mathcal{S}|=0.2$ (B), $|\mathcal{S}_{train}|/|\mathcal{S}|=0.3$ (C), $|\mathcal{S}_{train}|/|\mathcal{S}|=0.4$ (D), $|\mathcal{S}_{train}|/|\mathcal{S}|=0.5$ (E), $|\mathcal{S}_{train}|/|\mathcal{S}|=0.6$ (F), $|\mathcal{S}_{train}|/|\mathcal{S}|=0.7$ (G), $|\mathcal{S}_{train}|/|\mathcal{S}|=0.8$ (H), $|\mathcal{S}_{train}|/|\mathcal{S}|=0.9$ (I). Dash line represents the expected error of optimal classier based on Eq. \ref{min_delta}.  }
\label{figS14-9}
\end{figure}

\begin{figure}
\centering
\includegraphics[width=.85\linewidth]{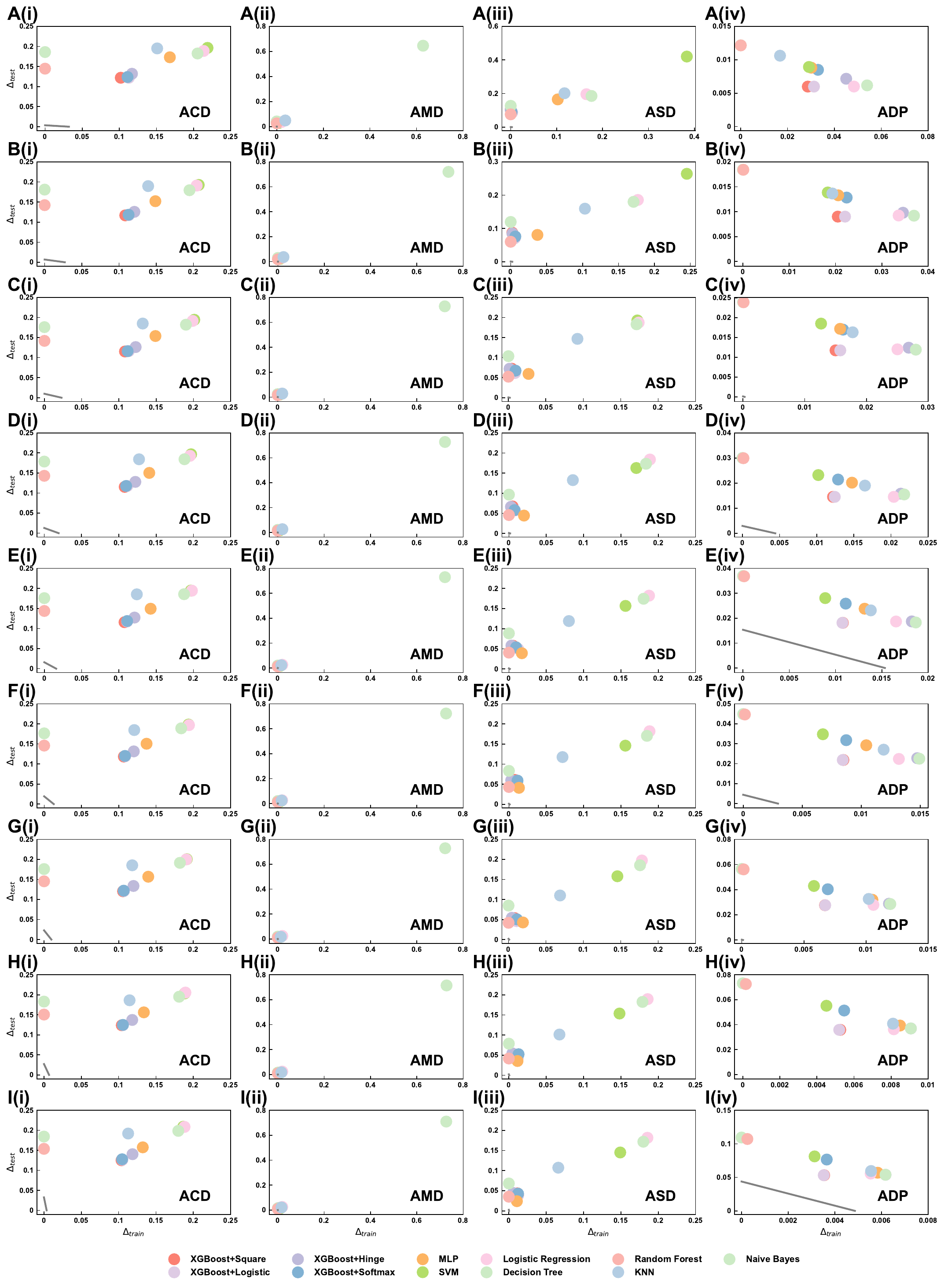}%
\caption{The loss errors of for 4 additional datasets (ACD, AMD, ASD and ADP) in training ($\Delta_{train}^{f}$) and test sets ($\Delta_{test}^{f}$) when $|\mathcal{S}_{train}|/|\mathcal{S}|=0.1$ (A), $|\mathcal{S}_{train}|/|\mathcal{S}|=0.2$ (B), $|\mathcal{S}_{train}|/|\mathcal{S}|=0.3$ (C), $|\mathcal{S}_{train}|/|\mathcal{S}|=0.4$ (D), $|\mathcal{S}_{train}|/|\mathcal{S}|=0.5$ (E), $|\mathcal{S}_{train}|/|\mathcal{S}|=0.6$ (F), $|\mathcal{S}_{train}|/|\mathcal{S}|=0.7$ (G), $|\mathcal{S}_{train}|/|\mathcal{S}|=0.8$ (H), $|\mathcal{S}_{train}|/|\mathcal{S}|=0.9$ (I). Gray line represents the expected error of optimal classier based on Eq. \ref{min_delta}.}
\label{figS15-1}
\end{figure}

\begin{figure}
\centering
\includegraphics[width=.85\linewidth]{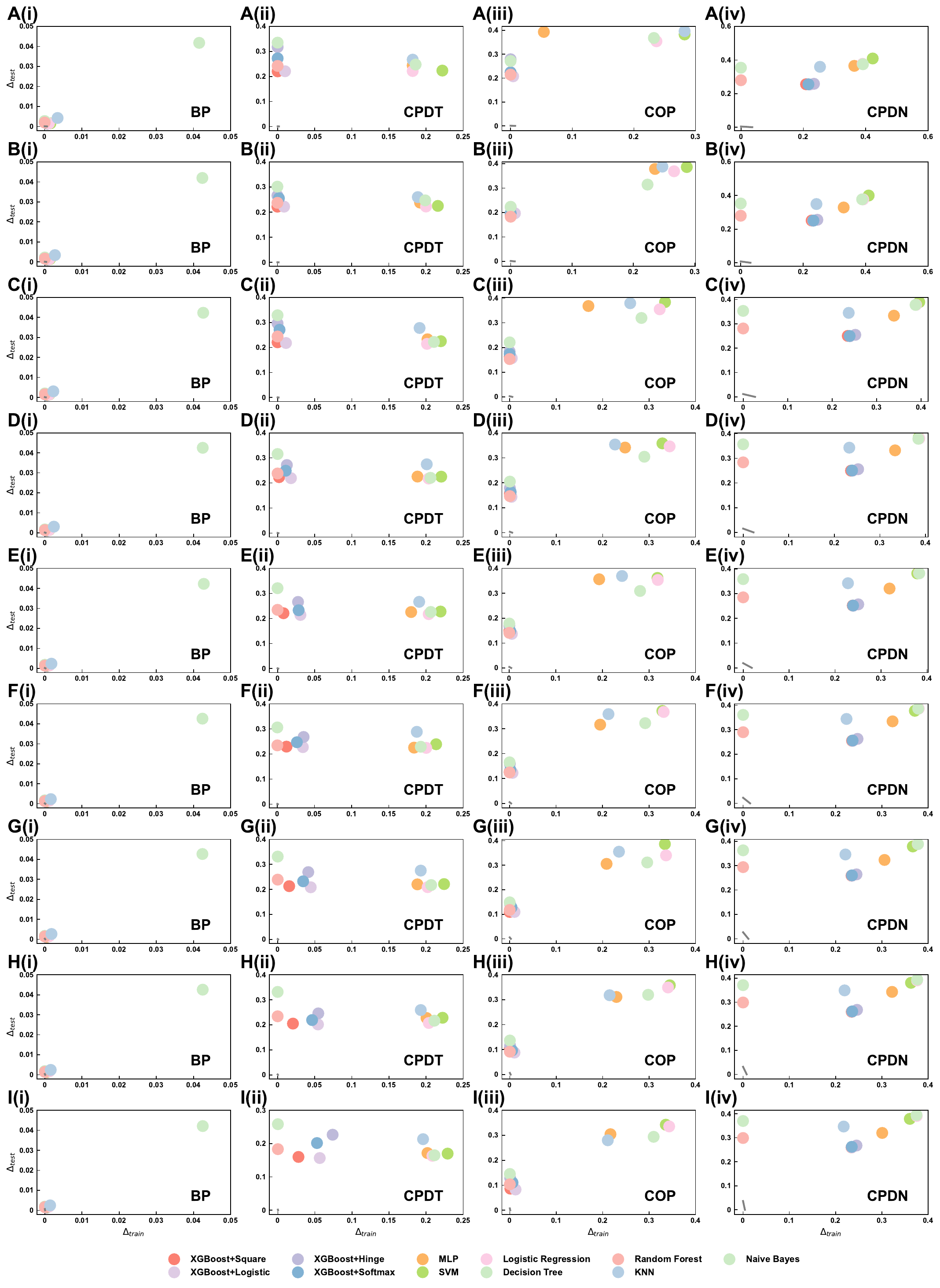}%
\caption{The loss errors of for 4 additional datasets (BP, CPDT, COP and CPDN) in training ($\Delta_{train}^{f}$) and test sets ($\Delta_{test}^{f}$) when $|\mathcal{S}_{train}|/|\mathcal{S}|=0.1$ (A), $|\mathcal{S}_{train}|/|\mathcal{S}|=0.2$ (B), $|\mathcal{S}_{train}|/|\mathcal{S}|=0.3$ (C), $|\mathcal{S}_{train}|/|\mathcal{S}|=0.4$ (D), $|\mathcal{S}_{train}|/|\mathcal{S}|=0.5$ (E), $|\mathcal{S}_{train}|/|\mathcal{S}|=0.6$ (F), $|\mathcal{S}_{train}|/|\mathcal{S}|=0.7$ (G), $|\mathcal{S}_{train}|/|\mathcal{S}|=0.8$ (H), $|\mathcal{S}_{train}|/|\mathcal{S}|=0.9$ (I). Gray line represents the expected error of optimal classier based on Eq. \ref{min_delta}.}
\label{figS15-2}
\end{figure}

\begin{figure}
\centering
\includegraphics[width=.85\linewidth]{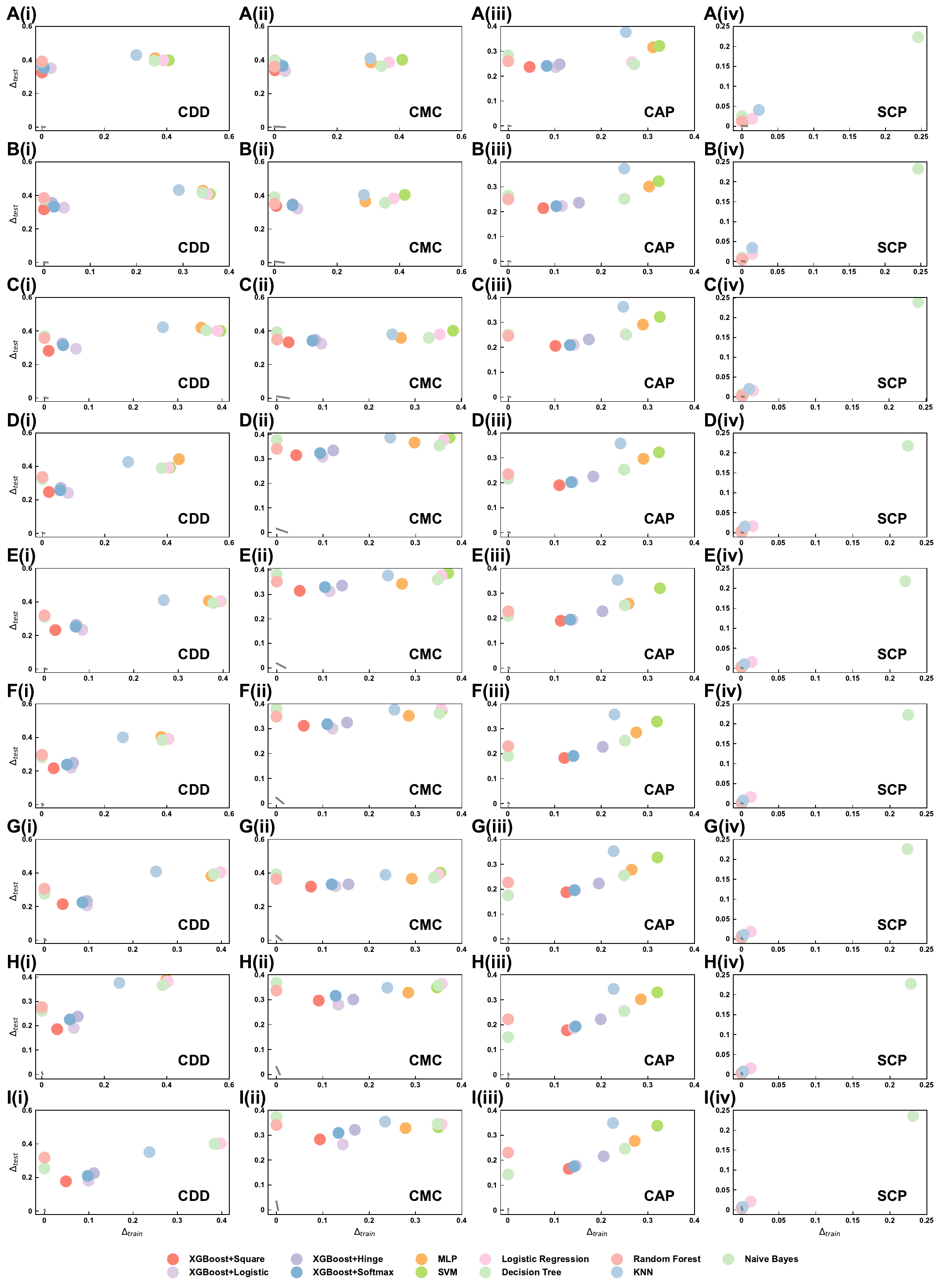}%
\caption{The loss errors of for 4 additional datasets (CDD, CMC, CAP and SCP) in training ($\Delta_{train}^{f}$) and test sets ($\Delta_{test}^{f}$) when $|\mathcal{S}_{train}|/|\mathcal{S}|=0.1$ (A), $|\mathcal{S}_{train}|/|\mathcal{S}|=0.2$ (B), $|\mathcal{S}_{train}|/|\mathcal{S}|=0.3$ (C), $|\mathcal{S}_{train}|/|\mathcal{S}|=0.4$ (D), $|\mathcal{S}_{train}|/|\mathcal{S}|=0.5$ (E), $|\mathcal{S}_{train}|/|\mathcal{S}|=0.6$ (F), $|\mathcal{S}_{train}|/|\mathcal{S}|=0.7$ (G), $|\mathcal{S}_{train}|/|\mathcal{S}|=0.8$ (H), $|\mathcal{S}_{train}|/|\mathcal{S}|=0.9$ (I). Gray line represents the expected error of optimal classier based on Eq. \ref{min_delta}.}
\label{figS15-3}
\end{figure}

\begin{figure}
\centering
\includegraphics[width=.85\linewidth]{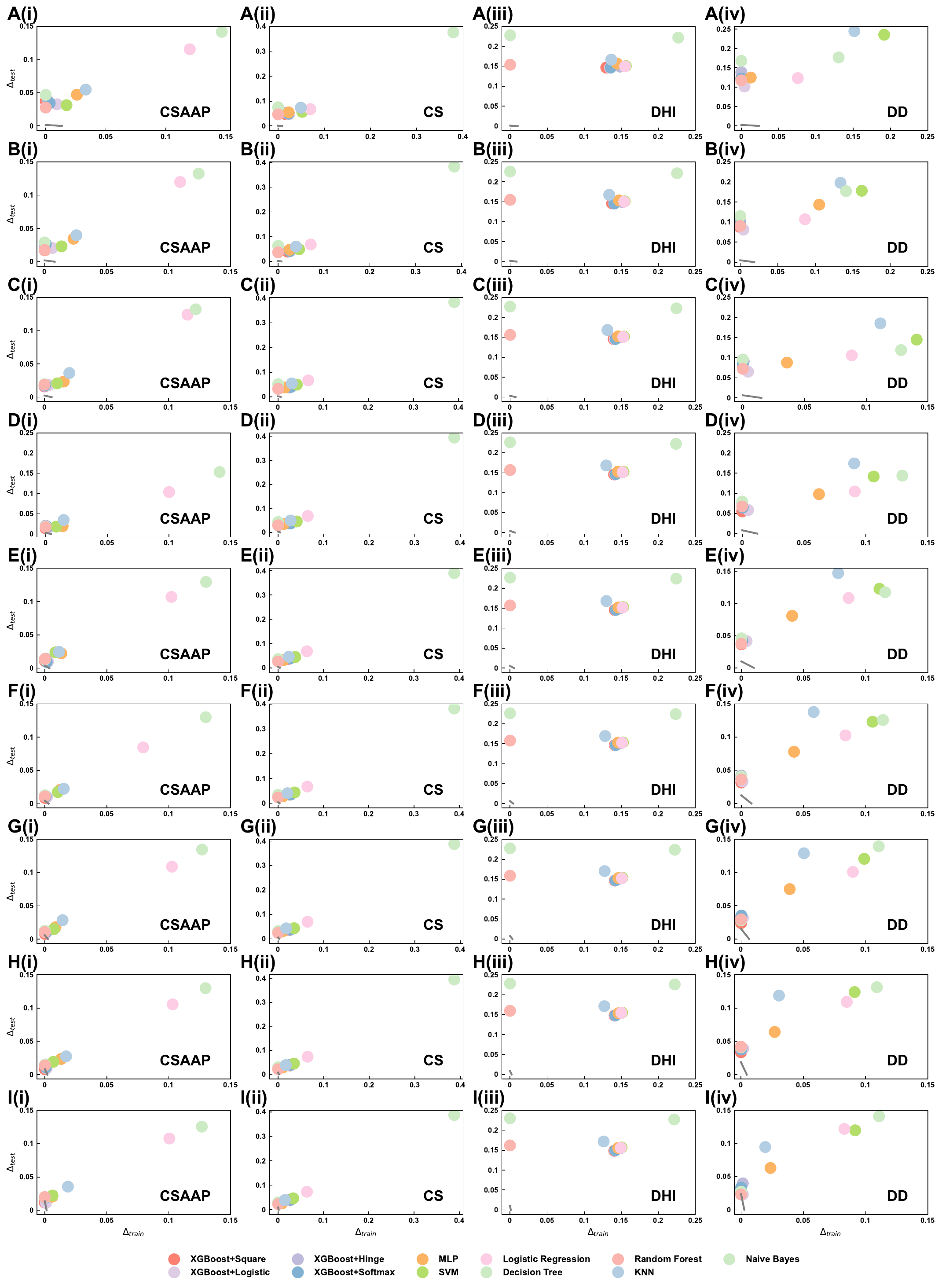}%
\caption{The loss errors of for 4 additional datasets (CSAAP, CS, DHI and DD) in training ($\Delta_{train}^{f}$) and test sets ($\Delta_{test}^{f}$) when $|\mathcal{S}_{train}|/|\mathcal{S}|=0.1$ (A), $|\mathcal{S}_{train}|/|\mathcal{S}|=0.2$ (B), $|\mathcal{S}_{train}|/|\mathcal{S}|=0.3$ (C), $|\mathcal{S}_{train}|/|\mathcal{S}|=0.4$ (D), $|\mathcal{S}_{train}|/|\mathcal{S}|=0.5$ (E), $|\mathcal{S}_{train}|/|\mathcal{S}|=0.6$ (F), $|\mathcal{S}_{train}|/|\mathcal{S}|=0.7$ (G), $|\mathcal{S}_{train}|/|\mathcal{S}|=0.8$ (H), $|\mathcal{S}_{train}|/|\mathcal{S}|=0.9$ (I). Gray line represents the expected error of optimal classier based on Eq. \ref{min_delta}.}
\label{figS15-4}
\end{figure}

\begin{figure}
\centering
\includegraphics[width=.85\linewidth]{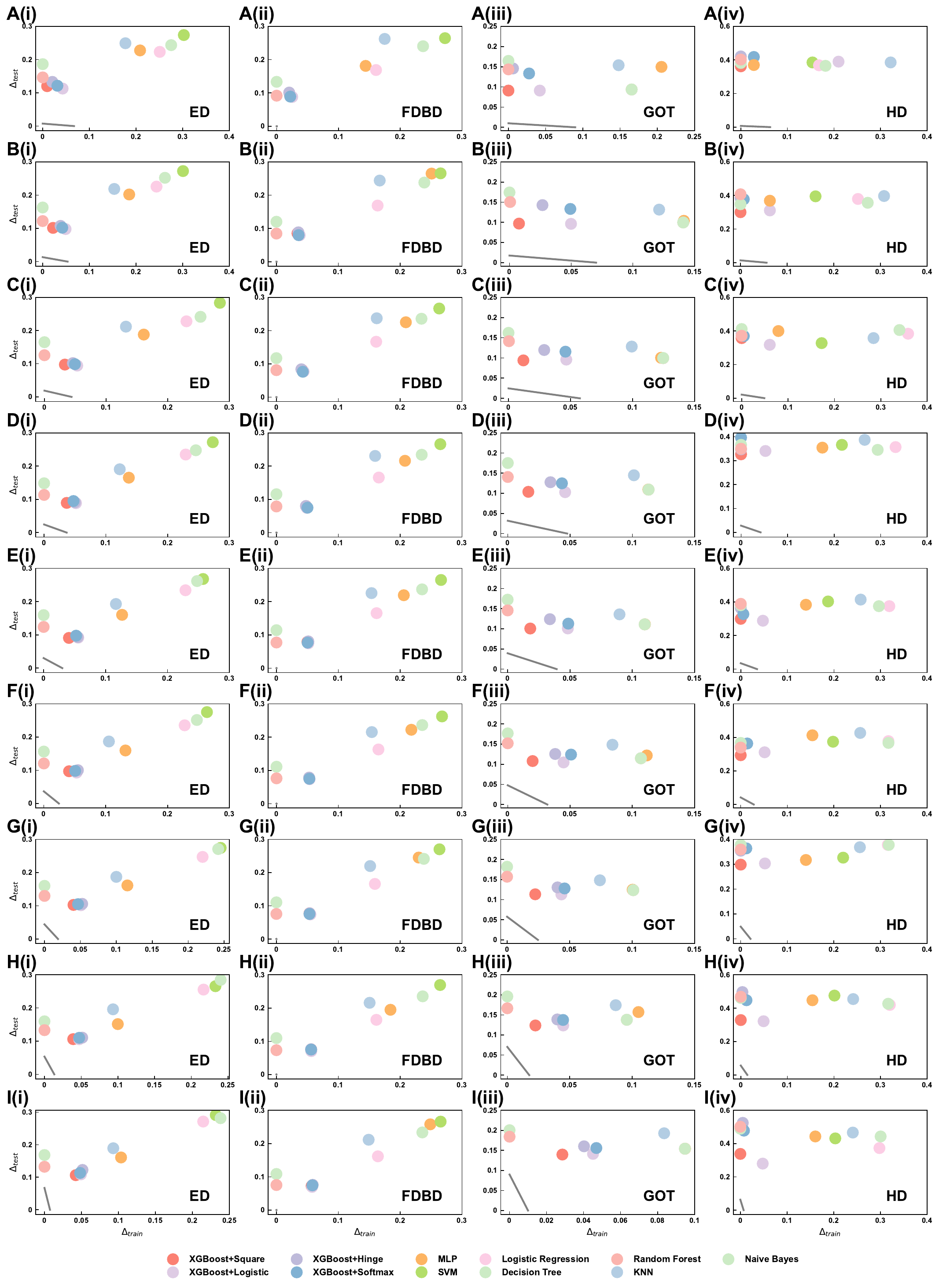}%
\caption{The loss errors of for 4 additional datasets (ED, FDBD, GOT and HD) in training ($\Delta_{train}^{f}$) and test sets ($\Delta_{test}^{f}$) when $|\mathcal{S}_{train}|/|\mathcal{S}|=0.1$ (A), $|\mathcal{S}_{train}|/|\mathcal{S}|=0.2$ (B), $|\mathcal{S}_{train}|/|\mathcal{S}|=0.3$ (C), $|\mathcal{S}_{train}|/|\mathcal{S}|=0.4$ (D), $|\mathcal{S}_{train}|/|\mathcal{S}|=0.5$ (E), $|\mathcal{S}_{train}|/|\mathcal{S}|=0.6$ (F), $|\mathcal{S}_{train}|/|\mathcal{S}|=0.7$ (G), $|\mathcal{S}_{train}|/|\mathcal{S}|=0.8$ (H), $|\mathcal{S}_{train}|/|\mathcal{S}|=0.9$ (I). Gray line represents the expected error of optimal classier based on Eq. \ref{min_delta}.}
\label{figS15-5}
\end{figure}

\begin{figure}
\centering
\includegraphics[width=.85\linewidth]{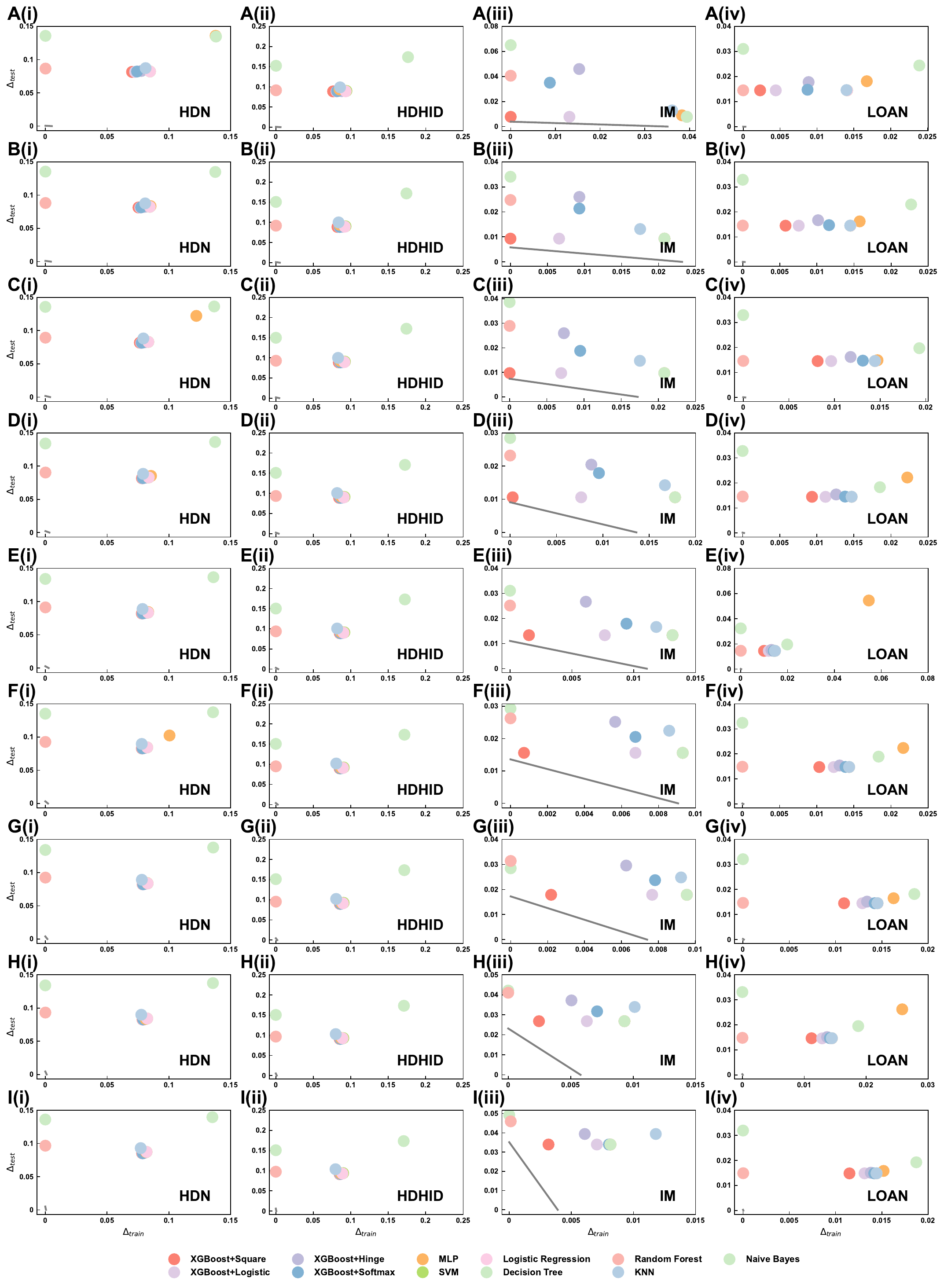}%
\caption{The loss errors of for 4 additional datasets (HDN, HDHID, IM and LOAN) in training ($\Delta_{train}^{f}$) and test sets ($\Delta_{test}^{f}$) when $|\mathcal{S}_{train}|/|\mathcal{S}|=0.1$ (A), $|\mathcal{S}_{train}|/|\mathcal{S}|=0.2$ (B), $|\mathcal{S}_{train}|/|\mathcal{S}|=0.3$ (C), $|\mathcal{S}_{train}|/|\mathcal{S}|=0.4$ (D), $|\mathcal{S}_{train}|/|\mathcal{S}|=0.5$ (E), $|\mathcal{S}_{train}|/|\mathcal{S}|=0.6$ (F), $|\mathcal{S}_{train}|/|\mathcal{S}|=0.7$ (G), $|\mathcal{S}_{train}|/|\mathcal{S}|=0.8$ (H), $|\mathcal{S}_{train}|/|\mathcal{S}|=0.9$ (I). Gray line represents the expected error of optimal classier based on Eq. \ref{min_delta}.}
\label{figS15-6}
\end{figure}

\begin{figure}
\centering
\includegraphics[width=.85\linewidth]{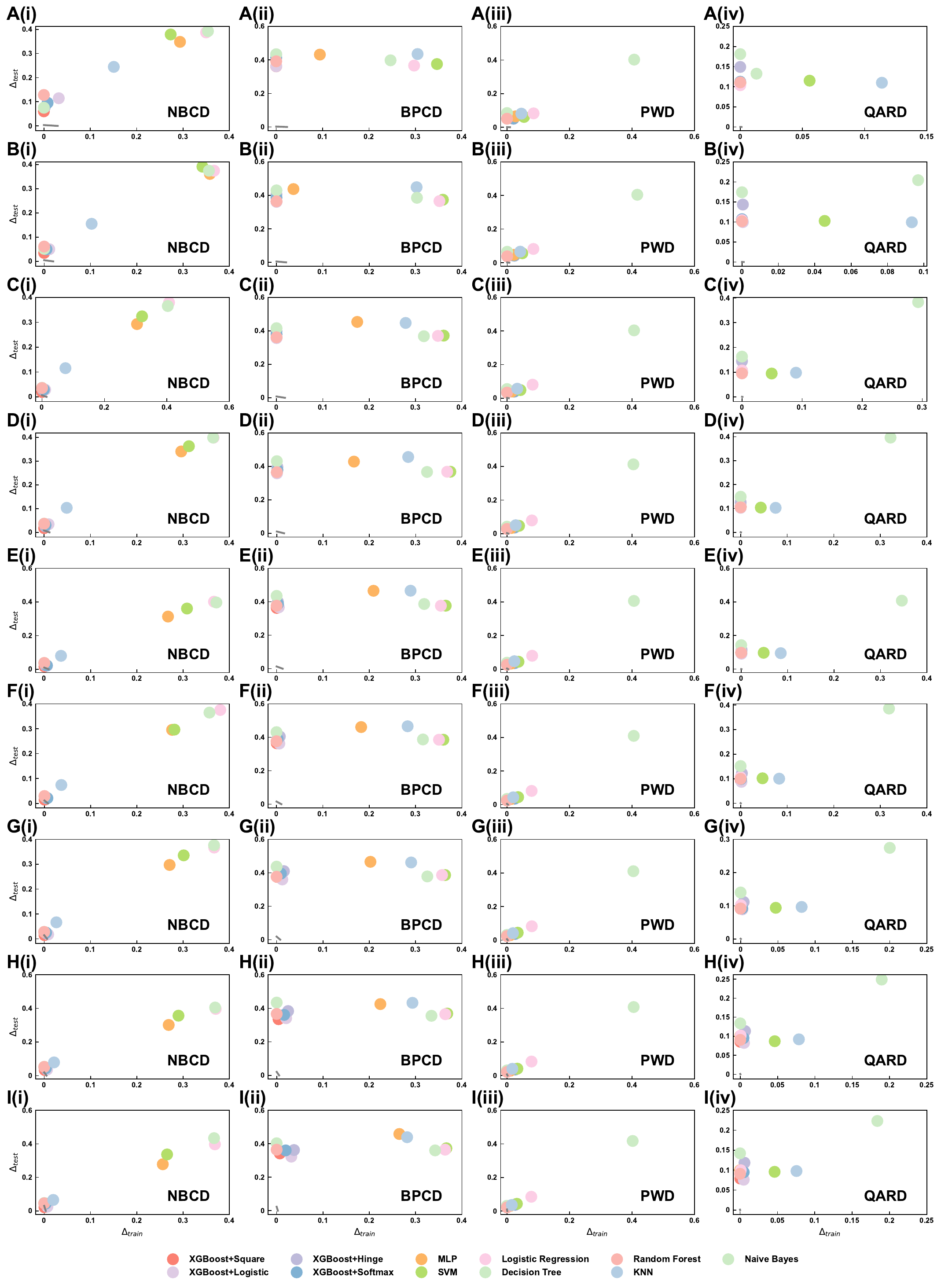}%
\caption{The loss errors of for 4 additional datasets (NBCD, BPCD, PWD and QARD) in training ($\Delta_{train}^{f}$) and test sets ($\Delta_{test}^{f}$) when $|\mathcal{S}_{train}|/|\mathcal{S}|=0.1$ (A), $|\mathcal{S}_{train}|/|\mathcal{S}|=0.2$ (B), $|\mathcal{S}_{train}|/|\mathcal{S}|=0.3$ (C), $|\mathcal{S}_{train}|/|\mathcal{S}|=0.4$ (D), $|\mathcal{S}_{train}|/|\mathcal{S}|=0.5$ (E), $|\mathcal{S}_{train}|/|\mathcal{S}|=0.6$ (F), $|\mathcal{S}_{train}|/|\mathcal{S}|=0.7$ (G), $|\mathcal{S}_{train}|/|\mathcal{S}|=0.8$ (H), $|\mathcal{S}_{train}|/|\mathcal{S}|=0.9$ (I). Gray line represents the expected error of optimal classier based on Eq. \ref{min_delta}.}
\label{figS15-7}
\end{figure}

\begin{figure}
\centering
\includegraphics[width=.85\linewidth]{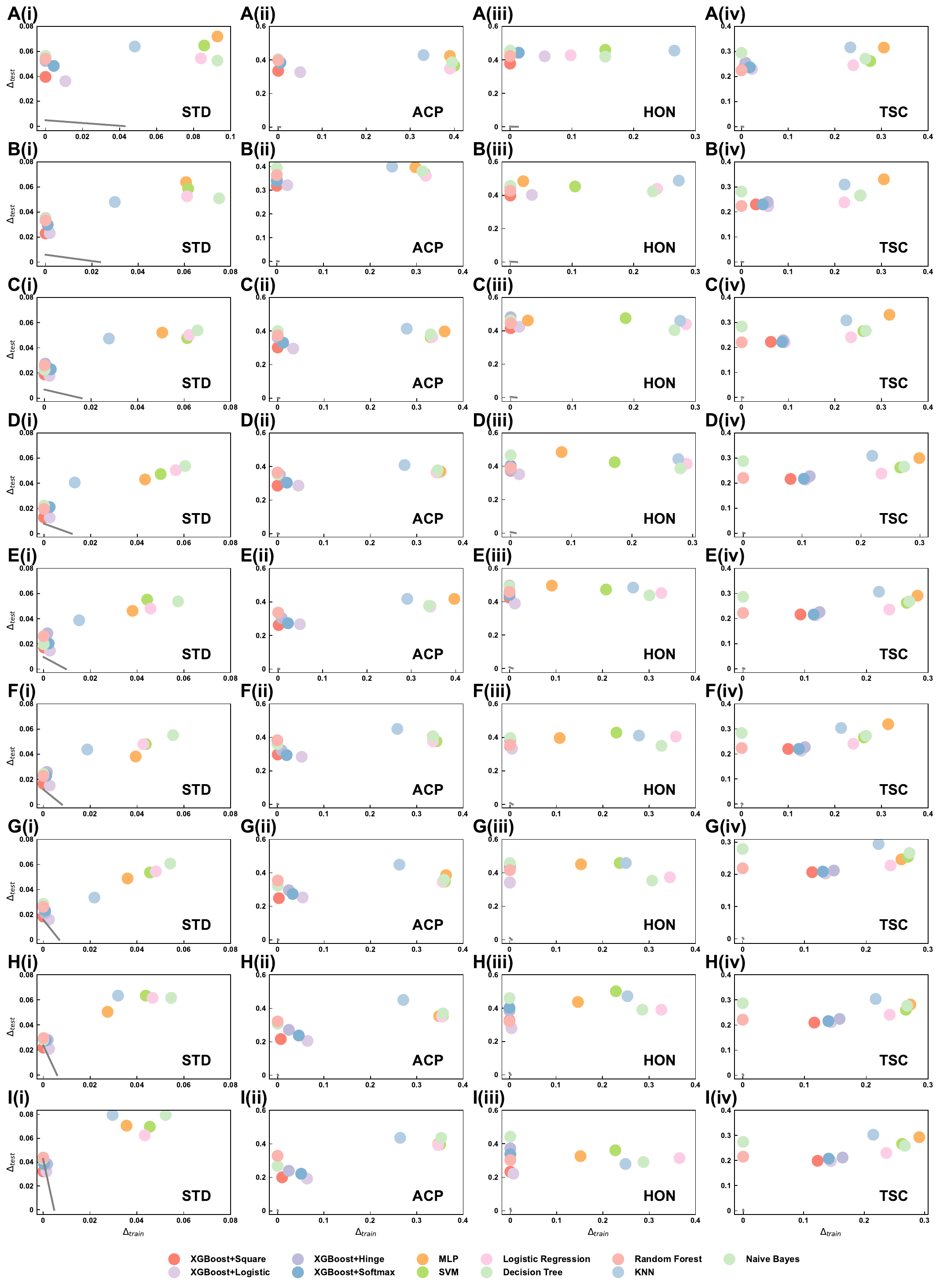}%
\caption{The loss errors of for 4 additional datasets (STD, ACP, HON and TSC) in training ($\Delta_{train}^{f}$) and test sets ($\Delta_{test}^{f}$) when $|\mathcal{S}_{train}|/|\mathcal{S}|=0.1$ (A), $|\mathcal{S}_{train}|/|\mathcal{S}|=0.2$ (B), $|\mathcal{S}_{train}|/|\mathcal{S}|=0.3$ (C), $|\mathcal{S}_{train}|/|\mathcal{S}|=0.4$ (D), $|\mathcal{S}_{train}|/|\mathcal{S}|=0.5$ (E), $|\mathcal{S}_{train}|/|\mathcal{S}|=0.6$ (F), $|\mathcal{S}_{train}|/|\mathcal{S}|=0.7$ (G), $|\mathcal{S}_{train}|/|\mathcal{S}|=0.8$ (H), $|\mathcal{S}_{train}|/|\mathcal{S}|=0.9$ (I). Gray line represents the expected error of optimal classier based on Eq. \ref{min_delta}.}
\label{figS15-8}
\end{figure}

\begin{figure}
\centering
\includegraphics[width=.8\linewidth]{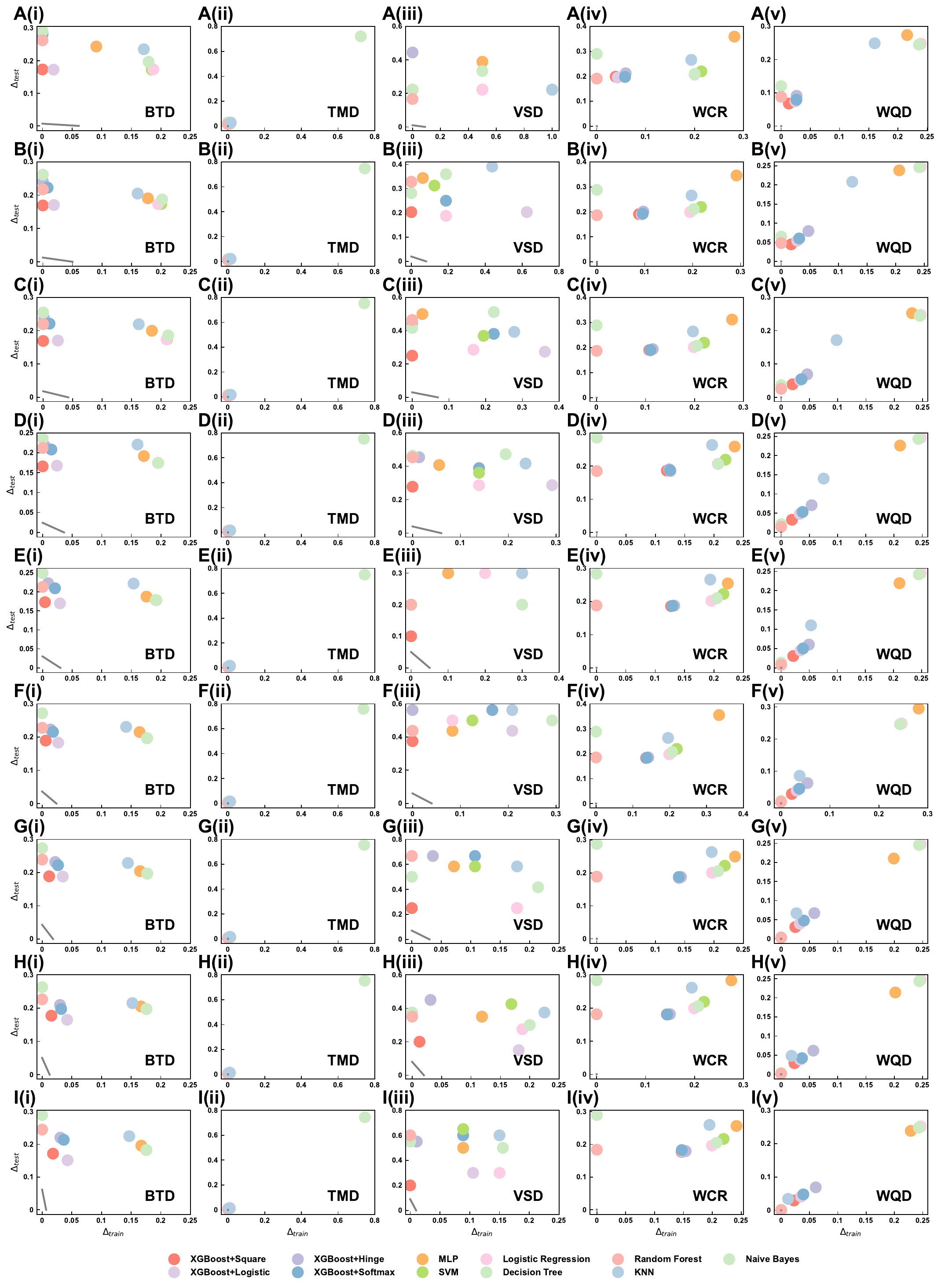}%
\caption{The loss errors of for 5 additional datasets (BTD, TMD, VSD, WCR and WQD) in training ($\Delta_{train}^{f}$) and test sets ($\Delta_{test}^{f}$) when $|\mathcal{S}_{train}|/|\mathcal{S}|=0.1$ (A), $|\mathcal{S}_{train}|/|\mathcal{S}|=0.2$ (B), $|\mathcal{S}_{train}|/|\mathcal{S}|=0.3$ (C), $|\mathcal{S}_{train}|/|\mathcal{S}|=0.4$ (D), $|\mathcal{S}_{train}|/|\mathcal{S}|=0.5$ (E), $|\mathcal{S}_{train}|/|\mathcal{S}|=0.6$ (F), $|\mathcal{S}_{train}|/|\mathcal{S}|=0.7$ (G), $|\mathcal{S}_{train}|/|\mathcal{S}|=0.8$ (H), $|\mathcal{S}_{train}|/|\mathcal{S}|=0.9$ (I). Gray line represents the expected error of optimal classier based on Eq. \ref{min_delta}.}
\label{figS15-9}
\end{figure}

\begin{figure}
\centering
\includegraphics[width=0.85\linewidth]{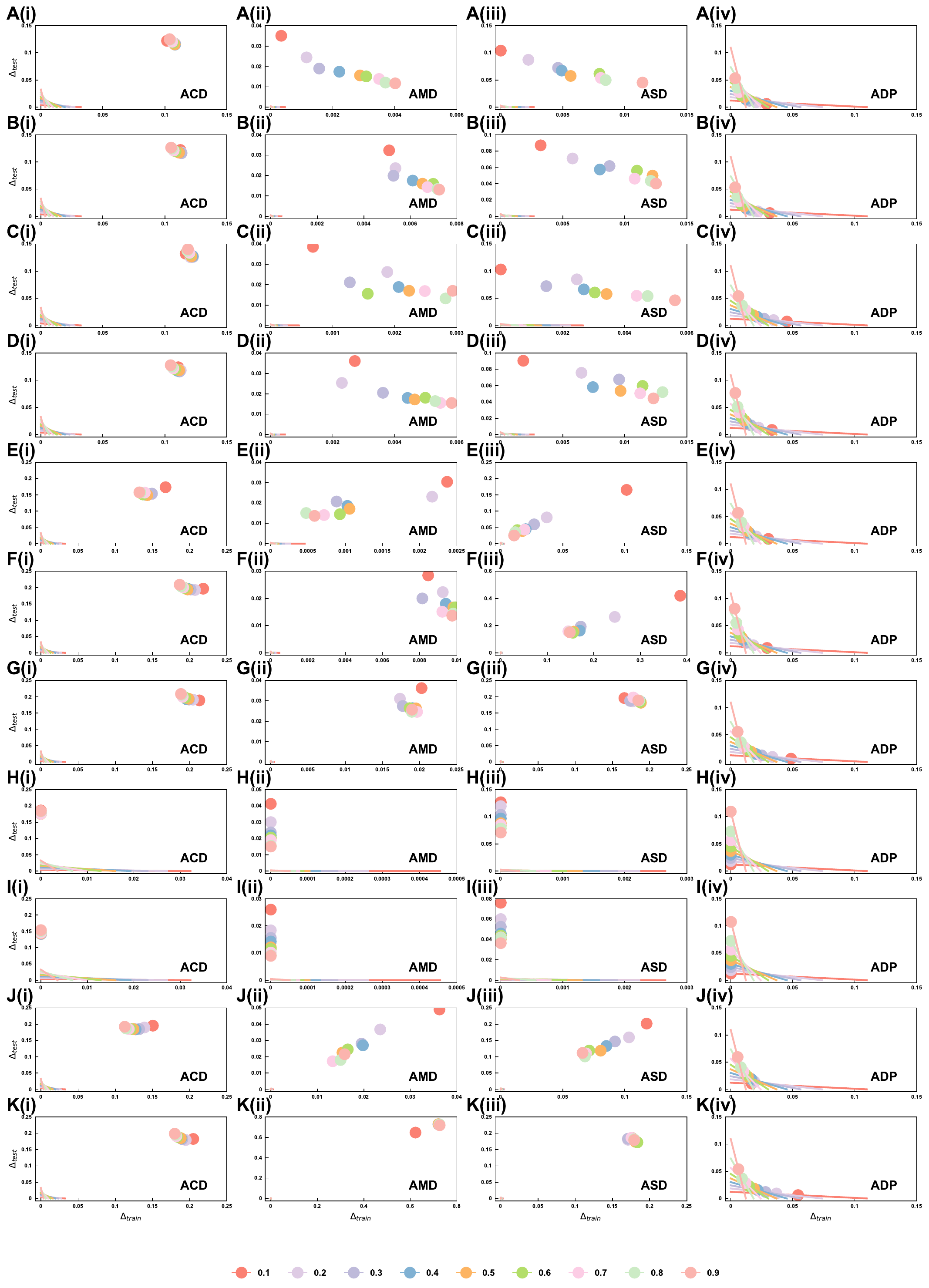}
\caption{The loss errors of four additional datasets (ACD, AMD, ASD and ADP) in training ($\Delta_{train}^{f}$) and test sets ($\Delta_{test}^{f}$) of different binary classifiers, including XGBoost with four classical objectives (A-D), MLP (E), SVM (F), Logistic Regression (G), Decision Tree (H), Random Forest (I), KNN (J). Colorful dots and lines represent different $|\mathcal{S}_{train}|/|\mathcal{S}|$ ranging from $0.1$ to $0.9$.}
\label{figS16-1}
\end{figure}

\clearpage
\begin{figure}
\centering
\includegraphics[width=.85\linewidth]{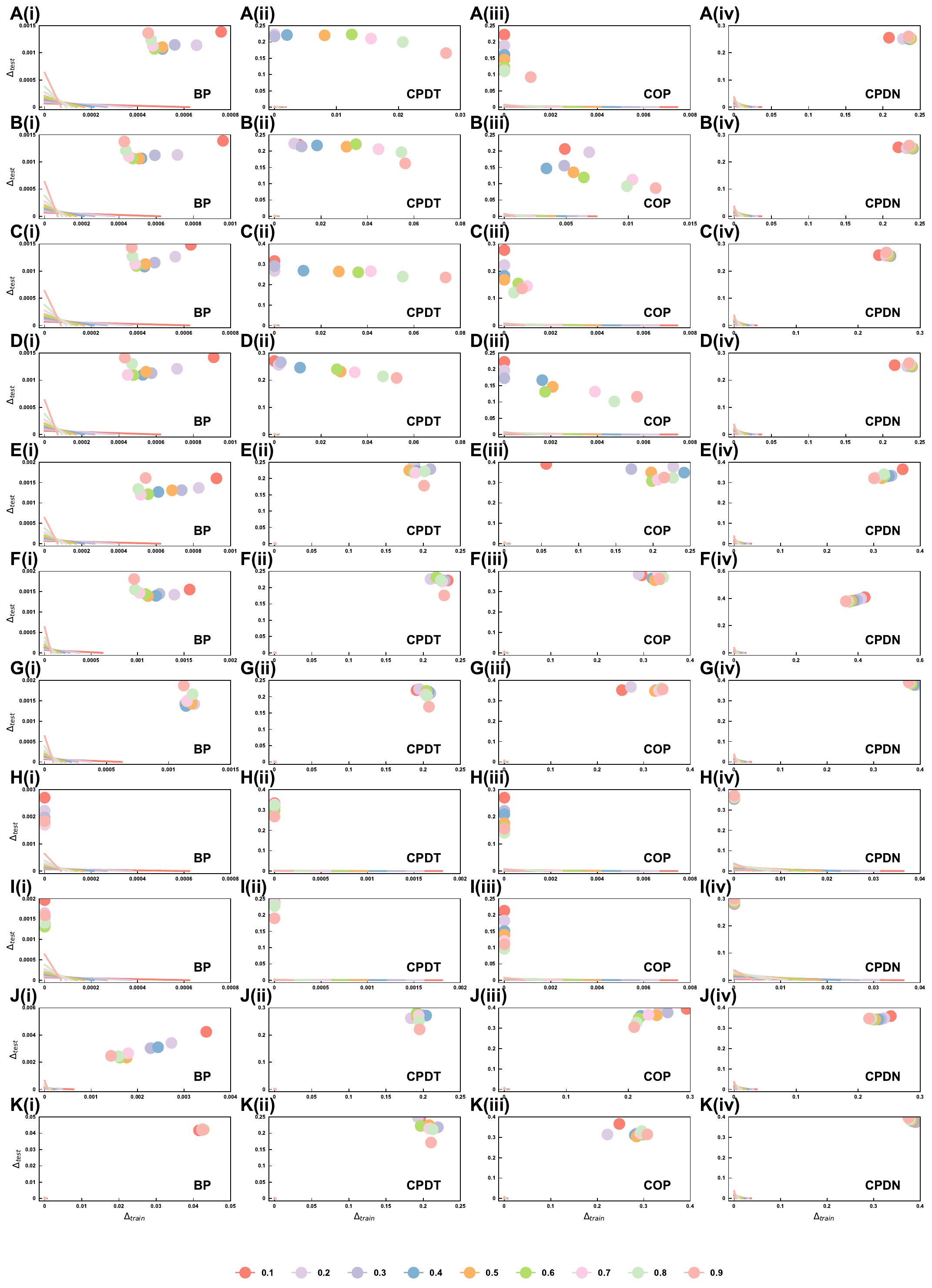}
\caption{The loss errors of four additional datasets (BP, CPDT, COP and CPDN) in training ($\Delta_{train}^{f}$) and test sets ($\Delta_{test}^{f}$) of different binary classifiers, including XGBoost with four classical objectives (A-D), MLP (E), SVM (F), Logistic Regression (G), Decision Tree (H), Random Forest (I), KNN (J). Colorful dots and lines represent different $|\mathcal{S}_{train}|/|\mathcal{S}|$ ranging from $0.1$ to $0.9$.}
\label{figS16-2}
\end{figure}

\begin{figure}
\centering
\includegraphics[width=.85\linewidth]{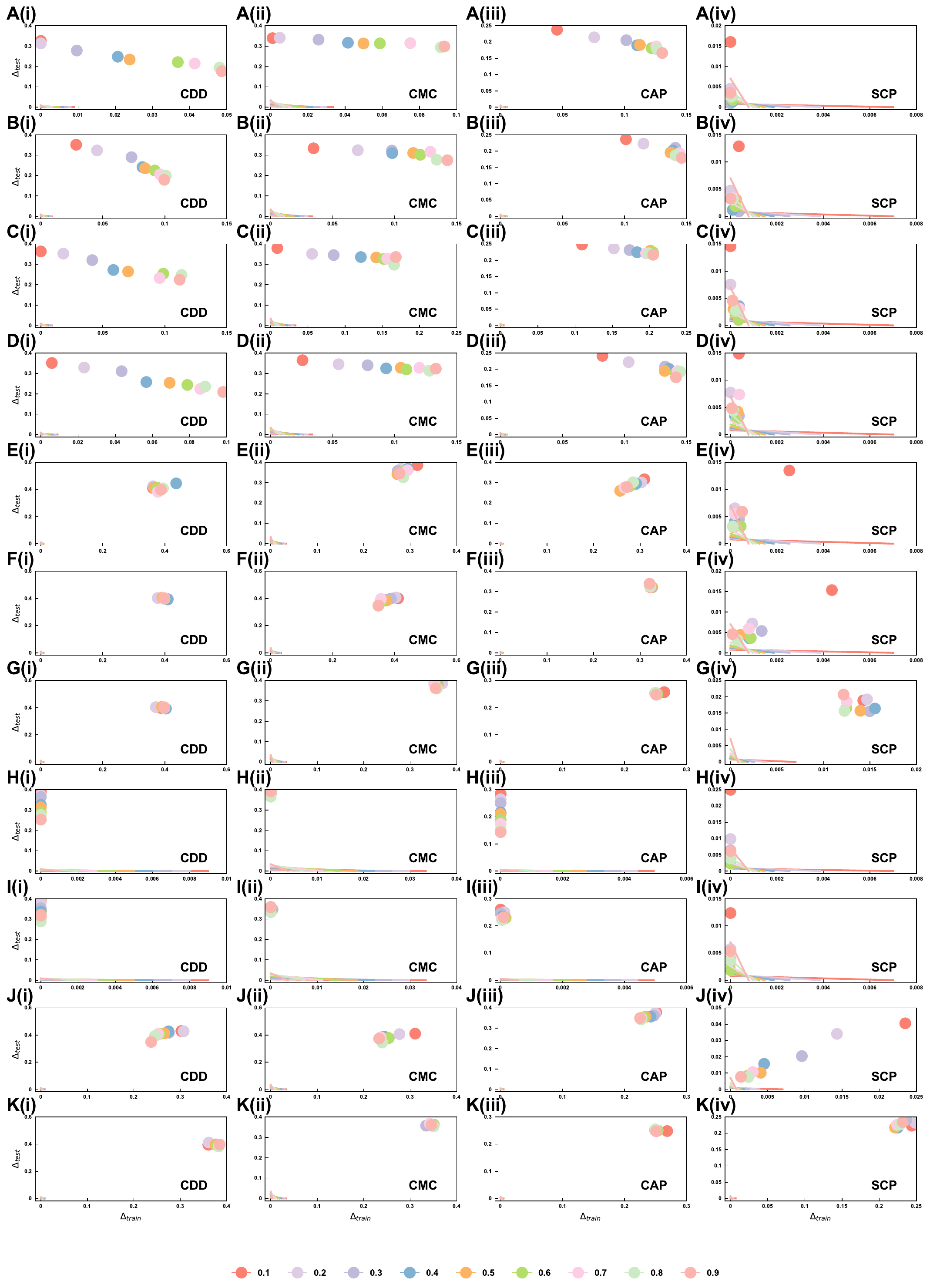}%
\caption{The loss errors of four additional datasets (CDD, CMC, CAP and SCP) in training ($\Delta_{train}^{f}$) and test sets ($\Delta_{test}^{f}$) of different binary classifiers, including XGBoost with four classical objectives (A-D), MLP (E), SVM (F), Logistic Regression (G), Decision Tree (H), Random Forest (I), KNN (J). Colorful dots and lines represent different $|\mathcal{S}_{train}|/|\mathcal{S}|$ ranging from $0.1$ to $0.9$.}
\label{figS16-3}
\end{figure}

\begin{figure}
\centering
\includegraphics[width=.85\linewidth]{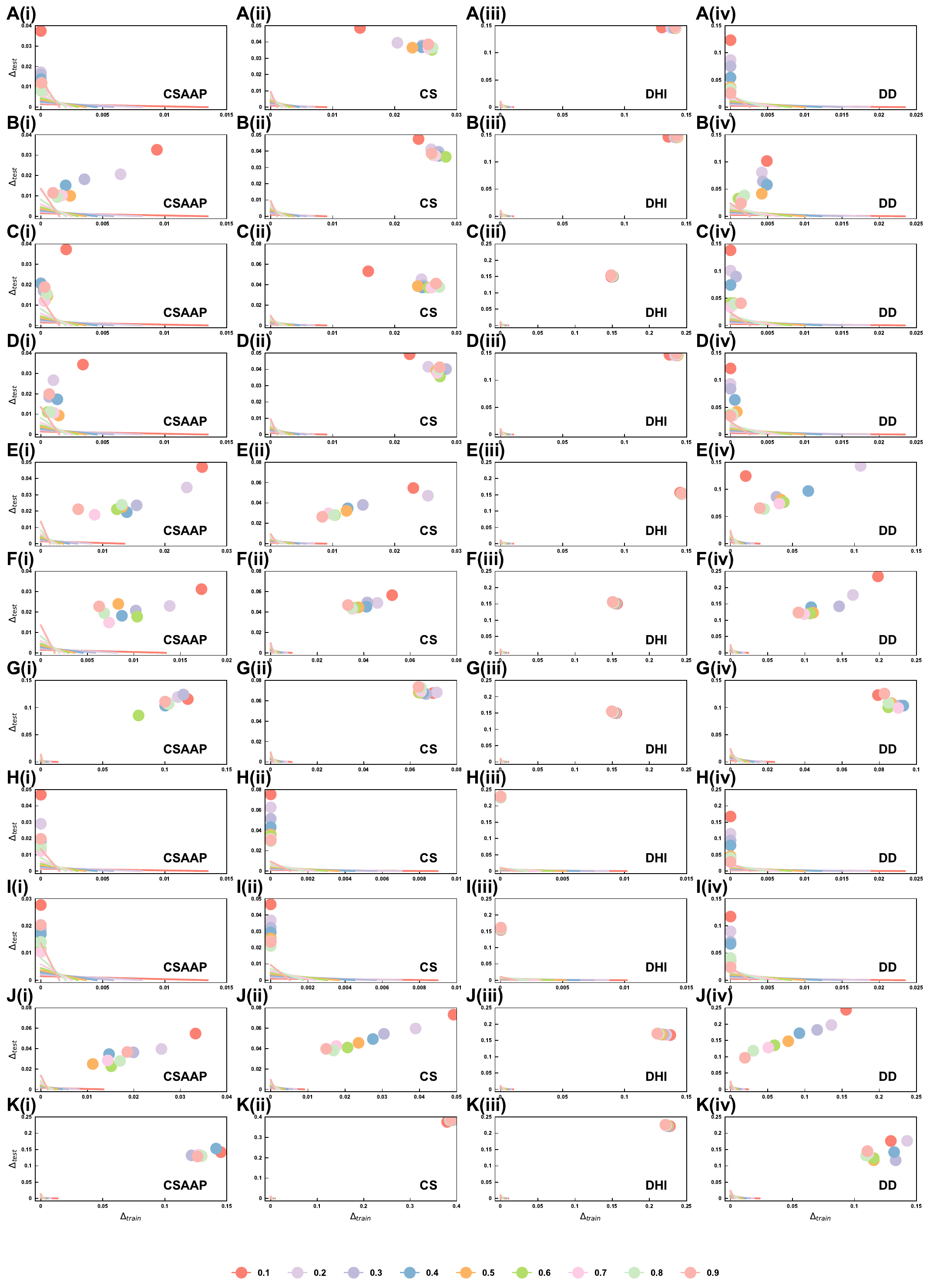}%
\caption{The loss errors of four additional datasets (CSAAP, CS, DHI and DD) in training ($\Delta_{train}^{f}$) and test sets ($\Delta_{test}^{f}$) of different binary classifiers, including XGBoost with four classical objectives (A-D), MLP (E), SVM (F), Logistic Regression (G), Decision Tree (H), Random Forest (I), KNN (J). Colorful dots and lines represent different $|\mathcal{S}_{train}|/|\mathcal{S}|$ ranging from $0.1$ to $0.9$.}
\label{figS16-4}
\end{figure}

\begin{figure}
\centering
\includegraphics[width=.85\linewidth]{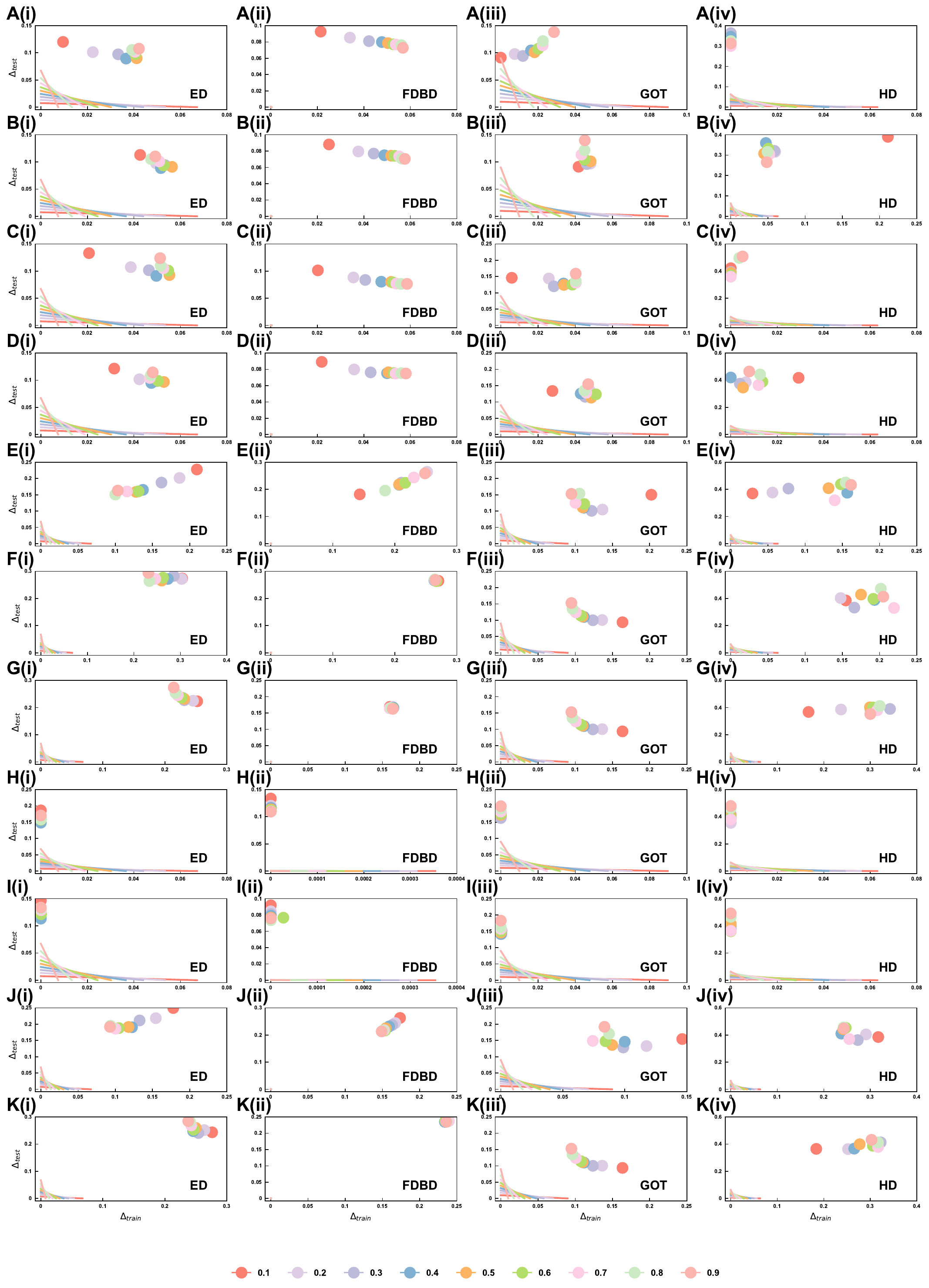}%
\caption{The loss errors of four additional datasets (ED, FDBD, GOT and HD) in training ($\Delta_{train}^{f}$) and test sets ($\Delta_{test}^{f}$) of different binary classifiers, including XGBoost with four classical objectives (A-D), MLP (E), SVM (F), Logistic Regression (G), Decision Tree (H), Random Forest (I), KNN (J). Colorful dots and lines represent different $|\mathcal{S}_{train}|/|\mathcal{S}|$ ranging from $0.1$ to $0.9$.}
\label{figS16-5}
\end{figure}

\begin{figure}
\centering
\includegraphics[width=.85\linewidth]{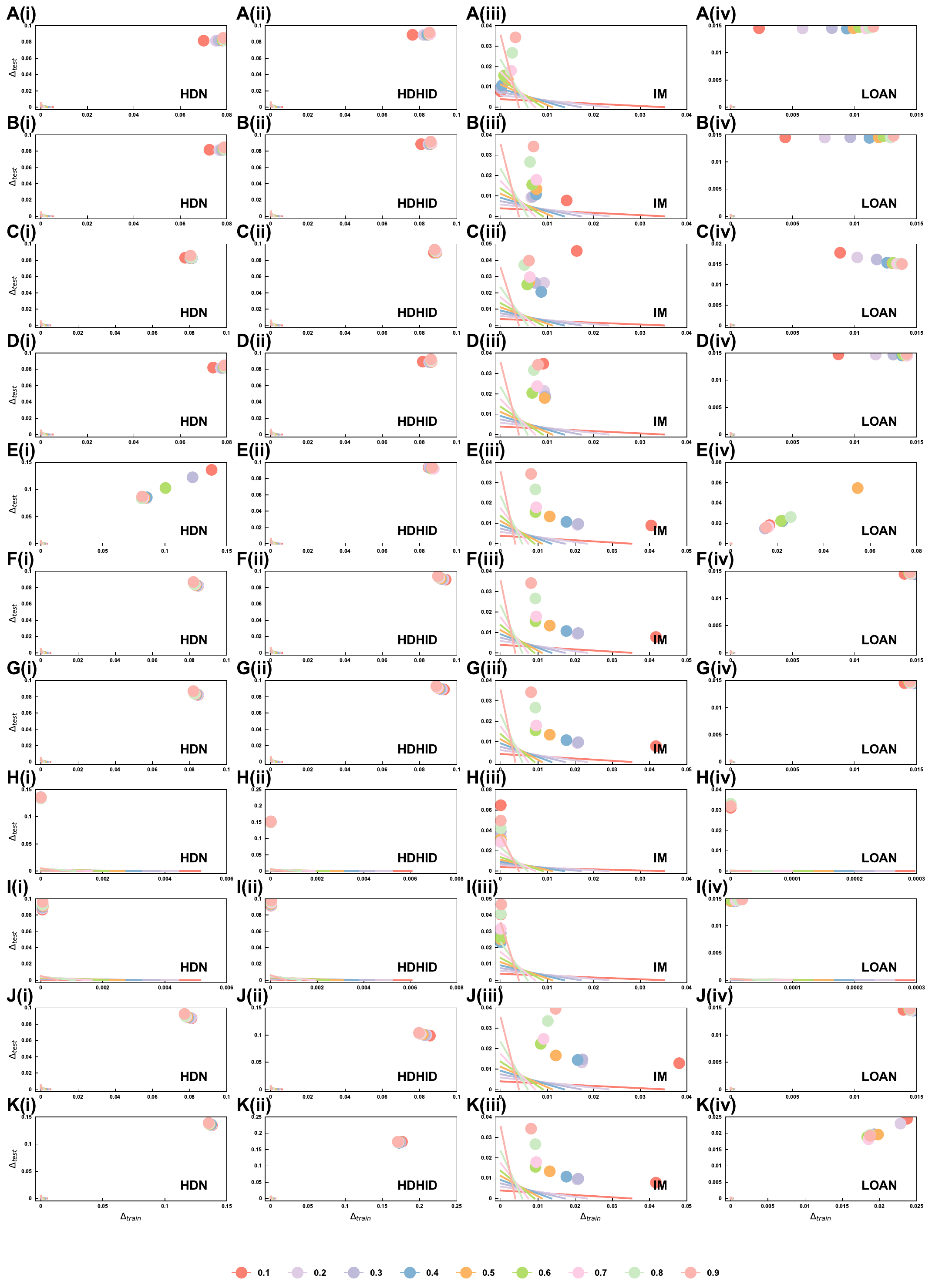}%
\caption{The loss errors of four additional datasets (HDN, HDHID, IM and LOAN) in training ($\Delta_{train}^{f}$) and test sets ($\Delta_{test}^{f}$) of different binary classifiers, including XGBoost with four classical objectives (A-D), MLP (E), SVM (F), Logistic Regression (G), Decision Tree (H), Random Forest (I), KNN (J). Colorful dots and lines represent different $|\mathcal{S}_{train}|/|\mathcal{S}|$ ranging from $0.1$ to $0.9$.}
\label{figS16-6}
\end{figure}

\begin{figure}
\centering
\includegraphics[width=.85\linewidth]{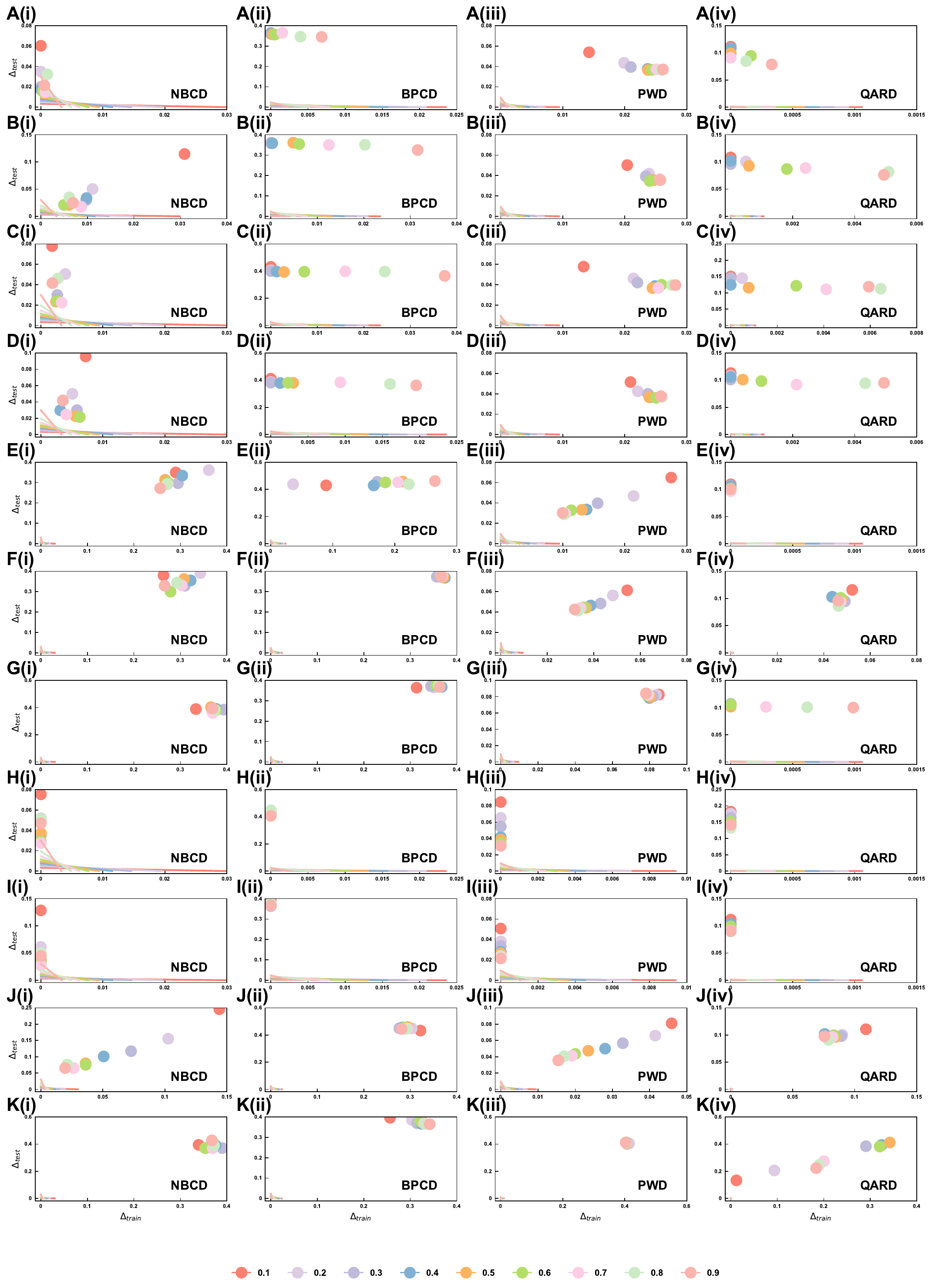}%
\caption{The loss errors of four additional datasets (NBCD, BPCD, PWD and QARD) in training ($\Delta_{train}^{f}$) and test sets ($\Delta_{test}^{f}$) of different binary classifiers, including XGBoost with four classical objectives (A-D), MLP (E), SVM (F), Logistic Regression (G), Decision Tree (H), Random Forest (I), KNN (J). Colorful dots and lines represent different $|\mathcal{S}_{train}|/|\mathcal{S}|$ ranging from $0.1$ to $0.9$.}
\label{figS16-7}
\end{figure}

\begin{figure}
\centering
\includegraphics[width=.85\linewidth]{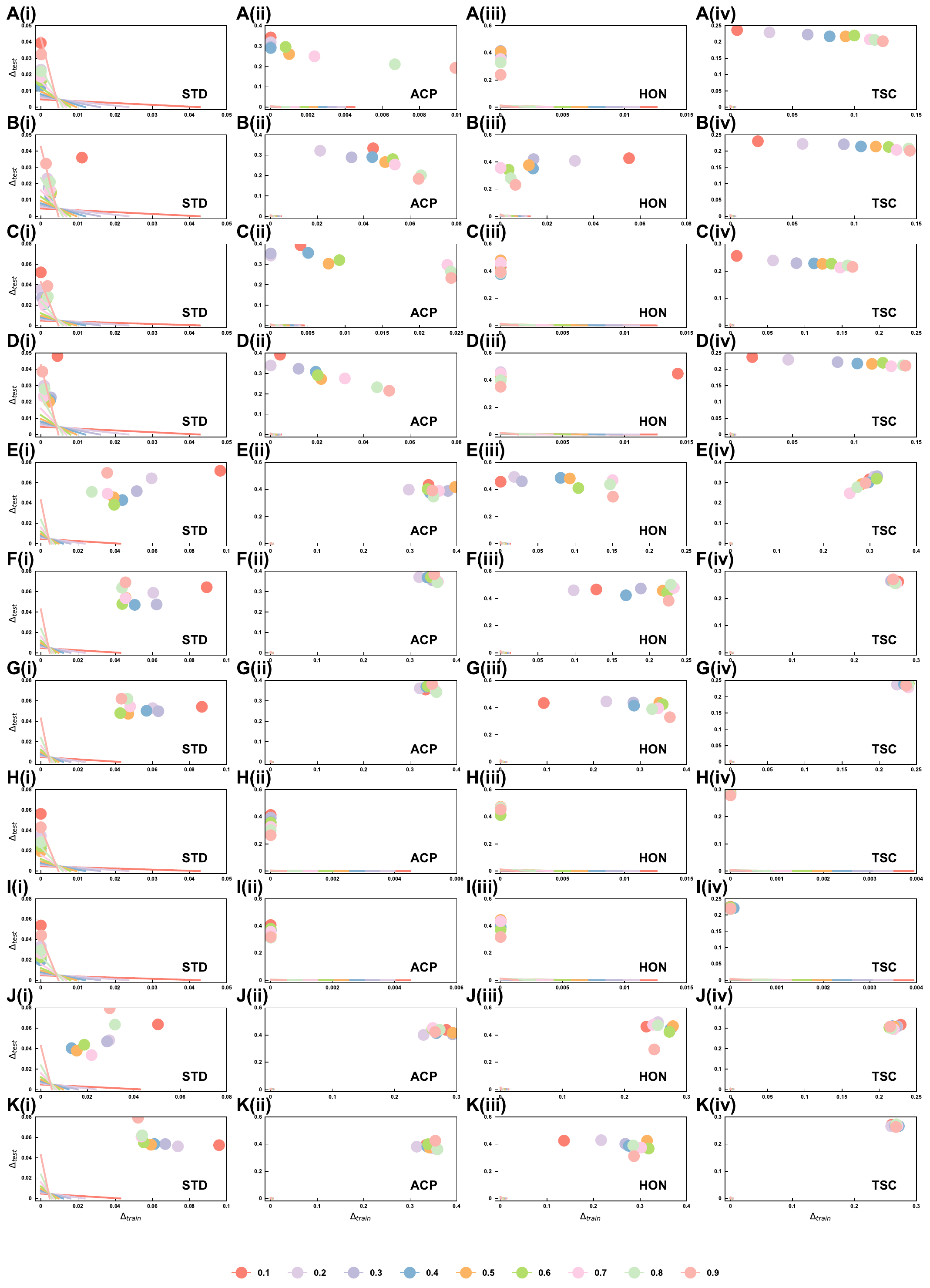}%
\caption{The loss errors of four additional datasets (STD, ACP, HON and TSC) in training ($\Delta_{train}^{f}$) and test sets ($\Delta_{test}^{f}$) of different binary classifiers, including XGBoost with four classical objectives (A-D), MLP (E), SVM (F), Logistic Regression (G), Decision Tree (H), Random Forest (I), KNN (J). Colorful dots and lines represent different $|\mathcal{S}_{train}|/|\mathcal{S}|$ ranging from $0.1$ to $0.9$.}
\label{figS16-8}
\end{figure}

\begin{figure}
\centering
\includegraphics[width=.8\linewidth]{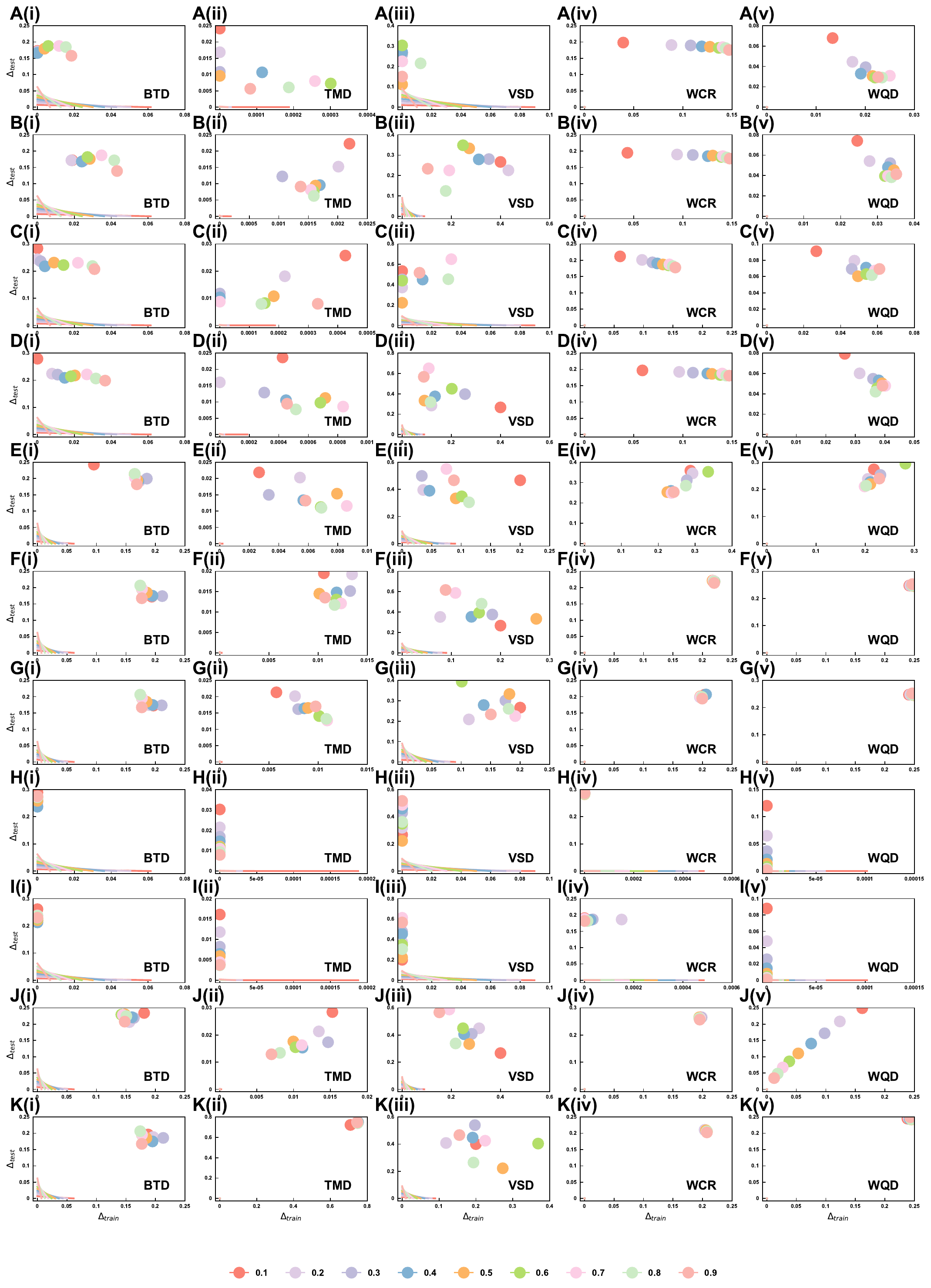}%
\caption{The loss errors of five additional datasets (BTD, TMD, VSD, WCR and WQD) in training ($\Delta_{train}^{f}$) and test sets ($\Delta_{test}^{f}$) of different binary classifiers, including XGBoost with four classical objectives (A-D), MLP (E), SVM (F), Logistic Regression (G), Decision Tree (H), Random Forest (I), KNN (J). Colorful dots and lines represent different $|\mathcal{S}_{train}|/|\mathcal{S}|$ ranging from $0.1$ to $0.9$.}
\label{figS16-9}
\end{figure}

\begin{figure}
    \centering
    \includegraphics[width=.9\linewidth]{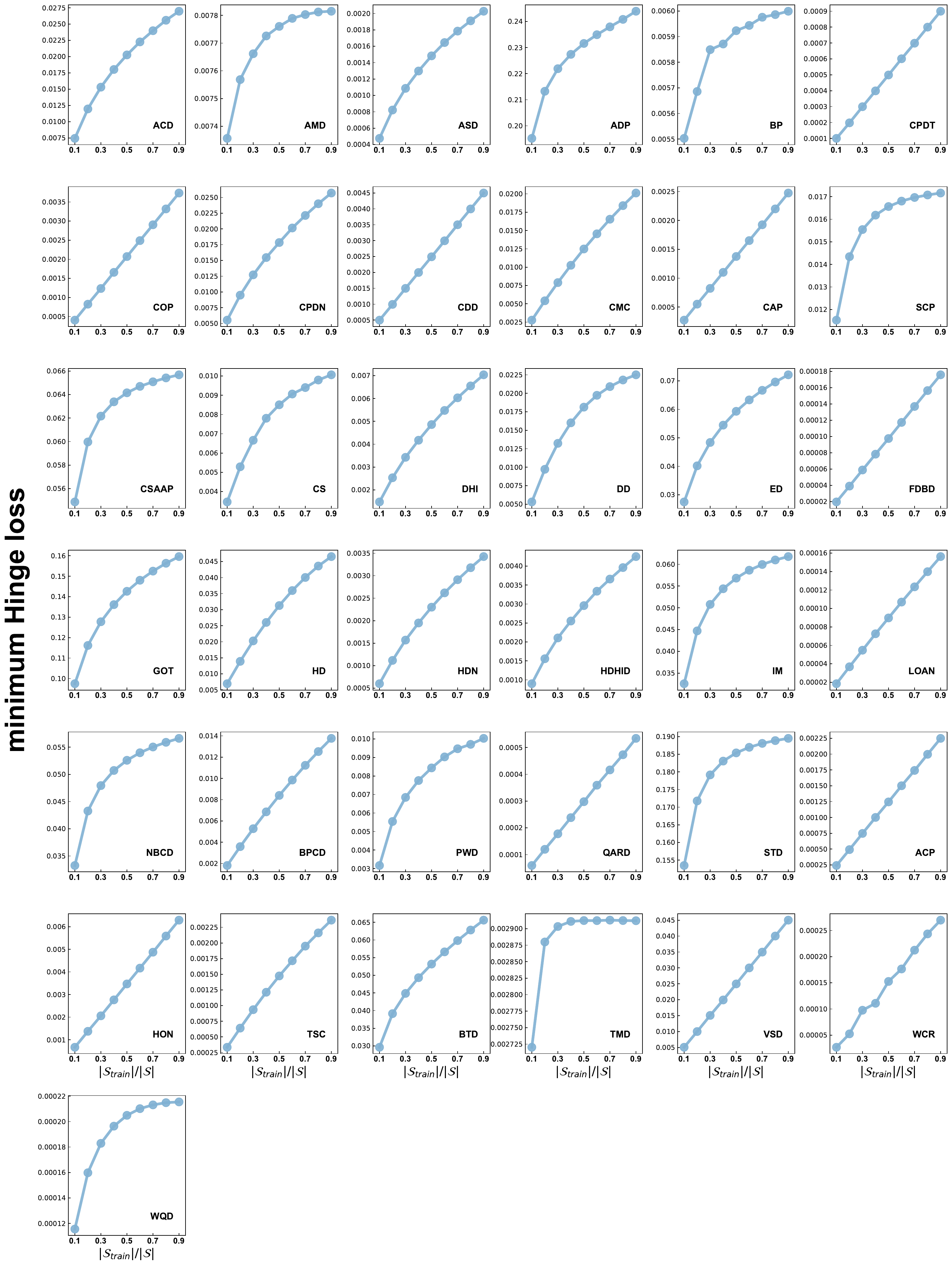}
    \caption{Minimum Hinge loss for 37 additional datasets under different random divisions In these panels, dots correspond to numerical results derived from the data divisions, while lines represent the theoretical predictions adjusted for the respective division ratios (see Eqs. 54 and 56).}
    \label{figS17}
\end{figure}

\begin{figure}
    \centering
    \includegraphics[width=.9\linewidth]{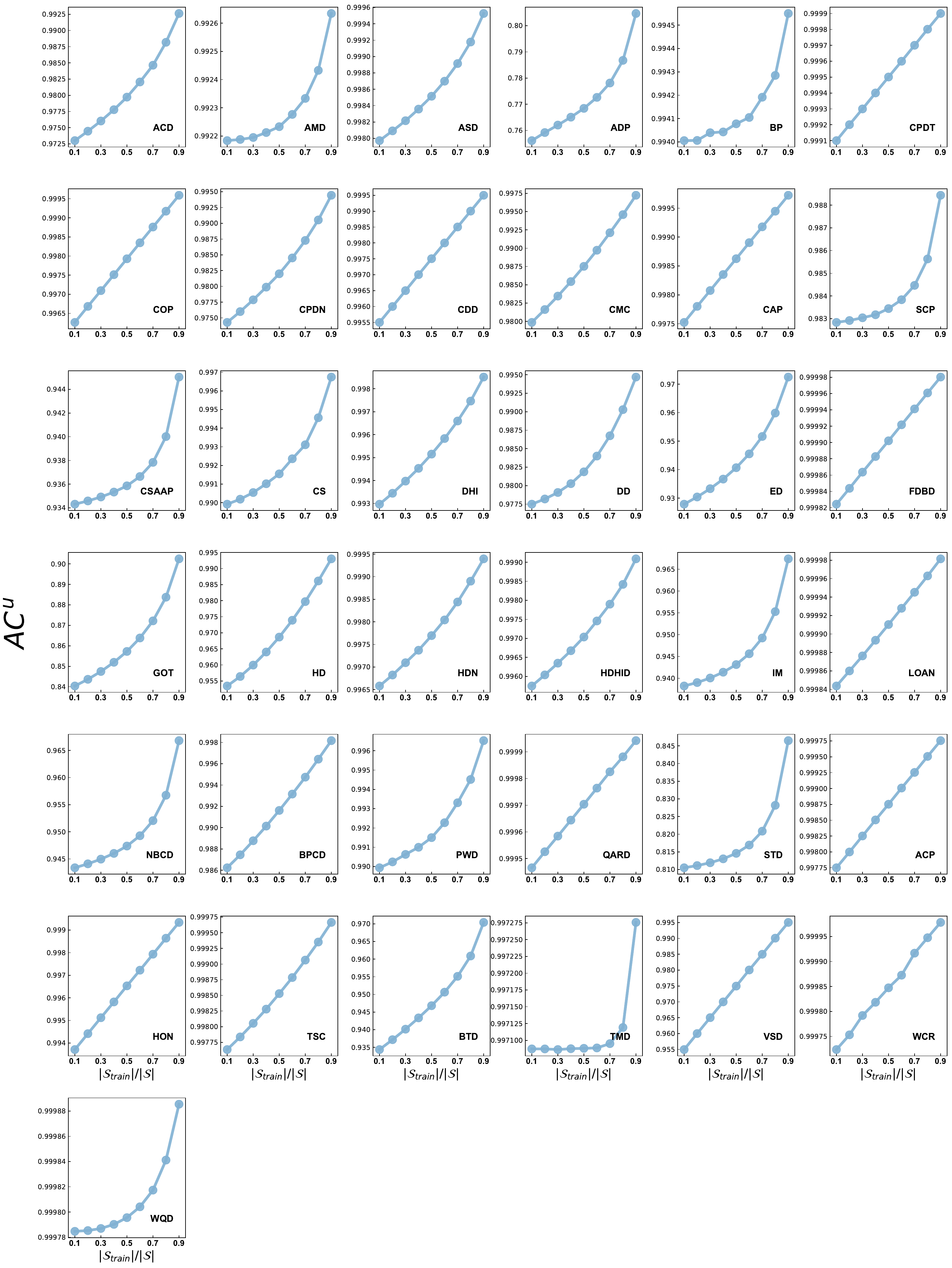}
    \caption{Upper bound of accuracy ($AC^{u}$) for 37 additional datasets under different random divisions In these panels, dots correspond to numerical results derived from the data divisions, while lines represent the theoretical predictions adjusted for the respective division ratios (see Eqs. 54 and 56).}
    \label{figS18}
\end{figure}

\begin{figure}
    \centering
    \includegraphics[width=.9\linewidth]{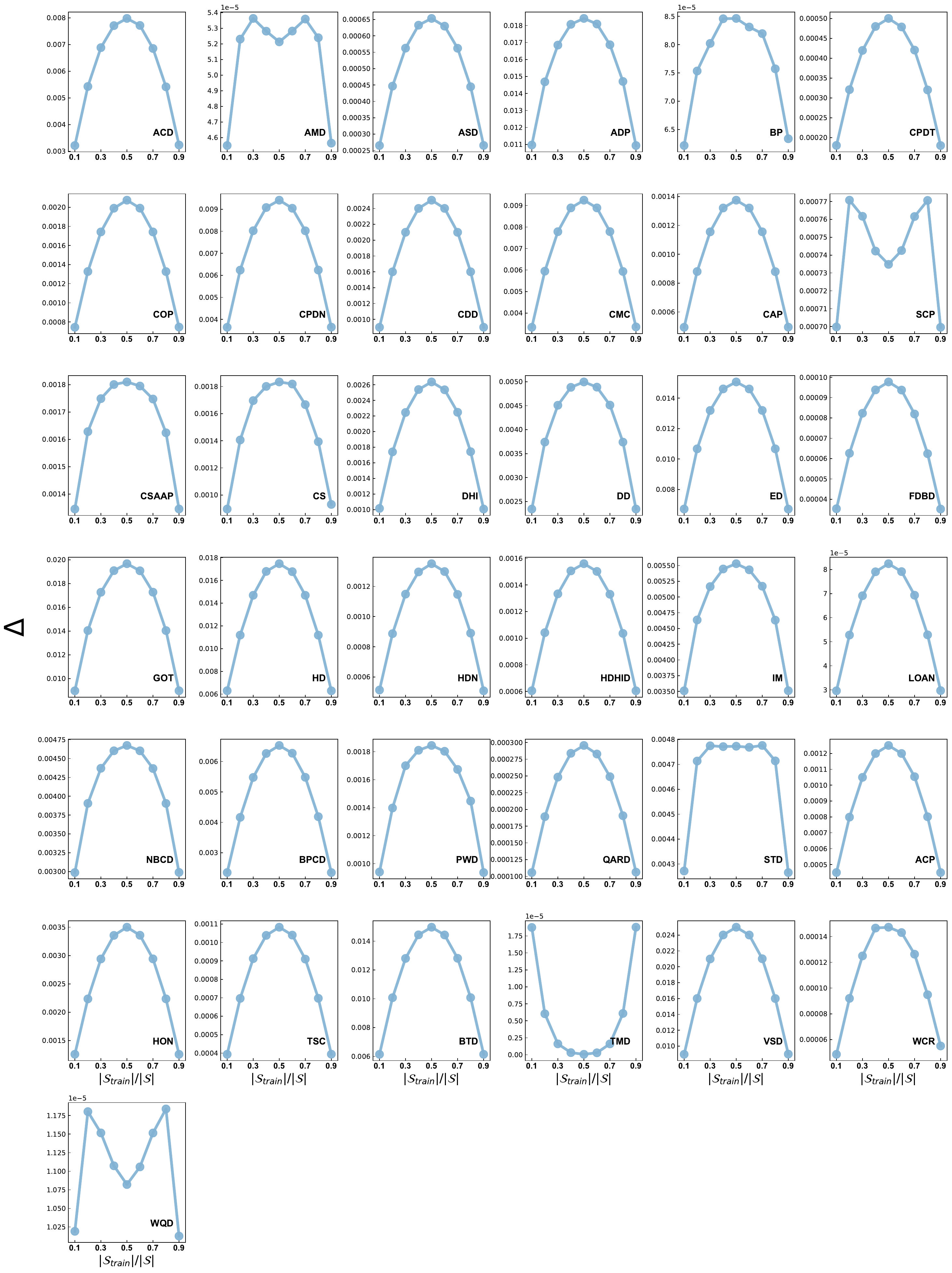}
    \caption{Anticipated optimal errors ($\Delta$) for 37 additional datasets under different random divisions In these panels, dots correspond to numerical results derived from the data divisions, while lines represent the theoretical predictions adjusted for the respective division ratios (see Eqs. 54 and 56).}
    \label{figS19}
\end{figure}

\begin{figure}
    \centering
    \includegraphics[width=.9\linewidth]{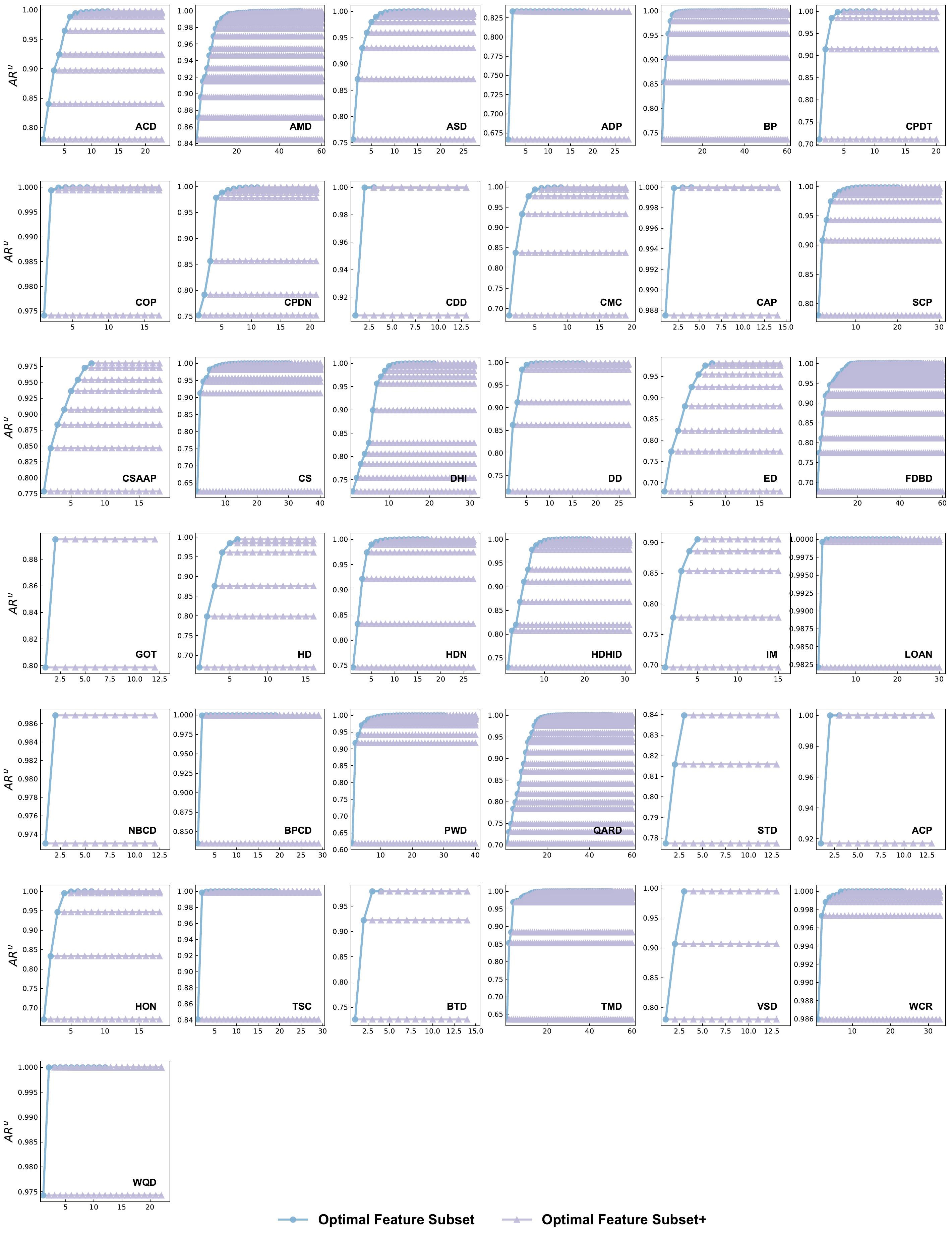}
    \caption{The $\text{AR}^u_{k_0}$ versus the optimal $k_0$ feature subset in feature selection (blue lines and dots) for 37 additional datasets. After we selected the optimal $k_0$ feature subset, we would use the feature extraction skill (LDA) to create new extracted features and add them into the original $k_0$ feature one by one (see red lines and dots).}
    \label{figS20}
\end{figure}

\begin{figure}
    \centering
    \includegraphics[width=.9\linewidth]{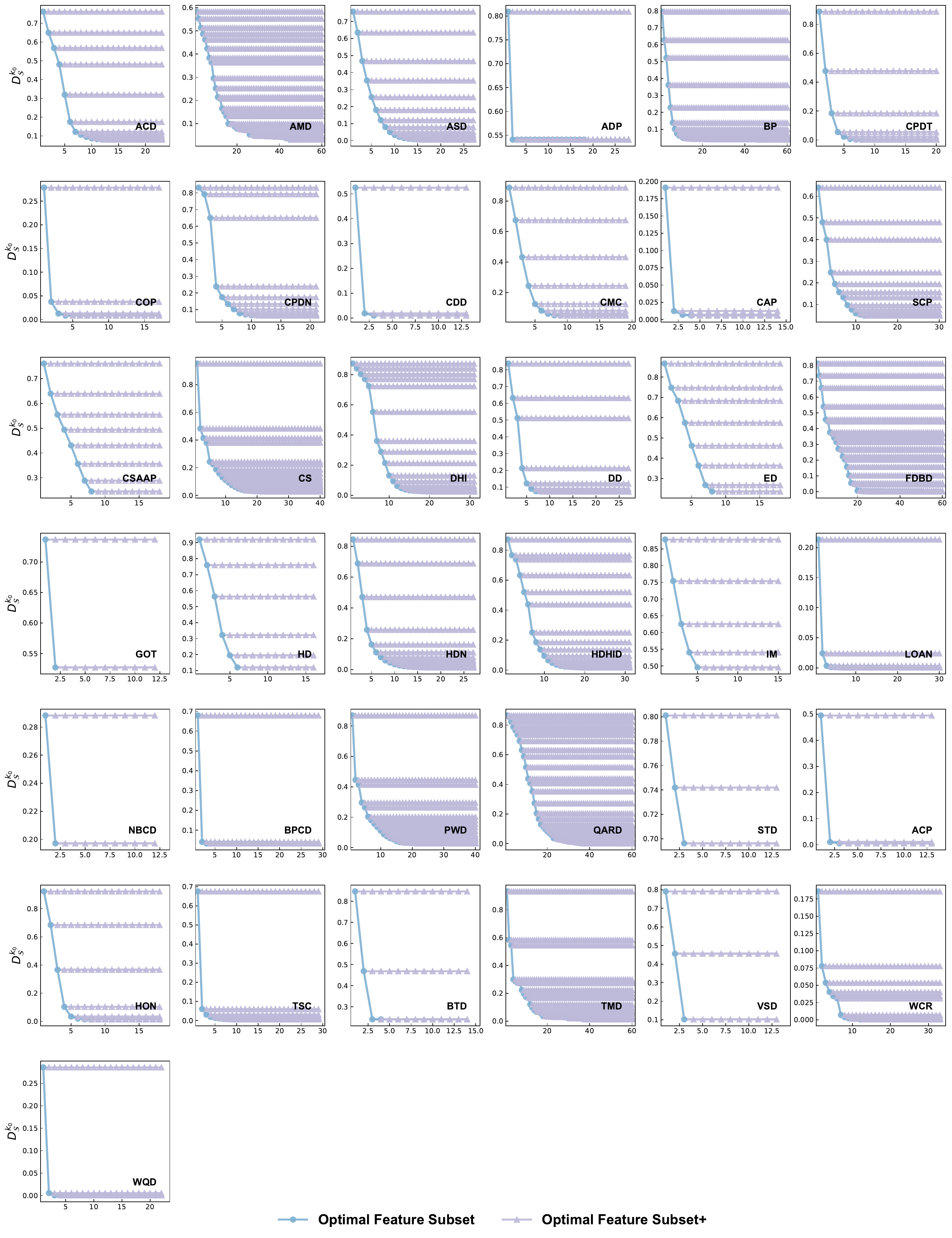}
    \caption{The $D^{k_0}_{\mathcal{S}}$ versus the optimal $k_0$ feature subset in feature selection (blue lines and dots) for 37 additional datasets. After we selected the optimal $k_0$ feature subset, we would use the feature extraction skill (LDA) to create new extracted features and add them into the original $k_0$ feature one by one (see red lines and dots)}
    \label{figS21}
\end{figure}